%% file: main.tex
\pdfoutput=1
\RequirePackage{fix-cm}
\documentclass[%
    twoside, openright, titlepage, numbers=noenddot,%
    cleardoublepage=empty,%
    BCOR=5.5mm, paper=a5, fontsize=10pt,
]{scrreprt}
\usepackage{etex}
\reserveinserts{10}

\usepackage{paralist}
\usepackage{mathtools}
\usepackage{dsfont}
\usepackage{enumerate}
\usepackage[abs]{overpic}
\usepackage[shortlabels]{enumitem}
\usepackage{bbm}
\usepackage{multicol}
\usepackage{subfiles}
\usepackage[dvipsnames,x11names]{xcolor}
\usepackage{color, colortbl}
\usepackage{pifont}

\input{preamble/general}

\usepackage[final]{pdfpages}

\usepackage{amsmath}
\usepackage{amssymb}

\input{preamble/math}

\begin{document}
\frenchspacing
\raggedbottom%
\selectlanguage{english}
\pagenumbering{roman}
\pagestyle{scrplain}

\definecolor{darkgreen}{RGB}{1, 128, 1}

%
%
%

%
%


\subfile{frontbackmatter/titlepage}

\subfile{frontbackmatter/titleback}

\cleardoublepage\subfile{frontbackmatter/dedication}
\cleardoublepage

\subfile{frontbackmatter/abstract}

\cleardoublepage\subfile{frontbackmatter/acknowledgments}
\pagestyle{scrheadings}
\cleardoublepage\subfile{frontbackmatter/contents}

\cleardoublepage%
\part{Introduction}
\cleardoublepage\pagenumbering{arabic}%

\subfile{chapters/01_introduction/introduction}

\cleardoublepage%

\subfile{chapters/02_related_work/related_work}

\cleardoublepage%
\subfile{chapters/03_background/background}

\part{Synthesis of Hand-Object Interaction}
\label{part:hoi}
\cleardoublepage%
\subfile{chapters/04_hoi_generation/hoi_generation}

\cleardoublepage%
\part{Human-to-Robot Handover}
\label{part:h2r}
\cleardoublepage%
\subfile{chapters/05_handover/handover}
\cleardoublepage%
\part{Conclusion}
\cleardoublepage%
\subfile{chapters/06_conclusion/conclusion}

\cleardoublepage

\subfile{frontbackmatter/bibliography}

\cleardoublepage%
\bookmarksetup{startatroot}
\pagenumbering{gobble}

\end{document}

%% file: preamble/general.tex
\usepackage[utf8]{inputenc}
\usepackage{ifthen}
\usepackage[main=english,ngerman]{babel}

\PassOptionsToPackage{%
  eulerchapternumbers,
  floatperchapter, pdfspacing,%
  beramono,%
  a5paper,%
}{classicthesis}

\input{preamble/meta}


\newcounter{dummy}

\newcommand{\ie}{i.\,e.\xspace}

\newcommand{\eg}{e.\,g.\xspace}

\usepackage{csquotes}  
\usepackage[T1]{fontenc}
\usepackage{xspace}
\usepackage{textcomp}
\usepackage{mparhack}
\usepackage{relsize}
\usepackage{amssymb}

\input{preamble/biblatex}

\let\origcite\cite%
\def\cite#1{\unskip~\origcite{#1}}

\usepackage{amsmath}
\usepackage{mathtools}
\usepackage{isomath}

\usepackage{tabularx}
\usepackage{ltablex}
\usepackage{floatrow}
\usepackage{caption}
\usepackage{subcaption}
\expandafter\def\csname ver@subfig.sty\endcsname{}

\usepackage{algorithmic}
\usepackage[ruled,vlined]{algorithm2e} 

\usepackage{blindtext}

\usepackage[dvipsnames]{xcolor}
\usepackage[%
  hyperfootnotes=false,%
  pdfpagelabels,%
]{hyperref}
\usepackage{hyperxmp}
\hypersetup{%
  colorlinks=true, linktocpage=true,%
  breaklinks=true, pdfpagemode=UseNone, pageanchor=true, pdfpagemode=UseOutlines,%
  plainpages=false, bookmarksnumbered, bookmarksopen=true, bookmarksopenlevel=1,%
  hypertexnames=true, pdfhighlight=/O,
  pdftitle={\myPlainTitle},%
  pdfauthor={\myName},%
  pdfcopyright={Copyright (C) \myTime, \myName},%
  pdfsubject={},%
  pdfkeywords={},%
  pdflang={en},%
}

\usepackage{graphicx}
\usepackage{rotating}
\usepackage{tikz}
\usepackage{pgfplots}
\pgfplotsset{compat=newest}
\usetikzlibrary{shapes.geometric}

\usepackage{cleveref}

\PassOptionsToPackage{printonlyused}{acronym}
\usepackage{acronym}

\usepackage{enumitem}

\usepackage{scrhack}
\usepackage{classicthesis}

\KOMAoptions{headinclude=true,footinclude=false}
\areaset[current]{\textwidth}{1.618034\textwidth}

\clearplainofpairofpagestyles
\ofoot[\pagemark]{}
\definecolor{chapter-color}{cmyk}{1, 0.50, 0, 0.25}
\definecolor{link-color}{cmyk}{1, 0.50, 0, 0.25}
\definecolor{cite-color}{cmyk}{0, 0.7, 0.9, 0.2}

\usepackage{bookmark}
\hypersetup{
  urlcolor=Black, linkcolor=Black, citecolor=Black, 
}

\let\chapterNumber\undefined%
\newfont{\chapterNumber}{eurb10 scaled 5500}

\usepackage{units}

\definecolor{niceblue}{HTML}{093458} 

\usepackage{etoolbox}
\makeatletter
\pretocmd{\chapter}{\addtocontents{toc}{\protect\addvspace{15\p@}}}{}{}
\makeatother

\titleformat{\chapter}[display]%
  {\relax}{\vspace*{-3\baselineskip}\makebox[\linewidth][r]{\color{halfgray}\chapterNumber\thechapter}}{10pt}%
  {\raggedright\spacedallcaps}[\normalsize\vspace*{.8\baselineskip}\titlerule]%

\def\chapterabstract#1{%
  \begingroup
  \baselineskip1.3em
  \leftskip1em
  \rightskip\leftskip\itshape#1
  \par
  \endgroup
}

\setkomafont{dictumtext}{\itshape\small}%
\setkomafont{dictumauthor}{\normalfont}

\usepackage{balance}       
\usepackage{graphics}      
\usepackage[T1]{fontenc}   
\usepackage{txfonts}
\usepackage{mathptmx}
\DeclareMathAlphabet{\mathcal}{OMS}{cmsy}{m}{n}
\usepackage{booktabs}
\usepackage{textcomp}
\usepackage{graphicx,import}

\usepackage{multirow}
\usepackage{makecell}
\usepackage{array}
\usepackage{bbm}
\usepackage{xcolor,colortbl}
\usepackage{mdframed}
\usepackage{ifthen}
\usepackage{mathtools}
\usepackage[export]{adjustbox}
\usepackage{xspace}

\usepackage[normalem]{ulem}

\input{preamble/macros}

\input{preamble/names}
\usepackage{bibentry}
\usepackage{tabularx}

\makeatletter
\DeclareRobustCommand\onedot{\futurelet\@let@token\@onedot}
\def\@onedot{\ifx\@let@token.\else.\null\fi\xspace}

\def\eg{\emph{e.g}\onedot} 
\def\ie{\emph{i.e}\onedot} 
 
 \def\vs{\emph{vs}\onedot}

\makeatother


%% file: preamble/meta.tex
\newcommand{\myTitle}{%
  Modeling Dynamic Hand-Object Interactions with Applications to Human-Robot Handovers
\xspace%
}

\newcommand{\myPlainTitle}{%
   modeling dynamic hand-object interactions with applications to human-robot handovers
\xspace%
}

\newcommand{\myDissNumber}{30710}
\newcommand{\myName}{Sammy Joe Christen\xspace}

\newcommand{\myTime}{2024\xspace}

\newcommand{\myDOI}{10.3929/ethz-b-000721218}

\usepackage{scrtime}


%% file: preamble/biblatex.tex
\usepackage[
  style=nature,%
  natbib=true,%
  clearlang=true,%
  backend=biber,%
]{biblatex}

\AtEveryBibitem{%
  \clearfield{day}%
  \clearfield{month}%
  \clearfield{endday}%
  \clearfield{endmonth}%
}

\DeclareRedundantLanguages{en,EN,English}{english}

\DeclareFieldFormat{pages}{\mkfirstpage{#1}}

%% file: preamble/macros.tex
\definecolor{darkgreen}{RGB}{1, 128, 1}
\definecolor{darkred}{RGB}{128, 1, 1}
\definecolor{teal}{RGB}{1, 160, 160}

\newcommand{\newparagraph}[1]{\\\mbox{}\\}

\newcommand{\figref}[1]{Figure~\ref{#1}}
\newcommand{\tabref}[1]{Table~\ref{#1}}

\newcommand{\secref}[1]{Section~\ref{#1}}
\newcommand{\chapref}[1]{Chapter~\ref{#1}}

\newcommand{\contribution}[1]{
\begin{mdframed}[backgroundcolor=gray!20, hidealllines=true]
\vspace{1em}
\chapterabstract{#1}
\vspace{1em}
\end{mdframed}
\clearpage}

\newcommand{\challenges}[1]{
\begin{mdframed}[backgroundcolor=Apricot!8, hidealllines=true]
\vspace{1em}
\chapterabstract{#1}
\vspace{1em}
\end{mdframed}}

\newcommand{\hypotheses}[1]{
\begin{mdframed}[backgroundcolor=CadetBlue!8, hidealllines=true]
\vspace{1em}
\chapterabstract{#1}
\vspace{1em}
\end{mdframed}
}

\DeclarePairedDelimiterX{\norm}[1]{\lVert}{\rVert}{#1}

\newcommand{\Fig}[1]{Fig.~\ref{fig:#1}}

\newcommand{\Tab}[1]{Tab.~\ref{tab:#1}}

\newcommand{\Eq}[1]{Eq.~\ref{eq:#1}}

\newcommand{\Sec}[1]{Sec.~\ref{sec:#1}}

\newcommand{\deliui}[1]{}
\newcommand{\del}[1]{}

\makeatletter
\DeclareCiteCommand{\fullcite}
  {\defcounter{maxnames}{\blx@maxbibnames}%
    \usebibmacro{prenote}}
  {\usedriver
     {\DeclareNameAlias{sortname}{default}}
     {\thefield{entrytype}}}
  {\multicitedelim}
  {\usebibmacro{postnote}}
\makeatother

\definecolor{hexorange}{HTML}{FF9900}
\definecolor{hexblue}{HTML}{3C78D8}
\DeclareRobustCommand*\circledorange[1]{\tikz[baseline=(char.base)]{
            \node[hexorange,shape=circle,draw,inner sep=1pt, thick=8pt, scale=0.8] (char) {#1};}}
       
\DeclareRobustCommand*\circledblue[1]{\tikz[baseline=(char.base)]{
            \node[hexblue,shape=circle,draw,inner sep=1pt, thick=8pt, scale=0.8] (char) {#1};}}

\newcommand{\gc}{\cellcolor[HTML]{E4E4E4}}
\definecolor{Gray}{gray}{0.9}

\newcolumntype{C}[1]{>{\centering\let\newline\\\arraybackslash\hspace{0pt}}m{#1}}
\newcolumntype{L}[1]{>{\raggedright\let\newline\\\arraybackslash\hspace{0pt}}m{#1}}
\newcolumntype{R}[1]{>{\raggedleft\let\newline\\\arraybackslash\hspace{0pt}}m{#1}}

\makeatletter
\def\and{
  \end{tabular}%
  \hskip 2.85em \@plus.17fil%
  \begin{tabular}[t]{c}}
\makeatother

\newcommand{\cmark}{\ding{51}}%
\newcommand{\xmark}{\ding{55}}%

%% file: preamble/names.tex
\newcommand{\taskname}{dynamic grasp synthesis}
\newcommand{\dgrasp}{\textit{D-Grasp}\xspace}
\newcommand{\artigrasp}{\textit{ArtiGrasp}\xspace}
\newcommand{\taskartigrasp}{relocation and articulation}
\newcommand{\raisim}{RaiSim\xspace}
\newcommand{\dgraspTitle}{D-Grasp: Physically Plausible Dynamic Grasp Synthesis for Hand-Object Interactions}
\newcommand{\artigraspTitle}{ArtiGrasp: Physically Plausible Synthesis of Bi-Manual Dexterous Grasping and Articulation}
\newcommand{\handoverTitle}{Learning Human-to-Robot Handovers from Point Clouds}
\newcommand{\synhrTitle}{SynH2R: Synthesizing Hand-Object Motions for Learning Human-to-Robot Handovers}

%% file: preamble/math.tex
\newcommand{\policy}{\boldsymbol{\pi}}
\DeclareMathOperator*{\argmax}{arg\,max}

\newcommand{\mdp}{\mathcal{M}}
\newcommand{\StatesSet}{\mathcal{S}}
\newcommand{\ActionsSet}{\mathcal{A}}
\newcommand{\TransitionSet}{\mathcal{T}}

\newcommand{\goals}{\mathbf{G}}

\newcommand{\actions}{\mathbf{a}}
\newcommand{\state}{\mathbf{s}}
\newcommand{\discountRate}{\gamma}
\newcommand{\forcevec}{\mathbf{f}}

\newcommand{\posesix}{\mathbf{T}}
\newcommand{\posvec}{\mathbf{x}}
\newcommand{\posevec}{\mathbf{q}_h}
\newcommand{\posevecgen}{\mathbf{q}}

\newcommand{\grasplabel}{\mathbf{D}}
\newcommand{\torques}{\boldsymbol{\tau}}
\newcommand{\policyvec}{\boldsymbol{\pi}}
\newcommand{\gpolicyvec}{\boldsymbol{\pi}_{g}}

\newcommand{\V}[1]{\mathbf{#1}} 

\newcommand{\grasppred}{\boldsymbol{\sigma}}
\newcommand{\pcvec}{\mathbf{P}}
\newcommand{\policypre}{\boldsymbol{\pi_{\text{pre}}}}
\newcommand{\criticpre}{Q_{\text{pre}}}
\newcommand{\transitionvec}{\mathbf{d}}
\newcommand{\replay}{\mathbf{\mathcal{D}}}
\newcommand{\policyexp}{\boldsymbol{\pi_{\text{exp}}}}
\newcommand{\policyft}{\boldsymbol{\pi}_*}
\newcommand{\criticft}{Q_*}
\newcommand{\rewardvec}{\mathbf{r}}
\newcommand{\goalvec}{\mathbf{g}}
\newcommand{\expertvec}{\mathbf{e}}

%% file: frontbackmatter/titlepage.tex
\begin{titlepage}
    \begin{center}
        \large
        \begingroup
            \spacedlowsmallcaps{Diss. ETH No. \myDissNumber}
        \endgroup

        \hfill

        \vfill

        \begingroup
            \spacedallcaps{\myTitle}
        \endgroup

        \vfill

        \begingroup
            A thesis submitted to attain the degree of\\
            \vspace{0.5em}
            \spacedlowsmallcaps{Doctor of Sciences}\\
            (Dr.\ sc.\ ETH Zurich)
        \endgroup


        \begingroup
            \vspace{3.5em}
            presented by\\
            \vspace{3.5em}
            \spacedlowsmallcaps{\myName}\\
            M.Sc., ETH Zurich\\
            \vspace{0.5em}
            born on 26.07.1991\\
        \endgroup

        \vfill

        \begingroup
            accepted on the recommendation of\\
            \vspace{0.5em}
            Prof.\ Dr.\ Stelian Coros \\
            Prof.\ Dr.\ Jemin Hwangbo \\
            Prof.\ Dr.\ Li Yi \\
        \endgroup

        \vfill

        \myTime%

        \vfill
    \end{center}
\end{titlepage}

%% file: frontbackmatter/titleback.tex
\thispagestyle{empty}

\hfill

\vfill

\noindent\myName: \textit{\myTitle} 
\textcopyright\ \myTime

\bigskip

\noindent\spacedlowsmallcaps{DOI}: \myDOI

%
%
%
%
%

%% file: frontbackmatter/dedication.tex
\thispagestyle{empty}
\refstepcounter{dummy}

\vspace*{3cm}

\par\vspace*{.35\textheight}{\centering \textit{Dedicated to my parents}\par}

\medskip

%% file: frontbackmatter/abstract.tex
\pdfbookmark[1]{Abstract}{Abstract}
\begingroup
\let\clearpage\relax
\let\cleardoublepage\relax
\let\cleardoublepage\relax

\chapter*{Abstract}

Humans constantly grasp, manipulate, and move objects in their daily lives. Interactive systems aim to assist humans in performing these tasks, both in the real and virtual world. Building systems capable of understanding how humans interact with objects and generating such hand-object interactions could enable new applications, such as simulating human behavior for Embodied AI and human-robot interaction, producing animations in virtual reality settings, and augmenting pose estimation models with synthetic data during training. However, current methods for synthesizing human hand-object interaction either neglect the dynamic aspects of these interactions, such as static prediction of a hand-object grasp, or rely on ground-truth hand and object poses during inference. To make these models truly effective in assisting humans, we require scalable solutions that go beyond the static prediction of grasps.

In the first part of this dissertation, we introduce two novel tasks for modeling dynamic hand-object interactions in 4D (3D space + 1D time). First, we introduce the problem of dynamic grasp synthesis, going beyond the static generation of hand-object grasps. Dynamic grasp synthesis involves learning to grasp rigid objects with a single human hand and move them to a 6D target pose. We approach this task using physical simulation and reinforcement learning. Our approach involves splitting the interaction into a grasping stage and a motion synthesis stage, and a general reward function that combines incentives for grasp stability and human-like grasping. Second, since humans frequently perform bi-manual manipulation with articulated objects, we extend our approach to include both hands and complex articulations, which requires more fine-grained grasping and coordination between two hands. To address this, we introduce a learning curriculum and expand the observation space. Our experiments demonstrate that our methods significantly outperform baselines on these novel tasks in terms of grasp stability and physics metrics.

Building on this foundation, the second part of this dissertation explores the application of hand-object interaction synthesis to the challenge of human-to-robot handovers. This task remains challenging due to the difficulty of accurately simulating realistic human behavior. To address this, our approach integrates captured human-object motion data into a physical simulation environment, enabling the simulation of realistic human motions. We then introduce the first framework for end-to-end training of robotic handover policies using a simulated human-in-the-loop. Our system employs a two-stage student-teacher training framework that gradually learns to adapt to human motions. Our experiments show that this approach significantly outperforms existing learning-based solutions in both simulated and real-world settings. One key limitation of using captured data of human-object motions is the restricted amount of available data. To address this, we integrate our method for synthesizing hand-object interactions with the training of robotic policies for human-to-robot handovers. Specifically, we increase the diversity in training objects for the robot by 100x and create a large scale synthetic test set. Our experiments reveal that training the robot on a broader distribution of objects and human motions leads to improved success rates in grasping unseen objects in simulation compared to our previous work. Moreover, we find in a qualitative user study that users cannot distinguish robotic policies trained on purely synthetic human motions and those trained with real human motions. 

This dissertation demonstrates the capability of generating dynamic 4D hand-object interactions. This paves the way to better understand and assist humans in performing such interactions, which we showcase in the context of human-to-robot handovers. To that end, we use hand-object motions generated through our models for training handover policies in simulation and transferring them to the real system. This highlights the potential of synthetic data for scalable, human-aware robotic systems in the future.

\endgroup

\cleardoublepage%

\begingroup
\let\clearpage\relax
\let\cleardoublepage\relax
\let\cleardoublepage\relax

\begin{otherlanguage}{ngerman}
\pdfbookmark[1]{Zusammenfassung}{Zusammenfassung}
\chapter*{Zusammenfassung}

Menschen greifen, manipulieren und bewegen ständig Objekte in ihrem täglichen Leben. Um solche Interaktionen zu unterstützen, zielen interaktive Systeme darauf ab, Menschen bei der Lösung solcher Aufgaben zu helfen. Der Aufbau von Systemen, die in der Lage sind, solche Hand-Objekt-Interaktionen zu verstehen und zu generieren, eröffnet neue Anwendungsmöglichkeiten, wie die Erstellung von Animationen, die Simulation menschlichen Verhaltens für Embodied AI-Aufgaben und Mensch-Roboter-Interaktionen sowie die Ergänzung von Pose-Estimation Modellen mit synthetischen Daten während des Trainings. Aktuelle Methoden zur Synthese von Hand-Objekt-Interaktionen vernachlässigen jedoch entweder die dynamischen Aspekte dieser Interaktionen, wie die statische Vorhersage eines Hand-Objekt-Griffs, oder sie verlassen sich auf kostspielige menschliche Demonstrationen für das Training. Um diese Modelle wirklich auf Anwendungen, die Menschen unterstützen, anwendbar zu machen, benötigen wir skalierbare Lösungen, ohne auf teure menschliche Demonstrationen angewiesen zu sein.

Im ersten Teil dieser Dissertation stellen wir zwei neuartige Probleme zur Modellierung dynamischer Hand-Objekt-Interaktionen in 4D (3D Raum + 1D Zeit) vor. Zunächst führen wir das Problem der dynamischen Griff-Synthese ein, das über die statische Generierung von Griffen hinausgeht. Die dynamische Griff-Synthese umfasst Objekte mit einer menschlichen Hand zu greifen und sie in eine 6D-Zielpose zu bewegen. Wir gehen dieses Problem mit physikalischer Simulation und Reinforcement Learning an. Unser Ansatz teilt die Interaktion in eine Greifphase und eine Bewegungssynthesephase auf und verwendet eine generelle Reward Funktion, die Anreize für Griffstabilität und menschliches Greifen kombiniert. Zweitens, da Menschen häufig bimanuelle Manipulationen mit artikulierten Objekten durchführen, erweitern wir unseren Ansatz auf beide Hände und komplexe Artikulationen, was ein feineres Greifen und eine bessere Koordination zwischen beiden Händen erfordert. Um dies zu bewältigen, führen wir ein Curriculum ein und erweitern den Observation Space. Unsere Experimente zeigen, dass unsere Methoden in Bezug auf Griffstabilität und physikalische Metriken die Baseline-Modelle bei diesen neuartigen Aufgaben deutlich übertreffen.

Aufbauend auf dieser Grundlage untersucht der zweite Teil dieser Dissertation die Anwendung von Hand-Objekt-Interaktionsmodellen auf die Herausforderung der Übergabe von Objekten zwischen Mensch und Roboter. Dieses Problem bleibt aufgrund der Schwierigkeit, realistisches menschliches Verhalten zu simulieren, eine Herausforderung. Um dies zu bewältigen, stellen wir die erste Methode für das End-to-End-Training von Robotern unter Verwendung eines simulierten menschlichen Systems vor. Unser Ansatz integriert erfasste menschliche Daten in eine physikalische Simulationsumgebung und ermöglicht so die Simulation realistischer menschlicher Bewegungen. Unser System verwendet einen zweistufigen Lehrer-Schüler Ansatz, der schrittweise lernt, sich an menschliche Bewegungen anzupassen. Unsere Experimente zeigen, dass dieser Ansatz bestehende, lernbasierte Lösungen sowohl in simulierten als auch in realen Umgebungen deutlich übertrifft. Eine Einschränkung bei der Verwendung erfasster Daten menschlicher Daten ist die begrenzte Menge an verfügbaren Daten. Um dies zu beheben, integrieren wir unsere Methode zur Synthese von Hand-Objekt-Interaktionen in das Training von Robotern für Mensch-zu-Roboter Objekt Übergaben. Insbesondere erhöhen wir die Vielfalt der Trainingsobjekte für den Roboter um das 100-fache und erstellen ein gross angelegtes synthetisches Testset. Unsere Experimente zeigen, dass das Training des Roboters auf einer breiteren Verteilung von Objekten und menschlichen Bewegungen zu höheren Erfolgsraten beim Greifen unbekannter Objekte in der Simulation führt, verglichen mit unserer früheren Arbeit. Darüber hinaus zeigt eine qualitative Nutzerstudie, dass Nutzer keinen Unterschied zwischen Robotern, die auf rein synthetischen menschlichen Bewegungen trainiert wurden, und solchen, die mit realen menschlichen Bewegungen trainiert wurden, feststellen können.

Unsere Ergebnisse zeigen die Fähigkeit zur Generierung dynamischer 3D-Hand-Objekt-Interaktionen. Dies ebnet den Weg zu einem besseren Verständnis und einer besseren Unterstützung von Menschen bei der Durchführung solcher Interaktionen, was wir im Kontext der Mensch-zu-Roboter Objekt-Übergabe demonstrieren. Zu diesem Zweck verwenden wir durch unsere Modelle generierte Hand-Objekt-Bewegungen, um Roboter in der Simulation zu trainieren und auf das reale System zu transferieren. Dies unterstreicht das Potenzial synthetischer Daten für skalierbare, menschenzentrierte Robotiksysteme in der Zukunft.

\end{otherlanguage}

\endgroup

\vfill

%% file: frontbackmatter/acknowledgments.tex
\pdfbookmark[1]{Acknowledgements}{acknowledgements}

\bigskip

\begingroup
\let\clearpage\relax
\let\cleardoublepage\relax
\let\cleardoublepage\relax
\chapter*{Acknowledgements}

\def\thanks#1{%
\begingroup
\leftskip1em
\noindent #1
\par
\endgroup
}

First and foremost, I would like to thank my doctoral advisor Otmar Hilliges for his unwavering support and guidance throughout my doctorate. Without his encouragement and believe in my skills I would likely not have started a doctorate. I learned a great deal from Otmar, especially when it comes to structuring research projects and writing papers. His guidance has allowed me to excel in these tasks. Otmar, I wish to see you get back to full health and keep on pushing the boundaries of research. I could not have asked for a better advisor.

I want to thank my committee for agreeing to review this dissertation. Thanks to Stelian for stepping in for Otmar and helping me with the organization of the final part of my doctorate. I'm grateful to Jemin for his support with the grasping projects, and especially for the help with physical simulation. I'd like to thank Li for inviting me to present my work in his group and the nice discussions we had.

I would like to thank Emre for being a mentor, friend, and inspiration. You helped my PhD get on the right track early on! 
I am very grateful for Jie's support, for helping me master the most difficult time of my PhD and figure out a fruitful research direction to work on, which lead to many follow-up projects and student supervisions. You are such a huge and humble contributor to the success of the AIT lab! Thanks to Juan, for the great barbeques and for taking over so many administrative tasks of the AIT lab lately. Thank you Monica for handling the administrative side of AIT and helping me deal with HR. Thank goes out to Stefan, for the many cool projects we did together and also for his role as advisor in my master thesis. I want to thank Christoph for the good times, for teaching me what contextual bandits are and being a living example of a good work-life-balance. 
Thanks to Thomas for the many laughs, the beers at Kennedys, the (successful) pub quizes, and sticking together through the good and bad times of our PhDs. I would like to thank Marcel for being a great sports buddy, the fun times skating together, and the great backyard parties. I want to thank Alexis for being an awesome office mate and the fun times we had with Huggiebot. Thanks to Mert for carrying on with RL at AIT. Thanks to Velko for providing us with food in reading groups week after week and showing us what modern, cool HCI projects look like. Thanks to Muhammed for the support with D-Grasp. Thanks to Xu for not inviting me to the Chinese new year party, but also to the inspiration of how one can be a successful researcher despite constantly watching Dota tournaments on one screen. I want to thank Alex for the collaboration on some of our hand projects. I'd like to thank Manuel for being the arguably best ITC in ITC history, addressing issues before I had a chance to see them. I'd like to thank Wookie for being a role model and doing the best voice-overs before AI generated text-to-speech became a thing. Thanks to Artur and Hsuan-I for being the next generation of ITCs. I am grateful for Adrian helping me with related work on hand-object, although mostly citing his own papers. Thanks to Yufeng for the many fun lunches and I already wish you all the best in the future as professor. Thanks to Morteza for the nice conversations about research, Chen for teaching us coding, and Lixin for managing our webpage. Thank you Xucong for being the coolest postdoc to have beers with! Thank you to Xi for being a great office mate. I would like to thank all my amazing students that helped push forward my research direction. Lukas for the collaboration on the HiDe project to get my research going, Nikola for the collaboration on the manipulation project. I want to thank Jona for his devotion to make the grasping project into a success. Thanks to Marco for the nice and successful collaboration on the SFP project. I also want to thank Hui for taking over the ArtiGrasp project and the collaboration on the projects that followed. Thanks to the rest of the AIT lab for the good times, the conversations during lunch and coffee, the breaks jumping into the Limmat, and showing up at my wedding! 

I want to say thanks to the whole team at Meta Zurich that made my internship experience unique. First, thanks to my main advisor Bugra for his support throughout my project. Also, I want to personally thank Shreyas, Fadime, Edo, Stephanie, Eric, Cihan, Shitai, Anna, Evin, Tomas, Shugao, and Andrei for the great times during my internship. Also thanks to Muchen and Korrawe for the late night discussions about which metrics to use and how to improve my project. Lastly, thanks to Zen for the nice collaboration on the grasping project.

I am thankful for the people at NVIDIA Robotics, I thoroughly enjoyed my time there. I want to thank my main advisor Yu-Wei for the close collaboration, having a hands-on advisor was great! I am grateful for the support of Wei, the guidance throughout the project and the help with all the real world experiments. Thanks to Claudia for being a great collaborator, and the fun discussions about embodied AI. Lastly, I want to thank Dieter for the opportunity to be part of his team.

None of this would have been possible without my friends and family, who have supported me throughout all these years, stood by my side, and provided advice whenever needed! Thanks especially to my parents Martin and Hildegard and my sister Jennifer, for always believing in me and supporting what I do. I am thankful to my grand-parents, Anna, Josef, Adolf and Margrith, my extended family Peter, Anja, Thomas, Susann, Silvia. At the heart of this, I am eternally grateful to my wife Caroline for all the patience, taking tasks off my back during deadline times, keeping our apartment intact, putting up with my work schedule and the extra shifts while on vacation. Your unconditional love and support in everything I do continue to help me grow into a better person. Thank you.

\endgroup

%% file: frontbackmatter/contents.tex
\refstepcounter{dummy}
\pdfbookmark[1]{\contentsname}{tableofcontents}
\setcounter{tocdepth}{1} 
\setcounter{secnumdepth}{3} 
\manualmark%
\markboth{\spacedlowsmallcaps{\contentsname}}{\spacedlowsmallcaps{\contentsname}}
\makeatletter
\renewcommand{\toclevel@part}{10}
\makeatother
\bookmarksetup{level=part}
\tableofcontents
\automark[section]{chapter}
\renewcommand{\chaptermark}[1]{\markboth{\spacedlowsmallcaps{#1}}{\spacedlowsmallcaps{#1}}}
\renewcommand{\sectionmark}[1]{\markright{\thesection\enspace\spacedlowsmallcaps{#1}}}

%% file: chapters/01_introduction/introduction.tex
\def\dir{chapters/01_introduction}

\chapter{Introduction}
\label{ch:introduction}

\input{\dir/content/motivation}

\input{\dir/content/approach}

\clearpage
\input{\dir/content/contributions}
\clearpage
\input{\dir/content/structure_of_thesis}
\clearpage
\input{frontbackmatter/publications}

%% file: chapters/01_introduction/content/motivation.tex
\section{Motivation}
\label{sec:motivation}

Humans interacting with their environment is a fundamental aspect of everyday life, with object and tool manipulation being among the most common forms of interaction. Interactive systems aim to support humans in their ability to effectively perform such interactions. Enabling these systems to understand and perform human-object interactions in 4D (3D space + 1D time) can enable effective interactions with the physical world and assist humans in various domains. For example, in the setting of a home-assisted living robot, a robot may need to support a person by taking a glass from their hand and placing it into a dishwasher. Successfully executing this task requires the robot to understand the human's intent, including how the glass is being held and positioned, and to use effective algorithms to securely grasp the glass and transfer it to the dishwasher. In this scenario, an environment with a human-in-the-loop could enable a realistic training ground for the robot. Therefore, a model capable of generating a large diversity of human hand-object interactions would provide a scalable and efficient solution to simulate human behavior. The ability to generate 4D hand-object interactions cannot only enhance intelligent systems to better understand human behavior but also provides an affordable source of synthetic data for training in various other applications, making this problem particularly interesting.

Despite growing efforts to collect datasets of human hand-object interactions, a critical gap remains; much of the existing research has focused primarily on predicting hand poses from images or generating static grasps given an object mesh \cite{jiang2021hand, karunratanakul2020grasping, corona2020ganhand}. However, to truly understand and effectively assist humans, it is essential to model 4D human-object interactions. This gap limits the potential application of such models in downstream tasks. For instance, in augmented or virtual reality (AR/VR), these models could enhance immersiveness by correcting physically implausible grasps or synthesizing realistic hand movements when the hand is not visible to the cameras. Furthermore, combining generated 4D human-object interaction data with photorealistic rendering techniques could provide affordable means of augmenting data for hand and object pose estimation. One particularly promising application is the simulation of human behavior for Embodied AI tasks and human-robot interaction (HRI). Currently, the standard approach for training robots is to rely on physical simulation, as they provide a more safe and efficient environment for development. However, existing methods for human-robot interaction assume humans are static elements of the environment \cite{yang2021reactive} or neglect the human completely \cite{wang2021goal}. By synthesizing dynamic human behavior in 4D, we can significantly increase the realism of these simulations, allowing robots to interact with more accurate human models during training.

The use of existing hand-object interaction datasets has led to in the development of frameworks that generate static grasps from object meshes \cite{karunratanakul2020grasping, jiang2021hand, corona2020ganhand}. These frameworks introduced various interesting model architectures based on generative adversarial networks \cite{corona2020ganhand}, implicit representations \cite{karunratanakul2020grasping}, or conditional variational auto-encoders \cite{jiang2021hand}. These data-driven methods typically rely on static grasp datasets, which are more abundant than datasets containing sequential data (cf. Table \ref{tab:compare_datasets}). However, such models are known to be data-hungry and prone to overfitting. Applying them to limited sequential data in current datasets (less than 1000 sequences) results in generative models that struggle to synthesize data beyond their training distribution \cite{ghosh2023imos}. For instance, an effective generative model of hand-object interactions should be capable of achieving task goals provided at inference time, such as moving objects to various target poses unseen during training. Additionally, in 4D generation, hand and object poses need to be synchronized, physically plausible, and human-like. The inherent dexterity of the human hand further increases the complexity of this task. However, collecting more sequential data to train these models is costly and requires extensive engineering efforts. Addressing these challenges therefore requires new methodologies   capable of generating sequences that go beyond the constraints of existing datasets.

\emph{In this thesis, we investigate the novel problem of effectively generating hand-object interactions in 4D and explore the utility of such models in aiding the understanding and training of algorithms, specifically in the context of human-to-robot handovers.} To address the synthesis of 4D hand-object interactions, we introduce two novel tasks and propose a framework based on reinforcement learning and physical simulation to solve these tasks (Part \ref{part:hoi}). This approach helps mitigate the data scarcity and learn models that can flexibly generate human-object interactions based on simple task goals, such as reaching a target object pose, while maintaining natural interactions by integrating human-like priors from single-frame grasping datasets. To demonstrate the practical value of our model for improving the understanding of human motions and training algorithms in downstream domains, we explore the application of human-to-robot handovers (Part \ref{part:h2r}). We first propose to embed captured human motions into simulation to train a robotic policy capable of grasping objects from humans. We then extend this work by leveraging hand-object interaction synthesis to increase the diversity of objects and human motions available for robot training. The remainder of this chapter introduces the problem settings and challenges associated with both the synthesis of hand-object interactions and the domain-specific application of human-to-robot handovers.

%% file: chapters/01_introduction/content/approach.tex
\begin{figure*}[t]
\begin{center}
   \includegraphics[width=\textwidth]{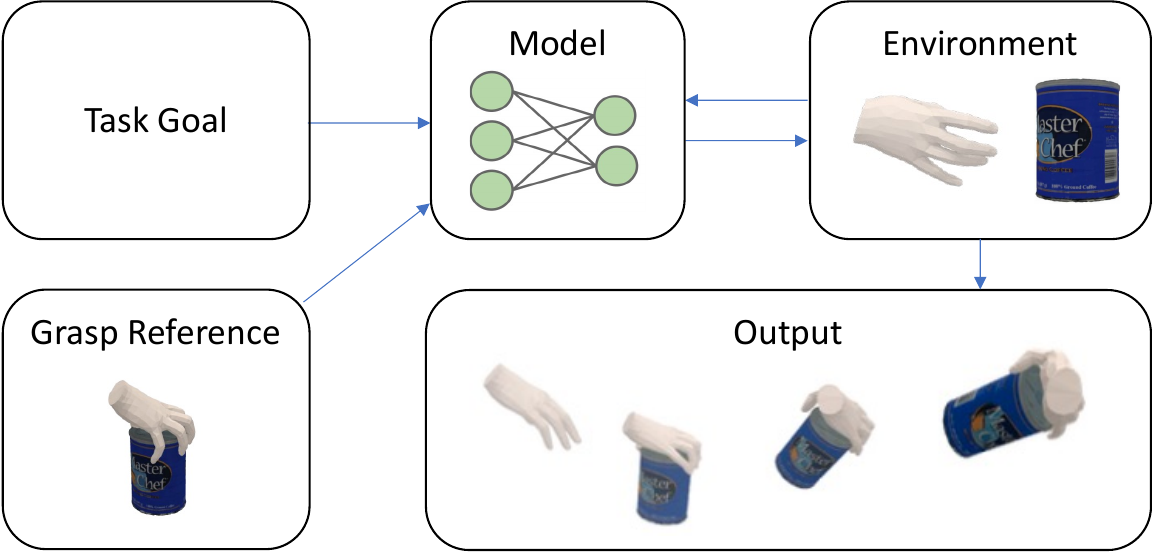}
\end{center}

   \caption{\textbf{Dynamic Grasp Synthesis Overview.} We are given a task goal, such as a target 6D object pose, and a grasp reference of a static hand-object pose, which are passed to a model. The aim of dynamic grasp synthesis is to generate a sequence of hand and object poses that fulfill the task objective. We assume that the model's predictions are passed to an environment which returns an updated state of the hand and object.}
\label{fig:intro:hoi}
\end{figure*}

\section{Problem Setting and Challenges}
\label{sec:approach}
This dissertation explores the synthesis of 4D hand-object interactions and their application to human-to-robot handovers. We introduce the problem setting and the challenges associated with hand-object interaction synthesis and human-to-robot handovers, respectively. 

\subsection{Synthesis of Hand-Object Interaction}

\paragraph{Problem Setting}
Our general framework for synthesizing hand-object interactions is illustrated in Figure \ref{fig:intro:hoi}. We start with a task goal, which could involve specifying a target location for moving the object, setting an articulation angle if the object is articulated, or providing another form of instruction, such as a text prompt. The objective is to train a model, using a neural network, that accomplishes the given task by interacting with an environment. This environment includes one or two hand models along with the object geometry. The model operates in a closed-loop manner, sequentially generating actions within the environment to produce the desired output. Additionally, we assume access to a human grasp reference, which consists of a single frame of an object-relative hand pose indicating the desired way to grasp an object, serving as guidance for the model.

\paragraph{Challenges} 
\label{sec:intro:hoi_challenges}
Previous work on synthesizing hand-object grasping has primarily focused on generating static grasps \cite{jiang2021hand, corona2020ganhand, karunratanakul2020grasping} or predicting local hand poses given a global hand and object pose sequence \cite{ye2012synthesis, zhang2021manipnet}. However, none of these approaches address the synthesis of complete sequences for both hand and object poses. Synthesizing natural hand-object interactions in 4D is challenging due to several factors. First, the motions must be temporally plausible, ensuring synchronization between hand and object movements while maintaining natural dynamics. Second, the grasp needs to be stable, with the hand contacts achieving force closure to securely hold the object. Third, a model should generalize across various object geometries. Finally, the method should extend beyond single-handed grasping of rigid objects and enable articulated objects and bi-manual manipulation. In summary, the primary challenges involved are:
\challenges{
\begin{enumerate}[label=$\textbf{{C}}$\textbf{1.\arabic*}]
    \item Generating sequences of hand and object poses that are human-like, temporally coherent, and physical plausible.
    \item Developing a robust model that generalizes across different object geometries.
    \item Adapting a model to handle complex manipulation scenarios such as bi-manual manipulation and articulated objects. 
\end{enumerate}}
Since this constitutes a novel task with no existing solutions in the literature, we introduce both the task and a specific solution for it. To address the challenges outlined in \textbf{C.1.1}, we propose leveraging reinforcement learning to learn a hand policy within a physical simulation environment. This approach allows us to generate temporally coherent and physically plausible hand-object motions by enforcing constraints on interactions between the hand and object. The object’s pose naturally results from forces applied by the hand, ensuring realistic motion dynamics. To achieve human-like grasping (\textbf{C1.1}) and generalization across different object geometries (\textbf{C1.2}), we introduce the concept of a static hand-object grasp reference, which provides guidance to our model. At inference time, this allows the policy to be guided toward unseen grasp references, provided a grasp reference is available. We show that such references can be obtained from static grasp synthesis \cite{jiang2021hand} or image-based pose estimation in our experiments. We propose a general reward function that combines adhering to this grasp reference with grasp stability metrics.
In our work D-Grasp (Chapter \ref{ch:hoi_generation:dgrasp}), we therefore hypothesize that:
\hypotheses{
\begin{enumerate}[label=\textbf{H1.\arabic*}]
    \item Reinforcement learning within a physical simulation can be used to learn a policy that generates temporally coherent and physically plausible hand-object interactions (\textbf{C1.1}). 
    \item Incorporating static hand-object grasp references into the model’s state space and reward function facilitates generalization to diverse objects (\textbf{C1.2}) and results in human-like grasping behavior (\textbf{C1.1}).
\end{enumerate}}

To go beyond single handed grasping and address \textbf{C1.3}, we propose extending our framework to handle articulated objects and bi-manual manipulation. Building on our previous framework, we suggest employing reinforcement learning within a physical simulation to tackle these challenges. However, directly applying the proposed framework to this problem is non-trivial, as learning two-handed manipulations and articulations requires careful handling. 

For articulation, hand poses often require a high level of precision, as only a specific part of the object should be grasped to achieve successful articulation. For example, opening a laptop requires grasping its thin top part, and even a slight deviation in the grasp, such as grasping the bottom part, can result in failure. In two-handed manipulation, directly training both hands in the same environment can cause one hand to unintentionally push the object away before the other hand can interact with it, leading to unstable training.  Additionally, the policy must be aware of the object's articulation to achieve the given task goal. 

To address these issues, we propose a learning curriculum that begins with single-handed grasping and object articulation of stationary objects, then progresses to training both hands jointly with non-stationary objects. We suggest that this enables a policy to learn high precision grasping and articulation first. We also introduce articulation-specific information into the observation space and add rewards that incentivize correct articulation. Unlike our previous framework D-Grasp, which used a non-learning-based control scheme for global hand pose during motion synthesis, we now propose employing a learning-based strategy to manage these more complex scenarios. In our work ArtiGrasp (Chapter \ref{ch:hoi_generation:artigrasp}), we therefore hypothesize the following:
\hypotheses{
\begin{enumerate}[label=\textbf{H1.3}]
    \item Bi-manual and articulation of objects can be achieved by adapting our model to incorporate articulation information in both the observation space and reward function, supported by a training curriculum and a learning-based control strategy for the global hand poses (\textbf{C1.3}). 
\end{enumerate}}
In Part \ref{part:hoi}, we begin by introducing the novel problem setting of dynamic grasp synthesis and present a framework to solve it (Chapter \ref{ch:hoi_generation:dgrasp}, published in \cite{christen:2022:dgrasp}). This addresses hypotheses \textbf{H1.1} and \textbf{H1.2}. Building on this, we extend the task to include bi-manual object manipulation of articulated objects and propose a solution for hypothesis \textbf{H1.3} (Chapter \ref{ch:hoi_generation:artigrasp}, published in \cite{zhang2024artigrasp}). These two works advance the field by demonstrating that dynamic and diverse hand-object interactions can be synthesized in 4D with minimal reliance on human priors, which paves the way for utilizing such models in downstream applications. One such application is human-to-robot handovers, which we explore next.

\subsection{Human-to-Robot Handover}
\label{sec:intro:handovers}
\begin{figure*}[t]
\begin{center}
   \includegraphics[width=\textwidth]{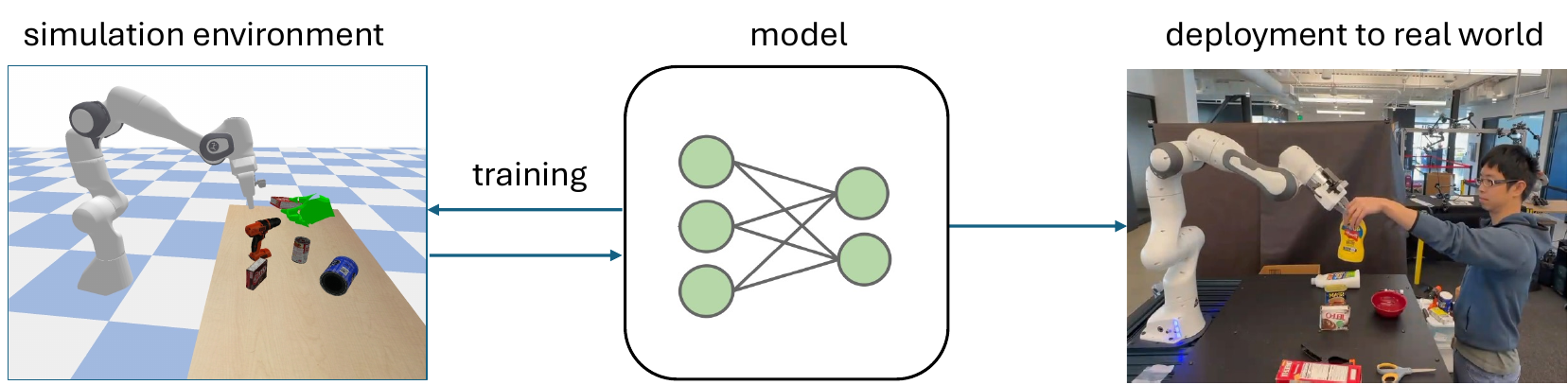}
\end{center}
   \caption{\textbf{Human-to-Robot Handover Overview.} We set up a simulation environment (left) that contains a tabletop setting with different objects, a human hand model (green), and a robotic manipulator. The human grasps an object and moves it into a handover pose, from which the robot should securely grasp it and move it to a target location. A model (middle) is trained in this simulation environment before deployment to a real robotic platform (right).}
\label{fig:intro:h2r}
\end{figure*}

\paragraph{Problem Setting} The task of human-to-robot handovers involves a human picking up an object from a table and moving it into a handover pose for the robot to grasp. The robot's goal is to accurately determine the handover pose, safely grasp the object from the human without causing collisions or dropping it, and then transport the object to a designated target location. Unlike previous work, we consider a more dynamic scenario where the human and robot move simultaneously. This requires the robot to continuously adapt to the human’s motion in real-time. The robotic platform consists of an arm equipped with a two-fingered gripper and a wrist-mounted RGB-D camera. We assume access to proprioceptive sensing, such as the joint configuration of the arm and end-effector, alongside RGB-D images. Our goal is to develop a policy that maps sensory inputs to actions for effectively controlling the robot. To facilitate efficient training, we assume access to a physical simulation environment that accurately models this task setup, as illustrated in Figure \ref{fig:intro:h2r}.

\paragraph{Challenges}
Previous systems for human-to-robot handovers make assumptions that limit their reproducability and deployment to new environments; some methods only run in open-loop fashion, requiring a human to remain stationary once the robot starts moving~\cite{rosenberger2021object}, while others need hand-designed cost functions for grasp and motion planning~\cite{yang2022model, marturi2019dynamic}. A learning-based method \cite{wang2021learning}, using state inputs, enables simultaneous motion of the human and robot in simulation. However, an open challenge remains in training policies that can transfer from simulation to the real world and adapt to human movements. Progress is primarily hindered by the difficulties of simulating humans accurately, creating a framework that adapts to human dynamics, and selecting representations that transfer effectively to real-world systems.
Therefore, the main challenges involved are:

\challenges{
\begin{enumerate}[label=\textbf{C2.\arabic*}]
    \item Building a learning framework for human-to-robot handovers in simulation that can be transferred to a real robotic platform.
    \item Training a model that generalizes well to dynamic human motions and object geometry.
\end{enumerate}
}
In our work, we propose leveraging recent advances in human simulation for handover evaluation \cite{chao2022handoversim} to learn policies with a human in the loop. While this benchmark environment has solely been used for evaluation, we aim to adapt it into a learning environment. This transformation requires developing a model with a suitable representation for real-world transfer and a training procedure that explicitly accounts for human motion.
We hypothesize that:
\hypotheses{
\begin{enumerate}[label=\textbf{H2.\arabic*}]
    \item Embedding human motions into the physical simulation and explicitly considering them in the model design and training procedure will enable learning a policy that effectively transfers to real systems with real human interactions. 
\end{enumerate}
}
One key disadvantage of using the human motions from the evaluation benchmark \cite{chao2022handoversim} for training is its limited size. Increasing the diversity of human behaviors by capturing more data of humans handing over objects is inherently limited by the cost of collecting data. Our goal is to develop a framework that further improves generalization to object geometry (\textbf{C2.2}) without having to rely on data capture. Therefore, we propose to leverage our developments in hand-object interaction synthesis to increase the diversity of objects in the human handover motions. Additionally, this will allow us to increase the diversity of test scenarios in the available benchmark \cite{chao2022handoversim}, which is currently limited to 144 handover scenarios. Our hypothesis is the following:
\hypotheses{
\begin{enumerate}[label=\textbf{H2.2}]
    \item Hand-object interaction synthesis can be used to generate more diverse human handover motions to increase the generalizability of handover policies.
\end{enumerate}}
In Part \ref{part:h2r}, we first address hypothesis \textbf{H2.1} by proposing the first end-to-end framework to learn human-to-robot handovers from point clouds (Chapter \ref{ch:handovers:handoversim2real}, published in \cite{christen:2023:handoversim2real}). Our framework is trained purely in simulation with a human in the loop and can successfully be transferred to the real system. To address hypothesis \textbf{H2.2}, we extend this framework by leveraging human handover motions generated by our work on hand-object interaction synthesis for training (Chapter \ref{ch:handovers:synh2r}, published in \cite{christen:2024:synh2r}). These two works introduce a novel framework for modeling human-to-robot handovers, providing significant improvements in the development of learning-based methods for more complex human-robot interactions. By addressing the difficulty of modeling and accounting for human behavior in simulation based training, we pave the way for adaptive, real-time interaction models that have the potential to be adapted to broader applications in human-robot collaboration in the future.

%% file: chapters/01_introduction/content/contributions.tex
\chapter{Contributions}
We summarize the four main contributions of our dissertation. 

\section*{1. \dgraspTitle}
We introduce the new task of dynamic grasp synthesis, where the goal is to generate human grasping motions that approach, grasp, and relocate an object to a target 6D pose. We propose \textbf{D-Grasp, a method to solve dynamic grasp synthesis by leveraging physical simulation and reinforcement learning.} To this end, our method is conditioned on a single frame hand grasp reference. Our policy learns to generate a dynamic sequence that follows the grasp reference and moves the object to the target pose. Our main technical contribution comprises a two-stage framework that separately models grasping and object relocation. Our general reward function incentivizes following the grasp reference and perform stable grasping, which enables both human-like and physically plausible grasping. Our experimental results justify our design choices and demonstrate that our method works with grasp references from capture systems, grasp synthesis, and image-based pose estimation.

\section*{2. \artigraspTitle}
Our previous work on dynamic grasp synthesis, D-Grasp, focuses on single-handed grasping and rigid objects. However, humans often interact with objects using both their hands. Moreover, a lot of objects are articulated. To that end, we present \textbf{ArtiGrasp, a novel method to synthesize bi-manual hand-object interactions that include grasping and articulation.} This task is challenging due to the diversity of the global wrist motions and the precise finger control that are necessary to articulate objects. ArtiGrasp leverages reinforcement learning and physics simulations to train a policy that controls the global and local hand pose. Our framework unifies grasping and articulation within a single policy guided by a single hand pose reference. Moreover, to facilitate the training of the precise finger control required for articulation, we present a learning curriculum with increasing difficulty. It starts with single-hand manipulation of stationary objects and continues with multi-agent training including both hands and non-stationary objects. To evaluate our method, we introduce Dynamic Object Grasping and Articulation, a task that involves bringing an object into a target articulated pose. This task requires grasping, relocation, and articulation. We show our method's efficacy towards this task and improvements over baselines. 

\section*{3. \handoverTitle}
Human-to-robot handovers are a critical task for human-robot interaction. 
While research in Embodied AI has made significant progress in training robot agents in simulated environments, interacting with humans remains challenging due to the difficulties of simulating humans. Fortunately, recent research has developed realistic simulated environments for handovers. In this contribution, we propose \textbf{the first framework to train control policies for vision-based human-to-robot handovers.} Leveraging this result, we introduce a method that is trained with a human-in-the-loop via a two-stage teacher-student framework that uses motion and grasp planning, reinforcement learning, and self-supervision. In our experiments, we show a significant performance gain over baselines on a simulation benchmark, sim-to-sim transfer, and sim-to-real transfer.

\section*{4. \synhrTitle}
In our previous contribution, we train robot policies by interacting with dynamic humans in simulated environments, where the policies can later be transferred to the real world. However, a major bottleneck is the reliance on human motion capture data, which is expensive to acquire and difficult to scale to arbitrary objects and human grasping motions. In this contribution, we introduce \textbf{SynH2R, a framework that can generate plausible human handover motions suitable for training human-to-robot handovers.} To achieve this, we build on our contributions in hand-object interaction synthesis and propose a method that is designed to generate handover-friendly motions similar to humans. This allows us to generate synthetic training and testing data with 100x more objects than previous work that relies on captured data. In our experiments, we show that our method trained purely with synthetic data is competitive with state-of-the-art methods that use real human motion data, both in simulation and on a real system. Additionally, we can perform evaluations on a larger scale compared to prior work. With our newly introduced synthetic test set, we show that our model can better scale to a large variety of unseen objects and human motions compared to the baselines.

%% file: chapters/01_introduction/content/structure_of_thesis.tex
\section{Structure of Dissertation}
We describe the structure of this doctoral dissertation as follows:  

\begin{tabularx}{\textwidth}{lX}
\multicolumn{2}{l}{\color{CTtitle}\textsc{\MakeLowercase{I. Introduction}}} \\
\textsc{\MakeLowercase{\chapref{ch:related_work}}} & introduces the related work on hand-object modeling and human-to-robot handovers. \\
\textsc{\MakeLowercase{\chapref{ch:background}}} & establishes the theoretical background on reinforcement learning and presents the algorithms used throughout this thesis.  \\
\multicolumn{2}{l}{\color{CTtitle}\textsc{\MakeLowercase{II. Synthesis of Hand-Object Interaction}}} \\ 
\textsc{\MakeLowercase{\chapref{ch:hoi_generation:dgrasp}}} & \textbf{Contribution 1: }\emph{\dgraspTitle.}\\
\textsc{\MakeLowercase{\chapref{ch:hoi_generation:artigrasp}}}& \textbf{Contribution 2: }\emph{\artigraspTitle.}\\
\textsc{\MakeLowercase{\chapref{ch:hoi:conclusion}}} & summarizes, discusses the implications, and highlights the limitations of our work on hand-object interaction. \\
\multicolumn{2}{l}{\color{CTtitle}\textsc{\MakeLowercase{III. Human-to-Robot Handover}}} \\ 

\textsc{\MakeLowercase{\chapref{ch:handovers:handoversim2real}}} & \textbf{Contribution 3: }\emph{\handoverTitle.}\\
\textsc{\MakeLowercase{\chapref{ch:handovers:synh2r}}}& \textbf{Contribution 4: }\emph{\synhrTitle.}\\
\textsc{\MakeLowercase{\chapref{ch:handover:conclusion}}} & summarizes, discusses the implications, and highlights the limitations of our work on human-to-robot handovers. \\
\multicolumn{2}{l}{\color{CTtitle}\textsc{\MakeLowercase{IV. Conclusion}}} \\ 
\textsc{\MakeLowercase{\chapref{ch:summary}}}& summarizes the contributions of this dissertation.  \\
\textsc{\MakeLowercase{\chapref{ch:outlook}}} & discusses future research directions.
\end{tabularx}

%% file: frontbackmatter/publications.tex
\pdfbookmark[1]{Publications}{publications}
\section{Publications}

\noindent
The contributions of this thesis are based on the following publications, in order of appearance:

\begin{enumerate}
    \item \fullcite{christen:2022:dgrasp:bold}
    \item \fullcite{zhang:2024:artigrasp:bold}
    \item \fullcite{christen:2023:handoversim2real:bold}
    \item \fullcite{christen:2024:synh2r:bold}   
\end{enumerate}

\noindent
Further publications that were conducted during the course of my PhD research but are out of scope of this thesis are listed below, in order of publication:
\begin{enumerate}
    \item \fullcite{christen:2019:drlhs:bold}
    \item \fullcite{stevsic:2020:learning:bold}
    \item \fullcite{christen:2021:learning:bold}
    \item \fullcite{vulin:2021:intrinsic:bold}
    \item \fullcite{langerak:2021:generalizing:bold}
    \item \fullcite{block:2022:huggiebot:bold}
    \item \fullcite{bagatella:2022:sfp:bold}
    \item \fullcite{ziani:2022:tempclr:bold}
    \item \fullcite{block:2022:huggiebotdemo:bold}
    \item \fullcite{christen:2022:controllable:bold} \item \fullcite{christen:2023:varyingbodyshape:bold}
    \item \fullcite{braun:2023:physically:bold}
    \item \fullcite{langerak:2024:marlui:bold}
    \item \fullcite{zhang:2024:graspxl:bold}
    \item \fullcite{christen:2024:diffh2o:bold}
    \item \fullcite{yin:2024:egohdm:bold}
    \item \fullcite{luo:2024:grasping:bold}
    \item \fullcite{li:2024:latenthoi:bold}
    \item \fullcite{albaba:2024:rile:bold}
\end{enumerate}

\textbf{*} indicates equal contribution. \\ 

\noindent
The technical contributions of this thesis have lead to one
patent application:
\begin{enumerate}
    \item \fullcite{christen:2024:patent:bold}
\end{enumerate}

%% file: chapters/02_related_work/related_work.tex
\def\dir{chapters/02_related_work}

\chapter{State of the Art}
\label{ch:related_work}

In this chapter, we will cover the state-of-the-art in human-object modeling and human-to-robot handovers. We start by describing the recent efforts in collecting datasets of human-object interactions, which have lead to the development of methods for both hand-object reconstruction and the synthesis of hand-object interaction. Since our focus is on the latter, we will only briefly cover hand-object reconstruction methods in this chapter and refer the reader to \cite{fan2024benchmarks} for a recent overview. We then explore the state-of-the-art in synthesizing grasps for object meshes. Thereafter, we move towards a dynamic setting of generating hand-object interactions for both disembodied hands and the whole body. In the second section of this chapter, we review the progress in human-to-robot handovers and describe the state-of-the-art in policy learning for grasping.  

\section{Human-Object Interaction}

\input{\dir/content/datasets}

\input{\dir/content/static_grasp}
\input{\dir/content/hand_object_motion}
\input{\dir/content/whole_body_motion}

\section{Human-Robot Handovers}

\input{\dir/content/human_robot_handovers}
\input{\dir/content/policy_grasping}

\clearpage

%% file: chapters/02_related_work/tables/datasets.tex
\begin{table*}[t]
\centering
\resizebox{1.00\linewidth}{!}{
\begin{tabular}{lcccccccc|cccc|c}

\toprule
\multicolumn{1}{c}{dataset} & real/ & seq. & real & \multicolumn{4}{c}{\# number of:} & \multicolumn{1}{c|}{ego-} & articulated & both & human &  \multicolumn{1}{c|}{dexterous} & \multicolumn{1}{c}{annotation} \\ \cline{5-8}
 & syn. & data & images & \multicolumn{1}{c}{seq.} & \multicolumn{1}{c}{img} & \multicolumn{1}{c}{view} & \multicolumn{1}{c}{objects} & \multicolumn{1}{c|}{centric} & \multicolumn{1}{c}{objects} & \multicolumn{1}{c}{hands} & \multicolumn{1}{c}{body} & \multicolumn{1}{c|}{manipulation} & \multicolumn{1}{c}{type} \\ \hline

ObMan~\cite{hasson2019learning} & syn. & \xmark & \xmark & - & 154k & 1 & 2772 & \xmark & \xmark & \xmark & \xmark & \xmark & simulate \\
\rowcolor{Gray}
ContactPose~\cite{brahmbhatt2020contactpose} & real & \xmark & \cmark & - & 2.9M & 3 & 25 & \xmark & \xmark & \xmark & \xmark & \xmark & multi-kinect \\
AffordPose~\cite{jian2023affordpose} & syn. & \xmark & \xmark & - & 26.7k & 3 & 641 & \xmark & \xmark & \xmark & \xmark & \xmark & multi-manual \\ 
\rowcolor{Gray}
ContactArt~\cite{zhu2024contactart} & real & \cmark & \xmark & - & 332k & - & 80 & \xmark & \cmark & \xmark & \xmark & \cmark & transfer \\ 
HO3D~\cite{hampali2020honnotate} & real & \cmark & \cmark & 68 & 78k & 1-5 & 10 & \xmark & \xmark & \xmark & \xmark & \xmark & multi-kinect \\
\rowcolor{Gray}
H2O-3D~\cite{hampali2022keypoint} & real & \cmark & \cmark & 17 & 76k & 5 & 10 & \xmark & \xmark & \cmark & \xmark & \xmark & multi-kinect \\
H2O~\cite{kwon2021h2o} & real & \cmark & \cmark & 192 & 571k & 5 & 8 & \cmark & \xmark & \cmark & \xmark & \xmark & multi-kinect \\ 
\rowcolor{Gray}
ARCTIC~\cite{fan2023arctic} & real & \cmark & \cmark & 339 & 2.1M & 9 & 11 & \cmark & \cmark & \cmark & \cmark & \cmark & mocap \\ 
OakInk~\cite{yang2022OakInk} & real & \cmark & \cmark & 792 & 230k & 4 & 100 & \xmark & \xmark & \xmark & \xmark & \cmark & multi-manual \\
\rowcolor{Gray}
OakInk2~\cite{zhan2024oakink2} & real & \cmark & \cmark & 627 & 4.01M & 4 & 75 & \cmark & \cmark & \cmark & \xmark & \cmark & mocap \\

DexYCB~\cite{chao2021dexycb} & real & \cmark & \cmark & 1000 & 582k & 8 & 20 & \cmark & \xmark & \xmark & \xmark & \xmark & multi-manual \\
\rowcolor{Gray}
Core4D~\cite{zhang2024core4d} & real & \cmark & \cmark & 1000 & - & 5 & 37 & \cmark & \xmark & \cmark & \cmark & \xmark & mocap \\ 
FHPA~\cite{garciahernando2018first} & real & \cmark & \cmark & 1200 & 105k & 1 & 4 & \cmark & \xmark & \xmark & \xmark & \xmark & magnetic \\
\rowcolor{Gray}
GRAB~\cite{taheri2020grab} & real & \cmark & \xmark & 1300 & - & - & 51 & - & \xmark & \cmark & \cmark & \xmark & mocap \\
TACO~\cite{liu2024taco} & real & \cmark & \cmark & 2500 & 5.2M & 13 & 196 & \cmark & \xmark & \cmark & \xmark & \cmark & auto \\ 
\rowcolor{Gray}
HOI4D~\cite{liu2022hoi4d} & real & \cmark & \cmark & 4000 & 2.4M & 1 & 800 & \cmark & \cmark & \xmark & \xmark & \xmark & single-manual \\ 
\bottomrule
\end{tabular}
}
\caption{\textbf{Dataset Overview.} Comparison of Datasets for Human-Object Interaction (adapted and extended from \cite{fan2023arctic}). We focus on datasets that include fine-grained object grasping and annotations for the hand pose. For a detailed overview of datasets that comprise interactions with larger objects, such as chairs or tables, we refer to \cite{zhang2024core4d} for an overview.} 

\label{tab:compare_datasets}
\end{table*}

%% file: chapters/02_related_work/content/static_grasp.tex
\subsection{Static Grasp Synthesis}
Leveraging the datasets presented in \Tab{compare_datasets}, a large number of methods attempt to estimate grasp parameters, such as the hand and object pose, directly from RGB images \cite{karunratanakul2020grasping, yang2021cpf, doosti2020hope, tekin2019h+, hasson2019learning, hasson2020leveraging, liu2021semi, cao2021reconstructing}. Some predict the mesh of the hand and the object directly \cite{hasson2019learning}, or assume a known object and predict its 6 degrees of freedom (DoF) pose in addition to the hand pose \cite{hasson2020leveraging, cao2021reconstructing, yang2021cpf, liu2021semi}. Others predict 3D keypoints and 6 DoF pose of the object \cite{doosti2020hope, tekin2019h+} or produce an implicit surface representation of the grasping hands \cite{karunratanakul2020grasping}. To improve the prediction accuracy of the grasp, many of these works incorporate additional contact losses \cite{karunratanakul2020grasping, hasson2019learning} or propose a contact-aware refinement step \cite{yang2021cpf, cao2021reconstructing}. 
More directly related to this thesis are methods that attempt to generate static grasps given an object mesh and sometimes also  information about the hand \cite{brahmbhatt2019contactdb, brahmbhatt2020contactpose, karunratanakul2021skeleton, zhu2021human, jiang2021hand, karunratanakul2020grasping, taheri2020grab}. Generally, these approaches either predict a contact map on the object \cite{brahmbhatt2019contactdb, brahmbhatt2020contactpose, jiang2021hand} or synthesize the joint-angle configuration of the grasping hand \cite{taheri2020grab, karunratanakul2020grasping, karunratanakul2021skeleton, zhu2021human}. \cite{jiang2021hand} propose a hybrid method, where predicted contact maps on objects are used to refine an initial grasp prediction. As for RGB-based methods, grasps can be represented as a mesh \cite{taheri2020grab, zhu2021human} or using an implicit surface representation \cite{karunratanakul2020grasping}. In GANHand \cite{corona2020ganhand}, a generative adversarial network \cite{goodfellow2014generative} is used to predict grasps from RGB images of objects.
ContactOpt has combined hand pose estimation with static grasp generation by leveraging contact information to post-process noisy hand pose predictions from RGB images \cite{grady2021contactopt}. 

Given the recent success of diffusion models, AffordanceDiffusion \cite{ye2023affordance} introduces a diffusion-guided framework that reconstructs hand poses given an object image. In a follow-up work, \cite{ye2023vhoi} use diffusion models to reconstruct hand-object pose sequences from videos. In G-HOP \cite{ye2023ghop}, the setting is extended to not only generate the hand grasp on an object, but also generate an object mesh along with it. While most works on static grasp synthesis focus only on the hand, FLEX \cite{tendulkar2023flex} trains a hand and body pose prior and later optimizes the priors to achieve diverse, static full-body grasps. In summary, all of these works focus on generating static grasps and are purely data-driven. In our work, however, we take into consideration the dynamic nature of human-object interaction and consider the physical plausibility of dynamic grasp-based hand-object interactions by leveraging a physics-driven simulation. 

%% file: chapters/02_related_work/content/hand_object_motion.tex
\subsection{Hand-Object Motion Generation}
We now turn to the generation of full motions of hand-object interaction instead of static grasps. Here we discuss methods that consider the hand in isolation from full-body, which has wide applications in VR and robotics. Due to its similarity to our work in modeling the human hand, we also include approaches that focus on a dexterous robotic hand in simulation. One prominent solution is to turn to physical simulation and RL. Some works use a stationary hand to learn fine-grained manipulation skills, such as reorientation of objects \cite{openai2018learning, chen2021simple}. Another approach is to  learn dexterous manipulation tasks in simulation from full human demonstration data either collected via teleoperation~\cite{rajeswaran2018learning} or from videos~\cite{garciahernando2020physics, qin2021dexmv}. \citet{rajeswaran2018learning} propose a teleoperation system that maps from the human hand to a robotic hand in simulation to collect expert demonstrations for tasks such as opening a door or hammering in a nail. These expert demonstrations are stored in a replay buffer and mixed with experiences collected with a reinforcement learning agent. The goal is to leverage the demonstrations to learn the complex manipulation tasks more robustly, even under changing initial conditions. In \citet{garciahernando2020physics}, hand poses are estimated from videos and corrected via residual reinforcement learning to achieve manipulation tasks. Improving upon this, DexMV \cite{qin2021dexmv} learns a robotic hand policy is trained via imitation learning with hand and object poses estimated from human demonstration videos. \citet{mandikal2021learning} propose a reward function that incentivizes dexterous robotic hand policies to grasp in the affordance region of objects. However, since the policy is only incentivized to grasp in a certain region, the fingers often end up in unnatural configurations. In their follow-up work \cite{mandikal2021dexvip}, the authors address this issue by formulating a reward based on a prior extracted from hand-object interaction videos. In these two works, however, a single policy is trained per object.

\citet{li2024task} propose a solution for grasping multiple objects at once. \citet{she2022learning} propose a new dynamic state representation to generate dexterous grasps, whereas other works leverage implicit representation \cite{karunratanakul2020grasping} or force-closure \cite{liu2021synthesizing}. The works presented in this thesis, D-Grasp (Chapter \ref{ch:hoi_generation:dgrasp}) and ArtiGrasp (Chapter \ref{ch:hoi_generation:artigrasp}), leverage RL and physics simulation to generate diverse hand-object interactions from sparse reference inputs. In the meantime, there have been new works that build on our concept. For example, Unidexgrasp~\cite{xu2023unidexgrasp} establishes robotic grasping on unseen object geometry using a VAE-based network to generate the reference inputs, which are passed to an RL-based policy. Their follow-up work Unidexgrasp++\cite{wan2023unidexgrasp++} proposes a solution that transfers from priviliged information to vision-based inputs, which may be used for sim-to-real transfer in robotic grasping. 
 
To model hand-object interactions, another solution is to use purely data-driven frameworks. For example, \cite{zhou2022toch, xueyiliu2024geneoh, taheri2024grip} propose methods that enable denoising hand-poses from noisy sequences of hand-object poses. \citet{ye2012synthesis} predict the local hand pose given full body and object motion. Similarly, ManipNet~\cite{zhang2021manipnet} predicts local hand poses for two-handed interactions based on 
wrist-object trajectories. Given object motion of articulated objects, CAMS~\cite{zheng2023cams} predicts one-handed poses that align with the object motion, whereas \cite{chen2022pregrasp} focus on grasping objects using dexterity in the environment. However, data-driven models are prone to suffer from physical artifacts that often require correction in post-processing. In our work, we leverage a physics-based approach to explicitly address this problem during synthesis.

%% file: chapters/02_related_work/content/whole_body_motion.tex
\subsection{Whole-Body Hand-Object Motion Generation}
Some studies focus on generating coarse full-body object interactions \cite{zhang2022couch,wang2021scene,lee2023locomotion, hassan2023synthesizing, luo2021dynamics, merel2020catch, xu2023interdiff}, such as carrying boxes. InterDiff~\cite{xu2023interdiff} generates human-object interactions with large objects via diffusion models. Recent works \cite{diller2023cghoi, li2023controllable, peng2023hoi} improve upon interaction quality by leveraging contact-based predictions in combination with inference time-guidance. Alternatively, \citet{hassan2023synthesizing} use physical simulation and adversarial imitation learning from large motion capture databases~\cite{peng2022ase} to learn interactions with large objects such as boxes and sofas. Unlike our work, these approaches focus primarily on the plausibility of the full-body motion and less on the fine-grained hand poses necessary for grasping everyday objects.  

Focusing on interactions with smaller objects, FLEX \cite{tendulkar2023flex} trains a hand and body pose prior and later optimizes the priors to achieve diverse, static full-body grasps. GOAL \cite{wang2021goal} and SAGA \cite{wu2022saga}  use CVAEs to generate approaching motions for full-body grasps, whereas TOHO \cite{li2024task} models both approaching and manipulation tasks via neural implicit representations. IMoS \cite{ghosh2023imos}, a two-stage method to generate hand-object interactions on seen objects based on action commands. Starting from a grasping state, they first generate body motions and then optimize for object trajectories using a heuristics-based optimization. However, these methods are data-driven and suffer from physical artifacts such as interpenetration. In contrast, our work generates physically plausible grasping by employing physical simulation, but neglects the full-body. Two follow-up works to our dissertation leverage physical simulation and reinforcement learning with a full-body model \cite{braun:2023:physically, luo:2024:grasping}. \citet{braun:2023:physically} first learn priors for the hand and body separately using adversarial imitation learning \cite{peng2022ase} from captured hand-object datasets \cite{taheri2020grab}, and later fine-tune a high-level policy for full-body grasping that samples from these priors. Similarly, OmniGrasp \cite{luo:2024:grasping} proposes a more general framework for full-body grasping that generalizes to over 1200 objects by training a prior on a large motion capture database \cite{mahmood2019amass}.

%% file: chapters/02_related_work/content/human_robot_handovers.tex
\subsection{Human-to-Robot Handovers}

Encouraging progress in hand and object pose estimation~\cite{hasson2019learning,liu2021semi,li2022artiboost} has been achieved, aided by the introduction of large hand-object interaction datasets~\cite{garciahernando2018first,hampali2020honnotate,brahmbhatt2020contactpose,moon2020interhand2.6m,taheri2020grab,chao2021dexycb,ye2021h2o,kwon2021h2o,liu2022hoi4d, fan2023arctic}. These developments enable applying model-based grasp planning~\cite{bicchi2000robotic,miller2004graspit,bohg2013data}, a well-studied approach in which full pose estimation and tracking are needed, to H2R handovers~\cite{sanchezmatilla2020benchmark,chao2021dexycb}. However, these methods require the 3D shape models of the object and cannot handle unseen objects.
Alternatively, some recent works~\cite{rosenberger2021object,yang2021reactive,yang2022model,duan2022learning, marturi2019dynamic} achieve H2R handover by employing learning-based grasp planners to generate grasps for novel objects from raw vision inputs such as images or point clouds~\cite{morrison2018closing,mousavian20196}. 
While promising results have been shown, these methods work only on an open-loop sequential setting in which the human hand has to remain still once the robot starts to move~\cite{rosenberger2021object}, or need complex hand-designed cost functions for grasp selection~\cite{yang2021reactive} and robot motion planning~\cite{yang2022model, marturi2019dynamic} for reactive handovers, which requires expertise in robot motion and control. 
Hence, these methods are difficult to reproduce and deploy to new environments.
Progress towards dynamic simultaneous motion has been shown by a learning-based method \cite{wang2021learning}, using state inputs and performing experiments only in simulation. Therefore, an open challenge remains for training policies that receive visual input directly.
In our work, we aim to learn control policies together with grasp prediction for handovers in an end-to-end manner from segmented point clouds with a deep neural network. 
To facilitate easy and fair comparisons among different handover methods, \cite{chao2022handoversim} propose a physics-simulated environment with diverse objects and realistic human handover behavior collected by a capture system~\cite{chao2021dexycb}. They provide benchmark results of several previous handover systems, including a learning-based grasping policy trained with static objects~\cite{wang2021goal}, but solely focus on evaluation.
Learning a safe and efficient handover policy in such an environment remains an open challenge, especially considering the dynamic human in the loop. In this thesis, we propose the first framework that addresses this challenge by integrating simulated human motions into the training environment.

%% file: chapters/02_related_work/content/policy_grasping.tex
\subsection{Policy Learning for Grasping}
Object grasping is an essential skill for many robot tasks, including handovers. 
Prior works usually generate grasp poses given a known 3D object geometry such as object shape or pose~\cite{bicchi2000robotic,miller2004graspit,bohg2013data, eppner2020acronym}. Using object pose information, optimal motion and grasp planning \cite{wang2020manipulation} can be used to plan robot motions for grasping. However, information such as pose or shape is nontrivial to obtain from real-world sensory input such as images or point clouds. 
To overcome this, recent works train deep neural networks to predict grasps directly from sensor data~\cite{kleeberger2020survey} and compute trajectories to reach the predicted grasp pose. 
Though 3D object geometry is no longer needed, the feasibility is not guaranteed since the grasp prediction and trajectory planning are computed separately. 
Some recent works directly learn grasping policies given raw sensor data. 
\cite{kalashnikov2018qt} propose a self-supervised RL framework based on RGB images to learn a deep Q-function from real-world grasps. 
To improve data efficiency, \cite{song2020grasping} use a low-cost handheld device to collect grasping demonstrations with a wrist-mounted camera. They train an RL-based 6-DoF closed-loop grasping policy with these demonstrations.
\cite{wang2021goal} combines imitation learning from expert data with RL to learn a control policy for object grasping from point clouds. 
Although this method performs well in HandoverSim \cite{chao2022handoversim} when the human hand is not moving, it has difficulty coordinating with a dynamic human hand since the policy is learned with static objects. 
Instead, our policy is directly learned from large-scale dynamic hand-object trajectories obtained from the real world.
To facilitate the training for the dynamic case, we propose a two-stage teacher-student framework, that is conceptually inspired by \cite{chen2021simple}, which has been proven critical through experiments.

%% file: chapters/03_background/background.tex
\def\dir{chapters/03_background}

\chapter{Background}
\label{ch:background}

In this section, we first explain and formalize reinforcement learning. We then describe actor-critic methods and introduce the two algorithms utilized throughout this thesis, namely Proximal Policy Optimization (PPO) \cite{schulman2017proximal} and TD3 \cite{fujimoto2018addressing}.

\input{chapters/03_background/content/rl}

%% file: chapters/03_background/content/rl.tex
\section{Reinforcement Learning}

\subsection{Intuition} 
\begin{figure*}[h!]
\begin{center}
   \includegraphics[width=1.0\textwidth]{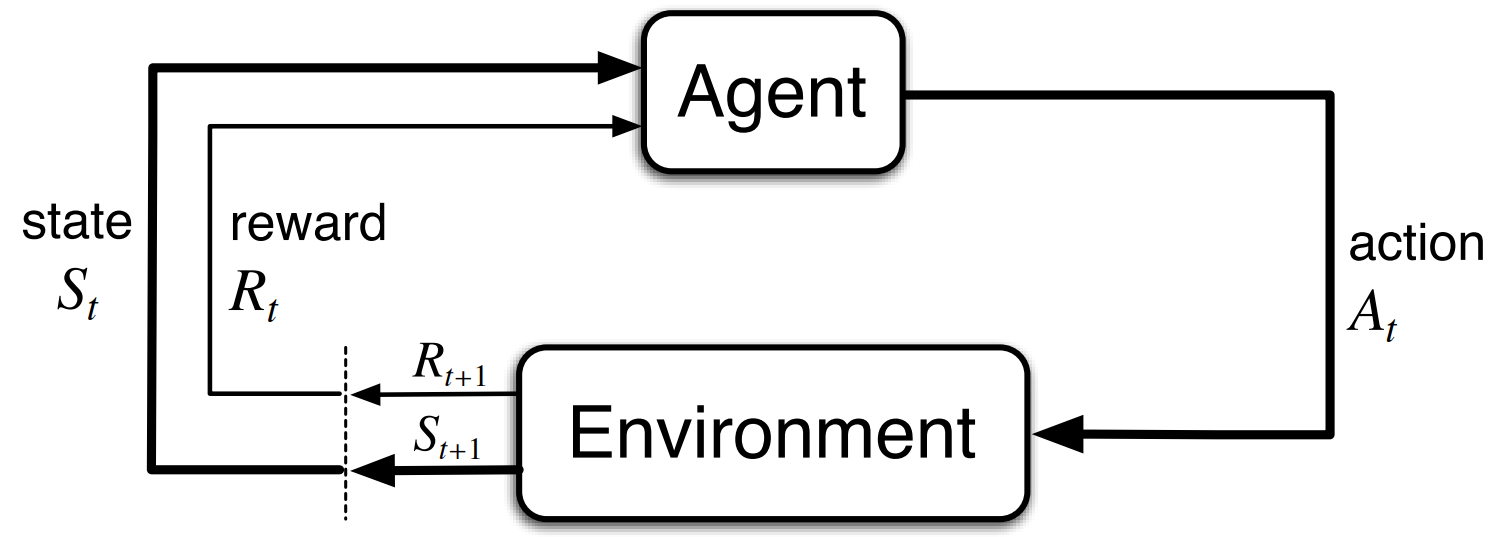}
\end{center}
   \vspace{-4mm}
   \caption{The interaction between agent and environment in RL \cite{sutton1998introduction}.}

\label{fig:rl_loop}
\end{figure*}

Reinforcement learning is a paradigm in machine learning that models the problem of sequential decision making. As shown in \figref{fig:rl_loop}, the problem involves an agent interacting with an unknown environment. The agent can take actions, which are applied to the environment and cause a change in the state. The underlying dynamics of the system, i.e., which state the agent transitions into given an action, are unknown. The agent gets an update of the state and decides for the next action to take. The goal of the agent is to maximize a numerical reward signal over the long-term. This reward signal can either be sparse, for example, only be non-zero upon finishing the task successfully, or dense, where some indicative reward is provided at every step. For example, let us assume a setting with a robot that has to find the exit in a maze without prior knowledge of the structure. In a sparse reward setting, the agent would be rewarded only when finding the exit, whereas with a dense reward, the agent could be rewarded higher the closer it gets to the exit. Over the course of training, the agent should learn which actions get it closer to finishing the maze until finally finding an optimal policy to solve the maze.

One of the main challenges of reinforcement learning is trading off exploiting actions that are known to be good with exploring new actions. In classical reinforcement learning, where an environment with discrete states and actions (tabular RL) is assumed, there are many algorithms with guaranteed proofs of convergence \cite{sutton1998introduction}. However, in this thesis, we do not study such scenarios, and focus on utilizing RL and its set of algorithms as tools to formulate the problems of hand-object interaction and human-to-robot handovers. In this section, we will therefore focus on first formalizing the problem, and then explain the Deep Reinforcement Learning (DRL) algorithms that will be relevant in the presented chapters.

\subsection{Markov Decision Processes} 
We follow the standard formulation of a Markov Decision Process (MDP), which is a mathematical framework to model decision-making problems. The MDP is defined as a tuple $\mdp = \{\StatesSet, \ActionsSet, , \mathcal{R},  \discountRate, \TransitionSet, \rho_0\}$, where $\StatesSet$ and $\ActionsSet$ are state and action spaces, respectively. $\mathcal{R} : \StatesSet \times \ActionsSet \to \mathbb{R}$ is the reward function discounted by factor $\gamma\in [0,1]$, $\TransitionSet : \StatesSet \times \ActionsSet \to \StatesSet$ the transition function of the environment and $\rho_0 = p(\state_0)$ the initial state distribution. An MDP is Markovian, meaning that transitions from a state into a new state solely depend on the current state and action irrespective of the history of states and actions. A policy $\policy$ is a mapping from state to action $\policy : \StatesSet \times \ActionsSet \to \mathbb{R}$ and determines which action to take in a given state. The expected discounted reward when following policy $\policy$ starting from state $\state \in \StatesSet$ is defined as value function $V$: 

\begin{equation}
V^{\policy}(\state) = \mathbb{E}_{\policy}\left[\sum_{k=0}^\infty \gamma^k R_{t+k+1} \mid S_t = \state \right].
\end{equation} 

\noindent Similarly, the Q-value function is defined as the expected discounted reward when taking actions $\actions \in \ActionsSet$ in state $\state \in \StatesSet$:
\begin{equation}
\label{eq:q_value}
    Q^{\policy}(\state,\actions) = \mathbb{E}_{\policy}\left[\sum_{k=0}^{\infty}\gamma^k R_{t+k+1}\mid S_t = \state, A_t = \actions \right].
\end{equation}

\noindent The value function $V$ and Q-value function $Q$ are related via:
\begin{equation}
V^{\policy}(\state) = \sum_\actions \policy(\actions|\state) Q^{\policy}(\state,\actions).
\end{equation}

\noindent To estimate the expected long-term reward, the Bellman equation defines the relationship between states and actions in a recursive manner:
\begin{equation}
\label{eq:bellman}
    Q^{\policy}(\state, \actions) = \sum_{\state'} T(\state'|\state,\actions)[R(\state',\state,\actions) + \gamma Q^{\policy}(\state',\policy(\state'))].
\end{equation}
It states that the expected reward given an action $\actions$ in state $\state$ is the immediate reward plus the discounted Q-value estimate from the subsequent state $\state'$. This is weighted by the transition probabilities of reaching state $\state'$ given the state-action pair $(\state,\actions)$. To solve the Bellman equation, an optimal policy can be computed as $\policy^* = \arg\max_{\actions} Q^{\policy}(\state,\actions)$. 

To maintain consistency with the notation used throughout the chapters in this thesis, we adopt bold lowercase notation for states and actions in the subsequent sections. While traditional MDP formulations often distinguish between random variables (uppercase) and their values (lowercase), this thesis uses  bold lowercase symbols to represent  both the general state and action variables as well as their specific realizations. For the reward function, we also adopt lowercase notation. This notation aligns with common conventions in modern reinforcement learning literature. To further simplify the notation and increase readability, we omit the policy $\policy$ term in the superscript of the value function $V$ and Q-value function $Q$ in the remainder of this chapter. 

\subsection{Reinforcement Learning} 
\label{sec:background_rl}
In scenarios where the transition function $\TransitionSet$ (the underlying dynamics of a system) is known, the Bellman equation (Eq. \ref{eq:bellman}) can be solved using dynamic programming approaches \cite{bertsekas2012dynamic}. Alternatively, in settings with large state spaces, where dynamic programming become computationally intractable, Monte Carlo Sampling techniques are often employed \cite{sutton1998introduction}. However, in many real world settings, the transition function is unknown, requiring methods that can operate under such uncertainty. Reinforcement learning addresses this problem by optimizing decision-making in environments with unknown dynamics. Two broad approaches to this problem are known: model-based and model-free reinforcement learning.  

\paragraph{Model-based vs. model-free RL} Model-based reinforcement learning learns the environment's dynamics (transition and reward functions) and then uses these learned models for planning optimal policies, often through techniques like dynamic programming or model predictive control (MPC) \cite{morari1999model}. This approach improves sample efficiency for policy optimization. However, learning an accurate model is challenging, and errors in the learned model can result in poor policy performance. Model-free reinforcement learning directly learns the optimal policy through interactions with the environment, without requiring knowledge of the environment's dynamics. These methods are advantageous in real-world applications, such as robotic control, where accurately modeling complex dynamics is difficult. However, model-free RL is less sample efficient than model-based methods, typically requiring more interactions with the environment and taking longer to converge. In this dissertation, we employ model-free algorithms, and refer the reader to \cite{moerland2023model} for an overview of model-based algorithms. We will first discuss strategies for model-free policies to balance exploring new actions with exploiting known actions that lead to high rewards.

\paragraph{Exploration vs. Exploitation} The exploration versus exploitation dilemma is one of the most fundamental challenges in reinforcement learning. Exploration refers to the agent trying a wide range of actions in order to discover an optimal policy. Exploitation, on the other hand, involves selecting actions that are known to yield high rewards based on previous experience. The dilemma arises from the need to balance these two behaviors to maximize long-term rewards. A simple exploration strategy is to randomly sample actions, but this often results in the agent frequently visiting states near the initial states, while distant states may be neglected or never reached.At the other extreme, always selecting actions with the highest Q-values, known as a greedy strategy, can quickly lead to high rewards but risks getting stuck in local optima. This may prevent the exploration of better policies. A simple, yet effective strategy is the $\varepsilon$-greedy strategy, where the agent chooses a random action with a small probability $\varepsilon$ and selectcs the greedy action (the highest Q-value) otherwise. Over time, as the policy improves, the probability $\varepsilon$ may also be gradually decreased. Despite its simplicity, this effective technique is often used in practice. There are, of course, many more sophisticated algorithms to deal with the exploration vs. exploitation dilemma. There are, of course, more advanced approaches to addressing the exploration-exploitation dilemma. These include methods that estimate the uncertainty of state values and incentivize visiting states with high uncertainty through intrinsic rewards \cite{stadie2016incentivizing}, maximizing information gain \cite{still2012information}, or adding noise to actions in continuous action spaces \cite{lillicrap2016ddpg}. However, since this dilemma is not the primary focus of this dissertation, we refer the reader to \cite{ladosz2022exploration} for a comprehensive overview of exploration-exploitation solutions.

\paragraph{Temporal Difference Learning} 
\label{sec:bg:td}
In model-free reinforcement learning, a policy is optimized through trial and error from interactions with the environment. A common strategy is to approximate the Q-value function, $Q(\state, \actions)$, which represents the expected reward for taking action $\actions$ in state $\state$. Unlike dynamic programming algorithms, which rely on a model of the environment dynamics and update the values of the entire state space in each iteration, temporal difference (TD) learning updates value estimates based on individual state-action interactions and does not assume knowledge of the transition dynamics. To perform updates on single states, we need to establish an update rule that improves upon the previous estimate. The TD update rule provides a way to improve the current estimate of the value function:
\begin{equation}
\begin{split}
    & \Delta V(\state) = r(\state,\actions) + \gamma V(\state') - V(\state) \\
    & V(\state) \leftarrow V(\state) + \alpha \Delta V(\state).
\end{split}
\end{equation}
This is known as the TD(0) update because it computes the difference between the current estimate $V(\state)$ and the sum of the immediate reward $R(\state, \actions)$ for one step and the discounted value of the next state $\gamma V(\state')$. This error term is then added to the previous estimate, weighted by a learning rate $\alpha$. The TD(0) update can be generalized to N-step updates by considering rewards from N consecutive steps before bootstrapping from the value of the state reached at step N. This update rule can also be applied to Q-values:
\begin{equation}
\label{eq:td_update}
\begin{split}
    & \Delta Q(\state, \actions) = r(\state,\actions) + \gamma Q(\state', \actions') - Q(\state, \actions) \\
    & Q(\state, \actions) \leftarrow Q(\state) + \alpha \Delta Q(\state).
\end{split}
\end{equation}
The next step is to define algorithms that interact with the environment and leverage these update rules to improve a policy. These algorithms can be categorized into two types: on-policy and off-policy, which we will discuss next.

\paragraph{On-Policy vs. Off-Policy} 
\label{sec:bg:onoffpolicy}
The key difference between on-policy and off-policy methods lies in how the subsequent action $\actions'$ is sampled during Q-value updates, as shown in Eq. \ref{eq:td_update}.  In on-policy methods, the same policy is used both to gather samples from the environment and to select the next action during learning. An example of an on-policy algorithm is SARSA, which stands for State $\rightarrow$ Action $\rightarrow$ Reward $\rightarrow$ State' $\rightarrow$ Action'. In SARSA, the policy used to interact with the environment is also used to select the next action $\actions'$. A common strategy is to utilize an $\varepsilon$-greedy strategy for the policy, which samples samples a random with a low probability and otherwise samples the greedy action. After convergence, the optimal policy is derived from the state-action pairs with the highest Q-values.
In off-policy methods, two separate policies are used: one for interacting with the environment (the behavior policy) and another for updating the Q-values (the target policy). The behavior policy, used for interaction, can employ strategies like random exploration or $\varepsilon$-greedy. The target policy follows a greedy approach for updating the Q-values, leading to the following update rule:
\begin{equation}
\label{eq:q_learning}
    \Delta Q(\state, \actions) = r(\state,\actions) + \gamma \max_{\actions'} Q(\state', \actions') - Q(\state, \actions).
\end{equation}

\noindent This approach is known as Q-learning. Similar to SARSA, the optimal policy in Q-learning is determined through the actions with the highest Q-values after convergence.

\section{Deep Reinforcement Learning}
So far, we have assumed that the environment an agent interacts with consists of a discrete set of state and actions. In many scenarios, however, the state space can be extremely large or even continuous. Applying classical reinforcement learning methods (see \secref{sec:background_rl}) to such environments introduces significant challenges.These methods learn about each state separately, which means there is no generalization across similar states. For example, in environments with large state spaces, such as image input, Q-learning is likely fail, as it will encounter states during testing that were never visited during training. Therefore, we need a way to approximate values of unvisited states.

Moreover, both the state and actions space can be continuous, for example, in robotic control, where the actions are joint torques and the states comprise sensor information. A common solution is to discretize the state and action spaces, however, this approach has several limitations. First, a coarse discretization may cause the agent to miss important actions, whereas a fine-grained resolution can be computationally intractable. Second, as the dimensionality of the observation and action spaces increases, the number of discrete state-action pairs grows exponentially. This curse of dimensionality makes learning infeasible, as the agent would need to visit an extremely large amount of to find an optimal policy. 

To address these limitations, we can leverage the representative power of neural networks as function approximators that estimate the Q-value function. The same approach can be applied to continuous action spaces, where a policy is a neural network that maps from states to continuous actions. This approach is generally known as deep reinforcement learning (DRL). We will begin by discuss Deep Q-learning which uses neural networks to approximate the Q-function. Next, we will introduce policy gradient methods, a family of methods designed to handle continuous action spaces effectively.    

\begin{figure}[t!]
    \centering
    \begin{subfigure}[b]{0.49\textwidth}
        \centering
        \includegraphics[width=\textwidth]{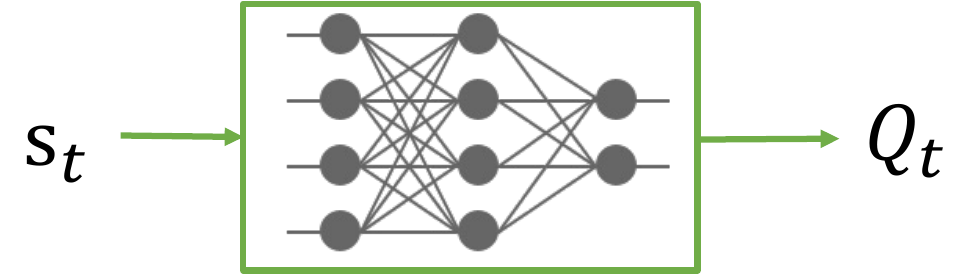}
        \caption{Q-value Function Approximation}
        \label{fig:bg:qvalue_nn}
    \end{subfigure}
    \hfill
    \begin{subfigure}[b]{0.48\textwidth}
        \centering
        \includegraphics[width=\textwidth]{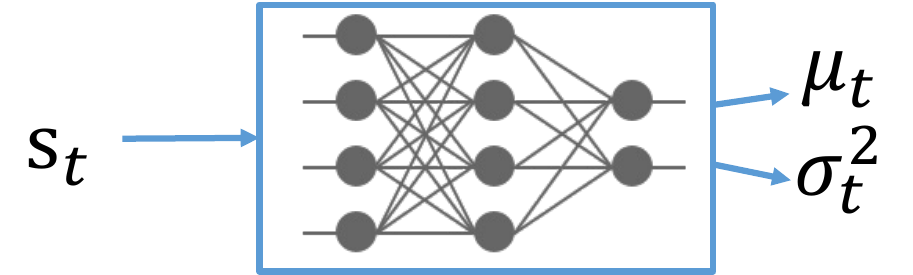}
        \caption{Policy Function Approximation}
        
        \label{fig:bg:policy_nn}
    \end{subfigure}
    \caption{In deep reinforcement learning, both the Q-values and the policy can be approximated with neural networks.}
\end{figure}

\subsection{Deep Q-learning}
\label{sec:dqn}
In standard Q-learning, the agent learns a separate Q-value for each state-action pair ($\state, \actions$). In Deep Q-learning, the key idea is to use a neural network to learn the mapping from state-action pairs to their Q-values (see \figref{fig:bg:qvalue_nn}). The Q-function is parameterized by network's weights denoted as $\theta$. This approach, combined with techniques to stabilize training, is called Deep Q-networks (DQN) \cite{mnih2015human}. To train the Q-network, stochastic gradient descent is used to minimize the temporal different error, also known as Bellman error, as defined by the loss function:

\begin{equation}
\label{eq:deep_q_learning}
    \mathcal{L}(\phi) = r(\state,\actions) + \gamma \max_{\actions'} Q_{\phi}(\state', \actions') - Q_{\phi}(\state, \actions).
\end{equation}

\noindent To stabilize training gradients are backpropagated only through the $Q_{\phi}(\state, \actions)$ term in Equation \ref{eq:deep_q_learning}. However, the updates in reinforcement learning are typically not independent and identically distributed (i.i.d), since states visited in a trajectory are strongly correlated. To address this, transitions ($\state, \actions, \state', r$) visited during exploration are stored in a replay buffer. During training, the Q-network is updated by sampling transitions from this buffer. Since Q-learning is an off-policy algorithm, we can reuse old samples, even from a previous policy, to update the Q-network (see \secref{sec:bg:onoffpolicy}). 

Another challenge arises because both terms in Equation \ref{eq:deep_q_learning} depend on the same network parameters $\phi$, meaning we are trying to update $Q_\phi$ such that it becomes closer to the target $r(\state,\actions) + \gamma \max_{\actions'} Q_{\phi}(\state', \actions')$, which itself is a function of $\phi$. This can lead to instability during training. To mitigate this, DQN proposes a target network, a separate network with parameters $\phi_{\text{target}}$, which is a copy of the Q-network but updated less frequently. This decouples the target value from the current paramters of the Q-network, stabilizing the training process. Therefore, the updated loss function becomes:

\begin{equation}
\label{eq:deep_q_learning_target}
        \mathcal{L}(\phi) = r(\state,\actions) + \gamma \max_{\actions'} Q_{\phi_\text{target}}(\state', \actions') - Q_{\phi}(\state, \actions).
\end{equation}

In this approach, DQN still relies on a discrete action space. After training, the optimal policy is determined by evaluating the Q-values for all possible actions in a given state and selecting the action with the highest value. DQN has demonstrated superhuman performance on Atari games with high-dimensional image input, validating the effectiveness of approximating the Q-function with neural networks. We now discuss algorithms that work with continuous action spaces. 

\subsection{Policy Gradient Algorithms}
While DQN is effective for discrete action spaces, it is not directly applicable to problems with continuous action spaces. Therefore, we now turn to poolicy gradient algorithms, a family of methods that directly optimize the policy. The key insight is to represent a policy as a neural network that maps from states to actions, allowing the agent to directly learn the correct behavio rather than relying on a value function. The challenge, however, lies in updating the policy network such that it improves the agent's performance. 

To this end, we define policy as a probability distribution $\policy_\theta(\actions_t|\state_t) = \mathcal{N}(\mu_t, \sigma^2_t|\state_t)$ that maps states to a mean $\mu_t$ and variance $\sigma^2_t$ and is parameterized by the network weights $\theta$, as shown in \figref{fig:bg:policy_nn}. The goal of policy gradients is to increase the likelihood of trajectories with high rewards and decrease the likelihood of trajectories with low rewards. To explore the environment, we sample actions from the policy action distribution at each time step. To evaluate and improve the policy, we define a performance measure $J$ that captures the expected long-term reward of the trajectories:

\begin{equation}
\label{eq:J_definition}
    J(\theta) = \mathbb{E}_{\tau \sim \policy_\theta}\left[\sum_{t} \gamma^t r(\state_t, \actions_t) \right],
\end{equation}
\noindent where $\tau$ represents a trajectory, i.e., a sequence of state and action pairs $(\state_0, \actions_0, \cdots, \\ \state_{T-1}, \actions_{T-1})$, and $r(\state_t, \actions_t)$ denotes the reward received at each time step. The probability of a trajectory is therefore given by:
\begin{equation}
    p(\tau) = p(\state_0, \actions_0, \cdots, \state_T, \actions_T) = p(\state_1) \prod_{t=0}^{T-1} \policy_\theta(\actions_t|\state_t)p(\state_{t+1}|\actions_t, \state_t). 
\end{equation}
The objective is to maximize the performance measure $J$ such that:
\begin{equation}
    \theta^* = \argmax_{\theta} J(\theta).
\end{equation}
\noindent To update the policy parameters $\theta$, we use gradient ascent:
\begin{equation}
\label{eq:J_update}
    \theta_{t+1} = \theta_t + \alpha \nabla_{\theta} J(\theta), 
\end{equation}
\noindent where $\alpha$ is the learning rate. It is important to ensure that the update step through gradient ascent improves the policy. Additionally, we need to compute the gradients of the performance measure $J$  without explicit knowledge of the environment dynamics. As a first step, we simplify the notation in Equation \ref{eq:J_definition} and convert the expectation to an integral form:
\begin{equation}
    J(\theta) = \mathbb{E}_{\tau \sim \policy}\left[\sum_{t} \gamma^t r(\state, \actions) \right] = \mathbb{E}_{\tau \sim \policy}[r(\tau)] = \int p(\tau) r(\tau) d\tau.
\end{equation}
\noindent Next, we compute the gradients of $J(\theta)$ using the log probability trick:
\begin{equation}
    \nabla_\theta J(\theta) = \int \nabla_\theta p(\tau) r(\tau) d\tau = \int p(\tau) \nabla_\theta \log p(\tau) r(\tau) d\tau.
\end{equation}
\noindent We then compute $\nabla_\theta \log p(\tau)$:
\begin{equation}
\begin{split}
   & \nabla_\theta \log p(\tau) = \nabla_\theta \log \left [ \prod_{t=0}^{T-1} \policy_\theta(\actions_t|\state_t)p(\state_{t+1}|\actions_t, \state_t) \right ] \\
   & = \nabla_\theta \left [ \log p(\state_1) + \sum_{t=0}^{T-1} \log \policy_\theta(\actions_t|\state_t) + \sum_{t=0}^{T-1} \log p(\state_{t+1}|\actions_t, \state_t) \right ] \\
   & = \nabla_\theta \sum_{t=0}^{T-1} \log \policy_\theta(\actions_t|\state_t).
\end{split}
\end{equation}
\noindent Because the initial state $\state_0$ and the environment dynamics are independent of the network parameters $\theta$, they do not affect the gradient computation. Therefore, the gradient of the performance measure simplifies to:
\begin{equation}
\label{eq:J_gradients}
   \nabla_\theta  J(\theta) = \mathbb{E}_{\tau \sim \policy_\theta} \left [ \left ( \nabla_\theta \sum_{t=0}^{T-1} \log \policy_\theta(\actions_t|\state_t) \right ) \left ( \sum_{t=0}^{T-1} \gamma^t r(\state_t, \actions_t) \right ) \right ].
\end{equation}
\noindent In this equation, the first term represents the gradient of the log-probability of the actions, which we can be computed using standard backpropagation techniques. The second term is a reward-based scaling factor. Intuitively, this means that we are adjusting the policy parameters in a way that scales gradients by rewards, making trajectories of high reward more likely. The remaining challenge is to implement a practical algorithm that leverages these policy updates and approximates the expected rewards. 

\paragraph{REINFORCE Algorithm} This algorithm generates trajectories $\tau_i$ by rolling out the policy, i.e., sampling actions from the policy action distribution until termination. The gradient estimate in Equation \ref{eq:J_gradients} is then  approximated by averaging over $N$ sampled trajectories:
\begin{equation}
   \nabla_\theta  J(\theta) = \frac{1}{N} \sum_{i=1}^N \left ( \sum_{t=0}^{T-1} \nabla_\theta \log \policy_\theta(\actions_t^i|\state_t^i) \right ) \left ( \sum_{t=0}^{T-1} \gamma^t r(\state_t^i, \actions_t^i) \right ).
\end{equation}
\noindent These gradients are then used to update the network parameters $\theta$ via gradient ascent, as shown in Equation \ref{eq:J_update}. However, the gradients can be noisy, since the rewards are averaged over a limited number of samples. To reduce this variance, a baseline term $b(\state_t)$, which is independent of the policy, can be introduced:
\begin{equation}
\label{eq:reinforce_baseline}
   \nabla_\theta  J(\theta) = \frac{1}{N} \sum_{i=1}^N \left ( \nabla_\theta \sum_{t=0}^{T-1} \log \policy_\theta(\actions_t^i|\state_t^i) \right ) \left ( \sum_{t=0}^{T-1} \gamma^t r(\state_t^i, \actions_t^i) - b(\state_t^i) \right ),
\end{equation}
\noindent where $b(\state_t)$ can for example be a running average reward. This baseline helps to reduce the variance of the gradient estimates while maintaining the an unbiased estimate of the gradient.

The next step is to explore whether we can compute even better estimates of the gradients, which brings us to actor-critic methods. We will discusse these methods in the following section. 

\section{Actor-Critic Methods}
A more accurate estimate for the baseline function given in Equation \ref{eq:reinforce_baseline} can further reduce the variance. As discussed in \secref{sec:bg:td}, temporal difference learning provides a way to estimate the expected long term reward from a given state, known as the value function. We can use this value function to replace both the baseline term $b(\state_t)$ and the sampled trajectory reward $\sum_{t=0}^{T-1} \gamma^t r(\state_t, \actions_t)$ in Equation \ref{eq:reinforce_baseline} as follows:

\begin{equation}
\label{eq:actor_critic}
   \nabla_\theta J(\theta) = \frac{1}{N} \sum_{i=1}^N \sum_{t=0}^{T-1} \nabla_\theta \log \policy_\theta(\actions_t|\state_t) \left ( r(\state_t, \actions_t) + \gamma V_\phi(\state_{t+1}^i) - V_\phi(\state_t) \right ).
\end{equation}
In this equation, the term $r(\state_t, \actions_t) + \gamma V_\phi(\state_{t+1}^i) - V_\phi(\state_t)$ is called advantage function. It measures how much better an action is compared to the average action at a given state: 
\begin{equation}
\label{eq:advantage}
   A(\state_t, \actions_t) = Q(\state_t, \actions_t) - V(\state_t) \approx r(\state_t, \actions_t) + \gamma V(\state_{t+1}) - V(\state_t).
\end{equation}
Note that the TD(0) single-step error term introduced in \secref{sec:bg:td} is an approximation of the advantage function.

This approach depends both on policy gradient updates and a value function estimate, where the policy (actor) and the value function (critic) are represented by separate neural networks. To update the policy network parameters $\theta$, the gradients are computed according to Equation \ref{eq:actor_critic} and updated via gradient ascent. The critic network parameters $\phi$ are updated using a similar approach to Equation \ref{eq:deep_q_learning_target}, but with a value function $V$ instead of the Q-value function. Most state-of-the-art approaches build upon this actor-critic framework. In the following sections, we will describe the two algorithms used in this dissertation.

\subsection{Proximal Policy Optimization}
\label{bg:ppo}
One of the main challenges with the previously introduced algorithms is selecting a suitable step size $\alpha$ for gradient ascent (Equation \ref{eq:J_update}). Unlike supervised learning, where step sizes are less sensitive, one large step in the wrong direction can significantly degrade the policy. To address this, the Trust Region Policy Optimization (TRPO) algorithm \cite{schulman15trpo} introduces a method to control the step size by constraining change in the action distribution. Therefore, TRPO formulates policy optimization as a constrained optimization problem:
\begin{equation}
\label{eq:trpo}
\begin{split}
     & \theta^* = \argmax{J(\theta, \theta_{\text{old}})}, \\
     & \text{s.t. } D_{KL}(\policy_\theta(\cdot|\state_t)||\policy_{\theta_{\text{old}}}(\cdot|\state_t)) \leq \delta.
\end{split}
\end{equation}

Here, the performance measure $J(\theta, \theta_{\text{old}})$ evaluates how the current policy $\policy_\theta$ performs relative to the old policy $\policy_{\theta_{\text{old}}}$, weighted by the advantage function $A$ (Equation \ref{eq:advantage}), also known as the surrogate advantage:
\begin{equation}
     J(\theta, \theta_{\text{old}}) = \mathbb{E}_{\tau \sim \policy_{\theta_{\text{old}}}} \left [ \sum_{t=0}^{T-1} \frac{\policy_{\theta}(\actions_t|\state_t)}{\policy_{\theta_{\text{old}}}(\actions_t|\state_t)} A(\state_t, \actions_t) \right ].
\end{equation}
The term $D_{KL}$ represents the KL-divergence between the old and current policy  distribution, which must remain smaller than a threshold $\delta$. Although the gradients of the surrogate advantage function turn out to be equal to the policy gradients established in Equation \ref{eq:actor_critic}, the added constraint complicates the computation of the gradients. TRPO simplifies the problem using Taylor Expansion, such that the optimization problem can be analytically solved via Lagrangian duality \cite{boyd2004convex}. Since this is a second order optimization problem, it requires computating the inverse of the Hessian matrix, which can be computationally expensive. To mitigate this, TRPO leverages the conjugate gradient algorithm \cite{hestenes1952methods} to approximate the inverse without calculating the full matrix. For further details on the optimization problem and its solution, we refer the reader to the original TRPO paper \cite{schulman15trpo}. \\

\noindent Proximal Policy Optimization (PPO) \cite{schulman2017proximal}, which we use in this dissertation, builds upon TRPO by relaxing the hard constraint of Equation \ref{eq:trpo} and turning the optimization into a first-order problem.  Instead of enforcing a hard constraint, PPO proposes a penalty term for deviations in the policy distribution:
\begin{equation}
\label{eq:ppo}
     \theta^* = \argmax{J(\theta, \theta_{\text{old}})} - \beta D_{KL}(\policy_\theta(\cdot|\state_t)||\policy_{\theta_{\text{old}}}(\cdot|\state_t)),
\end{equation}
\noindent where $\beta$ is a tunable parameter that controls the penalty term. This formulation allows for some larger deviations from the old policy while generally optimizing within a trust region. In practice, however, PPO often achieves the best performance by using a clipped objective function, which further simplifies the optimization process:
\begin{equation}
\begin{split}
\label{eq:ppo_clip}
      & J(\theta, \theta_{\text{old}}) = \mathbb{E}_{\tau \sim \policy_{\theta_{\text{old}}}} \left [ \sum_{t=0}^{T-1} \min( \rho_t, \text{clip}(\rho_t, 1-\varepsilon, 1+\varepsilon) A(\state_t, \actions_t)) \right ],\\
      & \text{where } \rho_t = \frac{\policy_{\theta}(\actions_t|\state_t)}{\policy_{\theta_{\text{old}}}(\actions_t|\state_t)}.
\end{split}
\end{equation}
\noindent Th term $\rho_t$ represents the ratio of the probabilities under the current and old policy, and $\text{clip}(\rho_t, 1-\varepsilon, 1+\varepsilon)$ restricts this ratio to the range $[1-\varepsilon, 1+\varepsilon]$. This clipping technique prevents large policy updates that could destabilize the policy.\\

\noindent Overall, the PPO algorithm is more straightforward to implement and computationally efficient due to its first order nature and relaxed constraints. Note that PPO and TRPO are on-policy algorithms, where each batch of collected trajectories is used to update the policy and then discarded. For the estimation of the advantage function, a critic network is typically used, which leads to more stable policy updates.

\subsection{TD3}
\label{bg:td3}
TD3 (Twin Delayed Deep Deterministic policy gradient) \cite{fujimoto2018addressing} is a widely used algorithm for continuous control that improves upon its predecessor, Deep Deterministic Policy Gradients (DDPG) \cite{lillicrap2016ddpg}. Like DDPG, it is an actor-critic method, consisting of a deterministic policy $\policy_\theta(\state)$ (actor) and a Q-function approximator $Q_\phi(\state,\actions)$ (critic), both represented by neural networks with parameters $\theta$ and $\phi$, respectively. To improve exploration with a deterministic policy, noise is added to actions in the exploration phase. As in DQN, TD3 is an off-policy algorithm that uses a replay buffer to store training transitions. 

During training, both the actor and critic are updated using samples from the buffer. Unlike DQN, where the action space is discrete, TD3 deals with continuous action spaces. Thus, computing the maximum over the continuous action space for the critic update in Equation \ref{eq:deep_q_learning_target} becomes more challenging. To address this, a similar technique of target network for the Q-function (see \secref{sec:dqn}) is applied to the policy network. Specifically, TD3 uses a policy target network $\policy_{\theta_{\text{target}}}$ which is an older version of the policy network that is less frequently updated than the policy. By sampling an action from the target network $\policy_{\theta_{\text{target}}}$ we get an approximation of the best action from a given state. Therefore, the critic is updated by minimizing the following loss function based on the Bellman error:
\begin{equation}
\label{eq:critic_target_ddpg}
        \mathcal{L}_\text{BE}(\phi) = \mathbb{E}_{\mdp } \left [ R(\state,\actions) + \gamma  Q_{\phi_\text{target}}(\state', \policy_{\theta_{\text{target}}}) - Q_{\phi}(\state, \actions) \right ], 
\end{equation}
\noindent where the transitions $\mdp$ are sampled from the replay buffer.
For the actor network, the policy parameters are updated to maximize the Q-values:

\begin{equation}
\label{eq:ddpg_update}
\mathcal{L}_{\text{DDPG}}(\theta) = \mathbb{E}_{\policy_\theta} \left [Q_{\phi}(\state_{t}, \actions_{t})    \vert \state_{t}, \actions_{t} = \policy_{\theta}(\state_{t}) \right ].
\end{equation} 
TD3 introduces several improvements over DDPG to increase training stability and improve performance.To mitigate overestimation of the value function, TD3 uses two separate critic networks and computes the minimum value between them to calculate the target value for the loss function. Additionally, TD3 adds noise to the actions during critic updates to smooth the Q-function along changes in actions. Furthermore, TD3 proposes to update the policy less frequently compared to the critic to the policy from being updated too quickly.

%% file: chapters/04_hoi_generation/hoi_generation.tex
\def\dir{chapters/04_hoi_generation}

\input{\dir/dgrasp/dgrasp}

\input{\dir/artigrasp/artigrasp}

\input{\dir/content/conclusion}

%% file: chapters/04_hoi_generation/dgrasp/dgrasp.tex
\chapter{D-Grasp: Physically Plausible Dynamic Grasp Synthesis for Hand-Object Interactions}
\chaptermark{D-Grasp: Physically Plausible Dynamic Grasp Synthesis for HOI}
\label{ch:hoi_generation:dgrasp}

\contribution{
To model dynamic hand-object interactions, we introduce the novel task of \taskname; given an object with a known 6D pose and a static grasp reference, our goal is to generate motions that move the object to a target 6D pose. This is challenging, because it requires reasoning about the complex articulation of the human hand and the intricate physical interaction with the object. We propose a novel method that frames this problem in the reinforcement learning framework and leverages a physics simulation, both to learn and to evaluate such dynamic interactions. 
A two-stage approach decomposes the task into grasping and motion synthesis. It can be used to generate novel hand sequences that approach, grasp, and move an object to a desired location. To learn grasping, our general reward function enables training a policy across many objects and retaining human-likeness through the use of a single frame grasp reference label. In our experiments, we demonstrate that our approach successfully generates realistic and physically plausible hand-object interactions. Furthermore, we show that a grasp reference can be acquired from grasp synthesis or image-based pose estimation.
}

\makeatletter \def\input@path{\dir/04_hoi_generation/dgrasp} \makeatother

\input{chapters/04_hoi_generation/dgrasp/sections/01_introduction}

\input{chapters/04_hoi_generation/dgrasp/sections/04_method}
\input{chapters/04_hoi_generation/dgrasp/sections/05_experiments}
\input{chapters/04_hoi_generation/dgrasp/sections/06_conclusion}

%% file: chapters/04_hoi_generation/dgrasp/sections/01_introduction.tex
\begin{figure*}[t]
\begin{center}
   \includegraphics[width=1.0\textwidth]{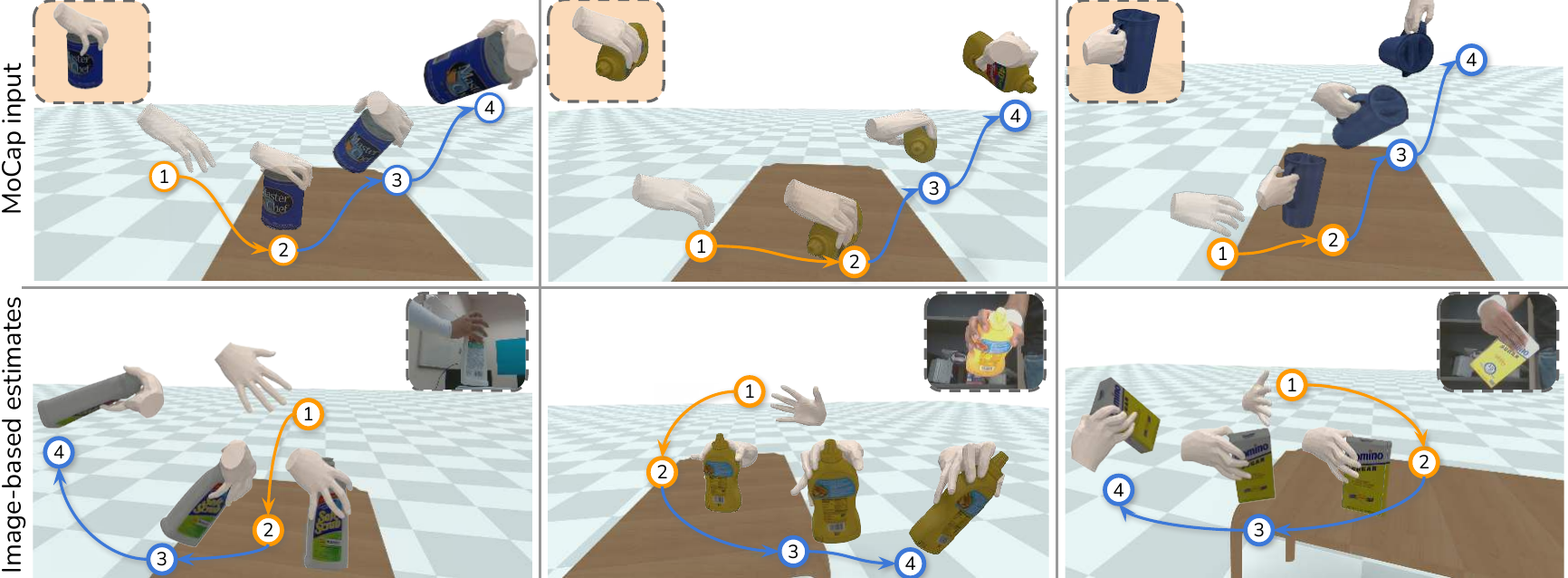}
\end{center}
   \caption{\textbf{Dynamic Grasp Synthesis}: 
    Our method learns diverse grasps from static grasp labels (shown in insets), originating from existing datasets, grasp synthesis or image-based estimates. Our approach can then synthesize diverse dynamic sequences with the objects in-hand. We decompose the task into: stable grasping \circledorange{\textcolor{black}{1}}-\circledorange{\textcolor{black}{2}}, followed by the synthesis of a 3D global motion to move the object into a 6D target pose \circledblue{\textcolor{black}{3}}-\circledblue{\textcolor{black}{4}}. The hand-pose is continuously adjusted to ensure a stable grasp, leading to physically plausible and human-like sequences.}
\label{fig:dgrasp:teaser}
\end{figure*}

\section{Introduction}

A key problem in computer vision is to understand how humans interact with their surroundings. Because hands are our primary means of manipulation with the physical world, there has been an intense interest in hand-object pose estimation \cite{tekin2019h+, hasson2019learning, hasson2020leveraging, chao2021dexycb, hampali2020honnotate, taheri2020grab, jiang2021hand} and the \emph{synthesis} of \emph{static} grasps for a given object \cite{jiang2021hand, karunratanakul2021skeleton, li2022artiboost, taheri2020grab}. However, human grasping is not limited to a single time instance, but involves a continuous interaction with objects in order to \emph{move} them. It requires maintaining a stable grasp throughout the interaction, introducing intricate dynamics to the task. This involves reasoning about the complex physical interactions between the dexterous hand and the manipulated object, including collisions, friction, and dynamics. A generative model that can synthesize realistic and physically plausible object manipulation sequences in 4D (3D space + 1D time) would have many downstream applications in AR/VR, robotics and human-computer interaction (HCI). 

We propose the new task of \textit{\taskname}. Given an object with a known 6D pose and a static grasp reference, our goal is to generate a grasping motion of a human hand and to move the object to a target 6D pose in a natural and physically-plausible way. This new setting adds several challenges. First, the object geometry and the spatial configuration of the object and the hand need to be considered in continuous interaction. Second, contacts between the hand and object are crucial in maintaining stability of the grasps, where even a small error in hand pose may lead to an object slipping. Moreover, contact is typically unobservable in images \cite{ehsani2020use} and measuring the stability of a grasp is very challenging in a static setting. Finally, synthesizing sequences of hand motion requires the generation of smooth and plausible trajectories. While prior work investigates the control of dexterous robotic hands by learning from full demonstration trajectories \cite{garciahernando2020physics,rajeswaran2018learning}, we address the generation of hand motion from only a single-frame grasp reference. This is a more challenging setting, because the generation of human-like hand-object interaction trajectories without dense supervision is not straightforward.

Taking a step towards this goal, we propose \dgrasp, which generates physically plausible grasping motions with only a single grasp reference as input (\Fig{dgrasp:teaser}). Concretely, we formulate the \textit{\taskname} task as a reinforcement learning (RL) problem and propose a policy learning approach that leverages a physics simulation. Our RL-based approach considers the underlying physical phenomena and compensates data scarcity via exploration in the physics simulation. This ensures physical plausibility, e.g., there is no hand-object interpenetration and the fingers exert enough force on the object to hold it without slipping. 

Specifically, we introduce a two-stage framework that consists of a  grasping phase and a motion synthesis phase. The grasping policy's purpose is to establish and maintain a stable grasp, whereas the motion synthesis module generates a motion to move the object to a user-specified target position. To guide the grasping policy, we require a single grasp label corresponding to a static hand pose, which can be obtained either from a hand-grasping dataset~\cite{chao2021dexycb,hampali2020honnotate}, a state-of-the-art grasp synthesis method~\cite{jiang2021hand} or via an image-based pose estimator \cite{grady2021contactopt}. Crucially, we propose a reward function that is parameterized by the grasp label to incentivize the fingers to follow the target hand pose, leading to human-like grasps. Additionally, we reward exerting force on the object to reach stable grasps.
Our motion synthesis module generates motions that move the hand and object to the final target pose. Importantly, the grasping policy continually controls the grasp to not drop the object. 

In our experiments, we first demonstrate that samples from motion capture, static grasp synthesis or image-based pose estimates often do not lead to stable grasps when evaluated in a physics  simulation (\figref{fig:dgrasp:grasp_qual}). We then present how our method can learn to produce physically plausible and stable grasps when guided by such labels. Next, we set out to generate motions with the object in-hand to reach a wide range of target poses. We provide an extensive ablation,  revealing the importance of the two-stage approach and the reward formulation for \taskname.
\pagebreak

In summary we contribute the following: 
\begin{itemize}
    \item The new task of \taskname that entails grasping an object with a simulated human hand and bringing it into a target 6D pose.
    \item \dgrasp, an RL-based method to synthesize physically-plausible and natural hand-object interactions.
    \item Extensive evaluations and ablations validating the effectiveness of our framework in generating physically plausible and natural hand-object interactions.
    \item Experiments showing that our method can generate grasp motions with static grasp references, which can originate from motion capture, static grasp synthesis or image-based pose estimation.
\end{itemize}

%% file: chapters/04_hoi_generation/dgrasp/sections/04_method.tex
\section{Method}
\label{sec:dgrasp:method}
\begin{figure*}[t]
\begin{center}
   \includegraphics[width=1.0\textwidth]{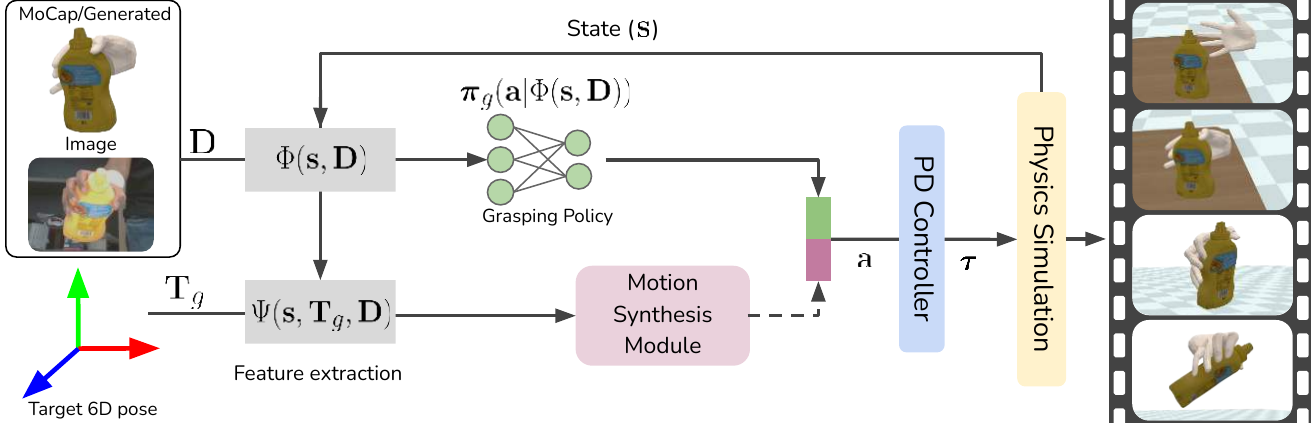}
\end{center}
   \caption{\textbf{Method Overview:} Taking a single, static grasp label  $\grasplabel$ and a target object 6D pose $\mathbf{T}_g$ as input (leftmost), \dgrasp produces sequences of dynamic hand-object interactions (rightmost). To do so, we propose a two-stage framework that consists of a grasping policy $\boldsymbol{\pi}_g(\cdot)$ and a motion synthesis module. In the \emph{grasping} phase, only the grasping policy is active and finds a stable grasp on the object. In the subsequent \emph{motion synthesis} phase, both the grasping policy and the motion synthesis module act concurrently. The actions $\mathbf{a}$ consist of joint targets. These are combined and passed to a PD-controller that computes the required torques $\boldsymbol{\tau}$ to control a MANO-based hand model in a physics simulation. The physics simulation updates the state $\mathbf{s}$ which serves as input to a reward formulation (\secref{sec:dgrasp:method_reward}) that forms our supervision signal and incentivizes the hand to approach and grasp the object and to move it to the target 6D pose. We introduce two feature extraction layers ($\Phi(\cdot)$ an $\Psi(\cdot)$) that utilize the environment state $\mathbf{s}$ and grasp label $\grasplabel$ to find a suitable representation for the grasping policy and the motion synthesis module.}

\label{fig:dgrasp:method_overview}
\end{figure*}

We propose \emph{D-Grasp}, an RL-based approach that leverages a physics simulation for the \emph{\taskname}~task (\Fig{dgrasp:method_overview}). Our model requires a static grasp label consisting of the hand's 6D global pose and local pose for the fingers. We split the task into two distinct phases, namely a \emph{grasping} and a \emph{motion synthesis} phase. In the \emph{grasping} phase, the hand needs to approach an object and find a physically-plausible and stable grasp. In the \emph{motion synthesis} phase, the hand has to bring the object into the 6D target pose while the grasping policy retains a stable grasp on the object. Therefore, the grasping policy and motion synthesis module act concurrently in this phase. To this end, we follow a  framework that functionally separates grasping from motion synthesis. 

In the next section, we define the task setting and provide details about the physics simulation. Thereafter, we present both the \textit{grasping} and \textit{motion synthesis} phases of our method in Sections \ref{sec:dgrasp:method_grasp} and \ref{sec:dgrasp:method_motion}, respectively.

\subsection{Task Setting}
In the \textit{\taskname} task, we are given a 6D global pose $\posesix_h$ and 3D local pose $\posevec$ of a hand, and an object pose $\posesix_o$, where the 6D poses consist of a rotation and translation component $\posesix=[\mathbf{q}|\mathbf{t}]$. Given a label of a static grasp $\grasplabel=(\overline{\mathbf{q}}_h, \overline{\posesix}_h, \overline{\posesix}_o)$, the goal is to grasp the object and move it into a 6D goal pose  $\posesix_g$. The grasp label consists of the 6D global pose of the hand $\overline{\posesix}_h$ and object  $\overline{\posesix}_o$, as well as the target hand pose $\overline{\mathbf{q}}_h$ at the instance of the static grasp. 

\begin{figure}[t]
\centering
\begin{subfigure}[t]{0.49\textwidth}
    \centering
    \includegraphics[width=\columnwidth]{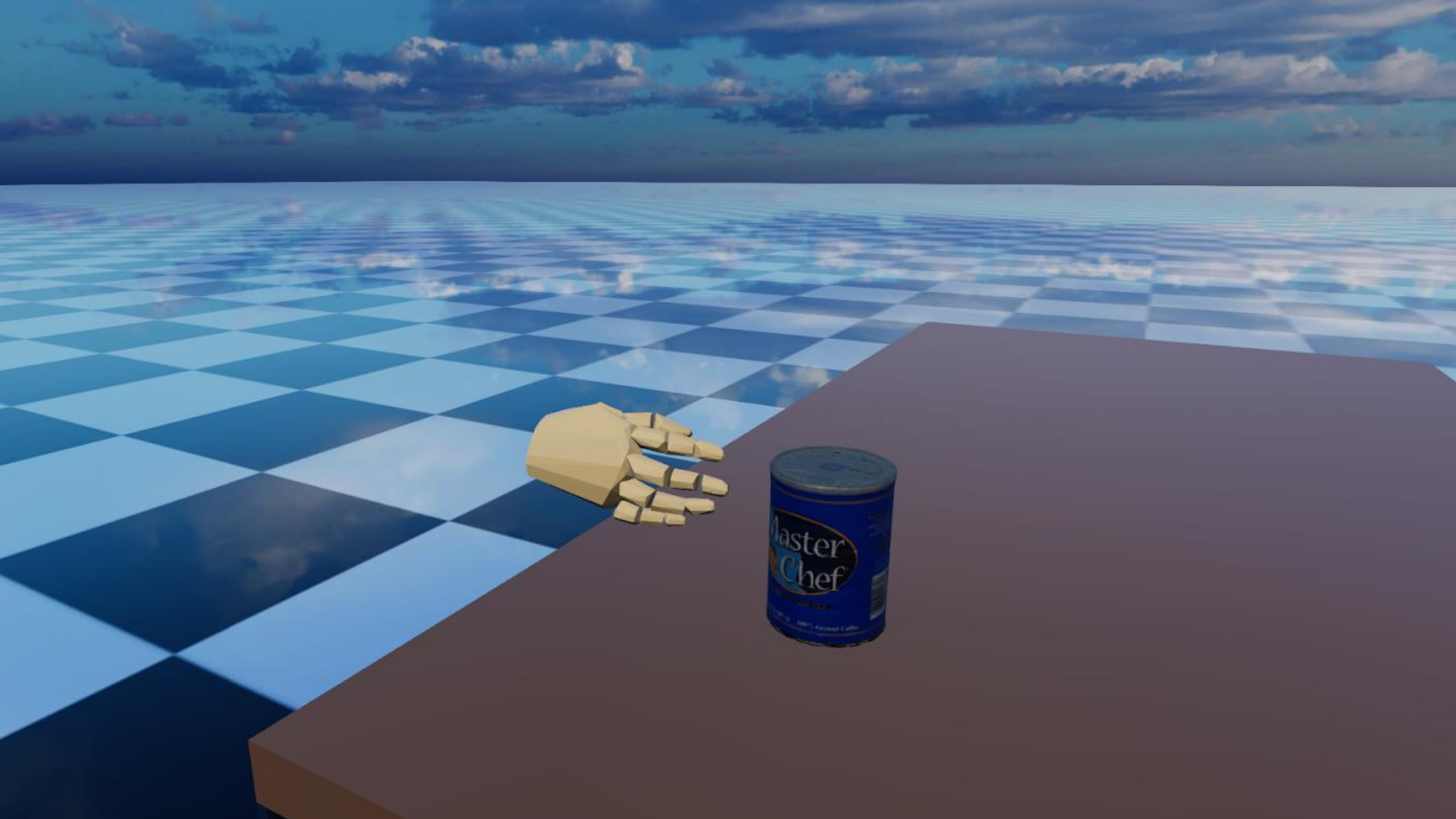}
    \caption{\textbf{Physics Simulation.} We create a controllable human hand model and deploy it in the RaiSim physics-engine \cite{hwangbo2018per} to provide information about contacts and dynamics.}
    \label{fig:hoi:dgrasp:raisim}
\end{subfigure}
\hfill
\begin{subfigure}[t]{0.49\textwidth}
    \centering
    \includegraphics[width=0.7\columnwidth]{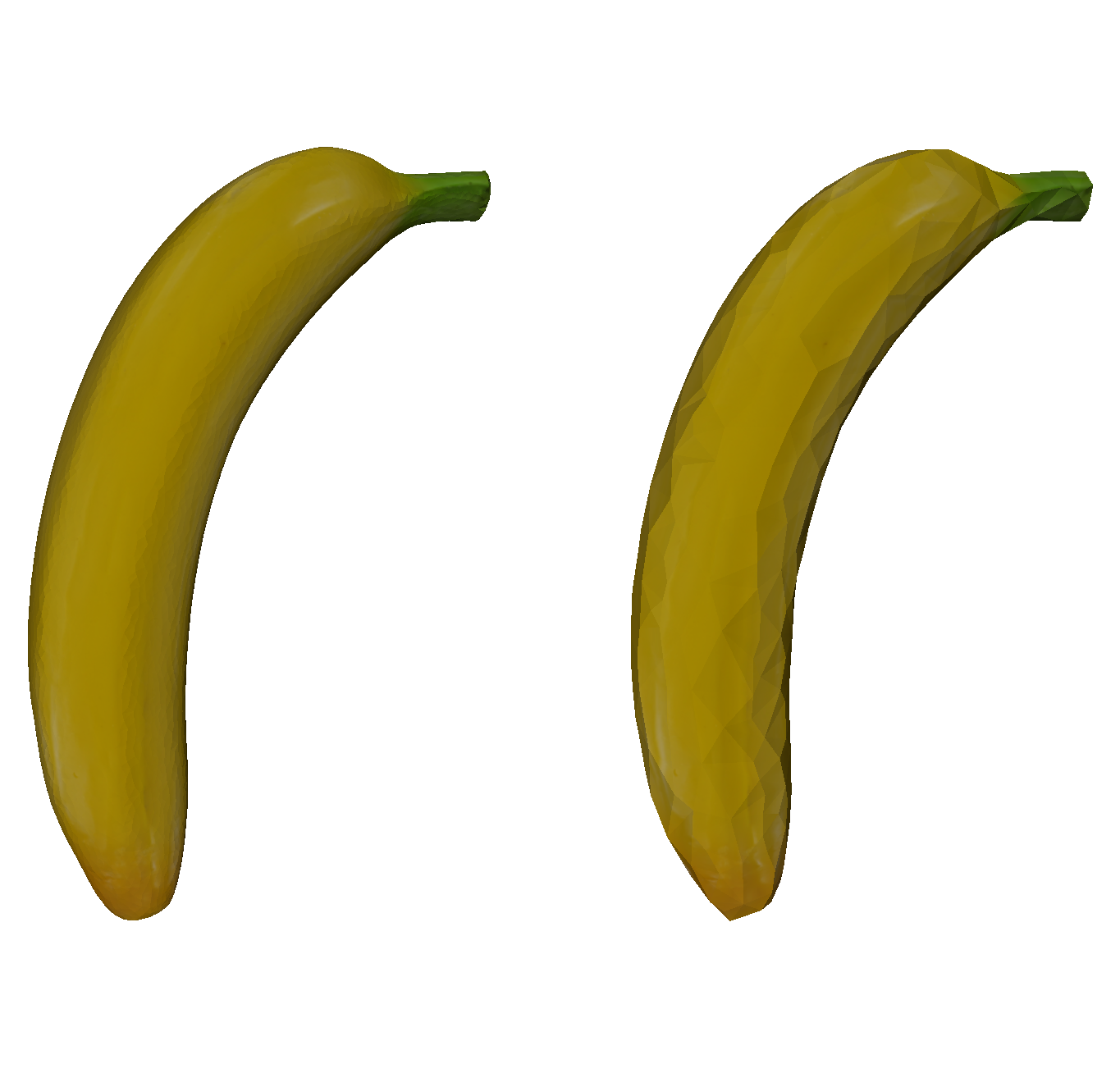}
    \caption{\textbf{Mesh Decimation.} We use mesh decimation to reduce the number of vertices of the object mesh. On the left is the original object mesh, on the right the decimated mesh. This helps to speed up the physics simulation during training.}
    \label{fig:hoi:dgrasp:mesh_decimation}
\end{subfigure}
\caption{Overview of the physics simulation and mesh decimation.}
\end{figure}

\paragraph{Simulation Setup}
\label{sec:dgrasp:simulation_setup}
As shown in \figref{fig:hoi:dgrasp:raisim}, we approximate a human-like hand in the physics engine by creating a controllable hand model and integrating information obtained from a statistical parametric hand model (i.e., MANO \cite{romero2017embodied}). Similar to \cite{yuan2021simpoe}, we compute the argmax of the skinning weights of MANO to assign each of the vertices to a body part. We then group the vertices accordingly and create a mesh for each body part. We extract the skeleton of the hand from MANO to get the relative joint positions and add joint actuators to control the hand. Finally, we restrict the joints to be within reasonable limits. In our implementation, we use a unified hand model corresponding to the mean MANO shape. We limit the joint range in a data-driven manner. Specifically, we estimate the joint limits by parsing the DexYCB dataset and acquiring the maximum joint range, similar to \cite{spurr2020weakly}. Since the data may not contain the full range of possible joint displacements, we increase this limit by a 20\%. In practice, we found that approximating the collision bodies with primitive shapes (i.e., the simple objects and the hand meshes) led to an order of magnitude increase in training speed. This is because the simulation time increases roughly quadratically with the number of collision points. Therefore, for more complex object meshes, we apply a decimation technique to reduce the number of vertices (see \figref{fig:hoi:dgrasp:mesh_decimation}). For the simpler meshes, we use primitive shapes and mesh alignment as an approximation, for example, a soup can is approximated by a cylinder. For training and evaluation, we therefore use the simplified meshes (except for the computation of the interpenetration metric).

\paragraph{State Space}
\label{sec:dgrasp:method_state}

The state $\state = (\posevec, \dot{\mathbf{q}}_h, \forcevec, \posesix_h, \dot{\posesix}_h, \posesix_o, \dot{\posesix}_o)$ entails proprioceptive information about the hand pose in the form of joint angles $\posevec$ and joint angular velocities $\dot{\posevecgen}_h$, the forces between the hand and object $\forcevec$, the 6D pose of the wrist $\posesix_h$ and the global 6D pose of the object $\posesix_o$ with their corresponding velocities $\dot{\posesix}_h$ and $\dot{\posesix}_o$. 
States are expressed with respect to a fixed global coordinate frame. 
We show experimentally that learning from the full state space can impede learning over several different grasp labels (\secref{sec:dgrasp:exp_ablation}). We therefore propose a representation that enables learning of the task in \secref{sec:dgrasp:method_feat_grasp}. 

\paragraph{Action Space}
\label{sec:dgrasp:method_action}
We define an action space to control the hand in the physics simulation. The fingers are controlled via one actuator per joint for a total of 45 actuators, to which we add 6 DoF to control the global pose. We employ PD-controllers that take reference joint angles $\posevecgen_{\text{ref}}$ as input and compute the torques that should be applied to the joints:
\begin{align}
    \torques &= k_p (\posevecgen_{\text{ref}} - \posevecgen) + k_d \dot{\posevecgen} \\
    \posevecgen_{\text{ref}} &= \posevecgen_b + \actions.
\end{align}
The policy $\policyvec$ outputs actions $\actions$, which are residual actions that change a bias term $\posevecgen_b$. For the finger joints and hand orientation, the bias term is equivalent to the current joint configuration $\posevecgen_b = \posevec$. We found this formulation to lead to smoother finger motion and therefore more stable grasps compared to the policy directly predicting $\posevecgen_{\text{ref}}$. Note that for simplicity's sake, we use the notation $\mathbf{q}_b$ for all joints, although the global hand position is controlled via translational joints.

\subsection{Physically Plausible Grasping}
\label{sec:dgrasp:method_grasp}

Here we discuss the \textit{grasping} phase. The goal is to approach an object and find a physically plausible grasp. A careful design of the model's input representation is key to learning a successful model for hand-object interactions~\cite{zhang2021manipnet}, which we show in our ablations (\secref{sec:dgrasp:exp_ablation}). Therefore, we introduce a feature extraction layer that converts the information from the physics simulation and grasp label into a suitable representation for model learning.

\paragraph{Feature Extraction for Grasping}
\label{sec:dgrasp:method_feat_grasp}
Rather than directly conditioning the policy on the state, we apply a feature extraction layer $\Phi(\state,\grasplabel)$ that takes the state and grasp label as input. For consistency, we can reformulate the policy as $\gpolicyvec(\actions|\Phi(\state,\grasplabel))$ (\figref{fig:dgrasp:method_overview}). 
The function $\Phi(\cdot)$ processes information from the grasp label, and applies coordinate frame transformations to achieve invariance w.r.t. global coordinates by transforming it to object-relative coordinates. To this end, the feature extraction layer receives the state  $\state = (\posevec, \dot{\mathbf{q}}_h, \forcevec, \posesix_h, \dot{\mathbf{T}}_h, \posesix_o, \dot{\posesix}_o)$ and grasp label $\grasplabel=(\overline{\mathbf{q}}_h, \overline{\posesix}_h, \overline{\posesix}_o)$ as input. Its output is defined as: 
\begin{equation}
     \Phi(\state,\grasplabel) =(\posevec, \dot{\mathbf{q}}_h, \forcevec, \widetilde{\posesix}_h, \widetilde{\posesix}_o, \dot{\widetilde{\posesix}}_o, \dot{\widetilde{\posesix}}_h, \widetilde{\posvec}_{o}, \widetilde{\posvec}_z, \goals).
\end{equation}

\begin{figure}[t]
\begin{center}
   \includegraphics[width=0.49\textwidth]{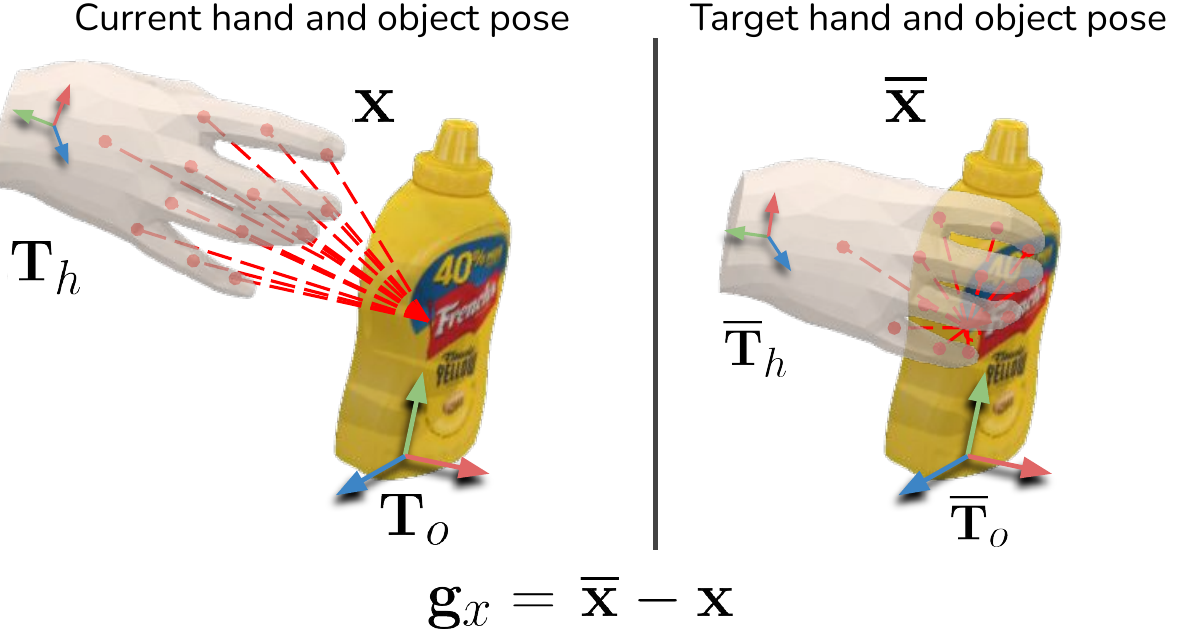}

\end{center}
   \caption{\textbf{Target Distance Component $\mathbf{g}_x$.} It incentivizes the policy to reach target points close to the grasp reference label $\grasplabel$. We extract the object-relative target 3D joint positions $\mathbf{\overline{x}}$ from $\grasplabel$ and compute the distance between $\mathbf{\overline{x}}$ and the current 3D joint positions $\mathbf{x}$ relative to the object's origin. We then convert  $\mathbf{g}_x$ into wrist-relative coordinates $\widetilde{\mathbf{g}}_x.$}
\label{fig:dgrasp:goal}
\end{figure}

The terms $\posevec$ and $\dot{\mathbf{q}}_h$ are the local joint angles and velocities, whereas $\forcevec$ represents contact force information. The remaining components are expressed in the wrist's reference frame (denoted by $\widetilde{\cdot}$ ): the object's 6D pose $\widetilde{\posesix}_o$ and its linear and angular velocities $\dot{\widetilde{\posesix}}_o$, the hand's 6D pose $\widetilde{\posesix}_h$ (relative to the initial wrist pose) and its linear and angular velocity $\dot{\widetilde{\posesix}}_h$, and the displacement of the object from its initial position $\widetilde{\posvec}_{o}$. Furthermore, $\widetilde{\posvec}_z$ introduces awareness of the vertical distance to the surface where the object rests. Lastly, we include the goal components $\goals=[\widetilde{\goalvec}_x|\widetilde{\goalvec}_q|\goalvec_c]$, which incentivize the model to reach contact points on the object. We show that these goal components are crucial for achieving stable grasps in \secref{sec:dgrasp:exp_ablation}. 
More specifically, the term $\widetilde{\goalvec}_x$ measures the 3D distance between the current and the target 3D positions of the hand (\figref{fig:dgrasp:goal}), $\posvec_h$ and $\overline{\posvec}_h$, respectively:
\begin{equation*}
   \widetilde{\goalvec}_x = \overline{\mathbf{x}}_h-\mathbf{x}_h.
\end{equation*}
Here, all joints and the fingertips are in the wrist's coordinate frame. Importantly, we compute object-relative target positions from the label $\grasplabel$ in order to be invariant to the object 6D pose during the grasping phase. 

Similarly, the term $\widetilde{\goalvec}_q$ represents the angular distance between the current rotations $\posevec$ and target rotations $\overline{\posevecgen}_h$ for the joints and the wrist orientation:
\begin{equation*}
   \widetilde{\goalvec}_q = \overline{\mathbf{q}}_h-\mathbf{q}_h.
\end{equation*}

Finally, $\goalvec_c$ includes the target contact vector $\goalvec_{c}$, i.e., which finger joints should be in contact with the object. In particular, the contact goal vector $\goalvec_c=(\overline{\goalvec}_c, \mathbb{I}_{\forcevec, 
\overline{\goalvec}_{c}>0})$ is the concatenation of two vectors, namely the desired contacts $\overline{\goalvec}_{c}$ and the term $\mathbb{I}_{\overline{\goalvec}_{c}>0}$ to indicate which of the desired contacts are satisfied. 
To get the desired contacts for each hand joint from the grasp label, we measure the distance between all of a joint's vertices of the created meshes (\secref{sec:dgrasp:simulation_setup})  and all the vertices of the object mesh, which can be computed from the grasp label $\grasplabel$. Hence, for each joint $j$, the desired contacts are then determined as follows: 
\begin{equation}
    \overline{\goalvec}_{c,j} =  \mathbb{I} \left [ \sum^{I}_{i = 1} \sum^{O}_{o = 1} \mathbb{I} [\|\overline{\mathbf{v}}_{\text{i}}-\overline{\mathbf{v}}_{o}\|^2<\epsilon] > 0\right].
\end{equation} 
If the distance between any vertex $\overline{\mathbf{v}}_{\text{i}}$ of a joint $j$ and an object vertex $\overline{\mathbf{v}}_{o}$ is below a small threshold $\epsilon$ (in our case 0.015m), we determine that the finger part should be in contact and hence the contact label should be equal to 1, otherwise 0.

\paragraph{Reward Function for Grasping}
\label{sec:dgrasp:method_reward}
To incentivize the policy to learn the desired behavior, we define a reward function as follows:
\begin{equation}
    r = w_{x} r_{x} +  w_{q} r_{q} +  w_{c} r_{c} +  w_{\text{reg}} r_{\text{reg}}.
\end{equation}
It comprises a combination between position, angle, contact and regularization terms, respectively. We weigh the reward components with the factors $w_{x},  w_{q},  w_{c},  w_{\text{reg}}$.\\
The position reward $r_{x}$ measures the weighted sum of distances between the target $\overline{\mathbf{x}}$ and the current 3D positions $\posvec$ for every joint (including the wrist):
\begin{equation}
    r_{x} = \sum^{J}_{j = 1} w_{x,j}\| \overline{\posvec}_j - \posvec_j\|^2.
\end{equation}
Similarly, the pose reward $r_q$ measures the distance between the current pose and the corresponding target pose in Euler angles and corresponds to the L2-norm of the feature $\widetilde{\goalvec}_q$:
\begin{equation}
   r_{q} = \|\widetilde{\goalvec}_q\|,
\end{equation}
The contact reward $r_{c}$ is extracted from the finger parts that should be in contact with the object. Specifically, it is computed as the sum of two terms. The first one represents the fraction of target contacts that the agent has achieved. The second term rewards the amount of force exerted on desired contact points, capped by a factor proportional to the object's weight $m_{o}$ through a factor $\lambda$:

\begin{equation}
    r_{c} = \frac{\widetilde{\goalvec}_c^\top \mathbf{I}_{\forcevec>0}}{\widetilde{\goalvec}_c^\top \widetilde{\goalvec}_c} + \min(\widetilde{\goalvec}_c^\top \forcevec, \lambda m_{o}).
\end{equation}
Finally, the reward $r^{\text{reg}}$ involves regularization terms on the hand's and object's linear and angular velocities:
\begin{equation}
r_{\text{reg}} =  w_{\text{reg},h} \| \dot{\widetilde{\posesix}}_h \|^2
+  w_{\text{reg},o} \| \dot{\widetilde{\posesix}}_o \|^2.
\end{equation} 

\paragraph{Wrist-Guidance Technique}
\label{sec:dgrasp:wrist-guidance}
To control the global pose during the grasping phase, we introduce a simple yet effective technique which we call \textit{wrist-guidance}. Intuitively, we bias the hand to approach the object. To achieve this, we leverage the object-relative target pose, of the hand on the object, obtained from the grasp label $\grasplabel$. We then use it as a bias term in the PD-controller of the global 3DoF position. In other words, we set the bias term of the first 3DoF (the translational joints) to $\posevecgen_b=\overline{\mathbf{x}}_h$ (\secref{sec:dgrasp:method_action}), where $\overline{\mathbf{x}}_h$ is the target position which we extract from the label. 
We find that this technique leads to better performance and faster convergence than using the previous joint positions as bias (\secref{sec:dgrasp:method_action}), which we show in ablations in \secref{sec:dgrasp:exp_ablation}.

\subsection{Motion Synthesis}
\label{sec:dgrasp:method_motion}
We now introduce the motion synthesis module, which is responsible for moving the object from an initial 6D pose into a target 6D pose. It controls only the movement of the wrist, i.e., the first 6DoF of the controllable hand model. In this phase, both the grasping policy described in \secref{sec:dgrasp:method_grasp} and the motion synthesis module are executed concurrently. While the grasping policy maintains a stable grasp, the motion synthesis module takes over the control of the 6D pose of the hand.  Similar to the grasping policy, we propose a feature extraction layer that incentivizes the model to move the hand to a target pose with the object in-hand.  

To control the global hand motion, we estimate a 6D target pose for the hand: $\posesix_\text{pd} = \Psi(\state, \posesix_g, \mathbf{D})$.
In particular, we compute the distance between the current and target 6D object pose $\Delta \posesix_o = (\posesix_g-\posesix_o)$. Assuming the object stays roughly stable within the hand, this term is then added to the current 6D hand pose and weighted by a factor $\beta$:
\begin{equation}
   \posesix_\text{pd} = \posesix_h + \beta \Delta \posesix_o.
\end{equation}
The term $\posesix_\text{pd}$ is passed to a PD-controller, which outputs torques that generate a motion to guide the hand to the estimated target pose. Note that in the \emph{motion synthesis} phase, this module replaces the control of the first 6DoF of the grasping policy. The displacement is recomputed after every action. 

%% file: chapters/04_hoi_generation/dgrasp/sections/05_experiments.tex
\section{Experiments}
We conduct several experiments to analyse the performance of our method. We first introduce the data and experimental details in Sections \ref{sec:dgrasp:exp_data} and \ref{sec:dgrasp:exp_setup}. Next, we show that our method can learn stable grasps and correct imperfect labels in \secref{sec:dgrasp:exp_grasp}. Lastly, we evaluate the motion synthesis task and provide ablations to highlight the importance of our method's components in Sections \ref{sec:dgrasp:exp_motion} and \ref{sec:dgrasp:exp_ablation}.
\subsection{Data}
\label{sec:dgrasp:exp_data}

\paragraph{DexYCB} We make use of the DexYCB dataset \cite{chao2021dexycb}. The dataset consists of 1000 sequences of object grasping, with 10 different subjects and 20 YCB objects \cite{calli2015ycb}. We filter out all left handed sequences and create a random 75\%/25\% train/test-split over all sequences and subjects.
The data sequence contains 6D global poses for the hand and objects in the camera frame and the local joint angles, hence providing sequences of 
$\begin{Bmatrix} (\overline{\mathbf{q}}_h, \overline{\posesix}_h, \overline{\posesix}_o)
\end{Bmatrix}_{t=1}^T$. The data also includes meshes for the hand and objects, and the camera parameters. We determine the grasp label based on the object's displacement with regards to its initial position. The time-step with an object displacement greater than a pre-determined threshold is chosen to be the target grasp $\grasplabel$. Furthermore, we use a recent state-of-the-art grasp synthesis method \cite{jiang2021hand} to generate grasp labels for all the objects in DexYCB and create a 400/200 label train/test-split.

\paragraph{HO3D}
We use generated grasp labels from static grasp synthesis \cite{jiang2021hand} or from an image-based pose estimator after offline optimization \cite{grady2021contactopt} for the  HO3D objects. We create a train/test-split that is proportional to the DexYCB split, which results in a 200/100 label train/test-split.

\subsection{Experimental Details}
\label{sec:dgrasp:exp_setup}
We train policies by using our implementation of PPO (see \secref{bg:ppo}) and run simulations in \raisim \cite{hwangbo2018per}. 
For each sequence, we initialize the environment with an object and a grasp label. We extract the initial hand pose from earlier steps in the data sequences at a minimum distance away from the object. First, we train the grasping policy with all training labels and objects. Then we continue with the motion synthesis component given the pretrained grasping policy. We evaluate physical plausibility of a grasp in terms of stability and interpenetration on a set of unseen grasp labels and unseen objects. When using grasp references from a static grasp synthesis method \cite{jiang2021hand}, we train with the objects used in DexYCB \cite{chao2021dexycb}. During evaluation, we report results on both the HO3D subset as done in \cite{jiang2021hand} and the objects from DexYCB. For the experiment with ContactOpt \cite{grady2021contactopt}, we train and test on the HO3D objects (except for 019\_pitcher\_base, which is not contained in the dataset). Note that since the models for grasp synthesis and the image-based pose estimates have no notion of physics in terms of where an object is positioned in space (in contrast to the data from DexYCB), we apply a small modification to the simulation to ensure a fair comparison. We place the object on a surface and allow the hand to approach from any direction, even penetrating the surface. We achieve this by disabling the collision response between the surface and the hand. In future work, an optimization could filter out poses that require approaching from beneath a surface. Also note that since we only have access to a single grasp reference and not a sequence for labels obtained from GraspTTA and ContactOpt, we start each sequence at a predefined distance away from the object in the mean MANO hand pose. For the evaluation of our method, we remove the surface (i.e. table) after the \emph{grasping} phase. The metrics are being measured from the moment the table is removed. For the baselines, we directly start the sequence in the target pose of both the hand and object (without a table present).

\paragraph{Metrics} We propose the following metrics: 
\label{sec:dgrasp:exp_metrics}

\noindent\textbf{Success Rate:} We define the success rate as the primary measure of physical plausibility. 
It is measured as the ratio of sequences which maintain a stable grasp, i.e., where the object does not slip and fall down for a period of a 5s window. We remove the surface in the simulation for this purpose. A success rate of 0.0 indicates no success, 1.0 means all trials were successful. \\
\textbf{Interpenetration (IV):} The amount of hand volume that penetrates the object. We use the MANO mesh \cite{romero2017embodied} and the high-resolution object mesh. To ensure a fair comparison against the static baseline, we choose the last time step of the grasping phase for our method and hence omit the approaching phase from the evaluation. \\  
\textbf{Simulated Distance (SD):}  Similar to the metric proposed in \cite{jiang2021hand}, we compute the mean displacement between the object and the hand's wrist. Instead of measuring the absolute displacement, we report the mean displacement in mm per second. We measure the displacement for a maximum window of 5s or stop whenever the object falls and hits the surface. \\
\textbf{Contact Ratio:} For the ablation study, we measure the ratio between the target contacts $\overline{\mathbf{g}}_c$ defined via the grasp label $\grasplabel$ and the contacts achieved in the physics simulation $\mathbb{I}[\mathbf{f}>0]$. We average over the whole sequence, therefore both the approaching and grasping phase are contained in this metric. \\
\textbf{MPE:} The mean position error between the object's position and target 3D position.\\
\textbf{Geodesic:} The angular distance between the object's current and target orientation.

\paragraph{Baselines} We propose the following baselines: 
\label{sec:dgrasp:exp_baselines}

\noindent \textbf{*+PD:} Similar to \cite{jiang2021hand}, we place the object into the hand via the grasp label. We then attempt to maintain the grasp using PD-control in the physics simulation.\\
\textbf{*+IK:} We employ an offline optimization to correct for imperfections (i.e., minor distances or penetrations) in the grasp label. The improved samples are passed to the PD-control. \\
\textbf{Flat-RL:} We employ an RL baseline that does not separate the grasping from the motion synthesis phase, but trains the full dynamic grasp synthesis task end-to-end.\\
\textbf{Ours+static grasp:} In this variant, we use our grasping policy for the grasping phase. During motion synthesis, we use PD-control to maintain the pose while the grasping policy is frozen and not actively interacting with the object.

\subsection{Grasping Objects}
\label{sec:dgrasp:exp_grasp}
In this experiment, we show that our method can learn to achieve stable grasps and that static grasp reference data is inherently bound to fail in a dynamic setting. We first train with labels from DexYCB \cite{chao2021dexycb} and further demonstrate that our approach also works with, and improves upon, labels obtained from a state-of-the-art grasp synthesis method \cite{jiang2021hand}, on both the DexYCB and HO3D object sets. Lastly, we present results using an image-based hand pose estimator on HO3D images and labels from ContactOpt \cite{grady2021contactopt}. 

\input{\dir/dgrasp/sections/tables/table_exp1_new}

\begin{figure}[h!]
\begin{center}
   \includegraphics[width=1.0\textwidth]{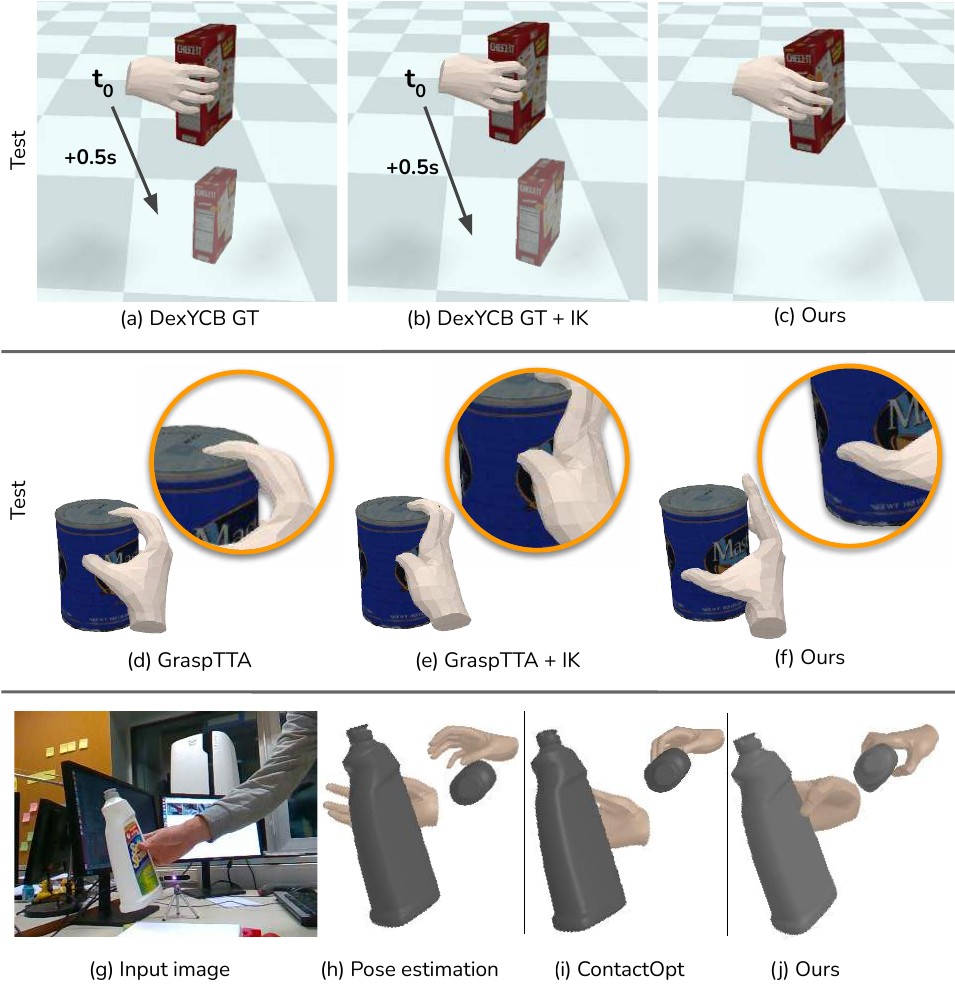}

\end{center}
   \caption{\textbf{Qualitative evaluation}. (a)-(c): static grasp labels often do not lead to stable grasps when evaluated in a physics simulation (a-b), which can be successfully corrected by our method (c). (d)-(f): showcases artifacts such as interpenetration when using a state-of-the-art grasp synthesis method \cite{jiang2021hand} (d-e). Our method (f) can correct such cases and generate physically-plausible grasps. (g)-(j): using images (g) to estimate an initial grasp (h). Physically implausible poses occur even with corrections via offline optimization (i), which can be corrected by our method (j).\\}
\label{fig:dgrasp:grasp_qual}
\end{figure}

We present quantitative evaluations in \tabref{tab:dgrasp:exp1_train} and qualitative results in \figref{fig:dgrasp:grasp_qual} and \figref{fig:dgrasp:app_qual_grasps}. Compared to the baselines, our method is able to achieve significantly better performance on all the metrics. Importantly, the grasping policy can improve the success rate, while minimizing interpenetration (an important metric in the grasp synthesis literature).  We note that our method achieves 0 interpenetration loss when evaluated in the physics simulation. In \tabref{tab:dgrasp:exp1_train}, however, we report interpenetration on the original MANO hand model and detailed object meshes.
For computational efficiency during training, the hand model and the object meshes are simplified in the physics simulation (see \secref{sec:dgrasp:simulation_setup}), limiting the performance of our model when evaluated in the original setting with regards to interpenetration. The offline optimization with IK slightly increases success rates in the motion capture setting, however, we found no improvement with IK for the generated (SYN) or image-based (IMG) experiments and hence omit it from the results. The improved performance in the image setup compared to other settings is due to the high-quality grasp references from \cite{grady2021contactopt}, which already optimizes for contact. In general, there is a performance drop when moving to unseen test labels. We also find that our approach may struggle with thin objects which are difficult to grasp on a surface, such as the scissor or marker objects.

\paragraph{Generalization to Unseen Objects}
To evaluate generalization performance on unseen objects, we train and test our model on six separate train/test splits with varying complexity. Each test set contains three objects from DexYCB. The remaining objects are used for training a policy. We average the results over all test sets and report the results in \tabref{tab:dgrasp:exp1_gen}. While there is room for improvement in overall success rate, our method outperforms the baseline in all metrics. 

\input{\dir/dgrasp/sections/tables/table_combined_gen_motion}

\subsection{Motion Synthesis}
\label{sec:dgrasp:exp_motion}
We now demonstrate our method's ability to synthesize motions with the grasped object in hand. The goal of this task is to grasp an object and generate a trajectory that brings the object to a target 6D pose. 
We use a subset of representative YCB objects and create a test set with 100 randomly sampled, out-of-distribution poses $\mathbf{T}_g$.
We compare against a standard RL baseline (Flat-RL) and a variant of our method that only \emph{maintains} the pose instead of actively grasping the object (Ours+static grasp). We also compare against a learning-based motion synthesis policy (Ours+learned policy).
As shown in \tabref{tab:dgrasp:motion}, the two-stage separation in our method is crucial for solving the task. Moreover, the decrease in performance when the hand pose is simply maintained (Ours+static grasp) solidifies the contribution of our approach. This implies that active control of the hand throughout the sequence is mandatory to maintain a stable grasp. Lastly, our method outperforms the learning-based variant (Ours+learned policy) of our motion synthesis module by a large margin on both metrics. 

\subsection{Ablations}
\label{sec:dgrasp:exp_ablation}
Here we analyze different components of our method and show that they are crucial for achieving stable grasps. To this end, we ablate our method with different feature spaces and reward functions. We select a subset of representative objects and evaluate on our train-split of DexYCB  (\secref{sec:dgrasp:exp_data}).
To validate our feature extraction layer and in particular the goal space (\secref{sec:dgrasp:method_feat_grasp}), we compare to a variant of our approach using the original state space (w/o FeatLayer) and a variant without the goal space (w/o GoalSpace). Furthermore, we evaluate our method without the contact reward (w/o ContactRew) and without the proposed wrist-guidance (w/o WristGuidance) as proposed in \secref{sec:dgrasp:wrist-guidance}. \Tab{ablation} shows that each component yields considerable performance improvement. We emphasize that the contact reward and a suitable feature representation are key for achieving stable grasps.

\input{\dir/dgrasp/sections/tables/table_ablation}

\subsection{Additional Qualitative Grasping Results}
\begin{figure}
\begin{center}
  \includegraphics[width=0.9\columnwidth]{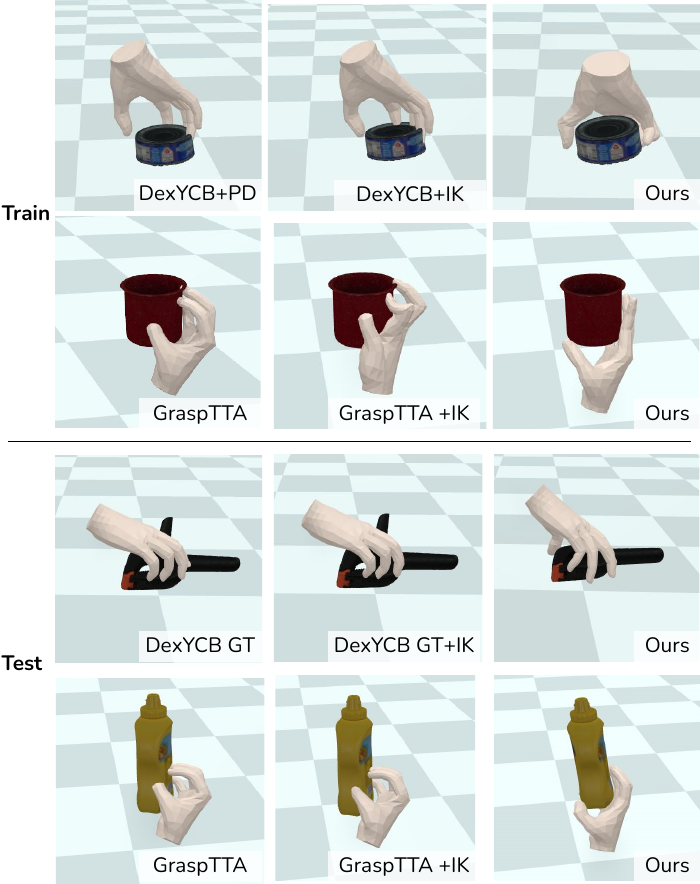}
\end{center}
  \caption{\textbf{Additional Qualitative Grasps.} We provide additional qualitative examples of grasps. Rows 1-2: Comparison of the grasps on the training-sets of DexYCB \cite{chao2021dexycb} and the generated grasps from \cite{jiang2021hand}. Rows 3-4: Comparison on the test-sets of DexYCB \cite{chao2021dexycb} and the generated grasps from \cite{jiang2021hand}. As shown, our method produces more physically plausible grasps, i.e., with less interpenetration and more realistic contacts than the baselines.}
\label{fig:long}
\label{fig:dgrasp:app_qual_grasps}
\end{figure}

We provide additional qualitative results in \figref{fig:dgrasp:app_qual_grasps}. Specifically, we include examples on the training sets of DexYCB \cite{chao2021dexycb} and the grasps generated with GraspTTA \cite{jiang2021hand}. Additionally, we present additional examples for both test-sets. As can be observed, the baselines suffer from interpenetration and grasps that cannot achieve force-closure. In contrast, our method can correct for interpenetration and achieve more realistic grasps. For more qualitative results, please see the supplementary videos on our project page\footnote{\href{https://eth-ait.github.io/d-grasp/}{eth-ait.github.io/d-grasp}}.

%% file: chapters/04_hoi_generation/dgrasp/sections/tables/table_exp1_new.tex
\begin{table*}[t]
	\centering
	\resizebox{1.0\textwidth}{!}{%
		\begin{tabular}{l|ll|ccc|ccc}
		
			\toprule
            & & \multirow{2}{*}{Models} & \multicolumn{3}{c}{Training set} & \multicolumn{3}{c}{Test set} \\
			& &  & \scriptsize{Success$\uparrow$} &  \scriptsize{SimDist [mm/s]$\downarrow$} &  \scriptsize{Interp. [$cm^3$]$\downarrow$} &  \scriptsize{Success$\uparrow$} &  \scriptsize{SimDist [mm/s]$\downarrow$} &  \scriptsize{Interp. [$cm^3$]$\downarrow$}  \\
			
			\midrule
            \parbox[t]{2mm}{\multirow{5}{*}{\rotatebox[origin=c]{90}{DexYCB}}}
			& \parbox[t]{2mm}{\multirow{3}{*}{\rotatebox[origin=c]{90}{\scriptsize{MC}}}}  
			& GT[1]+PD & 0.31 & $13.4 \pm 9.2$ & 4.59 & 0.35 & 13.1 $\pm 9.1$ & 4.41 \\

            & & \gc GT[1]+IK & \gc 0.39 & \gc $11.8 \pm 9.4$ & \gc 9.23 & \gc 0.50 & \gc $9.1 \pm 8.5 $ & \gc 9.74\\
            
            & & \textbf{Ours} & \textbf{0.70}  & $\mathbf{5.8 \pm 7.4}$  & \textbf{1.75} & \textbf{0.63}  & $\mathbf{8.0 \pm 8.1}$  & \textbf{1.77}\\

			\cmidrule{2-9}
			
		    & \parbox[t]{2mm}{\multirow{2}{*}{\rotatebox[origin=c]{90}{\scriptsize{SYN}}}} 
			& GraspTTA[2] +PD & 0.25 & $12.4 \pm 6.4$ &  4.92 & 0.24 & $12.7 \pm 6.5 $ &  4.94 \\

			& & \gc \textbf{Ours} & \gc \textbf{0.75} & \gc $\mathbf{3.9 \pm 7.2}$ & \gc \textbf{2.84} &  \gc \textbf{0.73} & \gc $\mathbf{4.6 \pm 6.7}$ & \gc\textbf{2.81} \\

			\midrule
			\parbox[t]{2mm}{\multirow{4}{*}{\rotatebox[origin=c]{90}{HO3D}}}
		    & \parbox[t]{2mm}{\multirow{2}{*}{\rotatebox[origin=c]{90}{\scriptsize{SYN}}}} 
			& GraspTTA[2] +PD & 0.31 & $ 10.0 \pm 6.6$ & 5.21 & 0.30 & $10.6 \pm 6.8 $ & 5.40\\

			& & \gc \textbf{Ours} & \gc \textbf{0.73} & \gc $\mathbf{4.4 \pm 7.4}$ & \gc \textbf{3.33} & \gc \textbf{0.71} & \gc $\mathbf{4.9 \pm 6.6}$ & \gc \textbf{3.40} \\

			\cmidrule{2-9}
			
		    & \parbox[t]{2mm}{\multirow{2}{*}{\rotatebox[origin=c]{90}{\scriptsize{IMG}}}} 
			& ContactOpt[3] +PD & 0.67 & $ 5.1 \pm 6.1$ & 14.94 & 0.60 & $6.5 \pm 5.8$ & 14.00 \\
			& & \gc \textbf{Ours} & \gc \textbf{0.88} & \gc $\mathbf{1.4 \pm 3.4}$ & \gc \textbf{2.67} & \gc \textbf{0.81}  & \gc $\mathbf{1.9 \pm 3.6}$ & \gc \textbf{2.08} \\

			\bottomrule
			
		\end{tabular}%
	}

	\caption{\textbf{Static grasp evaluation}. 
	We compare our model with grasp samples from the DexYCB dataset (MC), generated samples by a grasp synthesis method on the DexYCB and HO3D object sets (SYN), and samples extracted from an image-based hand pose estimator (IMG). We evaluate the baseline grasps in the simulation via PD-control (*+PD) directly or after de-noising via inverse kinematics (*+IK) for the motion capture data. We observe that our method outperforms the baselines in all metrics and conditions. The results indicate that static grasp references 1) will not lead to stable grasps when evaluated in a physics simulation and 2) suffer from interpenetration. Our method improves the interpenetration and learns stable grasps in a dynamic setting.
	}
	\label{tab:dgrasp:exp1_train}
\end{table*}{}

%% file: chapters/04_hoi_generation/dgrasp/sections/tables/table_combined_gen_motion.tex
\begin{table}[t]
    \centering
    \begin{tabular}{cc}
        \begin{subtable}[t]{0.47\textwidth}
            \centering
            \resizebox{0.9\linewidth}{!}{%
                \begin{tabular}{l|ccc}
                    \toprule
                    Models & Success $\uparrow$ & SD [mm/s] $\downarrow$ & IV [$cm^3$]$\downarrow$ \\
                    \midrule
                    GT+PD & 0.30 & $13.7 \pm 9.2$ & 4.41 \\
                    \gc GT+IK & \gc 0.38 & \gc $11.7 \pm 9.4$ & \gc 9.08 \\
                    \textbf{Ours} & \textbf{0.56} & $\mathbf{9.0 \pm 10.4}$ & \textbf{1.74} \\
                    \bottomrule
                \end{tabular}%
            }

            \caption{\textbf{Generalization}. We evaluate generalization to unseen objects and compare our model with the baselines. We create six different test sets of three objects each, which we leave out during training. We report the average performance over all test sets.}
            \label{tab:dgrasp:exp1_gen}
        \end{subtable} &
        \hfill
        \begin{subtable}[t]{0.47\textwidth}
            \centering
            \resizebox{0.85\linewidth}{!}{%
                \begin{tabular}{ll|cc}
                    \toprule
                    & Models & MPE [mm] $\downarrow$ & Geod. [rad.] $\downarrow$ \\
                    \midrule
                    & Flat-RL & 0.55 & 1.66 \\
                    & \gc Ours+static grasp & \gc 0.45 & \gc 1.46 \\
                    & Ours+learned policy & 0.30 & 0.92 \\
                    & \gc \textbf{Ours} & \gc \textbf{0.08}  & \gc \textbf{0.52}  \\
                    \bottomrule
                \end{tabular}%
            }
            \caption{\textbf{Evaluation of motion synthesis}. We compare our model with a standard RL baseline (Flat-RL) and different variants of our method. We observe that our hierarchical framework outperforms Flat-RL. Furthermore, an active grasping policy during motion synthesis is key to solving the task, as indicated by the performance drop for Ours+static grasp.}
            \label{tab:dgrasp:motion}
        \end{subtable}
    \end{tabular}
\end{table}

%% file: chapters/04_hoi_generation/dgrasp/sections/tables/table_ablation.tex
\begin{table}[t]
	\centering
	\resizebox{0.9\columnwidth}{!}{%
		\begin{tabular}{l|ccc}
		
			\toprule

			Models & Success $\uparrow$ & SimDist [mm/s] $\downarrow$ & Contact Ratio $\uparrow$  \\
			
			\midrule
            w/o ContactRew & 0.0 & $24.18 \pm 1.58$ & 0.02\\
            \rowcolor{Gray}
            w/o GoalSpace & 0.28 & $14.21 \pm 10.50$ & 0.18 \\
            w/o FeatLayer & 0.47 & $9.69 \pm 10.26$ & 0.21 \\
            \rowcolor{Gray}
			w/o WristGuidance & 0.58  & $7.88 \pm 10.57$ & 0.28 \\
            
            \textbf{Ours} & \textbf{0.89} & $\mathbf{4.83 \pm 1.71}$ & \textbf{0.43}\\
			
			\bottomrule
		\end{tabular}%
	}

	\caption{\textbf{Ablations}. We ablate our proposed components. All components together comprises our method. We observe that each component increases the performance significantly in all metrics.}
	\label{tab:ablation}
\end{table}{}

%% file: chapters/04_hoi_generation/dgrasp/sections/06_conclusion.tex
\section{Conclusion}
In this chapter we have made two distinct contributions. First, we have introduced the task of \textit{dynamic grasp synthesis} for human-object interactions. Second, to take a meaningful step into this direction, we have leveraged a physics simulation and reinforcement learning to generate sequences of hand-object interactions that are natural and physically plausible. By combining objectives that imitate a single frame grasp reference label with grasp stability rewards, we have ensured that the grasping is both human-like and stable.  In our experiments, we have demonstrated that our method can learn stable grasps and generate motions with the object-in hand without slipping. We have provided evidence that our method generalizes to unseen objects on a small scale experiment. While this proof of concept experiment indicates that our method works if a static hand pose reference for the unseen object is available, our method could be scaled to even larger object sets in the future. This indicates the potential of D-Grasp to enable the generation of 4D hand-object interactions for downstream applications. However, while we have focused on single-handed grasping of rigid objects, humans often interact with objects using both their hands, and some objects can be articulated. We will investigate this in the next chapter.

%% file: chapters/04_hoi_generation/artigrasp/artigrasp.tex
\chapter{ArtiGrasp: Synthesizing Physically Plausible Motion for Bi-Manual Dexterous Grasping and Articulation of Objects}
\chaptermark{Synthesizing Bi-Manual Grasping and Articulation of Objects}
\label{ch:hoi_generation:artigrasp}

\contribution{
In the previous chapter, we introduced the novel problem of synthesizing hand-object interactions of a single hand with a rigid object and developed a framework to solve it. In reality, humans often manipulate objects using both their hands, and many of these objects can be articulated. To accommodate such a setting, we propose ArtiGrasp, a method to generate bi-manual manipulation of articulated objects in a physically plausible manner. We introduce two key novelties over our previous framework. First, we add information about the articulation part of the object to the observation space and incentivize object articulation through the reward function. Second, to facilitate the training of the precise finger control required for articulation, we present a learning curriculum with increasing difficulty. In our experiments, we demonstrate that this enables the generation of human-like manipulations of objects, going beyond single-handed grasping and relocation. We show that our method can compose even longer sequences of multi-object manipulations, paving the way for more realistic human simulations in downstream applications such as human-robot interaction
}

\makeatletter \def\input@path{\dir/04_hoi_generation/artigrasp} \makeatother

\input{chapters/04_hoi_generation/artigrasp/sections/1_introduction}

\input{chapters/04_hoi_generation/artigrasp/sections/4_method}
\input{chapters/04_hoi_generation/artigrasp/sections/5_experiments}

\input{chapters/04_hoi_generation/artigrasp/sections/6_conclusion}

%% file: chapters/04_hoi_generation/artigrasp/sections/1_introduction.tex
\input{\dir/artigrasp/FIG/teaser}
\section{Introduction}
\label{sec:arti:introduction}

The ability to manipulate complex objects, such as operating a coffee machine, opening a laptop, or passing a box, is a fundamental part of everyday life. To generate such hand-object interactions, research has turned to synthesizing hand-object interactions using either data-driven~\cite{zhang2021manipnet,zheng2023cams} or physics-based methods (Chapter \ref{ch:hoi_generation:dgrasp}). While existing data-driven methods generate hand-object motions, including object articulation~\cite{zheng2023cams} and two-hand manipulation~\cite{zhang2021manipnet}, these methods typically rely on complete supervision from precise 3D motion data during training and require the object poses to be provided at inference time. 

In Chapter \ref{ch:hoi_generation:dgrasp}, we proposed D-Grasp, a physics-based methods that leverages reinforcement learning (RL) in a simulated environment. This approach reduces the data requirement for motion generation as they demand only a single hand pose reference per interaction. While this approach focuses on single-hand grasping motions for rigid objects, real world hand-object interactions are often bi-manual and include articulation. However, a framework for synthesizing bi-manual grasping and articulation of objects is still missing. 

In this chapter, we go beyond single-hand grasping interaction of rigid objects and present~\artigrasp, a novel method to synthesize dynamic bi-manual grasping and articulation of objects. Building on our previous work, we formulate this task as a reinforcement learning problem and leverage physical simulation. This allows our method to learn motions that adhere to physical plausibility, ensuring no object interpenetration occurs and that object articulation results from stable hand-object contacts and forces. We propose a general reward function and training scheme that enables grasping and articulation of a diverse set of objects without object- or task-specific retraining. 

Object articulation and bi-manual grasping present two key challenges compared to single-hand grasping. First, the articulation of different objects requires diverse wrist motions, making it challenging to define a general control strategy. For example, we show that a simple PD control scheme for relocation of objects after grasping, such as in D-Grasp, does not work well in this setting. To address this, we train an RL-based policy that learns to i) manipulate an object to a target articulation angle and ii) achieve natural interactions with the objects by utilizing only a single hand pose reference as input. The second key challenge is the precise finger control that is necessary to achieve successful articulation, where even small deviations from ideal positions on the target object impact performance. Furthermore, in the bi-manual manipulation setting, one hand can easily hinder the other hand from reaching its ideal position by moving the object. To address this challenge, we introduce a learning curriculum consisting of two phases. In the first phase, we fix the object base to the surface and create separate learning environments for each hand. This allows our policies to focus on learning precise finger control for articulation. In the second phase, we fine-tune the policies using non-fixed objects in a shared physics environment, allowing the hands to cooperate. 

In our experiments, we first assess both grasping and articulation separately, and then evaluate the \taskartigrasp~task, which involves transitioning an articulated object from its initial state into a target articulated object pose (see \figref{fig:arti:teaser}). To the best of our knowledge, there are no direct baselines for this task. Therefore, we adjust our previous work D-Grasp (Chapter \ref{ch:hoi_generation:dgrasp}) and show that simple adaptations lead to low task success rates. On the other hand, our method achieves performance gains of 5$\times$ over this baseline. We further demonstrate that our method can work with inputs from both motion capture data and noisy reconstructed poses from images with off-the-shelf hand-object pose estimation models. Finally, we ablate the main components of our framework.

In summary we contribute:
\begin{itemize}
    \item A method to achieve bi-manual manipulation and articulation of objects with simulated human hands.
    \item A general reward function to learn fine-grained wrist and finger control across different articulated objects.
    \item A learning curriculum with increasing difficulty to handle bi-manual manipulation of articulated objects
    \item Evaluations demonstrating that our method outperforms baselines and generates physically-plausible hand-object interactions.
\end{itemize}

%% file: chapters/04_hoi_generation/artigrasp/FIG/teaser.tex
\begin{figure}[H]
\hsize=\textwidth 
\centering
\includegraphics[width=0.95\textwidth]{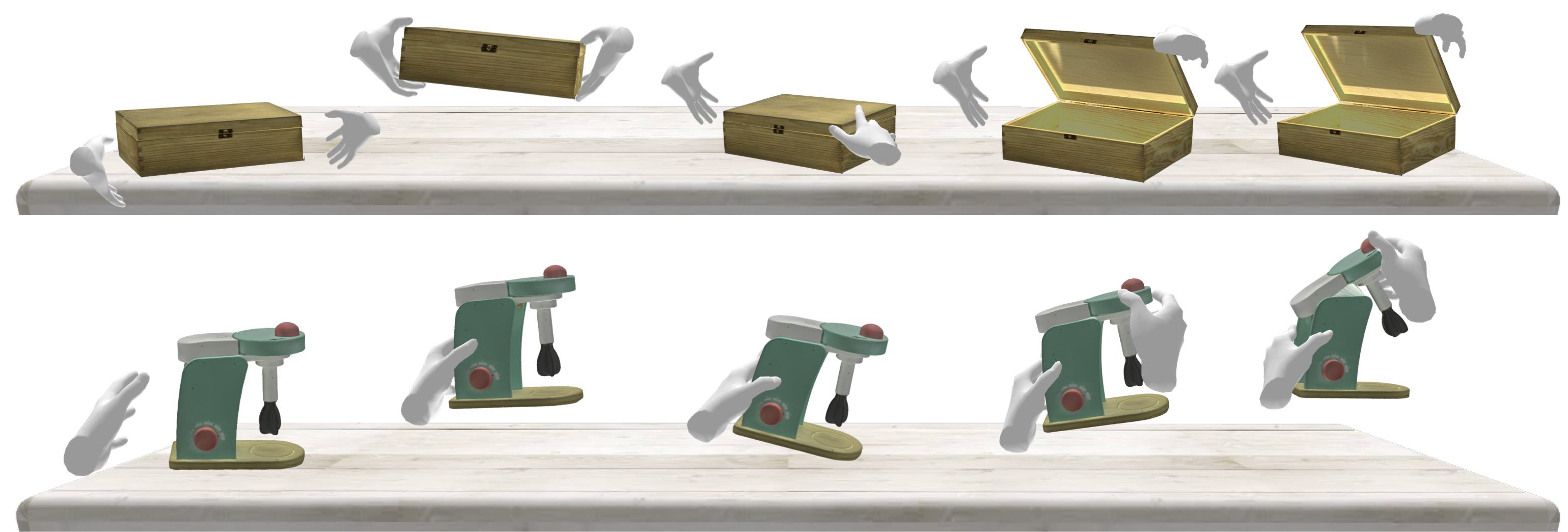} 
\caption{\textbf{Overview ArtiGrasp}. We present a method to synthesize physically plausible bi-manual manipulation of articulated objects. Our method can generate motion sequences such as grasping and relocating an object with one or two hands, and opening it to a target articulation angle.}
\label{fig:arti:teaser}
\end{figure}

%% file: chapters/04_hoi_generation/artigrasp/sections/4_method.tex
\begin{figure*}[h!]
  \includegraphics[width=1.0\textwidth]{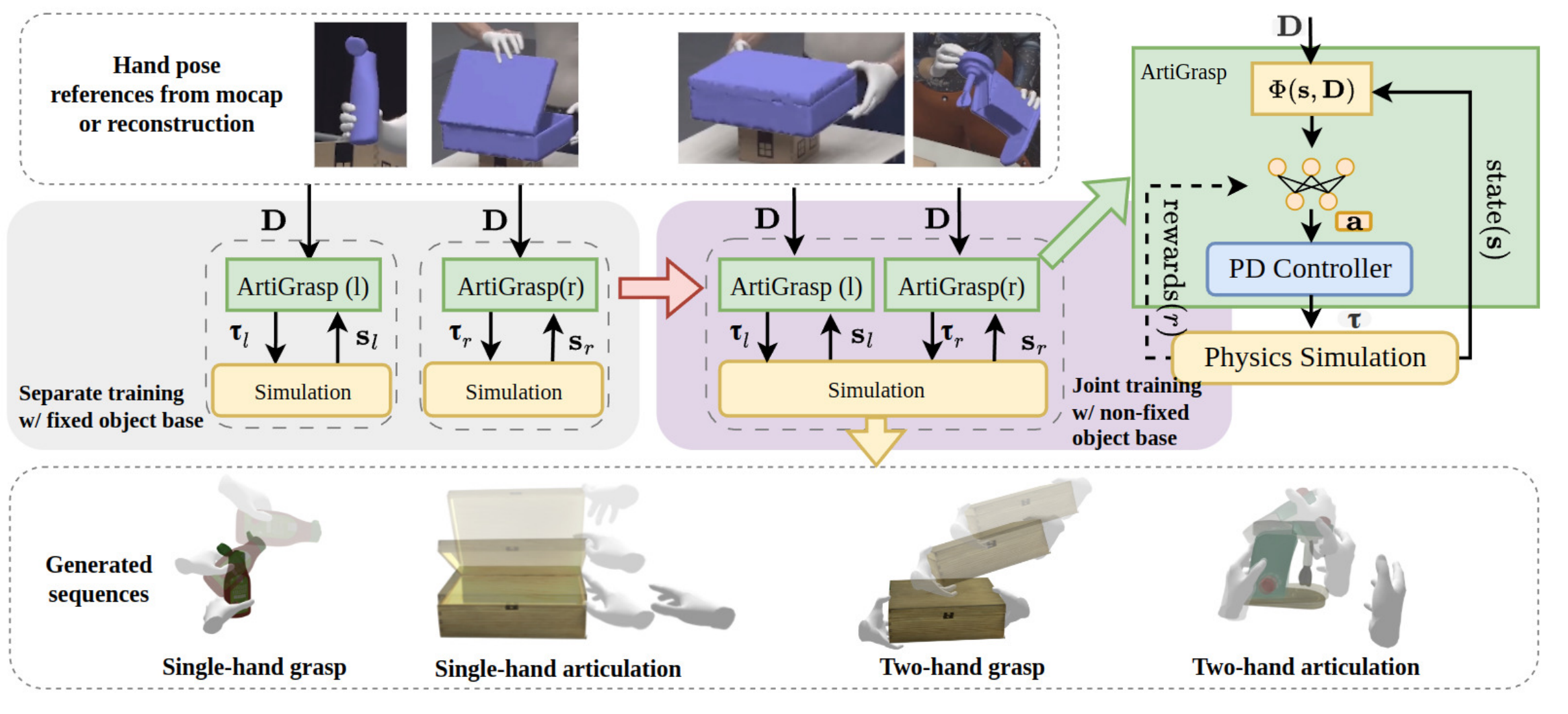}
  \centering
  \caption{\textbf{Grasping and Articulation Policy} Our method uses static hand pose references as input (top row) and generates dynamic sequences (bottom row, where higher transparency represents frames further along in time). We propose a curriculum that starts in a simplified setting with separate environments per hand and fixed-base objects (gray solid box on the left) and continues training in a shared environment with non-fixed object base (purple solid box in the middle). Our policies are trained using reinforcement learning and a physics simulation. Rewards are only used during training. The detailed structure of our policy is shown on the right.}
  \label{fig:artigrasp:structure}
\end{figure*}

\section{Task Definition}
\label{sec:arti:task}
The \taskartigrasp~task is illustrated in \figref{fig:arti:comparison}. We are given an articulated object that consist of two parts rotating about an axis $\V{q}_{\text{ax}}$, an initial articulated object pose $\boldsymbol{\Omega}^0$, a target articulated object pose $\overline{\boldsymbol{\Omega}}$, 
and two pairs of object-relative hand pose references $\V{D}$ (one for grasping and one for articulation). Our goal is to generate a sequence of one or two hands interacting with the object such that the initial object pose $\boldsymbol{\Omega}^0$ approaches the target pose $\overline{\boldsymbol{\Omega}}$. 
An articulated object pose $ \boldsymbol{\Omega}$ is defined by the 6 DoF global pose of the object base $\textbf{B}$ and the 1 DoF angle $\omega$ of its articulated joint. We define the output sequence as $\{(\mathbf{q}^t_l, \mathbf{T}^t_l, \mathbf{q}^t_r, \mathbf{T}^t_r, \boldsymbol{\Omega}^t)\}_{t=1}^{T}$, where $T$ is the number of time steps and $ \boldsymbol{\Omega}^t$ is the articulated object pose at time step $t$. The hand joint rotations and the global 6D hand pose are defined by $\V{q}_h^t$ and $\V{T}_h^t$ where $h \in \{l, r\}$. The hand pose references $\textbf{D}=(\overline{\textbf{q}}_l, \overline{\textbf{T}}_l, \overline{\textbf{q}}_r, \overline{\textbf{T}}_r)$ can be obtained from motion capture or grasp predictions~\cite{fan2023arctic} (see our experiments in \secref{sec:arti:experiment}).  

\section{Grasping and Articulation Policy}
\label{sec:arti:method}
We provide an overview of our policy learning framework in \figref{fig:artigrasp:structure}. Since we formulate the problem identically for both hands, we will omit the notation ``h'' for simplicity in this section.
 \artigrasp~is reinforcement learning based, and hence takes as input a state $\state$, provided by a physics simulation, and the hand pose reference $\textbf{D}$. A feature extraction layer $\Phi$ transforms these inputs and passes them to our policy network. We train a policy $\policyvec(\actions| \Phi(\textbf{s},\textbf{D}))$ for each hand. 
The policy predicts actions $\actions$ as PD-control targets, from which torques $\torques$ are computed. The torques are applied to our controllable hand model's joints in the physics simulation and the updated state is again fed to our feature extraction layer. In the physics simulation, we create controllable MANO hand models~\cite{romero2017embodied} with mean shape following \secref{sec:dgrasp:simulation_setup}. The models have 51 DoFs each and are represented by the local hand pose $\V{q} \in \mathbb{R}^{45}$ and global pose $\V{T} \in \mathbb{R}^{6}$. The objects are represented by meshes taken from the ARCTIC dataset~\cite{fan2023arctic}. Each mesh is split into a base and an articulation part with a single connecting joint. We now present details about RL, the feature extraction layer, the reward function, and our learning curriculum. 

\subsection{Feature Extraction}
The state $\state$ at a time step entails the current poses of the hands and object, as well as the contacts and forces per hand joint. We convert this information into features for the policy. Since we train a left-hand and a right-hand policy, the feature space is hand-specific, however, the overall structure is identical and defined as follows:

\begin{equation}
    \Phi(\textbf{s},\textbf{D}) = (\mathcal{H}, \mathcal{O}, \mathcal{G}), 
\end{equation}
\noindent where $\mathcal{H}$, $\mathcal{O}$, and $\mathcal{G}$ are the hand features, object features, and goal features, respectively. 

The hand features $\mathcal{H}$ are defined as $\mathcal{H}=(\textbf{q}, \dot{\textbf{q}}, \textbf{f}, \dot{\tilde{\textbf{T}}})$ where $\mathbf{q}$ and $\dot{\mathbf{q}}$ are the hand joints' local rotations and velocities, $\mathbf{f}$ are the net contact forces of each link of the hand, and $\dot{\tilde{\textbf{T}}}$ are the hand's linear and angular velocities in object-relative frame.

The object features are $\mathcal{O} = (\tilde{\boldsymbol{\Omega}}, \dot{\tilde{{\Omega}}}, \textbf{I}_{\text{art}})$. 
The terms $\tilde{{\Omega}}$ and $\dot{\tilde{\boldsymbol{\Omega}}}$ indicate the articulated object's 7 DoF pose and velocity expressed in wrist-relative frame. We convert global information into wrist-relative features (denoted by $\tilde{\cdot}$) to make the policy independent of the global state and prevent overfitting. 
To provide more information about the object's state with regards to articulation to our policy, we introduce the term $\textbf{I}_{\text{art}}=( \tilde{\textbf{q}}_{\text{ax}}, \tilde{\textbf{q}}_{\text{art}}, l_{\text{art}}, m_{\text{art}}, m_{\text{base}}, )$, where $\tilde{\textbf{q}}_{\text{ax}}$ and $\tilde{\textbf{q}}_{\text{art}}$ are the direction vector of the articulation axis and the direction vector from wrist to the axis, represented in wrist-relative frame. The terms $l_{\text{art}}$, $m_{\text{art}}$ and $m_{\text{base}}$ indicate the distance from wrist to the articulation axis and the weights of the object's parts, respectively. We ablate this component $\textbf{I}_{\text{art}}$ in \secref{sec:arti:ablation}. 

The goal features $\mathcal{G}$ guide the policy towards the hand pose reference and the target articulation angle. They are defined as: $\mathcal{G}=(\tilde{\textbf{g}}_q, \tilde{\textbf{g}}_x, \textbf{g}_c, \textbf{g}_a)$. In particular, $\tilde{\textbf{g}}_q = \overline{\textbf{q}}-\textbf{q}$ is the distance between the target and the current hand joint rotations (including wrist). The term $\tilde{\textbf{g}}_x = \overline{\textbf{x}}-\textbf{x}$ is the distance between the target and the current hand joint positions, which can be computed from the hand pose using forward kinematics. $\textbf{g}_c=[\overline{\textbf{c}}|\overline{\textbf{c}}-\textbf{c}]$ contains the target contacts and the difference between the target and the current binary contact vector. $\textbf{g}_a=\overline{\omega}-\omega$ is the difference between the the target and the current object articulation angle. The target position, pose, and contacts are extracted from the hand pose reference. The target articulation angle is set to zero for grasping and otherwise set to a random angle during training. The goal features are expressed in either the object's base or articulation coordinate frame, depending on the part that needs to be manipulated.

\subsection{Reward Function}
The individual time-step reward function should guide our policy towards a solution that imitates the reference pose and fulfills the task objectives at the same time. Therefore, we define it as follows:
\begin{equation}
    r = r_{\text{im}} + r_{\text{task}}, 
\end{equation}

\noindent where $r_{\text{im}}$ is the reward for imitating the reference pose and $r_{\text{task}}$ contains the task objective. The imitation reward is defined as:
\begin{equation}
    r_{\text{im}} = r_p + r_c + r_{\text{reg}}.
\end{equation}

The pose reward $r_p$ considers the joint position and joint angle error. The joint position error is the weighted sum of the distances between target and current positions $\overline{\textbf{x}}$ and $\textbf{x}$ of every joint, and the joint angle error measures the L2-norm of the differences between the target and current finger joint (and wrist) angles $\overline{\textbf{q}}$ and $\textbf{q}$:

\begin{equation}
    r_p = -\sum_{i=1}^Lw_{px}^i||\overline{\textbf{x}}^i-\textbf{x}^i||^2 - w_{pq}||\overline{\textbf{q}}-\textbf{q}||,
\end{equation}
\noindent where $w_{px}^i$ and $w_{pq}$ are weights for the respective terms.
The contact reward $r_c$ is composed of a relative contact term, which corresponds to the fraction of target contacts $\overline{\textbf{c}}$ the hand has achieved, and a contact impulse term, which encourages the amount of force $\textbf{f}$ applied on desired contact joints, capped by a factor proportional to the object's weight $m_o$:

\begin{equation}
    r_c = w_{cc}\frac{\overline{\textbf{c}}^T\textbf{I}_{f>0}}{\overline{\textbf{c}}^T\overline{\textbf{c}}} + w_{cf} \min(\overline{\textbf{c}}^T\textbf{f}, \lambda m_o),
\end{equation}
\noindent where $w_{cc}$ and $w_{cf}$ indicate the respective weights. The term $r_{\text{reg}}$ regularizes the linear and angular velocities of the hand and object:
\begin{equation}
    r_{\text{reg}} = -w_{rh}||\dot{\textbf{T}}||^2 - w_{ro}||\dot{\boldsymbol{\Omega}}||^2.
\end{equation}

\noindent The task reward $r_{\text{task}}$ consists of two incentives: manipulating the object to a target articulation angle and avoiding the movement of the object base from its initial pose:

\begin{equation}
    r_{\text{task}} = -w_{tq}||\overline{\omega}-\omega|| -w_{tx}||\posesix_o^0-\posesix_o||^2,
\end{equation}
where $\overline{\omega}$ and $\omega$ are the the target and the current articulation angle, $\posesix_o^0$ and $\posesix_o$ are the object's initial and current position. The weights $w_{tq}$ and $w_{tx}$ are used to balance the terms. 

\subsection{Curriculum}
\label{sec:arti:curriculum}
Training our policies with non-stationary objects from the beginning makes it difficult to learn the precise control necessary for fine-grained articulation. To address this, we introduce a learning curriculum that consists of two phases. In the first phase, we fix the objects to the table and train each hand separately in its own physics environment (grey shaded box in \figref{fig:artigrasp:structure}). This lets the policies learn precise finger movements and articulation. It also enables faster training, since the physics simulation speed scales roughly quadratically with the number of contacts in the environment. In the second phase, we move to the more complex setting where the object base is not fixed to the surface and the hands are both simulated in the same environment (purple shaded box in \figref{fig:artigrasp:structure}). In this setting, the policies need to learn to articulate the object without moving the object base or even tipping the whole object over. Additionally, the hands must collaborate, i.e., one hand should grasp the object without moving it too much, such that the other hand can successfully manipulate the object. In \secref{sec:arti:ablation}, we ablate the effectiveness of our curriculum. 

\section{Sequence Generation}
\label{sec:arti:sequence}
Given the unified policy per hand that can grasp and articulate objects, we now solve the \taskartigrasp~task (see \secref{sec:arti:task}) by combining the different subtasks. To achieve this, we use two pairs of hand pose references $\textbf{D}^{\text{grasp}}$ and $\textbf{D}^{\text{art}}$. In the first phase, the hand policies are executed until a stable grasp is reached. In this case, the target object articulation angle $\overline{\omega}$ is set to zero and $\textbf{D}^{\text{grasp}}$ is used as input. To move the object to its target 6D global pose, we use the policies to keep a stable grasp on the object and employ the motion synthesis module according to D-Grasp~\cite{christen:2022:dgrasp}. Note that in the case where the hand pose reference contains only single hand manipulation, we simply fix the other hand. After having relocated the object, we need to transition from grasping into pre-grasp poses for articulation. This is achieved through a heuristics-based control scheme. First, we release the grasps by bringing the fingers into open hand poses and moving them away from the object following the direction that points from the object center to the wrist. Next, we linearly interpolate a trajectory between the hand poses and pre-grasp poses for articulation $\textbf{D}^{\text{art}}$. The pre-grasp poses correspond to $\textbf{D}^{\text{art}}$ with a linear translation in global space. They are computed by setting them at a small distance away from direction of the object center to the wrist poses of the reference. In the last phase, we use our articulation policy to approach the object and articulate it to reach the target articulation angle.

%% file: chapters/04_hoi_generation/artigrasp/sections/5_experiments.tex
\section{Experiments}
\label{sec:arti:experiment}
We conduct several experiments to evaluate our framework. We first report experimental details in Section \ref{sec:arti:experiment_details}. We then conduct quantitative evaluations on grasping and articulation tasks in \secref{sec:arti:atomic} and \secref{sec:arti:recon}, including experiments with imperfect hand pose references from images. Finally, we demonstrate long motion sequence generation in Section \ref{sec:arti:long} and provide ablations to show the importance of our method's components in Section \ref{sec:arti:ablation}. For visual examples, please see the supplementary videos on our project page\footnote{\href{https://eth-ait.github.io/artigrasp/}{eth-ait.github.io/artigrasp}}.

\subsection{Experimental Details}
\label{sec:arti:experiment_details}

\paragraph{Implementation Details}
We use PPO (see \secref{bg:ppo}) for RL training and RaiSim~\cite{hwangbo2018per} for the physics simulation. We train all policies using a single Nvidia RTX 6000 GPU and 128 CPU cores. Training our method takes roughly three days. 

\paragraph{Dataset}
\label{sec:arti:dataset}
We use the ARCTIC dataset~\cite{fan2023arctic}, which contains fully annotated two-hand interaction sequences including dexterous grasping and manipulation of articulated objects. We separate all available sequences into the different interactions of grasping and articulation. For each interaction, we extract a single pair of hand pose references using heuristics. Since the ground-truth annotation for the test split is not released, we create a custom 65\%/35\% train/test-split over all sequences of the training and validation sets, with a total of 488 and 257 hand pose references for training and testing, respectively. For the experiments with image-based estimates, we only use the validation set consisting of 60 hand pose references, because the pose estimation model was trained on the ARCTIC training set (see \secref{sec:arti:recon}).

\paragraph{Metrics} 
\label{sec:arti:metrics} 
We mostly follow related work~\cite{jiang2021hand, christen:2022:dgrasp} and define three metrics for grasping (success rate, position and angle error), three metrics for articulation (success rate, simulated distance and angle error), and one additional metric for the \taskartigrasp~task. We omit the interpenetration metric since all of our baselines include a physics simulation which exhibits no interpenetration.

\noindent\textbf{Grasp Success Rate (Suc. G)}: A grasp is defined as success if the object is lifted higher than 0.1m and does not fall until the sequence terminates.

\noindent\textbf{Position Error (PE)}: The mean position error between the object's final and target 3D position in meters.

\noindent\textbf{Angle Error (AE)}: The mean angle error between the object's final and target base orientation measured as geodesic distance in radian.

\noindent\textbf{Articulation Success Rate (Suc. A)}: An articulation is defined as success if the hand can articulate the object for more than 0.3 rad and the articulated part does not slip until sequence termination.

\noindent\textbf{Articulation Angle Error (AAE)}: The mean error between the object final and target articulation angle in radian.

\noindent\textbf{Simulated Distance (SD)}: As articulation should not move the object base, we report average displacement of the object base in meters.

\noindent\textbf{Task Success Rate (Suc. T)}: We deem a task as success if the PE $< 0.05$m, the AE $<0.2$rad, and the AAE $<0.5$rad.

\paragraph{Baselines} 
\label{sec:arti:baseline}
We propose the following baselines:

\noindent\textbf{D-Grasp}: For grasping, we use vanilla D-Grasp and train the policies of the two hands directly with non-stationary objects. To compare D-Grasp to our method for articulation, we adjust the wrist control in D-Grasp. We first gradually increase the angle of the articulated joint and calculate the target 6D wrist pose with inverse kinematics by assuming that the wrist is fixed to the articulated part of the object. We then feed the wrist target pose to the PD controller.

\noindent\textbf{PD+IK}: We use the hand reference poses and set them as targets to the PD controller. The wrist for the articulation is controlled in the same way as in D-Grasp.

\pagebreak

\subsection{Evaluation}
\label{sec:arti:atomic}

\input{\dir/artigrasp/tables/composed}

\paragraph{Grasping and Articulation} For grasping, we pre-sample 30 6D target object poses randomly. To control the wrist movement for relocation after grasping the object with our method and the PD+IK baseline, we adopt the same motion synthesis module as in~\cite{christen:2022:dgrasp} (see \secref{sec:arti:sequence}). 
For articulation, we evaluate each hand pose reference on 5 target articulation angles: $\{0.5, 0.75, 1.0, 1.25, 1.5\}\ \text{rad}$. For both tasks, the initial hand poses are set at a pre-defined distance away in the direction that points from the object center to the wrist of the hand pose references, with partially opened hands. The quantitative results are shown in \tabref{tab:arti:grasp}. Our method significantly outperforms the PD+IK baseline on both grasping and articulation. Our policy has considerably better articulation performance and comparable grasping performance compared with D-Grasp. 

\begin{figure}[h!]
	\centering
 \includegraphics[width=0.95\textwidth]{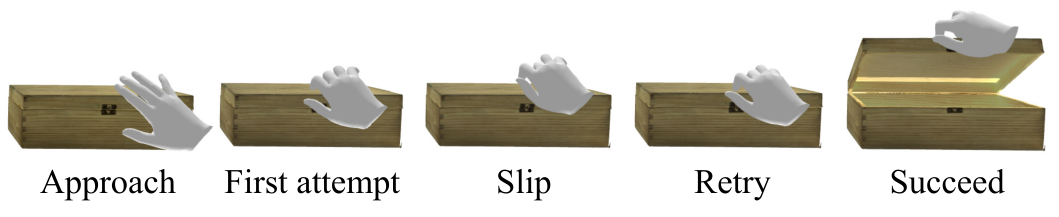}
  \vspace{-0.2cm}
\caption{\textbf{Recovery Behavior.} The hand shows some recovery ability from failure cases. Zoom in for details.}
\label{fig:arti:recovery}
\end{figure}

The results also show the difficulty of articulation and indicate that our learning-based wrist control is favorable for articulation compared to D-Grasp's non learning-based approach. Furthermore, as shown in the qualitative result \figref{fig:arti:recovery}, we observe some recovering behavior from failure cases, which indicates the robustness of our policy. In particular, the agent fails to grasp first but tries again to find a better grasp until it succeeds in articulating the object. Qualitative comparisons and more examples of generated sequences are presented in \figref{fig:arti:comparison_qual} and \figref{fig:arti:more_samples}. To evaluate the generalization ability of our framework, we conduct a proof-of-concept experiment with a single left-out object. The result indicates that our framework can generalize to an unseen object with about 15\% performance drop. We hypothesize that this is because the hand pose reference serves as a strong prior to the policy. However, more thorough evaluations need to be carried out once accurate 3D datasets with more articulated objects and hand-object poses become available.

\input{\dir/artigrasp/FIG/com_both}

\input{chapters/04_hoi_generation/artigrasp/FIG/more_qual}

\clearpage

\begin{figure*}[t!]
	\centering
 \includegraphics[width=1.0\textwidth]{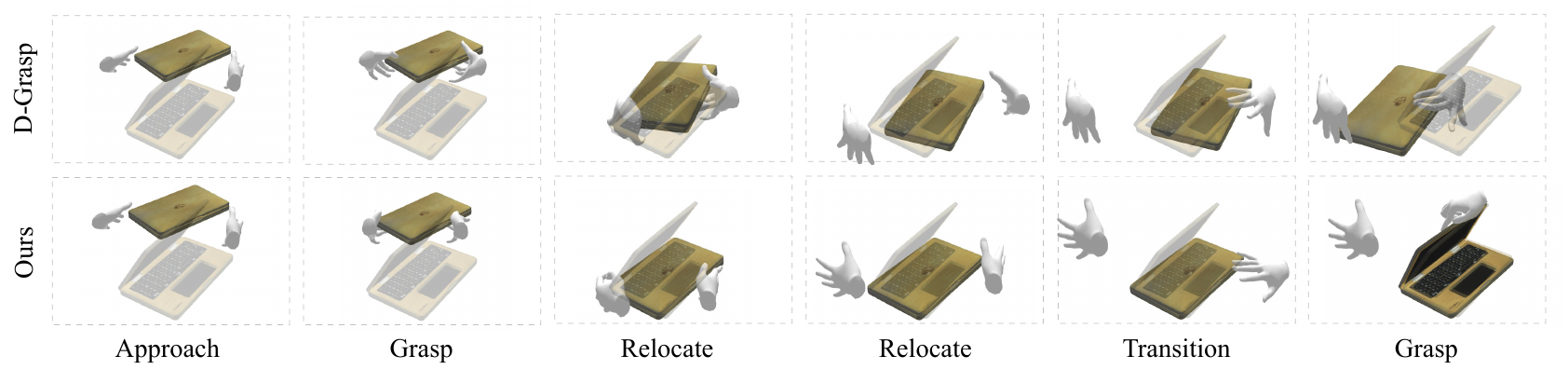}

\caption{\textbf{Qualitative evaluation of \taskartigrasp.} \dgrasp can grasp and relocate the object, but fails to articulate the object. Ours is more successful at tackling this task and can articulate the object after relocation.}
\label{fig:arti:comparison}
\end{figure*}

\begin{figure*}[t!]
	\centering
\includegraphics[width=1.0\textwidth]{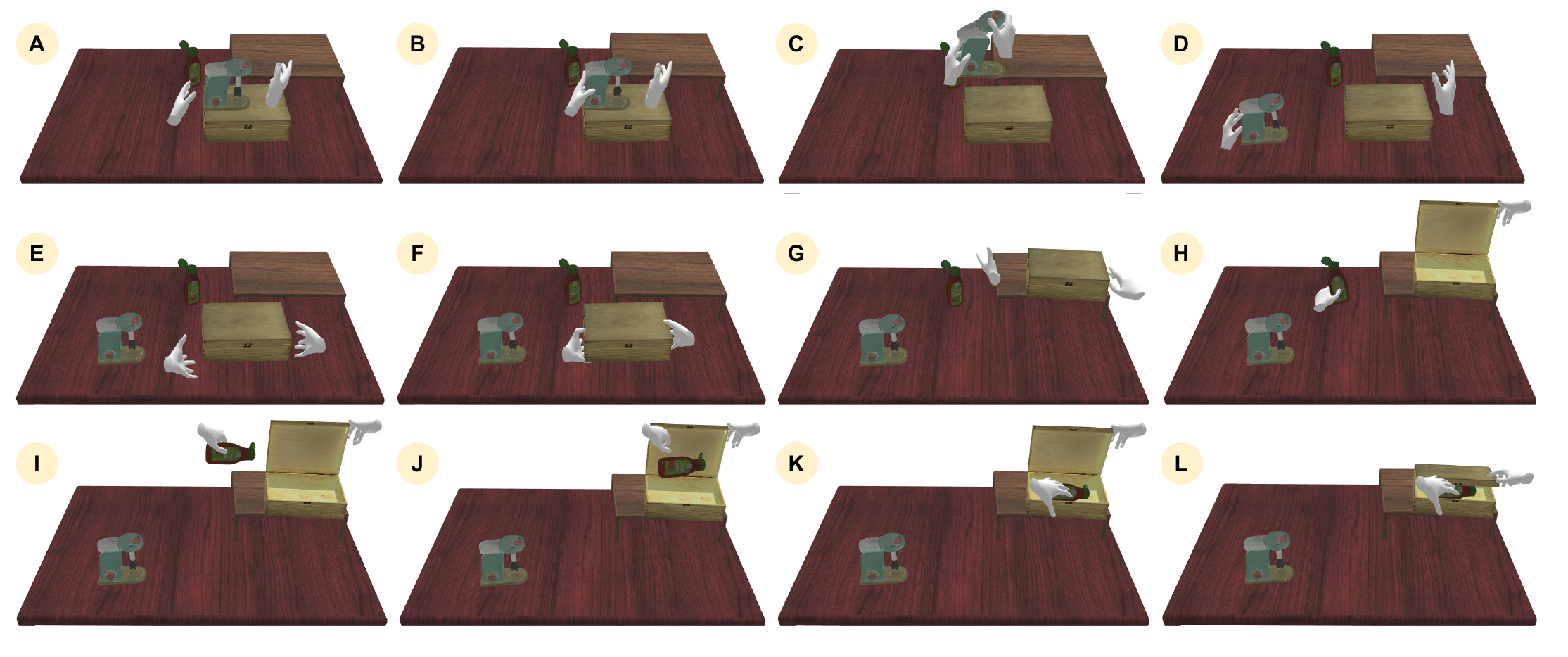}
\vspace{-6mm}
\caption{\textbf{Long sequence with multiple objects.} We show that our method can generate sequences of manipulating multiple objects. (A) Approaching the mixer with the left hand. (B) Grasping the mixer with the left hand. (C) Articulating the mixer with the right hand while the left hand is holding it. (D) Putting the mixer down on the table. (E) Approaching the box with both hands. (F) Grasping the box with both hands. (G) Relocating the box on the table and moving the left hand to the ketchup bottle. (H) Grasping the ketchup bottle with the left hand and opening the box with the right hand. (I) Relocating the ketchup bottle while the box is being held open. (J) Dropping the ketchup bottle into the box. (K) Moving the left hand away from the box. (L) Closing the box with the right hand.}

\label{fig:arti:long_sequence}

\end{figure*}

\paragraph{Relocation and Articulation Task} To evaluate this task (see \secref{sec:arti:task}), we combine all grasping hand pose references with all articulation hand pose references per object, and sample a random target articulated object 7 DoF pose per trial. This results in roughly 7500 evaluation trials in total. We report the average pose errors and the task success rate in \tabref{tab:arti:main_task}. 
Our method outperforms \dgrasp significantly in all metrics. For example, our method achieves a success rate of $5\times$  than that of D-Grasp. This shows that while D-Grasp can perform well in grasping when decoupled, it struggles in this composed task. We provide a qualitative comparison in \figref{fig:arti:comparison} and a demonstration of a longer sequence with multiple objects in  \figref{fig:arti:long_sequence}. 

\input{chapters/04_hoi_generation/artigrasp/tables/grasp_relocation}

\pagebreak

\begin{figure}[t!]

  \centering
    \includegraphics[width=0.9\linewidth]{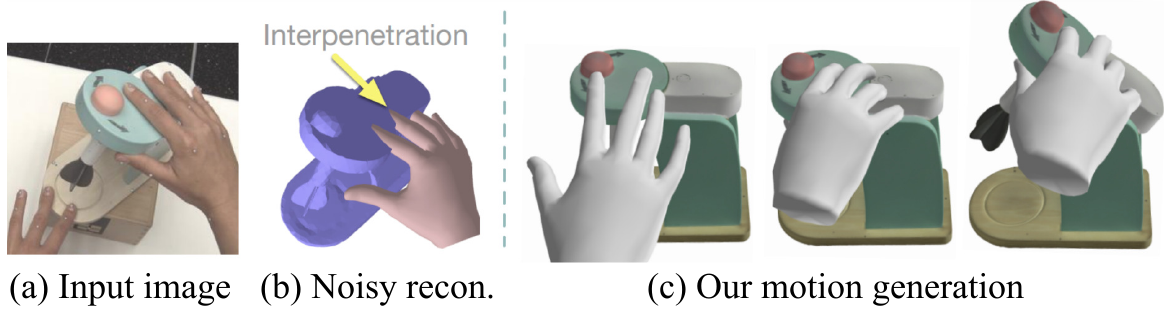}
  \vspace{-2mm}
  \caption{\textbf{Motion generation.} Our method synthesizes new motion sequences (c) with a noisy hand pose reference (b) reconstructed from a single RGB image (a).}
  \label{fig:arti:reco}

\end{figure}

\input{chapters/04_hoi_generation/artigrasp/tables/reconstructed}

\subsection{Generation with Reconstructed Hand Pose}
\label{sec:arti:recon}
We now evaluate our method with hand pose references obtained from image predictions via the off-the-shelf hand-object pose regressor from ARCTIC~\cite{fan2023arctic}. In particular, we estimate hand and object poses from images of the unseen validation subject in the ARCTIC and use the reconstructed results as grasp reference input to our method and baselines. We separate the evaluation into grasping and articulation and present the results in \tabref{tab:arti:recon}. Despite reconstruction noise such as hand-object interpenetration, our method can retain comparable performance as in the experiment with hand pose references from motion capture. This indicates the robustness of our method to prediction noise and its potential to synthesize new motions with hand-object pose references from single images.
An example of our generated motion is shown in \figref{fig:arti:reco}.

\pagebreak

\vspace{-2mm}
\subsection{Long Sequence with Multiple Objects}
\label{sec:arti:long}
Our method can generate long motion sequences in environments with multiple objects, which is shown in \figref{fig:arti:long_sequence}. We use a heuristics-based planner to compose the sequences. Specifically, we pre-determine the order of objects that should be interacted with along with the corresponding grasp references. After an interaction with an object is completed, we control the hand to move into a mean pose at a pre-defined distance away from the object. From this mean pose, we first move the hand into a pre-grasp pose close to the next object, before applying our ArtiGrasp policy. See \secref{sec:arti:sequence} for more details. Learning a high-level planning module to connect the different phases is an interesting direction to explore in the future. Note that while we propose a controlled setting to evaluate the \taskartigrasp~task, the order of manipulations can also be reversed. For example, an object can first be articulated, and then be moved to a different location. 

\input{\dir/artigrasp/tables/ablation}

\subsection{Ablations}
\label{sec:arti:ablation}
We ablate the impact of the introduced components on our framework. We compare our full method against i) training both hands cooperatively and with a non-stationary object from the start of training (\textit{w/o curriculum}) ii) training the hands separately and with a fixed-base object (\textit{w/o cooperation}). Additionally, we train our method without the articulation features $\textbf{I}_{art}$ (\textit{w/o art. features}, cf. \secref{sec:arti:method}). The results are presented in \tabref{tab:arti:ablation}. Without the curriculum, the policy achieves slightly better performance for grasping, but struggles with articulation. This is because grasping has different wrist motion than articulation which is easier to learn, which indicates the importance of a controlled setting to learn fine-grained articulation first. When training the hands separately without cooperation, grasping performance decreases because the hands cannot learn to collaborate for two-handed grasping. Lastly, the articulation features $\textbf{I}_{art}$ improve all articulation metrics, indicating that it provides important information about the object to the policy. 

\pagebreak

\input{\dir/artigrasp/FIG/qual_and_fail}

\subsection{Unnatural Poses}
As shown in \figref{fig:arti:unnatural}, our method can generate unnatural poses, which we argue occurs because of noisy pose references from ARCTIC~\cite{fan2023arctic}. We find that especially the pinky finger is often poorly labeled in the data, which translates to our policies. Developing hand pose priors to incentivize natural poses could be one way to mitigate this issue.

%% file: chapters/04_hoi_generation/artigrasp/tables/composed.tex
\begin{table}[t]
    \centering
        \centering
        \resizebox{0.9\linewidth}{!}{
            \begin{tabular}{l|ccc|ccc}
                \toprule
                & \multicolumn{3}{c|}{Grasping} & \multicolumn{3}{c}{Articulation} \\
                \midrule
                Model & Suc. G $\uparrow$ & PE $\downarrow$ & AE $\downarrow$ & Suc. A $\uparrow$ & AAE $\downarrow$ & SD $\downarrow$ \\
                \midrule
                PD+IK & 0.13 & 1.20 & 1.50 & 0.28 & 0.80 & 0.39 \\
                \rowcolor{Gray}
                D-Grasp & \textbf{0.72} & \textbf{0.12} & \textbf{0.62} & 0.22 & 0.93 & 0.49 \\
                Ours & 0.71 & 0.13 & 0.69 & \textbf{0.55} & \textbf{0.57} & \textbf{0.01} \\
                \bottomrule
            \end{tabular}
        }
        \caption{\textbf{Quantitative comparison for grasping and articulation tasks.} When the tasks are decoupled, our method outperforms the baselines on articulation and performs comparably to D-Grasp on grasping.}
        \label{tab:arti:grasp}
\end{table}






%% file: chapters/04_hoi_generation/artigrasp/FIG/com_both.tex
\begin{figure}[h!]
    \centering
    \begin{subfigure}[b]{0.49\textwidth}
        \centering
        \begin{subfigure}{1.0\linewidth}
            \centering
            {\includegraphics[width=0.3\textwidth]{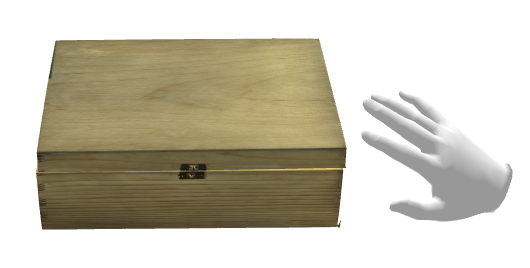} 
            \includegraphics[width=0.3\textwidth]{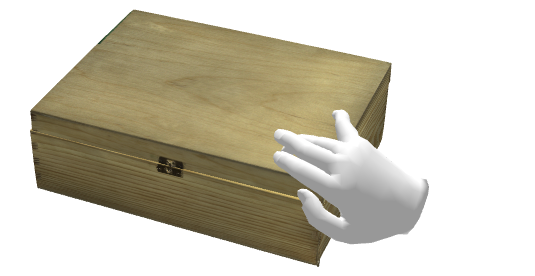}
            \includegraphics[width=0.3\textwidth]{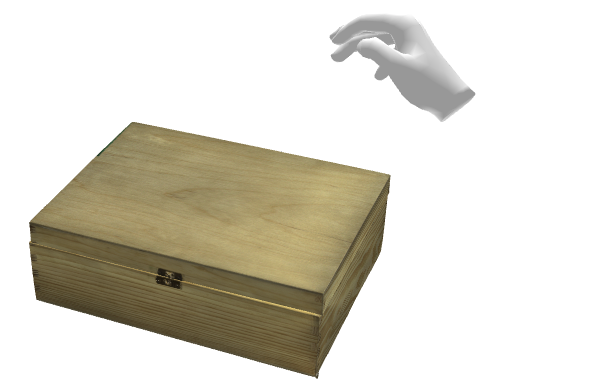}}
            \caption*{PD+IK}
        \end{subfigure}
        \hfill
        \begin{subfigure}{1.0\linewidth}
            \centering
            {\includegraphics[width=0.3\textwidth]{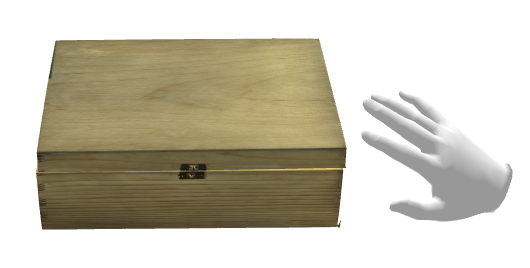} 
            \includegraphics[width=0.3\textwidth]{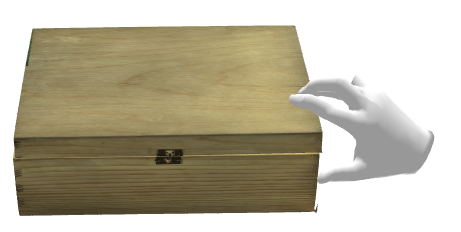}
            \includegraphics[width=0.3\textwidth]{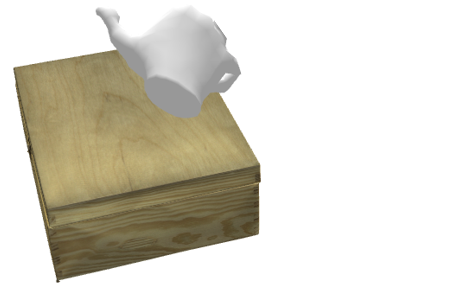}}
            \caption*{D-Grasp}
        \end{subfigure}
        \hfill
        \begin{subfigure}{1.0\linewidth}
            \centering
            {\includegraphics[width=0.3\textwidth]{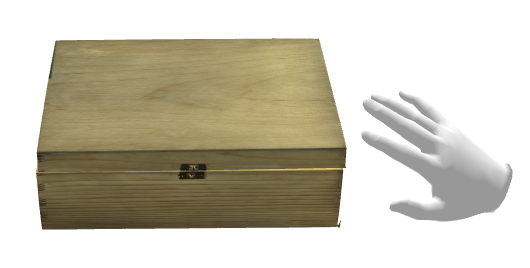} 
            \includegraphics[width=0.3\textwidth]{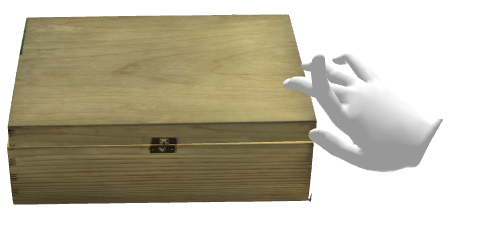}
            \includegraphics[width=0.3\textwidth]{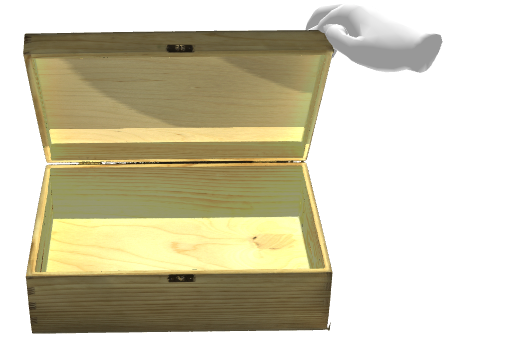}}
            \caption*{Ours}
        \end{subfigure}
        \caption{\textbf{Qualitative evaluation of articulation.} For articulation, both PD+IK and D-Grasp often fail at the task. Artigrasp articulates the object successfully.}
        \label{fig:arti:comparison_arti}
    \end{subfigure}
    \hfill
    \begin{subfigure}[b]{0.49\textwidth}
        \centering
        \begin{subfigure}{1.0\linewidth}
            \centering
            {\includegraphics[width=0.3\textwidth]{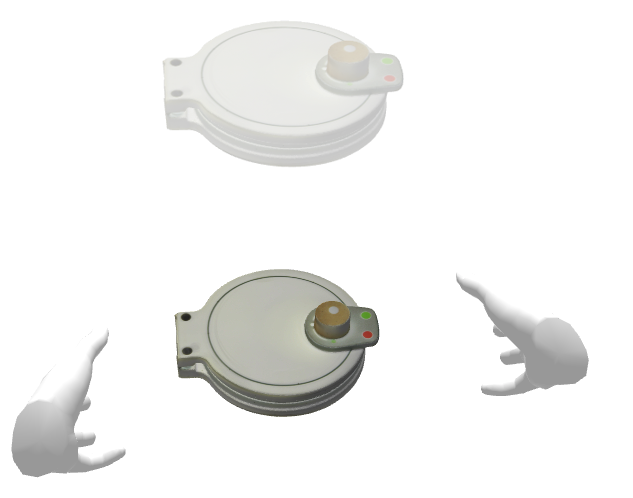}
            \includegraphics[width=0.3\textwidth]{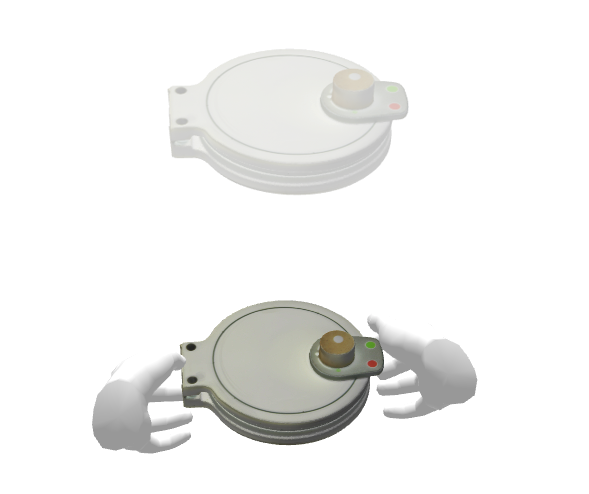}
            \includegraphics[width=0.3\textwidth]{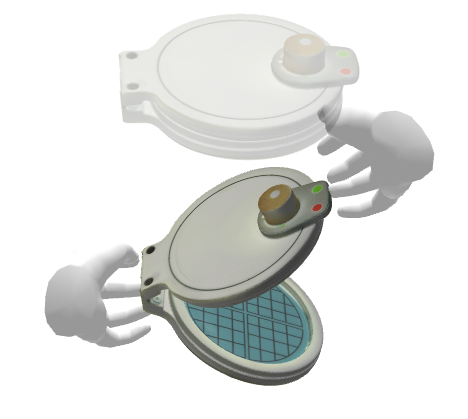}}
            \caption*{PD+IK}
        \end{subfigure}
        \hfill
        \begin{subfigure}{1.0\linewidth}
            \centering
            {\includegraphics[width=0.3\textwidth]{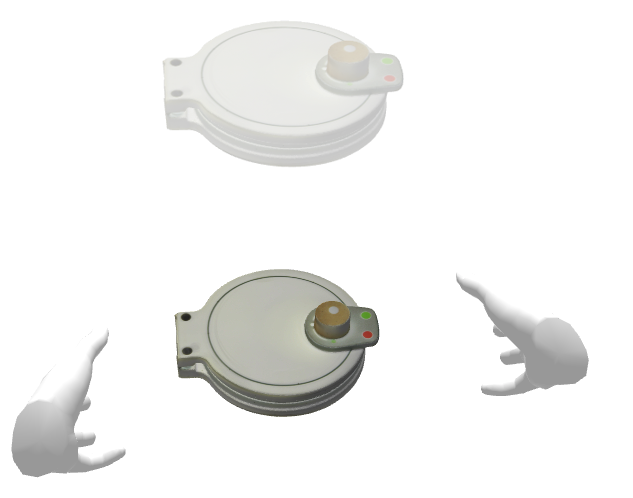} 
            \includegraphics[width=0.3\textwidth]{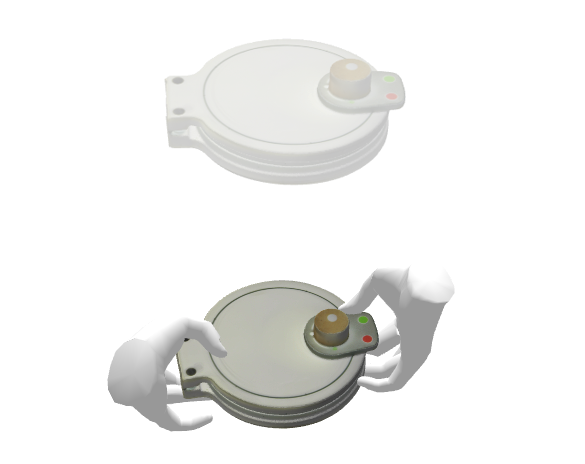}
            \includegraphics[width=0.3\textwidth]{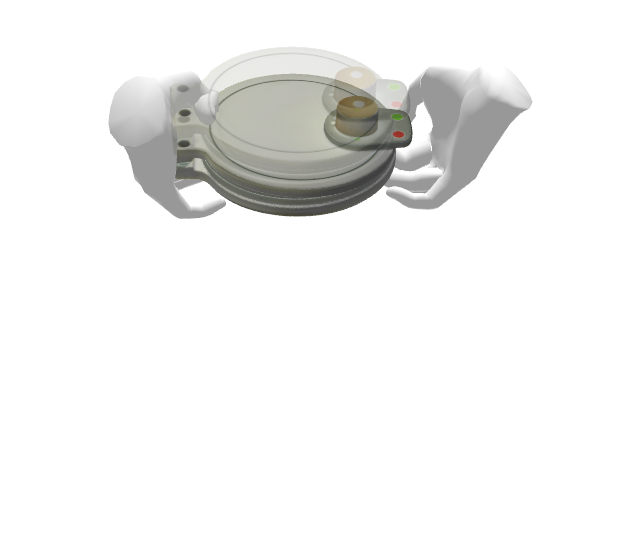}}
            \caption*{D-Grasp}
        \end{subfigure}
        \hfill
        \begin{subfigure}{1.0\linewidth}
            \centering
            {\includegraphics[width=0.3\textwidth]{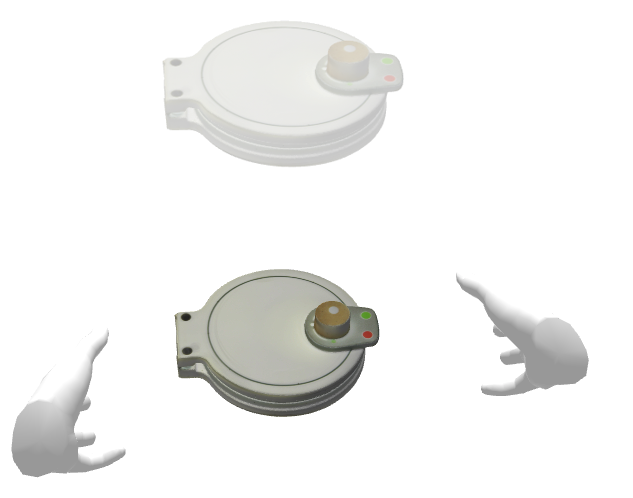}
            \includegraphics[width=0.3\textwidth]{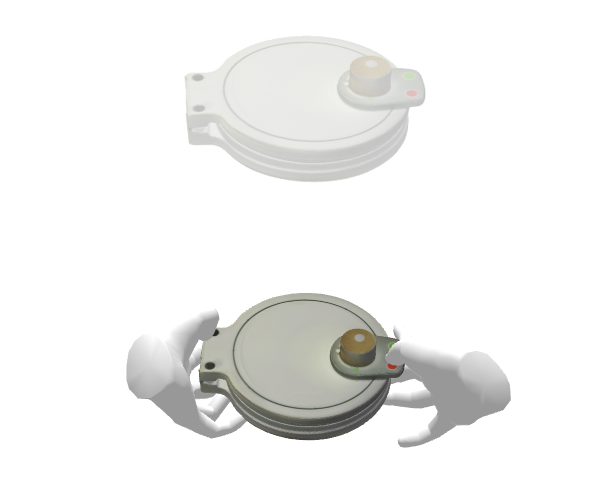}
            \includegraphics[width=0.3\textwidth]{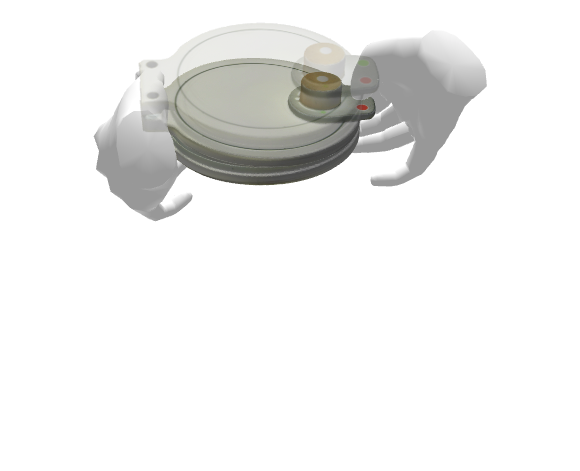}}
            \caption*{Ours}
        \end{subfigure}
        \caption{\textbf{Qualitative evaluation of grasping.} When evaluated on grasping, PD+IK often fails to successfully grasp the object. D-Grasp and ours succeed at the task.}
        \label{fig:arti:comparison_grasp}
    \end{subfigure}
    \caption{Comparison of articulation (left) and grasping (right) results.}
    \label{fig:arti:comparison_qual}
\end{figure}

%% file: chapters/04_hoi_generation/artigrasp/FIG/more_qual.tex
\begin{figure}[h!]
    \centering
    \begin{subfigure}[]{0.55\textwidth}
        \centering
        \includegraphics[width=\textwidth,trim=0 220 0 0,clip]{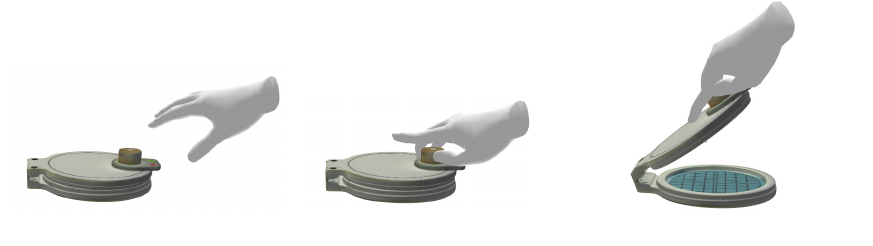}
    \end{subfigure}
    \hfill
    \begin{subfigure}[]{0.55\textwidth}
        \centering
        \includegraphics[width=\textwidth,trim=0 220 0 220,clip]{\dir/artigrasp/figures/seqs/individual_seqs/individual.pdf}
        \vspace{8mm}
    \end{subfigure}
    \caption{\textbf{Qualitative articulation result.} Our method articulating different objects.}
    \label{fig:arti:more_samples}
\end{figure}

%% file: chapters/04_hoi_generation/artigrasp/tables/grasp_relocation.tex
\begin{table}[t!]
    \centering
        \centering
        \resizebox{0.75\linewidth}{!}{
            \begin{tabular}{l|cccc}
                \toprule
                Models & Suc. T $\uparrow$ & PE $\downarrow$ & AE $\downarrow$ & AAE $\downarrow$ \\
                \midrule
                D-Grasp & 0.11 & 0.05 & 0.15 & 0.66 \\
                \rowcolor{Gray}
                Ours & \textbf{0.50} & \textbf{0.03} & \textbf{0.10} & \textbf{0.41} \\
                \bottomrule
            \end{tabular}
        }
        \vspace{1.5mm} 
        \caption{\textbf{Evaluation for our \textit{relocation and articulation} task.} ArtiGrasp outperforms D-Grasp on all metrics on the task of transitioning an articulated object into a target articulated object pose.}
        \label{tab:arti:main_task}
    

\end{table}

%% file: chapters/04_hoi_generation/artigrasp/tables/reconstructed.tex
\begin{table}[h!]
        \centering
        \resizebox{0.75\textwidth}{!}{
        \begin{tabular}{l|ccc|ccc}
           \toprule
           & \multicolumn{3}{c|}{Grasping} & \multicolumn{3}{c}{Articulation} \\
            \midrule
            Models & Suc.G $\uparrow$ & PE $\downarrow$ & AE $\downarrow$ & Suc.A $\uparrow$ & AAE $\downarrow$ & SD $\downarrow$ \\
           \midrule
            \dgrasp   & 0.60 & \textbf{0.16} & \textbf{0.78} & 0.20 & 1.07 & 0.63  \\
            \rowcolor{Gray}
            Ours    & \textbf{0.64} & \textbf{0.16} & 0.80 & \textbf{0.54} & \textbf{0.55} & \textbf{0.01}   \\
            \midrule
            Ours$^{*}$        & 0.67 & 0.14 & 0.95 & 0.54 & 0.53 & 0.01  \\ 
           \bottomrule
        \end{tabular}
        }
        \caption{\textbf{Evaluation with reconstructed hand pose references.} When evaluated with predictions from images, we observe a minor drop in performance for grasping and articulation compared to mocap data. However, the overall performance shows that our method can handle noisy estimates. The asterisk (*) denotes using hand pose references from mocap.}
        \label{tab:arti:recon}

\end{table}

%% file: chapters/04_hoi_generation/artigrasp/tables/ablation.tex
\begin{table*}[htbp]
    \centering
    \begin{minipage}{\textwidth}
        \centering
            \resizebox{\textwidth}{!}{
            \begin{tabular}{l|ccc|ccc}
               \toprule
                & \multicolumn{3}{c|}{Grasping} & \multicolumn{3}{c}{Articulation} \\
                \midrule
                Models & Suc. G $\uparrow$ & PE $\downarrow$ & AE $\downarrow$ & Suc. A $\uparrow$ & AAE $\downarrow$ & SD $\downarrow$ \\
               \midrule
                w/o curriculum     & \textbf{0.74} & \textbf{0.13} & \textbf{0.65} & 0.36 & 0.77 & 0.02  \\
                \rowcolor{Gray}
                w/o cooperation    & 0.21 & 0.32 & 1.43 & 0.48 & 0.65 & 0.02 \\
                w/o art. features  & 0.67 & 0.15 & 0.73 & 0.48 & 0.67 & 0.01 \\
                \midrule
                \rowcolor{Gray}
                Ours             & 0.71 & \textbf{0.13} & 0.69 & \textbf{0.55} & \textbf{0.57} & \textbf{0.01} \\
               \bottomrule
            \end{tabular}
            }
            \caption{\textbf{Ablations.} We ablate our curriculum, cooperative training, and the articulation features. All components are important aspects to achieve grasping and articulation with a single policy.}
            \label{tab:arti:ablation}
    \end{minipage}
\end{table*}

%% file: chapters/04_hoi_generation/artigrasp/FIG/qual_and_fail.tex
\begin{figure}[h!]
    \centering
    \begin{subfigure}[b]{0.49\textwidth}
        \centering
        \includegraphics[width=\textwidth]{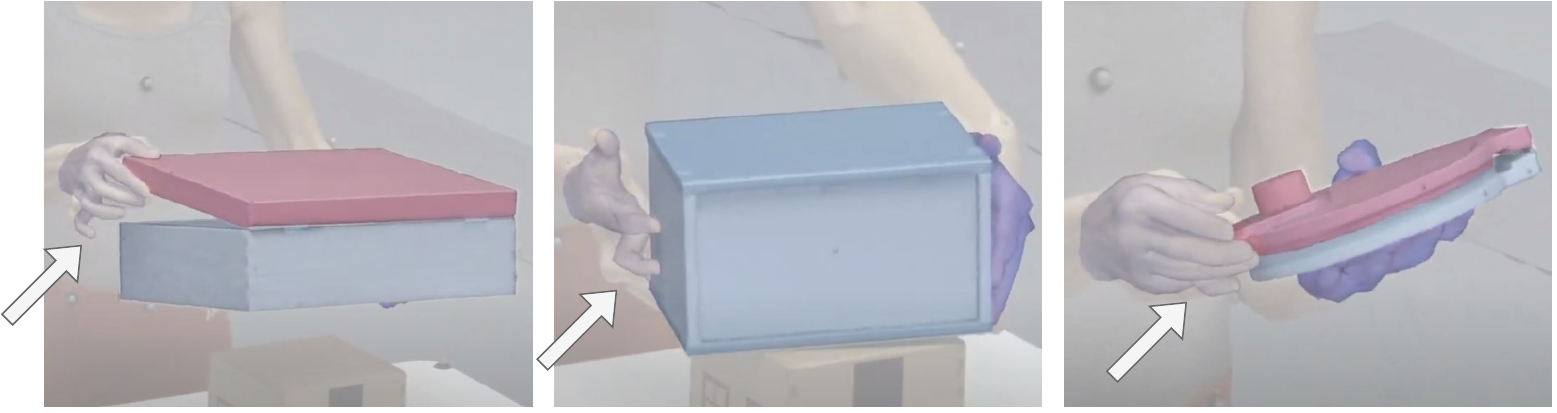}
        \caption{Unnatural hand pose references}
    \end{subfigure}
    \hfill
    \begin{subfigure}[b]{0.49\textwidth}
        \centering
        \includegraphics[width=\textwidth]{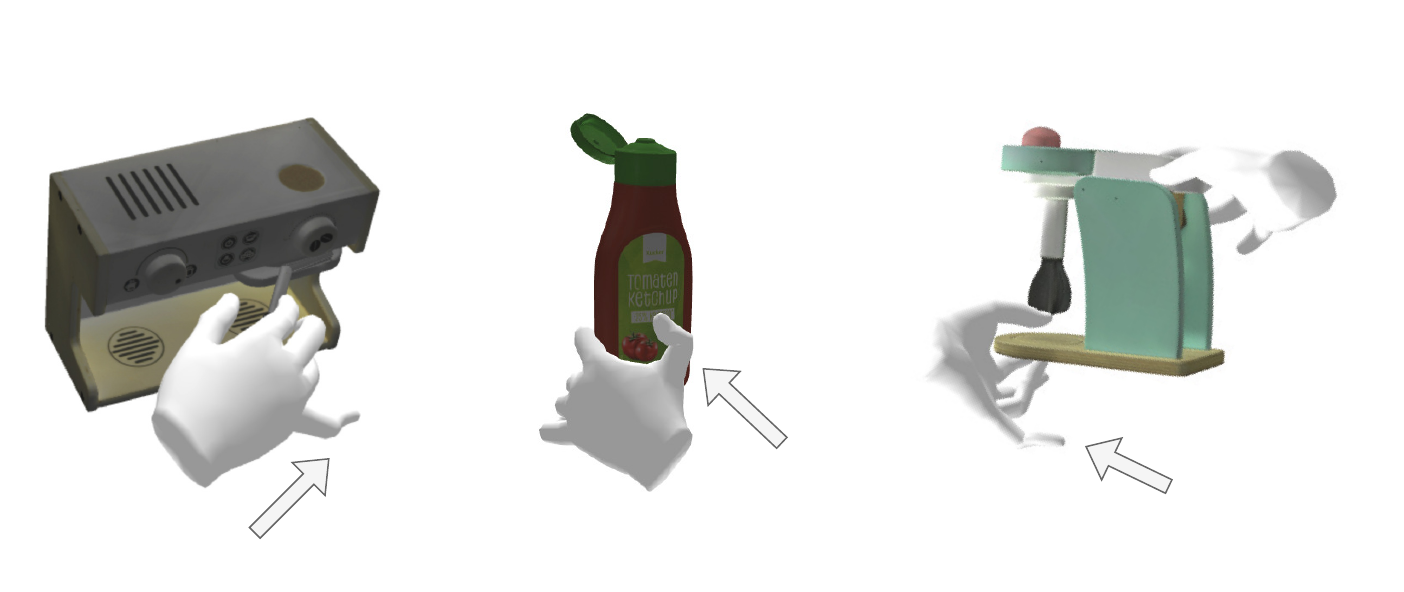}
        \caption{Unnatural generated hand poses}

    \end{subfigure}
    \caption{\textbf{Unnatural hand poses} (a) Some of the hand pose references we extract from the ARCTIC dataset contain unnatural hand poses. (b) Our method can output some unnatural hand poses, which can be due to noise in the hand pose references or because of the trade-off in the task objective.}
    \label{fig:arti:unnatural}
\end{figure}


        

%% file: chapters/04_hoi_generation/artigrasp/sections/6_conclusion.tex
\section{Conclusion}
In this chapter, we have presented a method to synthesize physically plausible bi-manual grasping and articulation of objects with a single policy. We have introduced an RL-based method that learns hand-object interactions in a physics simulation from static hand pose references. To address the difficulty in learning precise control for articulation, we have extracted articulation features and propose a curriculum with increasing task difficulty. We have shown that our method presents a first step towards the \taskartigrasp~task, where an object is moved from an initial pose into a target 7 DoF pose. Furthermore, we have demonstrated that noisy hand-object pose estimates obtained from individual RGB images can be used as input to our method. In a proof-of-concept we have shown that our policy has the potential to generalize to unseen objects, and better generalization may be achieved in the future with larger and more diverse grasp reference datasets. A limitation of our method is that it sometimes generates unnatural poses caused by noisy grasp references and the trade-off between the task and imitation reward, which is shown in \figref{fig:arti:unnatural}. This may be improved by integrating bio-mechanical constraints or hand pose priors obtained into our framework.

%% file: chapters/05_handover/content/conclusion.tex
\chapter{Summary \& Insights}
\label{ch:handover:conclusion}

\paragraph{Summary}
In this part, we introduced two works for human-to-robot handovers. In the Chapter \ref{ch:handovers:handoversim2real}, we presented the first learning-based framework for human-to-robot handovers from vision input. Our system consists of a simulated handover environment with human handover motions and uses a combination of imitation and reinforcement learning to train a robot policy. We employed a two-stage teacher-student training framework: the first stage assumes a stationary human and  gathers expert demonstrations via optimal grasp and motion planning, which are used for training a robot policy via behavior cloning.  In the second stage, where both the human and robot move simultaneously, grasp and motion planning is no longer feasible. Therefore, we fine-tuned the robot policy using reinforcement learning. By using point cloud representation in training, our policies transfer more effectively to real-world systems compared to using image-based inputs. 
In the Chapter \ref{ch:handovers:synh2r}, we addressed the limited amount of human motion data for training by adapting our hand-object interaction synthesis framework to human handover motion generation. This has allowed us to scale the amount of available training objects for handovers by 100x. To achieve this, we improved D-Grasp to robustly handle unseen objects and control the grasping approach direction, imitating human handover behavior. We trained our system solely on generated human handover motions and demonstrated performance comparable to a system trained with real human handover motions. Furthermore, a user study has revealed that real users cannot distinguish between the two system.

\paragraph{Implications} In the two presented works, we developed a novel methodology of training robotic policies for human-robot interaction that effectively transfer from simulation to a real-world system. Unlike previous approaches, our framework allows training policies with a human in the loop in simulation. We demonstrated that by creating a faithful simulation of human behavior, we can train robots to interact effectively with the real humans. This has several important implications. First, training in simulation is safe and provides an environment for both training and testing before transferring to real-world systems. Second, sim-to-real transfer for human-robot interaction can be achieved with an appropriate choice of representation, in our case segmented point clouds. Our models learns to distinguish between the desired area for grasping the objects and regions to avoid, such as the human hand. Lastly, collecting real human data is inherently expensive and limited in scale. In SynH2R, we demonstrated that a generative model can be used to bootstrap human motion data, addressing these limitations. Further improvements of generative human models will allow the exploration of even more complex human-robot interactions.

\paragraph{Limitations and Future Work}
While the proposed setting provides a convenient framework to test methods that encompass vision, control, and human-robot interaction in a controlled manner, there are limitations to our frameworks. We find that most human trajectories in HandoverSim have roughly the same length. Similarly, in SynH2R we focused on increasing the diversity in object shapes, while the generated human motions are of similar length. A future direction can therefore include exploring a wider variety of human motion behaviors. For example, in a realistic, interactive setting the human may be constantly moving, and the robot should only take objects from the human once it wants to hand them over. Anticipating the intent and future states of the human could provide a more natural system. We also noticed that humans start adapting to the robot once they learn how it behaves. Therefore, introducing a multi-agent training scheme where both the simulated human and the robot are trained jointly \cite{langerak2022marlui} is interesting. 

There are currently two bottlenecks that hinder the real system from achieving faster and better performance. First, the perception pipeline, which we adopted \cite{yang2021reactive}, suffers from slow processing times. Different modules of this pipline could be updated with faster and improved models, such as FastSAM \cite{zhao2023fast}. Second, the segmentation network was trained from third-person view. Hence, on the real system, we use a third-person view camera and transform the captured point cloud into the coordinate frame of the camera that is attached to the robot end-effector. Ideally, we could leverage the point cloud captured directly from the first-person view camera, with a segmentation network trained on such viewpoints. As we showed in our experiments, using third-person view cameras lead to lower success rates in handovers due to potential occlusions.

Lastly, our framework can potentially serve as a real world application framework to evaluate perception pipelines. We demonstrated that our policies achieve a certain robustness against sensor noise and perturbations. Hence, we can evaluate how well the vision pipeline or its modules perform when plugged into our framework. For example, we may test hand and object pose estimation estimation pipelines \cite{ziani2022tempclr,tse2022collaborative} or segmentation models \cite{fan2021learning} in such a manner.

%% file: chapters/05_handover/handover.tex
\def\dir{chapters/05_handover}

\input{\dir/handoversim2real/handoversim2real}
\input{\dir/synh2r/synh2r}
\input{\dir/content/conclusion}

%% file: chapters/05_handover/handoversim2real/handoversim2real.tex
\chapter{Learning Human-to-Robot Handovers from Point Clouds}
\label{ch:handovers:handoversim2real}

\contribution{
    In the previous part, we introduced two models to generate hand-object interactions in 4D, which enable potential use cases in training human-robot interactions. In this chapter, we explore the utilization of simulated human motions in the context of vision-based human-to-robot handovers, a critical task for human-robot interaction. While research in Embodied AI has made significant progress in training robot agents in simulated environments, interacting with humans remains challenging due to the difficulties of simulating humans. Fortunately, recent research has developed realistic simulated environments for evaluating human-to-robot handovers. Leveraging this result, we develop a learning environment and training procedure that enables training robotics policies with a simulated human in the loop. In empirical evaluation in simulation, we show that our approach outperforms baselines on the HandoverSim benchmark. We further evaluate sim-to-sim and sim-to-real transfer and find that our method transfers more robustly compared to baselines. A qualitative user study validates that participants prefer our system over the closest learning-based baseline.
}

\input{\dir/handoversim2real/sections/01_introduction.tex}

\input{\dir/handoversim2real/sections/03_background.tex}

\input{\dir/handoversim2real/sections/04_method.tex}

\input{\dir/handoversim2real/sections/05_experiments.tex}

\input{\dir/handoversim2real/sections/06_conclusion.tex}

%% file: chapters/05_handover/handoversim2real/sections/01_introduction.tex
\begin{figure*}[t]
\begin{center}
   \includegraphics[width=1.0\textwidth]{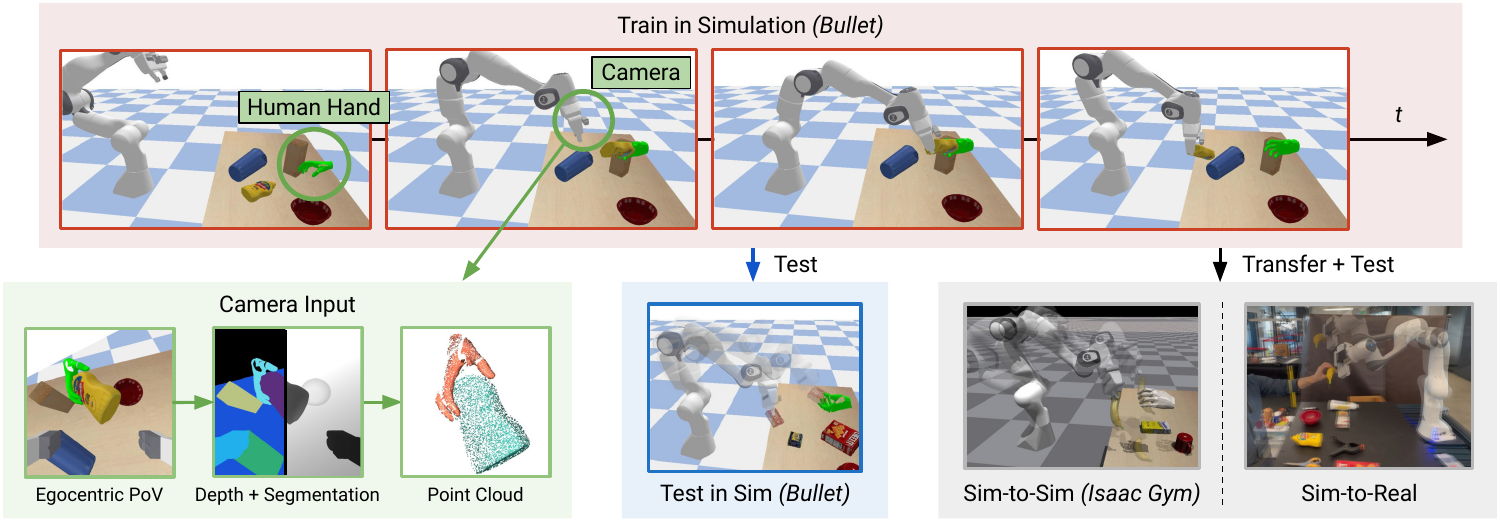}
\end{center}
\vspace{-6mm}
   \caption{\textbf{Overview.} We introduce a framework to learn human-to-robot handover policies from point cloud input. Our policies take input from a wrist mounted camera and directly generate action output for the robot's end effector. We train our policies in a simulated handover environment, and evaluate on unseen handover motion and poses. We further transfer the model across physics simulators and to a real robotic platform.}

\label{fig:h2r:teaser}
\end{figure*}

\section{Introduction}

Handing over objects between humans and robots is an important tasks for human-robot interaction (HRI)~\cite{ortenzi2021object}. It allows robots to assist humans in daily collaborative activities, such as helping to prepare a meal or exchanging tools with human collaborators in manufacturing. To complete these tasks successfully and safely, intricate coordination between the human and robot is required. This is challenging, because the robot has to react to human behavior, while only having access to sparse sensory inputs such as a single camera with limited field of view. Therefore, a need for methods that solve tasks such as handovers purely from vision input arises. 

Bootstrapping robot training in the real world can be unsafe and time-consuming. Therefore, recent trends in Embodied AI have focused on training agents to act and interact in simulated (sim) environments~\cite{deitke2020robothor,xiang2020sapien,shridhar2020alfred,srivastava2021behavior,szot2021habitat,gan2021threedworld,deitke2022retrospectives}. With advances in rendering and physics simulation, models have been trained to map raw sensory input to action output, and can even be directly transferred from simulation to the real world~\cite{anderson2021sim,shen2021igibson}. Many successes have been achieved particularly around the suite of tasks of robot navigation, manipulation, or a combination of both. In contrast to these areas, little progress has been made around tasks pertained to HRI. This is largely hindered by the challenges in embedding realistic human agents in these environments, since modeling and simulating realistic humans is challenging. 

Despite the challenges, an increasing number of works have attempted to embed realistic human agents in simulated environments~\cite{christen:2019:drlhs,erickson2020assistive,puig2021watch,perezd’arpino2021robot,pang2021towards,wang2021learning,chao2022handoversim}. Notably, a recent work has introduced a simulation environment (``HandoverSim'') for human-to-robot handover (H2R)~\cite{chao2022handoversim}. To ensure a realistic human handover motion, they use a large motion capture dataset~\cite{chao2021dexycb} to drive the movements of a virtual human in simulation. However, despite the great potential for training robots, the work of~\cite{chao2022handoversim} only evaluates off-the-shelf models from prior work, and has not explored any policy training with humans in the loop in their environment.

We aim to close this gap by introducing a vision-based learning framework for H2R handovers that is trained with a human in the loop (see \figref{fig:h2r:teaser}). In particular, we propose a novel mixed imitation learning (IL) and reinforcement learning (RL) based approach, trained by interacting with the humans in HandoverSim. Our approach draws inspiration from a recent method for learning polices for grasping static objects from point clouds~\cite{wang2021goal}, but proposes several key changes to address the challenges in H2R handovers. In contrast to static object grasping, where the policy only requires object information, we additionally encode human hand information in the policy's input. Also, compared to static grasping without a human, we explicitly take human collisions into account in the supervision of training. Finally, the key distinction between static object grasping and handovers is the dynamic nature of the hand and object during handover. To excel on the task, the robot needs to react to dynamic human behavior. Prior work typically relies on open-loop motion planners~\cite{wang2020manipulation} to generate expert demonstrations, which may result in suboptimal supervision for dynamic cases. To this end, we propose a two-stage training framework. In the first stage, we fix the humans to be stationary and train an RL policy that is partially guided by expert demonstrations obtained from a motion and grasp planner. In the second stage, we finetune the RL policy in the original dynamic setting where the human and robot move simultaneously. Instead of relying on a planner, we propose a self-supervision scheme, where the pre-trained RL policy serves as a teacher to the downstream policy.

We evaluate our method in three ``worlds'' (see \Fig{h2r:teaser}). First, we evaluate on the ``native'' test scenes in HandoverSim~\cite{chao2022handoversim}, which use the same backend physics simulator (Bullet~\cite{coumans20162021pybullet}) as training but unseen handover motions from the simulated humans. Next, we perform sim-to-sim evaluation on the test scenes implemented with a different physics simulator (Isaac Gym~\cite{makoviychuk2021isaac}). Lastly, we investigate sim-to-real transfer by evaluating polices on a real robotic system and demonstrate the benefits of our method. 

Im summary, we contribute:
\begin{itemize}
\vspace{-2mm}
    \item The first framework to train human-to-robot handover policies from vision input with a human in the loop.
    \vspace{-2mm}
    \item A novel teacher-student method that enables training of a jointly moving human and robot.
    \vspace{-2mm}
    \item An empirical evaluation showing that our approach outperforms baselines on the HandoverSim benchmark, sim-to-sim transfer, and sim-to-real transfer.
\end{itemize}

%% file: chapters/05_handover/handoversim2real/sections/03_background.tex
\section{Background}

\begin{figure*}[h]
\begin{center}
   \includegraphics[width=0.8\textwidth]{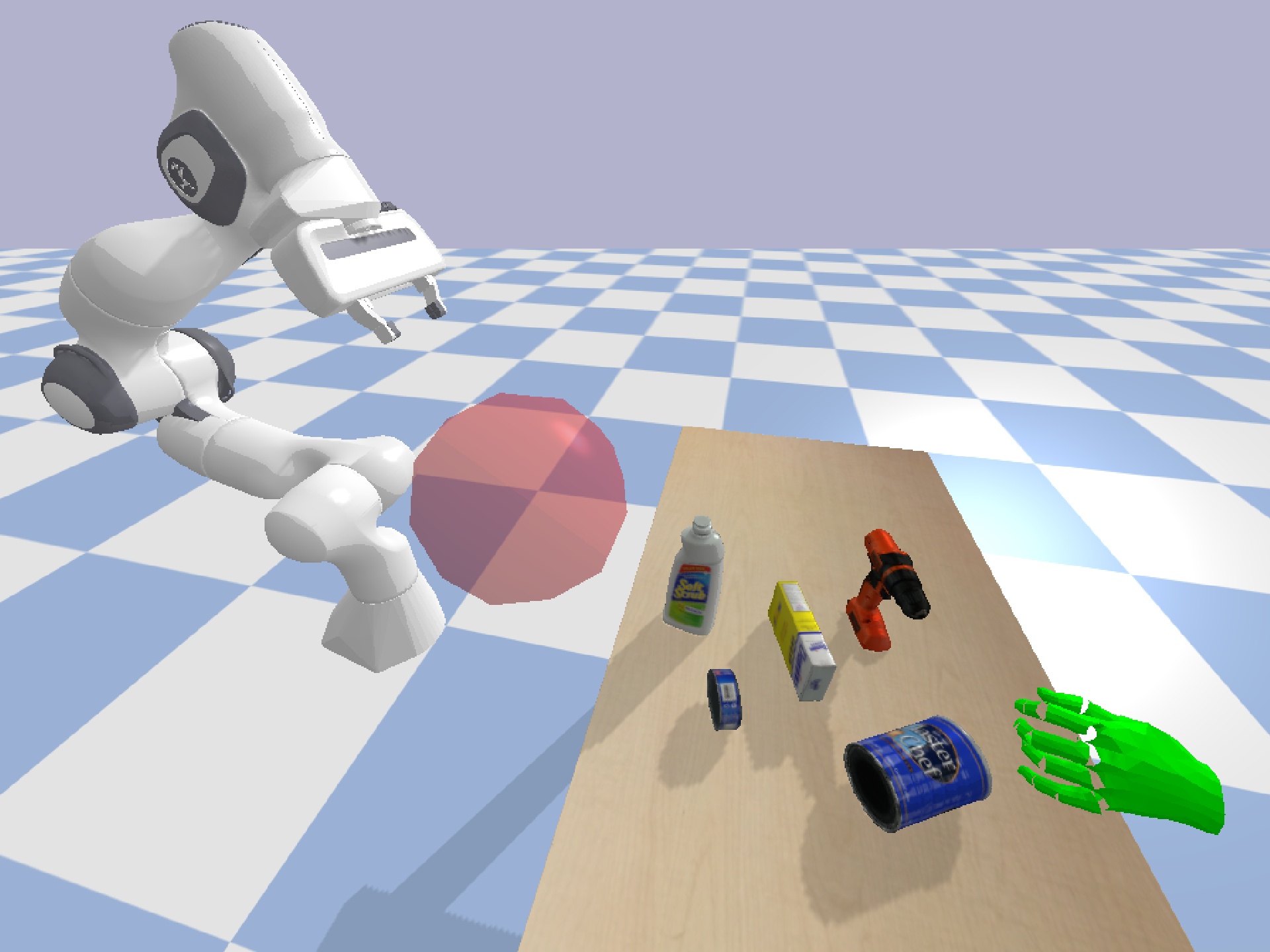}
\end{center}
   \caption{\textbf{HandoverSim Environment}. The red sphere indicates the goal region, the humand hand is colored in green.
   }

\label{fig:h2r:handoversim}
\end{figure*}

\subsection{HandoverSim Benchmark}
\label{sec:h2r:handoversim}
HandoverSim \cite{chao2022handoversim} is a benchmark for evaluating H2R handover policies in simulation. The task setting consists of a tabletop with different objects, a Panda 7DoF robotic arm with a gripper and a wrist-mounted RGB-D camera, and a simulated human hand (see \Fig{h2r:handoversim}). The task starts with the human grasping an object and moving it to a handover pose. The robot should move to the object and grasp it. The task is successful if the object has been grasped from the human without collision and brought to a designated position without dropping. To accurately model the human, trajectories from the DexYCB dataset \cite{chao2021dexycb}, which comprises a large amount of human-object interaction sequences, are replayed in simulation. Several baselines~\cite{wang2020manipulation,yang2021reactive,wang2021goal} are provided for comparison. However, the setup in HandoverSim has only been used for handover performance evaluation purposes, whereas in this work we utilize it as a learning environment.

%% file: chapters/05_handover/handoversim2real/sections/04_method.tex
\section{Method}

\begin{figure*}[t]
\begin{center}
   \includegraphics[width=1.0\textwidth]{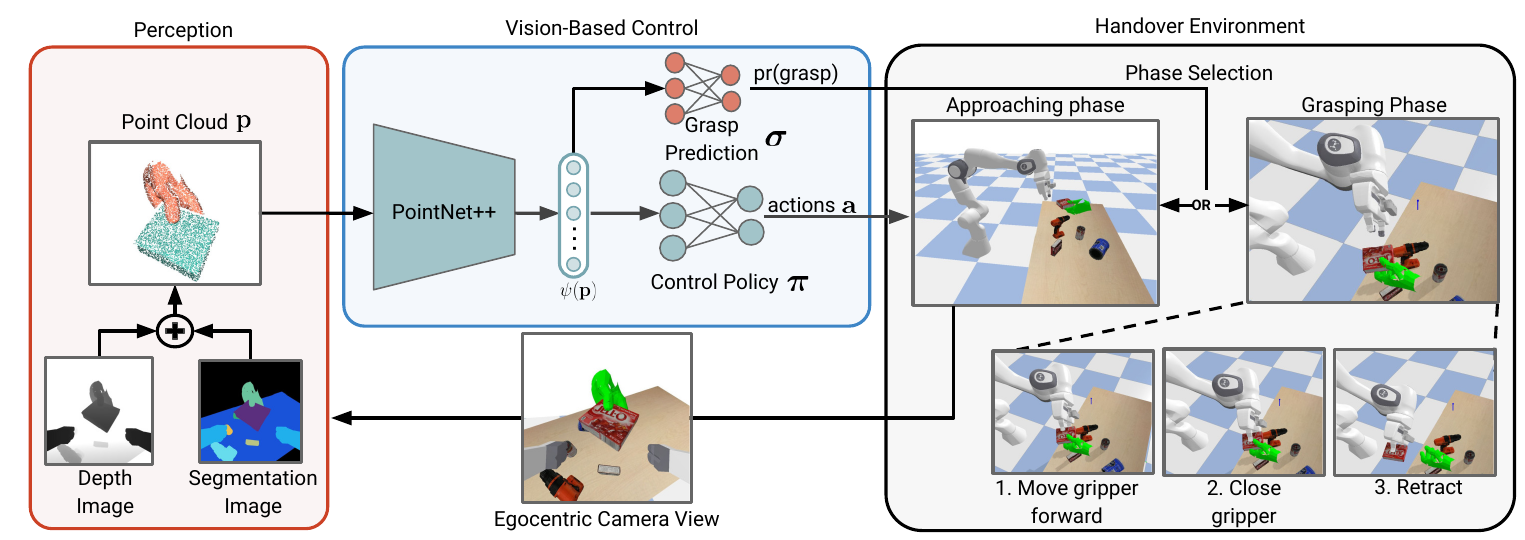}
\end{center}
   \caption{\textbf{Method Overview}. The \textbf{Perception} module takes egocentric RGB-D and segmentation images from the environment and outputs a hand/object segmented point cloud. Next, the segmented point cloud is passed to the the \textbf{Vision-based Control} module and processed by PointNet++\cite{qi2017pointnet++} to obtain a lower-dimensional representation. This embedding is used as input to both the control policy and the grasp predictor. Each task episode in the \textbf{Handover Environment} follows two phases: during the approaching phase, the robot moves towards a pre-grasp pose, driven by the control policy $\policy$ that outputs end-effector actions $\actions$. A learned grasp predictor  $\sigma$ monitors the motion and determines when the robot should switch into the grasping phase, which follows the steps: 1. moving the gripper forward from a pre-grasp to a grasping pose 2. closing the gripper 3. retracting the object to a designated location, after which the episode ends. }
\label{fig:h2r:method_overview}
\end{figure*}

The overall pipeline is depicted in \Fig{h2r:method_overview} and consists of three different modules: perception, vision-based control, and the handover environment. The perception module receives egocentric visual information from the handover environment and processes it into segmented point clouds. The vision-based control module receives the point clouds and predicts the next action for the robot and whether to approach or to grasp the object. This information is passed to the handover environment, which updates the robot state and sends the new visual information to the perception module. Note that the input to our method comes from the wrist-mounted camera, i.e., there is no explicit information, such as object or hand pose, provided to the agent. We will now explain each of the modules of our method in more detail.

\subsection{Handover Environment}
\label{sec:h2r:handover_env}
We split the handover task into two distinct phases (see \Fig{h2r:method_overview}).
First, during the \emph{approaching phase}, the robot moves to a pre-grasp pose that is close to the object by running the learned control policy $\policy$. A learned grasp predictor $\grasppred$ continuously computes a grasp probability to determine when the system can proceed to the second phase. Once the pre-grasp pose is reached and the grasp prediction is confident to take over the object from the human, the task will switch to the \emph{grasping phase}, in which the end-effector moves forward to the final grasp pose in open-loop fashion and closes the gripper to grasp the object. Finally, after object grasping, the robot follows a predetermined trajectory to retract to a base position and complete the episode. This task logic is used in both our simulation environment and the real robot deployment.
Sequencing based on a pre-grasp pose is widely used in literature for dynamic grasping~\cite{akinola2021dynamic}.

We follow the HandoverSim task setup \cite{chao2022handoversim}, where the human hand and objects are simulated by replaying data from the DexYCB dataset \cite{chao2021dexycb} (see \Sec{h2r:handoversim}).  
First, actions $\actions$ in the form of the next 6DoF end-effector pose (translation and rotation) are received from the policy $\policy(\actions| \state)$. We then convert the end-effector pose into a target robot configuration using inverse kinematics. Thereafter, we use PD-controllers to compute torques, which are applied to the robot. Finally, the visual information is rendered from the robot's wrist-mounted RGB-D camera and sent to the perception module. 

\subsection{Perception}
Our policy network takes a segmented hand and object point cloud as input. 
In the handover environment, we first render an egocentric RGB-D image from the wrist camera. Then we obtain the object point cloud $\pcvec_o$ and hand point cloud $\pcvec_h$ by overlaying the ground-truth segmentation mask with the RGB-D image.

Since the hand and object may not always be visible from the current egocentric view, we keep track of the last available point clouds. The latest available point clouds are then sent to the control module.

\subsection{Vision-Based Control}

\paragraph{Input Representation}
Depending on the amount of points contained in the hand point cloud $\pcvec_h$ and object point cloud $\pcvec_o$, we down- or upsample them into constant size. Next, we concatenate the two point clouds into a single point cloud $\pcvec$ and add two one-hot-encoded vectors to indicate the locations of object and hand points within $\pcvec$. We then encode the point cloud into a lower dimensional representation $\psi(\pcvec)$ by passing it through PointNet++ \cite{qi2017pointnet++}. Finally, the lower dimensional encoding $\psi(\pcvec)$ is passed on to the control policy $\policy$ and the grasp prediction network $\grasppred$. 

\paragraph{Control Policy} The policy network $\policy(\actions|\psi(\pcvec))$ is a small, two-layered MLP that takes the PointNet++ embedding as input state ($\state=\psi(\pcvec)$) and predicts actions $\actions$ that correspond to the change in 6DoF end-effector pose. These are passed on to the handover environment. 

\paragraph{Grasp Prediction}
We introduce a grasp prediction network $\grasppred(\psi(\pcvec))$ that predicts when the robot should switch from approaching to executing the grasping motion (cf. \Fig{h2r:method_overview}). We model grasp prediction as a binary classification task. The input corresponds to the PointNet++ embedding $\psi(\pcvec)$, which is fed through a 3-layered MLP. The output is a probability that indicates the likelihood of a successful grasp given the current point cloud feature. If the probability is above a tunable threshold, we execute an open-loop grasping motion. The model is trained offline with pre-grasp poses attained from \cite{eppner2020acronym}. We augment the dataset by adding random noise to pre-grasp poses. To determine the labels, we initialize the robot with the pre-grasp poses in the physics simulation and execute the forward grasping motion. The label is one if the grasp is successful, and zero otherwise. We use a binary cross-entropy loss for training.

\begin{figure*}[t!]
\begin{center}
   \includegraphics[width=1.0\textwidth]{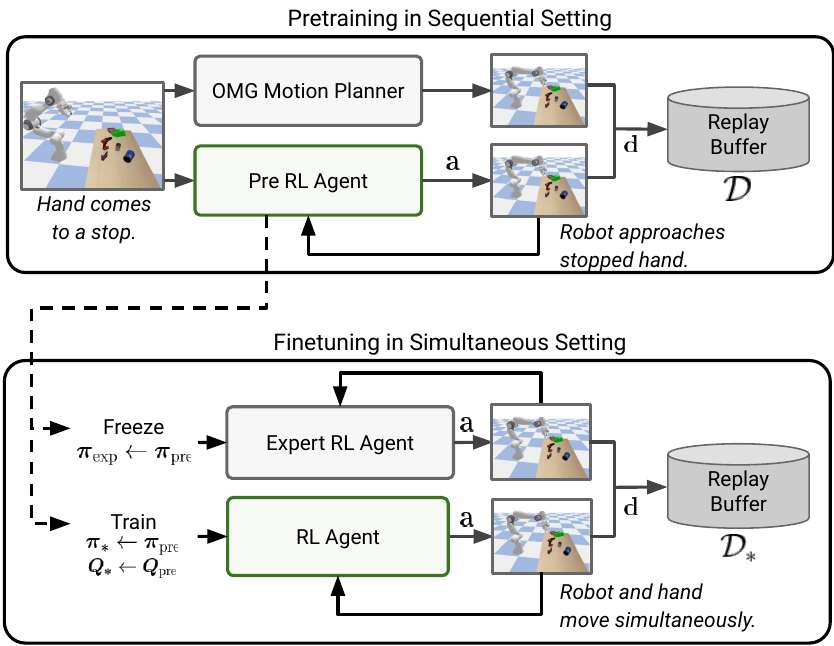}
\end{center}
   \caption{\textbf{Training Procedure}. In the \textbf{pretraining stage} (top box), the human hand is stationary. We alternate between collecting expert demonstrations via motion planning and exploration data with the reinforcement learning policy $\policypre$. In the \textbf{finetuning stage} (bottom box), the human and robot move concurrently. The expert motion planner is replaced by the expert policy $\policyexp$, which shares the weights of the pretrained policy $\policypre$. This policy network will be kept frozen for the rest of training and serves as a regularizer for the reinforcement learning agent. The reinforcement learning agent's actor network $\policyft$ and critic network $\criticft$ are also initialized with the weights of pretrained agent's networks, but the model will be updated during finetuning. In this stage, transitions are stored in a new replay buffer $\replay_*$. Data is sampled solely from this buffer during finetuning.
   }
\label{fig:h2r:training_procedure_left}
\vspace{-2mm}
\end{figure*}

\subsection{Two-Stage Teacher-Student Training}
\label{sec:h2r:training_procedure}
We aim at training a handover policy capable of moving simultaneously with the human. Training this policy directly in the setting of dynamic motion is challenging because expert demonstrations with open-loop planners to guide training can only be obtained when the human is stationary. A key contribution of our work is a two-stage training scheme for handovers that incrementally trains the policy to alleviate this challenge. In the first stage, we pretrain in a setting where the robot only starts moving once the human has stopped (\textbf{sequential}), gathering expert demonstrations via grasp and motion planning. This pretrained policy is further finetuned in the second stage in which the human and robot move simultaneously (\textbf{simultaneous}).

\vspace{-2mm}
\paragraph{Pretraining in Sequential Setting}
\label{sec:h2r:sequential}
In the sequential setting, the robot starts moving once the human has come to a stop (see \figref{fig:h2r:training_procedure_left}, top). To grasp the object from the stationary human hand, we 
 leverage motion planning to provide expert demonstrations \cite{wang2020manipulation} based on state information. During data collection, we alternate between motion planning and RL-based exploration. In both cases, we store the transitions $ \transitionvec_{t}= \{ \pcvec_t,\actions_t, \goalvec_t, \rewardvec_t, \pcvec_{t+1}, \expertvec_t \}$ in a replay buffer $\replay$, from which we sample during network training. The term $\pcvec_t$ and $\pcvec_{t+1}$ indicate the point cloud and the next point cloud, $\actions_t$ the action, $\goalvec_t$ the pre-grasp goal pose, $\rewardvec_t$ the reward, and $\expertvec_t$ an indicator of whether the transition is from the expert or from RL exploration. 
 
Inspired by~\cite{wang2021goal}, we collect expert trajectories with the OMG planner \cite{wang2020manipulation} that leverages ground-truth states, however, we also store the point cloud information for vision-based RL training of our policy. Note that some expert trajectories generated by the planner result in collision with the hand, which is why we introduce an offline pre-filtering scheme. We first parse the ACRONYM dataset \cite{eppner2020acronym} for potential grasps. We then run collision checking to filter out grasps where the robot and human hand collide. For the set of remaining collision-free grasps, we plan trajectories to grasp the object and execute them in open-loop fashion. 
On the other hand, the RL policy $\policypre$ explores the environment and receives a sparse reward, i.e., the reward is one if the task is completed successfully, otherwise zero. Hence, collisions with the human will get implicitly penalized by not receiving any positive reward. 

We also apply a variety of different techniques to make the policies more robust. We alternate between initializing the robot in a home position and random poses (that have the object and hand in its view). As proposed in \cite{wang2021goal}, we occasionally compute optimal actions using DAGGER \cite{ross2011reduction} during RL exploration and use them to supervise the policy's actions. Additionally, we start episodes of the RL agent with a random amount of initial actions proposed by the expert to further guide the training process.

\begin{figure*}[t!]
\begin{center}
   \includegraphics[width=0.8\textwidth]{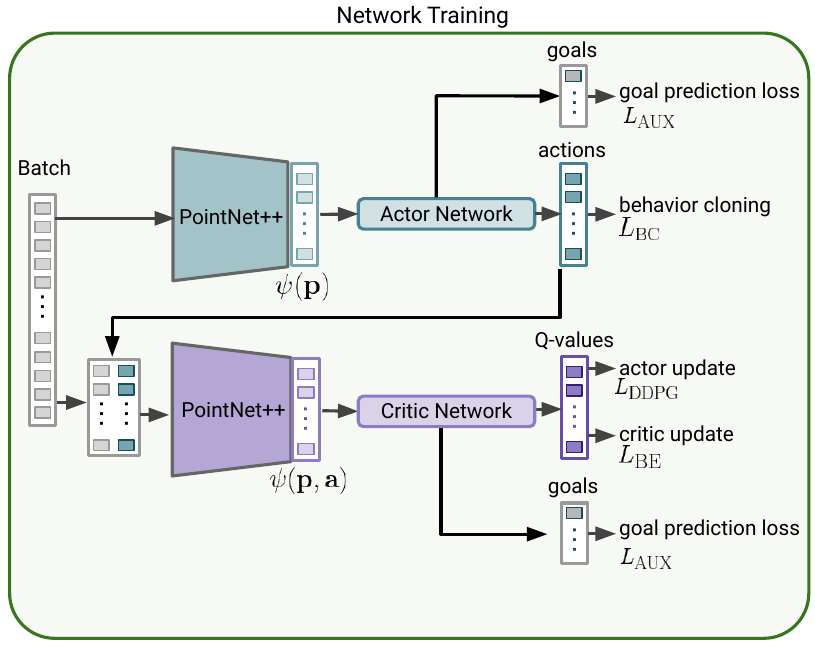}
\end{center}
\vspace{-3mm}
   \caption{\textbf{Network Architecture} Our model follows the actor-critic framework, where a policy (actor) predicts actions to control the robot and a critic network estimates the expected long term reward. Transitions for training are sampled from a replay buffer.
   }
\label{fig:h2r:training_procedure_right}
\end{figure*}

\paragraph{Finetuning in Simultaneous Setting}
In this setting, the human and robot move at the same time. Hence, we cannot rely on motion and grasp planning to guide the policy. On the other hand, we found that simply taking the pretrained policy $\policypre$ from the sequential setting and continuing to train it without an expert leads to an immediate drop in performance. Hence, we introduce a self-supervision scheme for stability reasons, i.e., we want to keep the finetuning policy close to the pretrained policy. To this end, we replace the expert planner from the sequential setting by an expert policy $\policyexp$, which is initialized with the weights of the pretrained policy $\policypre$ and frozen. This policy already provides a reasonable prior policy that successfully performs simple handovers in the simultaneous setting (see \Fig{h2r:training_procedure_right} bottom). We store successful trials with the pretrained policy in the replay buffer. Therefore, we have two policies that explore the environment. First, the expert policy $\policyexp$ as proxy for the motion and grasping planner to provide successful handover demonstrations. Second, the finetuning policy $\policyft$ and critic $\criticft$, which are initialized with the weights of the pretrained policy $\policypre$ and critic $\criticpre$, respectively. We proceed to train these two networks using the loss functions described in \secref{sec:handoversim2real:network_training}. In this stage, we drop most of the training techniques from the pre-training stage and start all the episodes from home position. We do not use DAGGER anymore and rollouts from the RL agent are not started with actions proposed by the expert.

\paragraph{Network Training}
\label{sec:handoversim2real:network_training}
The reinforcement learning algorithm we use is TD3, as described in \secref{bg:td3} and illustrated in \figref{fig:h2r:training_procedure_right}. During training, we sample a batch of random transitions from the replay buffer. The policy network is trained using a combination of behavior cloning, RL-based losses and an auxiliary objective. In particular, the policy is updated using the following loss function:
\begin{equation}
    \mathcal{L}(\theta) = \lambda \mathcal{L}_{\text{BC}}+(1-\lambda)\mathcal{L}_{\text{DDPG}}+\mathcal{L}_{\text{AUX}},
    \label{eq:h2r:loss_actor}
\end{equation}

\noindent where $\mathcal{L}_{\text{BC}}$ is a behavior cloning loss that keeps the policy close to the expert policy, $\mathcal{L}_{\text{DDPG}}$ is the actor-critic loss described in \Eq{critic_target_ddpg}, and $\mathcal{L}_{\text{AUX}}$ is an auxiliary objective that predicts the grasping goal pose of the end-effector \cite{wang2021goal}. The coefficient $\lambda$ balances the behavior cloning and the RL objective. The critic loss is defined as:

\begin{equation}
\mathcal{L}(\phi) =  \mathcal{L}_{\text{BE}}+\mathcal{L}_{\text{AUX}},
\label{eq:h2r:loss_critic}
\end{equation}  

\noindent where $\mathcal{L}_{\text{BE}}$ indicates the Bellman error from \Eq{deep_q_learning} and $\mathcal{L}_{\text{AUX}}$ is the same auxiliary loss used in \Eq{h2r:loss_actor}.
We refer the reader to \cite{wang2021goal} for the specific details and definitions of the loss functions.

%% file: chapters/05_handover/handoversim2real/sections/05_experiments.tex
\pagebreak
\section{Experiments}

We first evaluate our approach in simulation using the HandoverSim benchmark (Sec.~\ref{sec:h2r:simulation_evaluation}). Next, we investigate the performance of sim-to-sim transfer by evaluating the trained models on the test environments powered by a different physics engine (Sec.~\ref{sec:h2r:sim2sim}). Finally, we apply the trained model to a real-world robotic system and analyze the performance of sim-to-real transfer (Sec.~\ref{sec:h2r:sim2real}). Please see our project page\footnote{\href{https://handover-sim2real.github.io}{handover-sim2real.github.io}} for qualitative videos of our method and baselines.

\subsection{Simulation Evaluation}
\label{sec:h2r:simulation_evaluation}

\paragraph{Setup}~HandoverSim~\cite{chao2022handoversim} contains 1,000 unique H2R handover scenes divided into train, validation, and test splits. Each scene contains a unique human handover motion. We evaluate on the ``s0'' setup which contains 720 training and 144 testing scenes. Following the evaluation of GA-DDPG~\cite{wang2021goal} in~\cite{chao2022handoversim}, we consider two settings: (1) the ``sequential'' setting where the robot is allowed to move only after the human hand reaches the handover location and remains static there (i.e., ``hold'' in~\cite{chao2022handoversim}), and (2) the ``simulataneous'' setting where the robot is allowed to move from the beginning of the episode (i.e., ``w/o hold'' in~\cite{chao2022handoversim}).

\paragraph{Metrics}~We follow the evaluation protocol in HandoverSim~\cite{chao2022handoversim}. A handover is considered successful if the robot grasps the object from the human hand and moves it to a designated location. A failure is claimed and the episode is terminated if any of the following three conditions occur: (1) the robot collides with the hand (\emph{contact}), (2) the robot drops the object (\emph{drop}), or (3) a maximum time limit is reached (\emph{timeout}). Besides efficacy, the benchmark also reports efficiency in time, which is further broken down into (1) the execution time (\emph{exec}), i.e., the time to physically move the robot, and (2) the planning time (\emph{plan}), i.e., the time spent on running the policy. All metrics are averaged over the rollouts on the test scenes.

\paragraph{Baselines}~Our primary baseline is GA-DDPG~\cite{wang2021goal}. Besides comparing with the original model (i.e., trained in~\cite{wang2021goal} for table-top grasping and evaluated in~\cite{chao2022handoversim}), we additionally compare with a variant finetuned on HandoverSim (``GA-DDPG~\cite{wang2021goal} finetuned''). For completeness, we also include two other baselines from~\cite{chao2022handoversim}: ``OMG Planner~\cite{wang2020manipulation}'' and ``Yang et al.~\cite{yang2021reactive}''. However, both of them are evaluated with ground-truth state input in~\cite{chao2022handoversim} and thus are not directly comparable with our method. Additionally, these methods assume a stationary human before the robot starts moving, hence, we only report their scores for the sequential setting.

\input{\dir/handoversim2real/tables/table_main_exp}

\begin{figure*}
    \centering
    \begin{subfigure}[t]{1.0\textwidth}
         \centering
            \includegraphics[width=\textwidth]{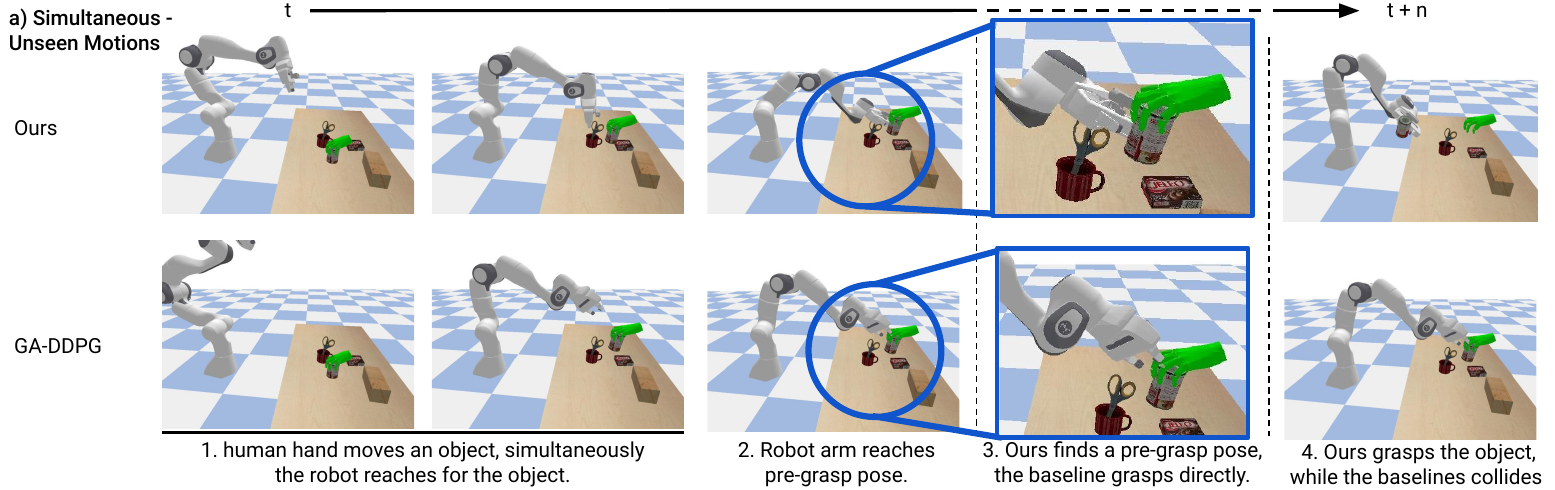}
            \label{fig:h2r:quali_a}
    \end{subfigure}
    \begin{subfigure}[b]{1.0\textwidth}
         \centering
            \includegraphics[width=\textwidth]{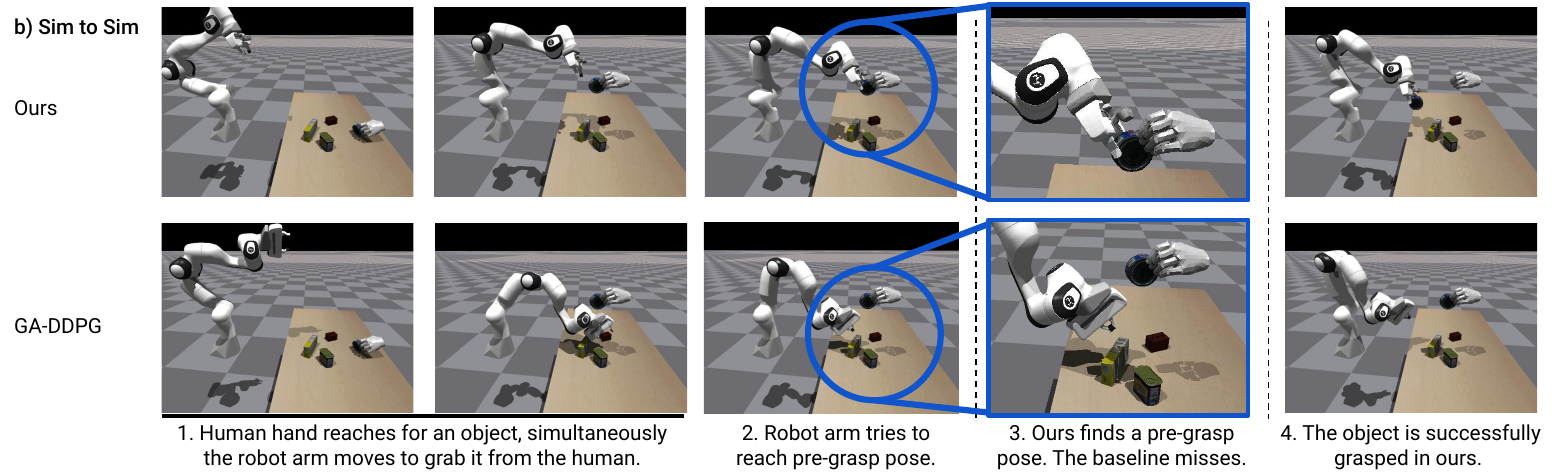}
            \label{fig:h2r:quali_b}
    \end{subfigure}
    \begin{subfigure}[b]{1.0\textwidth}
         \centering
            \includegraphics[width=\textwidth]{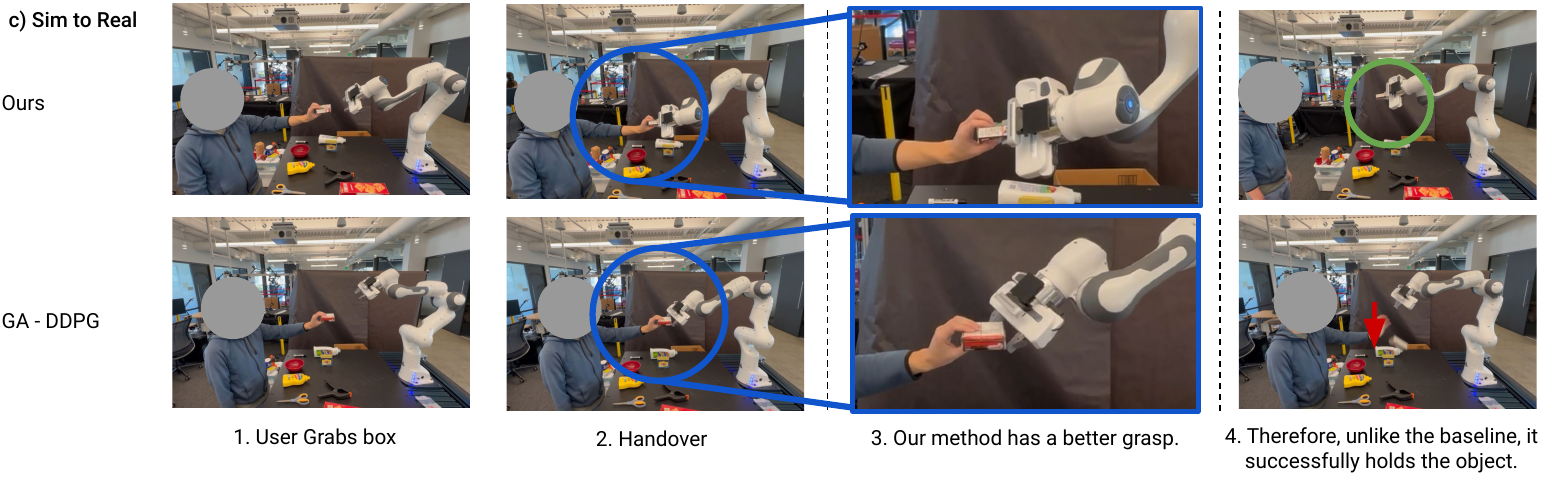}
            \label{fig:h2r:quali_c}
     \end{subfigure}
    \caption{\textbf{Qualitative results.} We provide a comparison to show our methods' advantages over GA-DDPG \cite{wang2021goal}. (a) Our method reacts to the moving human, while the baseline tries to go for a grasp directly, which leads to collision. (b) In the sim-to-sim transfer, we often find that the baseline does not find a grasp on the object. (c) In the sim-to-real experiment, GA-DDPG usually tries to get to a grasp directly, while our method adjusts the gripper into a stable grasping pose first. 
    }
    \label{fig:h2r:qualitative}
\end{figure*}

\paragraph{Results}~ \Tab{h2r:quantitative} reports the evaluation results on the test scenes. 
In the sequential setting, our method significantly outperforms all the baselines in terms of success rate, even compared to methods that use state-based input. Our method is slightly slower on average than GA-DDPG in terms of total time needed for handovers.
In the simultaneous setting, our method clearly outperforms GA-DDPG, which has low success rates. Qualitatively, we observe that GA-DDPG directly tries to grasp the object from the user while it is still moving, while our method follows the hand and finds a feasible grasp once the hand has come to a stop, resulting in a trade-off on the overall execution time. We provide a qualitative example of this behavior in \Fig{h2r:qualitative} (a). We also refer to the \ref{h2r:robustness} for a discussion of limitations and a robustness analysis of our pipeline under noisy observations.

\pagebreak

\paragraph{Handedness, Unseen Subjects, Unseen Objects} For completeness, we report the results of the remaining settings from HandoverSim \cite{chao2022handoversim}. Namely, we add the settings ``S1 - Unseen Subjects'' in \Tab{s1}, ``S2 - Unseen Handedness'' in \Tab{s2}, and ``S3 - Unseen Objects'' in \Tab{s3}. Overall, we observe that the results are consistent with the main paper. In general, the main baseline GA-DDPG \cite{wang2021goal} struggles in the simultaneous setting. Our method has significantly better performance in terms of overall success rates, while retaining a slightly slower mean accumulated time for successful handovers. This is because GA-DDPG often goes for a grasp in the most direct path, whereas our approach searches for a safe pre-grasp pose, from where the object can be grasped. For a qualitative demonstration of this behavior, we refer to the supplementary video. We also compare our final model with the pretrained versions (\emph{Ours w/o ft.}). The results further indicate that finetuning helps improve the model, especially in the simultaneous setting, e.g., the success rate in \Tab{s1} improves from 62.78\% to 73.33\% with the finetuned model.

Notably, results on S2 and S3 suggest that our method can generalize well to unseen subjects and unseen objects. This result is important because in unstructured real world environments, neither objects nor subjects have been encountered during training. However, there are only two unseen objects in the S3 benchmark split, thereby offering limited validation for generalization. We explore generalization to unseen objects in more detail in Chapter \ref{ch:handovers:synh2r}. 

\input{\dir/handoversim2real/tables/table_s1}
\input{\dir/handoversim2real/tables/table_s2}
\input{\dir/handoversim2real/tables/table_s3}

\clearpage

\input{\dir/handoversim2real/tables/table_ablations}

\paragraph{Ablations}~We evaluate our design choices in an ablation study and report the results in  \Tab{h2r:ablation}. We analyze the vision backbone by replacing PointNet++ with a ResNet18~\cite{he2016deep} that processes the RGB and depth/segmentation (DM) images. Similar to the findings in GA-DDPG, the PointNet++ backbone performs better. Next, we train our method from third person view instead of egocentric view and without active hand segmentation (\textit{w/o hand point cloud}), i.e., the policy only perceives the object point cloud but not the hand point cloud. We also ablate the auxiliary prediction (\textit{w/o aux prediction}) and evaluate a variant that directly learns to approach and grasp the object instead of using the two task phases of approaching and grasping (\textit{w/o standoff}). Lastly, we compare against our pretrained model, which was only trained in the sequential setting without finetuning (\textit{w/o finetuning}). We find that the ablated components comprise important elements of our method.  The results indicate an increased amount of hand collision or object drop in all ablations. A closer analysis in the simultaneous setting shows that our finetuned model outperforms the pretrained model.

\begin{figure}[t!]
 \centering
 \includegraphics[width=0.9\linewidth, trim={0cm 0cm 0cm 2cm}, clip]{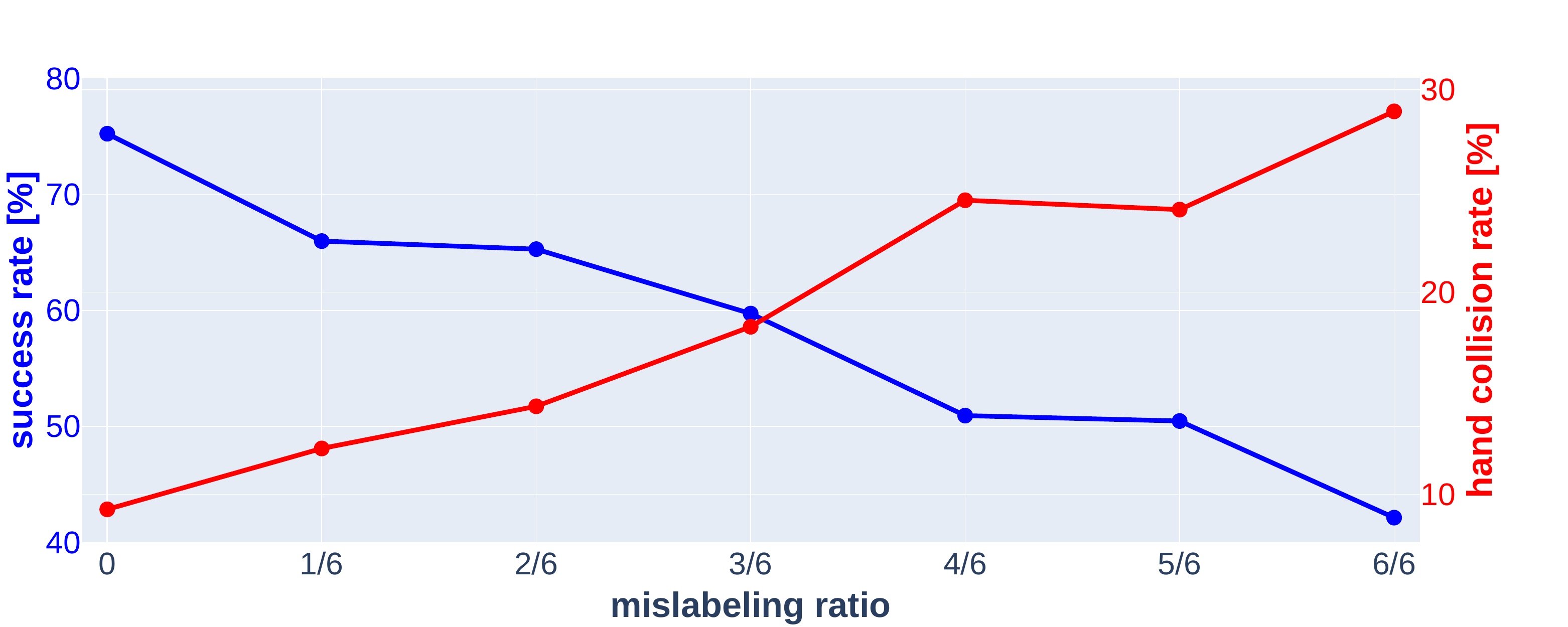}
     \caption{\textbf{Segmentation Mislabeling.} Our method's success (\textcolor{blue}{blue}) and hand collision rate (\textcolor{red}{red}) under increasing degree of mislabeling hand as object segments.}
 \label{fig:h2r:seg_err_plot}
\end{figure}

\paragraph{Robustness Analysis}
\label{h2r:robustness}
We evaluate in simulation how \textit{noisy observations} affect our pipeline by (1) adding simulated Kinect noise to depth images~\cite{handa2014benchmark} and (2) testing with imperfect hand segmentation. For (2), we divide the hand into 6 different parts (fingers and palm) and re-label a subset of parts as object in the segmentation mask. We vary the mislabeling ratio (``0/6'' no parts and ``6/6'' all parts) and sample randomly which parts will be mislabeled for a given episode. As expected, performance degrades with increasing noise in depth (\eg, a 59.49\% success rate in \Tab{h2r:kinect}) and increasing mislabeling ratio (\eg, decreasing success rate and increasing hand collisions in \Fig{h2r:seg_err_plot}). In the future, our model could be made more robust by introducing additional noise in training.

\input{\dir/handoversim2real/tables/table_kinect}

\clearpage

\subsection{Sim-to-Sim Transfer}
\label{sec:h2r:sim2sim}
Instead of directly transferring to the real world, we first evaluate the robustness of the models by transferring them to a different physics simulator. We re-implement the HandoverSim environment following the mechanism presented in~\cite{chao2022handoversim} except for replacing the backend physics engine from Bullet~\cite{coumans20162021pybullet} to Isaac Gym~\cite{makoviychuk2021isaac}. We then evaluate the models trained on the original Bullet-based environment on the test scenes powered by Isaac Gym. The results are presented in \Tab{h2r:sim2sim_singlecol}. We observe a significant drop for GA-DDPG on the success rates (i.e., to below 20\%) in both settings. Qualitatively, we see that grasps are often either missed completely or only partially grasped (see \Fig{h2r:qualitative} (b)). On the other hand, our method is able to retain higher success rates. Expectedly, it also suffers from a loss in performance. We analyze the influence of our grasp predictor on transfer performance and compare against a variant where we execute the grasping motion after a fixed amount of time (\emph{Ours w/o grasp pred.}), which will leave the robot enough time to find a pre-grasp pose. Part of the performance drop is caused by the grasp predictor initiating the grasping phase at the wrong time, which can be improved upon in future work.

\input{\dir/handoversim2real/tables/table_sim2sim_singlecol}

\pagebreak

\section{Sim-to-Real Transfer}
\label{sec:h2r:sim2real}

Finally, we deploy the models trained in HandoverSim on a real robotic platform. We follow the perception pipeline used in~\cite{yang2021reactive,wang2021goal} to generate segmented hand and object point clouds for the policy, and use the output to update the end effector's target position. We compare our method against GA-DDPG~\cite{wang2021goal} with two sets of experiments: (1) a pilot study with controlled handover poses and (2) a user evaluation with free-form handovers. 

\begin{figure*}[t!]
 \centering
 \includegraphics[width=0.9\linewidth]{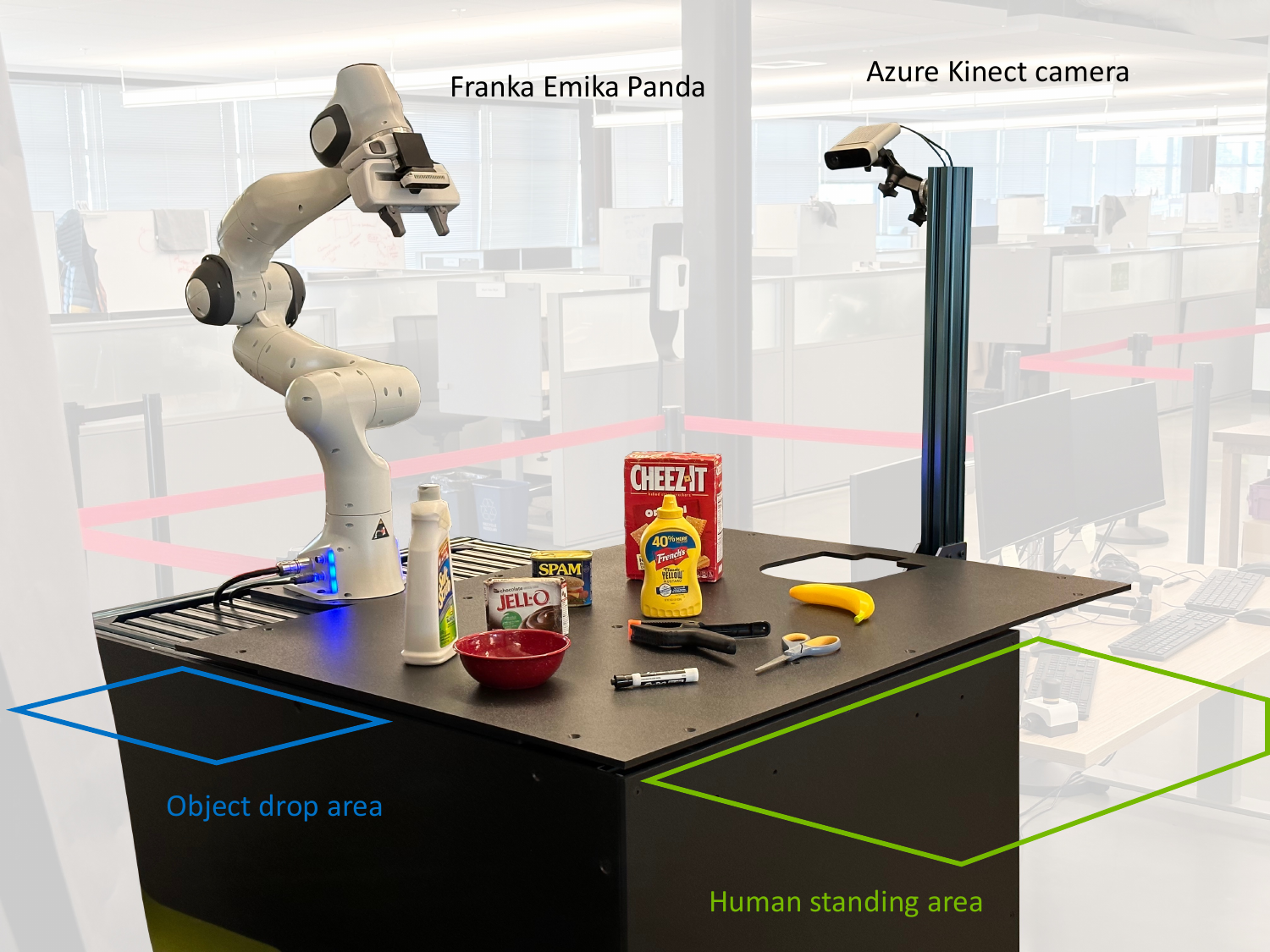}~
 \caption{\textbf{Real-world handover system setup.} A Franka Emika Panda robot and an Azure Kinect camera are rigidly mounted on a table. The human participant will stand across the table (in the green area), pick up objects, and attempt handovers to the robot. The robot will drop the object in a designated area (blue) after retrieving it from the handover.}

 \label{fig:h2r:setup}
\end{figure*}

\subsection{System Setup}
\label{sec:h2r:setup}

\Fig{h2r:setup} shows the setup of our real-world handover system. The setup consists of a Franka Emika Panda robot and a Azure Kinect camera, both rigidly mounted to a table. The Azure Kinect is mounted externally to the robot with the extrinsics calibrated, and is perceiving the scenes from a third-person view with an RGB-D stream. The objects for handover are initially placed on the table. During handovers, the human participant will stand on the opposite side of the table (in the green area), pick up the objects, and attempt handovers to the robot. If the robot successfully retrieves the object, it will move the end effector to a drop-off area (blue) and drop the object into a box.

Since our policy expects a segmented point cloud at the input, we follow the perception pipeline used in~\cite{yang2021reactive,wang2021goal} to generate segmented point clouds for the hand and object. The Azure Kinect is launched to provide a continuous stream of RGB images and point clouds. We first use Azure Kinect's Body Tracking SDK to track the 3D location of the wrist joint of the handover hand. At each time frame, we crop a sub-point cloud around the tracked joint location which includes points on both the hand and the held object. We additionally run a 2D hand segmentation model on the RGB image and use it to label the hand points in the cropped point cloud. We treat all the points not labeled as hand as the object. Since our policies are trained for wrist camera views and we use an external camera in the real-world system, we need to additionally compensate for the view point change. We transform the segmented point cloud from the external camera's frame to the wrist camera's frame using the calibrated robot-camera extrinsics and forward kinematics. This way we can simulate the segmented point cloud input which the policy observes during the training in simulation. Note that this perception pipeline can induce sim-to-real gaps from several sources: (1) noises in the point clouds from real cameras, (2) noises from body tracking and hand segmentation errors, (3) the change in view points (i.e., from the wrist to external camera), and (4) unseen human behavior.

Compared to GA-DDPG~\cite{wang2021goal}, we adapt the control flow of the policy to explicitly incorporate the pre-grasp mechanism in our method. To control the motion of Franka, we follow the pipeline used in~\cite{yang2021reactive,wang2021goal}. Given a target end effector pose at a new time step, we use Riemannian Motion Policies (RMPs)~\cite{ratliff2018riemannian} to generate a smooth trajectory for the robot arm. We use libfranka to control the Franka arm to follow the trajectory. The robot will start moving only when a segmented point cloud is perceived. Once it decides to grasp, we will execute a predefined motion where the robot closes the gripper, lifts the end effector, moves to the drop-off area, and opens the gripper. The robot will return to a standby pose and remain in that state if no segmented point cloud is perceived or after it drops off the object.
\pagebreak

\begin{figure}[t!]
 \centering
 \includegraphics[width=0.45\linewidth]{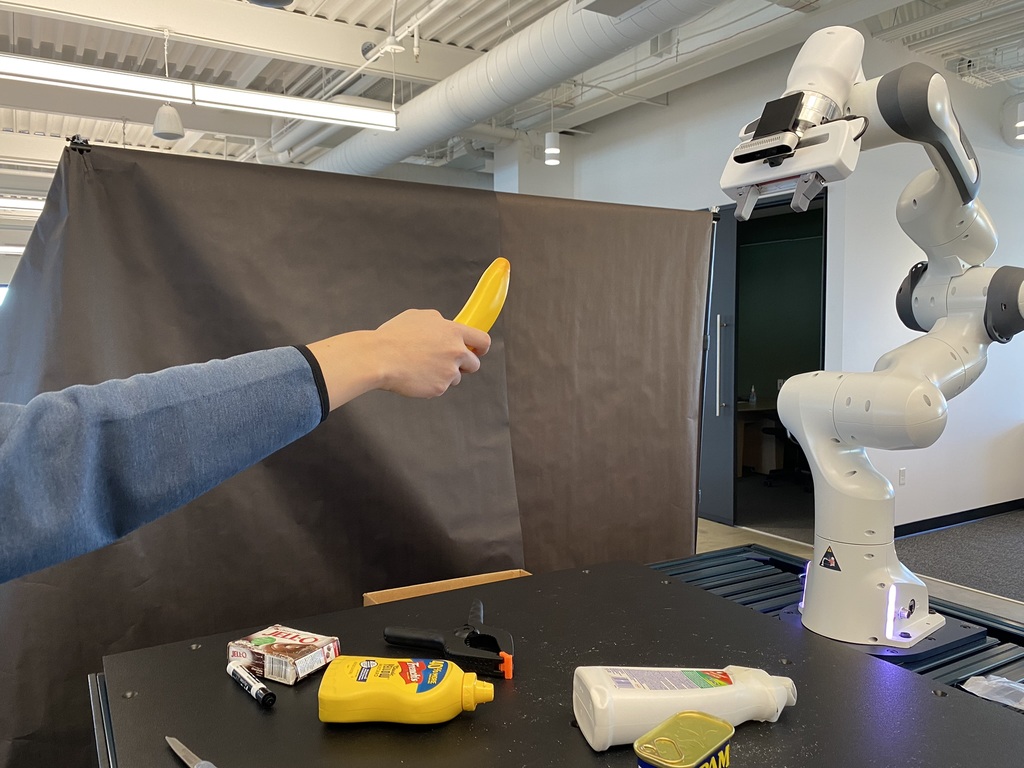}
 \caption{\textbf{Pilot study.} We conducted a pilot study by controlling the handover poses of the human subject.}
 \label{fig:h2r:pilot}

\end{figure}

\input{\dir/handoversim2real/figures/pilot_overview_l}

\vspace{-8mm}
\subsection{Pilot Study}
\paragraph{Evaluation Protocol} We select the following 10 objects from YCB ~\cite{calli2015ycb}:
\begin{multicols}{2}
 \begin{itemize}[noitemsep,topsep=0pt,parsep=0pt,partopsep=0pt]
  \item 011\_banana
  \item 037\_scissors
  \item 006\_mustard\_bottle
  \item 024\_bowl
  \item 040\_large\_marker
  \item 003\_cracker\_box
  \item 052\_extra\_large\_clamp
  \item 008\_pudding\_box
  \item 010\_potted\_meat\_can
  \item 021\_bleach\_cleanser
 \end{itemize}
\end{multicols}

For each object, we select 3 handover poses separately for the right and left hand, totaling 60 handover poses for both hands. The set of handover poses is selected to represent the handover task at different levels of difficulty: for each hand-object combination, we select one common handover pose (``pose 1''), one handover pose with the object held horizontally (``pose 2''), and one handover pose with severe hand occlusion by holding the object from the top (``pose 3''). \Fig{h2r:pilot} illustrates the setting where a subject holds an object in a controlled handover pose in front of the robot. \Fig{h2r:pilot_left} shows the selected handover poses for five objects.

For each subject, we iterate through the 60 handover poses and evaluate each pose once. A handover is considered failed if (1) the robot pinches (or is about to pinch) the subject's hand (in which case the subject may evade the grasp), (2) the robot drops the object during the handover, or (3) the robot has reached an irrecoverable state (e.g., a locked arm due to joint limits). A handover is successful if the robot retrieves the object from the subject's hand and successfully move it to the drop-off area without incurring any failures. We evaluate the same handover poses on two methods: GA-DDPG~\cite{wang2021goal} and ours. Therefore, each subject will perform 120 handover trials in total.

\paragraph{Results}
We conduct our pilot study with two subjects and report the results in  \Tab{h2r:pilot_study}. . The results indicate that our method outperforms GA-DDPG~\cite{wang2021goal} for both subjects on the overall success rate (i.e., 41/60 versus 21/60 for Subject 1). Qualitatively, we observe that GA-DDPG~\cite{wang2021goal} tends to fail more from unstable grasping as well as hand collision. Fig.~\ref{fig:h2r:qualitative} (c) shows two examples of the real world handover trials.

\input{\dir/handoversim2real/tables/table_pilot}

\subsection{User Evaluation}

\paragraph{Evaluation Protocol} We further recruited $6$ users to compare the two methods. We adopt the same 10 objects from the pilot study, and ask each user to hand over each object once with their right hand. We instruct the users to hand over objects ``in any way they like''. We compare the two methods (i.e., GA-DDPG~\cite{wang2021goal} and ours) by repeating the same process, i.e., we instruct the user to hand over the 10 objects to one system first, followed by to the other system. We counterbalance the order of the two systems throughout the user evaluation to avoid bias. During their experiments, the users are asked to fill out a questionnaire with Likert-scale and open-ended questions to provide feedback after they interact with each system.

\input{\dir/handoversim2real/figures/fig_user}

\paragraph{Results}

\begin{figure*}[t!]
    \centering
    \begin{subfigure}{\linewidth}
        \centering
        \includegraphics[width=\linewidth]{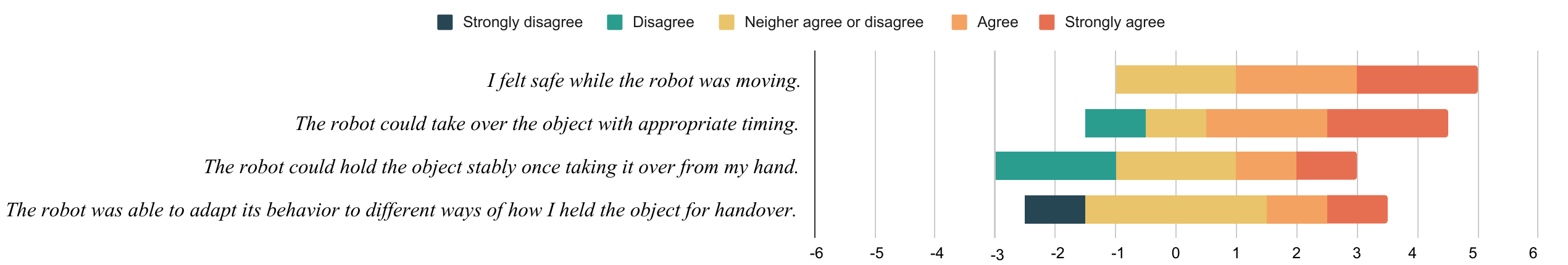}
        \includegraphics[width=\linewidth]{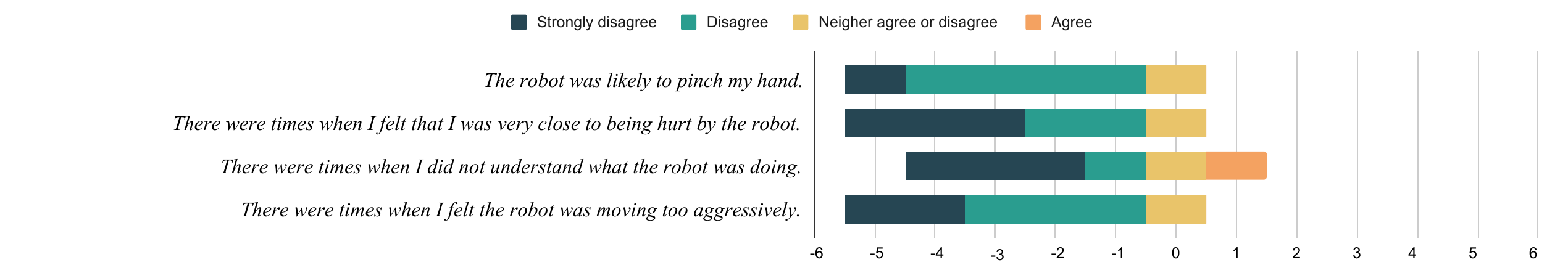}
        \caption{\small User's ranking with each statement for \textbf{GA-DDPG~\cite{wang2021goal}} in the user evaluation.}
        \label{fig:h2r:user_gaddpg}
    \end{subfigure}

    \begin{subfigure}{\linewidth}
        \centering
        \includegraphics[width=1.0\linewidth]{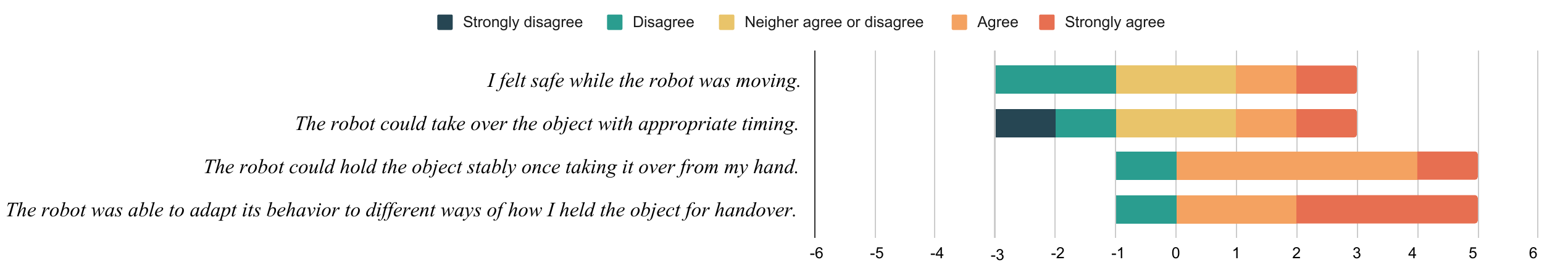}
        \includegraphics[width=1.0\linewidth]{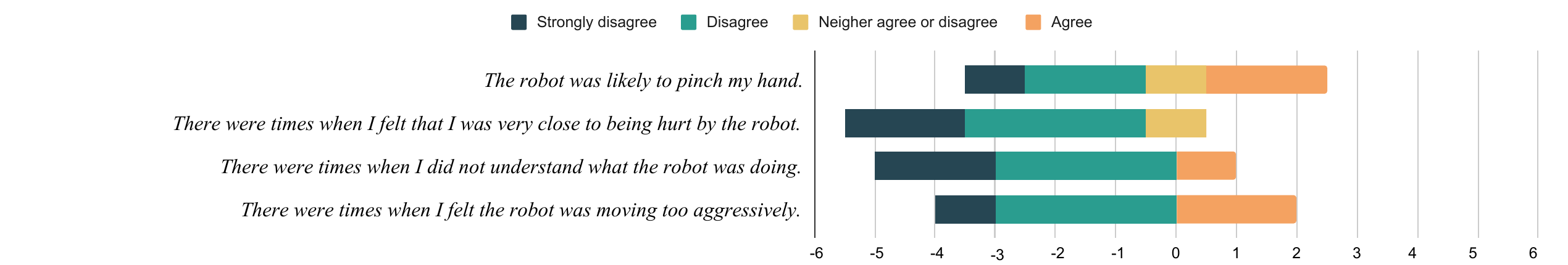}
        \caption{User's ranking with each statement for \textbf{our method} in the user evaluation.}
        \label{fig:h2r:user_ours}
    \end{subfigure}
    \caption{\textbf{User study questionnaire analysis.} User rankings with each statement for both \textbf{GA-DDPG} and \textbf{our method} in the user evaluation. Each color denotes a different degree of agreement. The length of the bar denotes the count of the users. For each bar, the center count of ``Neither agree or disagree'' is aligned with 0 on the horizontal axis. In the top figures, a higher agreement score (orange) indicates better performance, while in the bottom figures, a lower agreement score (green) indicates better performance.}
    \label{fig:h2r:combined}
\end{figure*}

We conduct our user evaluation with 6 users (\Fig{h2r:user}). The evaluation results are presented in  \Fig{h2r:user_gaddpg} for GA-DDPG~\cite{wang2021goal} and  \Fig{h2r:user_ours} for our method. Each figure shows the user's ranking with the statements queried in the questionnaire. For each statement, a user can rank their agreement level with one of the five options: ``Strongly disagree'' (1), ``Disagree'' (2), ``Neither agree or disagree'' (3), ``Agree'' (4), and ``Strongly agree'' (5) (see the color codes in  \Fig{h2r:user_gaddpg} and  \Fig{h2r:user_ours}). The length of each color bar denotes the count of the users. For each method, the statements are further grouped into two subfigures, where a higher agreement score indicates a better performance (top), and a lower agreement score indicates a better performance (bottom).

Overall, our method receives higher agreement scores over GA-DDPG~\cite{wang2021goal} for the statements ``\textit{The robot could hold the object stably once taking it over from my hand.}'' (i.e., (5,4,4,4,4,3) versus (5,4,3,3,3,2)) and ``\textit{The robot was able to adapt its behavior to different ways of how I held the object for handover.}'' (i.e., (5,5,5,4,4,3) versus (5,4,3,3,3,2)). This is congruent with our simulation evaluation results that our method can grasp objects more robustly by finding good pre-grasp poses around the object. This was also reflected in participants' comments. One said our method ``\textit{tends to explore more diverse grasp}", ``\textit{was much better at aligning the grasp}" and ``\textit{adjusts behavior for different objects in different poses}" when compared with GA-DDPG.  One pointed out that sometimes GA-DDPG ``\textit{grasped from the tip of the object}". 
The interpretability of the robot's motion was also acknowledged by their comments, \eg, it ``\textit{[was] safe and interpretable at all times}" and ``\textit{felt like we understood each other}".
Surprisingly, the users favor GA-DDPG~\cite{wang2021goal} more when it comes to safety related metrics, e.g., for the statement ``\textit{I felt safe while the robot was moving.}'' ((5,4,3,3,2,2) for ours versus (5,5,4,4,3,3) for GA-DDPG~\cite{wang2021goal}) and ``\textit{The robot was likely to pinch my hand.}'' ((1,2,2,3,4,4) for ours versus (1,2,2,2,2,3) for GA-DDPG~\cite{wang2021goal}). This can be attributed to GA-DDPG's tendency to grasp from the grasp points closest to the robot, and hence it often keeps a safe distance from the human hand. For our method, several users felt the robot hand pushing too much during grasping. One said it was ``\textit{flexible in grasp selection, but may be too close to my finger}''. Another said ``\textit{the forward movement ... put the gripper fairly close to me}''. This can potentially be addressed by incorporating force feedback in the grasping motion as well as taking gripper hand distance into account during training. The majority of participants agreed that the timing of our method is more appropriate, commenting the ``\textit{handover time was pretty seamless}" and ``\textit{didn't have to wait too long}".

Although the main objective in the user study was to let users interact freely with the system in a non-standardized manner, we additionally evaluate the user study quantitatively. We report the success rate and approach time (\ie, from the robot starting to move to grasp completion). Our method still compares favorably to GA-DDPG with a higher success rate ($88.9\%$ \vs $80.0\%$) and a shorter approach time ($6.40\pm2.27$s \vs $7.48\pm2.64$s). The better timing was noted by the majority of participants, who commented that the ``\textit{handover time was pretty seamless}" and ``\textit{didn't have to wait too long}". Interestingly, we observed in our user study that natural human-to-robot handovers are less susceptible to grasping failures in the real world, since the human partner would often help by agilely adjusting the object pose in the last mile to ensure a successful grasp, which is not reflected in the simulated human behavior.

\clearpage

\input{\dir/handoversim2real/figures/fig_failures}
\section{Limitations}
\label{sec:h2r:app_lim}

We will now discuss failure cases of our method. Some failures occur with smaller objects, where the human hand often encloses large parts of the object. For the robot to find grasps where the gripper does not touch the hand at all in such cases is extremely difficult, especially when only having access to point cloud input. The grasp prediction task that decides when to switch from approaching to grasping is quite challenging, because a small change in end-effector pose can already cause a handover to fail. We sometimes find that the grasp prediction triggers the grasp too early or in an instance where the object will eventually drop. Since the grasp prediction network is trained offline, it may be improved by finetuning in online fashion with experiences from policy rollouts. Furthermore, we investigate \textit{sensor-challenging} objects in real world transfers. Our depth sensor is vulnerable to transparent or dark objects, which may lead to failures of the policy (\Fig{h2r:real_failure}).

%% file: chapters/05_handover/handoversim2real/tables/table_main_exp.tex
\begin{table*}[t]
 \centering
 \small
  \resizebox{1.0\columnwidth}{!}{
 \begin{tabular}{l|l|cccc|cccc}

  \hline
  & & \multirow{2}{*}{success (\%)} & \multicolumn{3}{c|}{mean accum time (s)} & \multicolumn{4}{c}{failure (\%)} \\
  & & & exec & plan & total & contact & drop & timeout & total \\
  \hline
  \parbox[t]{2mm}{\multirow{5}{*}{\rotatebox[origin=c]{90}{Sequential}}}
  & OMG Planner~\cite{wang2020manipulation} $\dagger$        & 62.50 & 8.309          & 1.414          &  9.722          &  27.78          & ~~8.33 & ~~1.39 & 37.50 \\
  & \gc Yang et al.~\cite{yang2021reactive} $\dagger$  & \gc 64.58  & \gc 4.864  & \gc
  0.036& \gc 4.900          &  \gc 17.36          &  \gc 11.81          & \gc ~~6.25 & \gc 35.42         \\  
\cline{2-10}
  & GA-DDPG~\cite{wang2021goal}   & 50.00          & 7.139        & 0.142 & 7.281          & ~\textbf{4.86} &  19.44          &  25.69 & 50.00        \\
  & \gc GA-DDPG~\cite{wang2021goal} finetuned  & \gc 57.18         & \gc  \textbf{6.324}        & \gc \textbf{0.086} & \gc \textbf{6.411}      & \gc ~6.48 &  \gc 27.08          &  \gc ~9.26  & \gc 42.82      \\
 
  & Ours    & \textbf{75.23} & 7.743 & 0.177 & 7.922  & ~9.26 & \textbf{13.43} & ~\textbf{2.08}  & \textbf{24.77}    \\
\hline

  \parbox[t]{2mm}{\multirow{3}{*}{\rotatebox[origin=c]{90}{Simult.}}}
  
  & GA-DDPG~\cite{wang2021goal} &  36.81          & \textbf{4.664} & 0.132         & \textbf{4.796} &  ~9.03          &  25.00          &  29.17  &   63.19     \\
  & \gc GA-DDPG~\cite{wang2021goal} finetuned  & \gc 54.86          & \gc 4.832          & \gc \textbf{0.082} & \gc 4.914         & \gc ~\textbf{6.71} &  \gc 26.39          & \gc 12.04    & \gc 45.14      \\
 
  & Ours   & \textbf{68.75} & 6.232 & 0.178 & 6.411  & ~8.80 & \textbf{17.82} & ~\textbf{4.63} & \textbf{31.25}      \\
  
\bottomrule
  
 \end{tabular}
 }
 \caption{\textbf{HandoverSim Benchmark Evaluation.} Comparison of our method against various baselines from the HandoverSim benchmark \cite{chao2022handoversim}. In the sequential setting, we find that our baseline achieves better overall success rates than the baselines. In the simultaneous setting, we outperform the applicable baselines by large margins. The results for our method are averaged across 3 random seeds. $\dagger$: both methods \cite{wang2020manipulation},yang2021reactive are evaluated with ground-truth states in \cite{chao2022handoversim} and thus are not directly comparable with ours.}

 \label{tab:h2r:quantitative}
\end{table*}

%% file: chapters/05_handover/handoversim2real/tables/table_s1.tex
\begin{table*}[t!]
 \centering
  \resizebox{1.0\columnwidth}{!}{
 \begin{tabular}{l|l|cccc|cccc}

  \hline
  \multicolumn{10}{c}{S1: Unseen Subjects} \\
  \hline
  & & \multirow{2}{*}{success (\%)} & \multicolumn{3}{c|}{mean accum time (s)} & \multicolumn{4}{c}{failure (\%)} \\
  & & & exec & plan & total & contact & drop & timeout & total \\
  \hline
  \parbox[t]{2mm}{\multirow{5}{*}{\rotatebox[origin=c]{90}{Sequential}}}
  & OMG Planner~\cite{wang2020manipulation} $\dagger$  & 62.78  & 8.012 & 1.355 & 9.366 & 33.33 & ~~2.22 & ~~1.67 & 37.22   \\
  & \gc Yang et al.~\cite{yang2021reactive} $\dagger$  & \gc 62.78 & \gc 4.719  & \gc 0.039
 & \gc 4.758 &  \gc 14.44 &  \gc ~~7.78  & \gc 15.00 & \gc 37.22  \\  
\cline{2-10}
  & GA-DDPG~\cite{wang2021goal}   & 55.00 & \textbf{6.791} & \textbf{0.136} & \textbf{6.927} & ~~8.89 & 15.00 & 21.11 & 45.00 \\
  & \gc Ours w/o ft. & \gc 68.15  & \gc 7.151 & \gc 0.164
 & \gc 7.314          &  \gc    ~~6.85     &  \gc 12.96          & \gc 12.04 & \gc    31.85     \\  
  & Ours    & \textbf{75.00} & 7.108 & 0.159 & 7.267 & ~~\textbf{5.00} & \textbf{12.59} & ~~\textbf{7.41}  & \textbf{25.00}    \\
\hline

  \parbox[t]{2mm}{\multirow{3}{*}{\rotatebox[origin=c]{90}{Simult.}}}
  
  & GA-DDPG~\cite{wang2021goal} & 33.33 & \textbf{4.261} & \textbf{0.132} & \textbf{4.393} & 15.56 & 21.67 & 29.44 & 66.67   \\
    & \gc Ours w/o ft. & \gc 62.78 & \gc 5.695  & \gc
 0.164 & \gc 5.859   &  \gc     ~~5.93    &  \gc 17.59         & \gc 13.70 & \gc 37.22        \\  
  & Ours   & \textbf{73.33} & 5.633 & 0.158 & 5.791 & ~~\textbf{5.56} & \textbf{15.37} & ~~\textbf{5.74} & \textbf{26.67}  \\
  
\bottomrule
  
 \end{tabular}
 }
 \caption{\textbf{Unseen Subjects} Comparison of our method against various baselines from the HandoverSim benchmark \cite{chao2022handoversim} in the setting ``S1: Unseen Subjects''. The results of our method are averaged over 3 random seeds. $\dagger$: both methods \cite{wang2020manipulation,yang2021reactive} are evaluated with ground-truth states in \cite{chao2022handoversim} and thus are not directly comparable with ours.}
 \label{tab:s1}
\end{table*}

%% file: chapters/05_handover/handoversim2real/tables/table_s2.tex
\begin{table*}[t!]
 \centering
  \resizebox{1.0\columnwidth}{!}{
 \begin{tabular}{l|l|cccc|cccc}

  \hline
  \multicolumn{10}{c}{S2: Unseen Handedness} \\
  \hline
  & & \multirow{2}{*}{success (\%)} & \multicolumn{3}{c|}{mean accum time (s)} & \multicolumn{4}{c}{failure (\%)} \\
  & & & exec & plan & total & contact & drop & timeout & total \\
  \hline
  \parbox[t]{2mm}{\multirow{5}{*}{\rotatebox[origin=c]{90}{Sequential}}}
  & OMG Planner~\cite{wang2020manipulation} $\dagger$  & 62.78  & 8.275 & 1.481 & 9.755 & 30.56 & ~~3.89 & ~~2.78 & 37.22  \\
  & \gc Yang et al.~\cite{yang2021reactive} $\dagger$  & \gc 62.50 & \gc 4.808  & \gc 0.034
 & \gc 4.843 &  \gc 16.11  &  \gc  10.56   & \gc 10.83 & \gc 37.50 \\  
\cline{2-10}
  & GA-DDPG~\cite{wang2021goal}   & 55.00 & 7.145 & \textbf{0.129} & 7.274 & ~~\textbf{8.61} & 17.78 & 18.61 & 45.00  \\
  & \gc Ours w/o ft. & \gc 71.76  & \gc \textbf{7.045} & \gc 0.140
 & \gc   \textbf{7.185}        &  \gc  ~~8.80  &  \gc 14.72  & \gc ~~4.72 & \gc  28.24       \\  
  & Ours   & \textbf{72.96} & 7.101 & 0.144 & 7.245 & 11.29 & \textbf{12.69} & ~~\textbf{3.05}  & \textbf{27.04}    \\
\hline

  \parbox[t]{2mm}{\multirow{3}{*}{\rotatebox[origin=c]{90}{Simult.}}}
  
  & GA-DDPG~\cite{wang2021goal} & 28.33 & \textbf{4.747} & \textbf{0.133} & \textbf{4.881} & ~~9.17 & 34.44 & 28.06  & 71.67   \\
    & \gc Ours w/o ft. & \gc 64.81  & \gc 5.638  & \gc 0.144
 & \gc  5.783     &  \gc ~~\textbf{8.24} &  \gc 21.02  & \gc ~~5.93 & \gc 35.19        \\  
  & Ours  & \textbf{71.11} & 5.771 & 0.150 & 5.921 & 10.00 & \textbf{15.37} & ~~\textbf{3.61} & \textbf{28.89} \\
  
\bottomrule
  
 \end{tabular}
 }
 \caption{\textbf{Unseen Handedness.} Comparison of our method against various baselines from the HandoverSim benchmark \cite{chao2022handoversim} in the setting ``S2: Unseen Handedness''. The results of our method are averaged over 3 random seeds. $\dagger$: both methods \cite{wang2020manipulation,yang2021reactive} are evaluated with ground-truth states in \cite{chao2022handoversim} and are not directly comparable with ours.}
 \label{tab:s2}
\end{table*}

%% file: chapters/05_handover/handoversim2real/tables/table_s3.tex
\begin{table*}[t!]
 \centering
  \resizebox{1.0\columnwidth}{!}{
 \begin{tabular}{l|l|cccc|cccc}

  \hline
  \multicolumn{10}{c}{S3: Unseen Objects} \\
  \hline
  & & \multirow{2}{*}{success (\%)} & \multicolumn{3}{c|}{mean accum time (s)} & \multicolumn{4}{c}{failure (\%)} \\
  & & & exec & plan & total & contact & drop & timeout & total \\
  \hline
  \parbox[t]{2mm}{\multirow{5}{*}{\rotatebox[origin=c]{90}{Sequential}}}
  & OMG Planner~\cite{wang2020manipulation} $\dagger$ & 69.00 & 8.478 & 1.588 & 10.066 & 23.00 & ~~4.00 & ~~4.00 & 31.00   \\
  & \gc Yang et al.~\cite{yang2021reactive} $\dagger$  & \gc 62.00  & \gc 4.805 & \gc 0.031
 & \gc ~~4.837 &  \gc 18.00 &  \gc ~~9.00  & \gc 11.00 & \gc  38.00 \\  
\cline{2-10}
  & GA-DDPG~\cite{wang2021goal}   & 50.00 & \textbf{7.305} & \textbf{0.135} & ~~\textbf{7.440} & ~~\textbf{5.00} & 23.00 & 22.00 & 50.00       \\
  & \gc Ours w/o ft. & \gc  76.33 & \gc 7.410 & \gc 0.151
 & \gc  ~~7.565     &  \gc    ~~9.33     &  \gc   10.67   & \gc ~~\textbf{3.67} & \gc   23.67      \\  
  & Ours    & \textbf{79.67} & 7.499 & 0.156 & ~~7.656 & ~~6.33 & \textbf{10.33}  & ~~\textbf{3.67}  & \textbf{20.33}    \\
\hline

  \parbox[t]{2mm}{\multirow{3}{*}{\rotatebox[origin=c]{90}{Simult.}}}
  
  & GA-DDPG~\cite{wang2021goal} & 33.00 & \textbf{4.948} & \textbf{0.123} & ~~\textbf{5.071} &  10.00 & 33.00 & 24.00  & 67.00  \\
    & \gc Ours w/o ft. & \gc 72.00  & \gc 6.242 & \gc 0.168
 & \gc ~~6.410  &  \gc ~~7.33 &  \gc  13.67 & \gc ~~7.00 & \gc 28.00       \\  
  & Ours   & \textbf{75.67} & 6.153 & 0.160 & ~~6.314 & ~~\textbf{5.00} & \textbf{13.33} & ~~\textbf{6.00} & \textbf{24.33}  \\
  
\bottomrule
  
 \end{tabular}
 }
 \caption{\textbf{Unseen Object Evaluation.} Comparison of our method against baselines from the HandoverSim benchmark \cite{chao2022handoversim} in the setting ``S3: Unseen Objects''. The results of our method are averaged over 3 random seeds. $\dagger$: both methods \cite{wang2020manipulation,yang2021reactive} are evaluated with ground-truth states in \cite{chao2022handoversim} and are not directly comparable with ours.}
 \label{tab:s3}
\end{table*}

%% file: chapters/05_handover/handoversim2real/tables/table_ablations.tex
\begin{table}[t!]
 \vspace{-0mm}
 \centering
 \small

  \resizebox{1.0\columnwidth}{!}{%
 \begin{tabular}{l|cccc}
  \hline
  \multicolumn{5}{c}{Ablation Study} \\
  \hline
  & \multirow{2}{*}{success (\%)} &  \multicolumn{3}{c}{failure (\%)} \\
  & & contact & drop & timeout \\
  \hline
 w/ RGBDM + ResNet18 & 34.10  & ~~\textbf{6.20} &  45.80     &  13.90          \\
  \rowcolor{Gray}

    w/ third person view & 60.42  & ~~9,95 &  25.69     &  ~~3.94          \\

   w/o hand point cloud & 59.03 & 24.07 &  \textbf{11.58}  &  ~~5.32         \\

     \rowcolor{Gray}
   w/o aux prediction & 70.60  & 10.65 &  16.20     & ~~2.54          \\

   w/o standoff  & 52.55  & ~~7.87 &  36.80     &  ~~2.78          \\
      \rowcolor{Gray}
  w/o finetuning & 73.38 & ~~9.03 & 13.89 & ~~3.70      \\
  

  Ours & \textbf{75.23} & ~~9.26 & 13.43 & ~~\textbf{2.08} \\ 
  \hline
    \rowcolor{Gray}
  w/o finetuning simult. & 62.27 & 11.81 & 20.37 & ~~5.56      \\

  Ours simult. & \textbf{68.75} & \textbf{8.8} & \textbf{17.82} & ~~\textbf{4.63} \\ 
  \hline
 \end{tabular}
 }

 \caption{\textbf{Ablation.} We ablate the vision backbone, hand perception, and egocentric view. We also study the effect of finetuning, the auxiliary prediction, and splitting the task into two phases. All design choices are crucial aspects of our method with regards to overall performance. Results are averaged over 3 random seeds.
 }
 \vspace{-0mm}
 \label{tab:h2r:ablation}
\end{table}

%% file: chapters/05_handover/handoversim2real/tables/table_kinect.tex
\begin{table}[t]
 \vspace{-0mm}
 \centering
 \small
  \resizebox{0.8\columnwidth}{!}{%
 \begin{tabular}{l|cccc}
  \hline
  & \multirow{2}{*}{success (\%)} &  \multicolumn{3}{c}{failure (\%)} \\
  & & contact & drop & timeout \\
  \hline

   w/ Kinect noise & 59.49 & 13.19 &  19.68 &  ~~7.64         \\ 

  \rowcolor{Gray}
  Ours & \textbf{75.23} & ~~\textbf{9.26} & \textbf{13.43} & ~~\textbf{2.08} \\ 
\hline
 \end{tabular}
 }
 \vspace{-2mm}
 \caption{We analyze the effect of simulated Kinect noise~\cite{handa2014benchmark} on our model. Results are averaged over 3 random seeds.
 }
 \label{tab:h2r:kinect}
\end{table}

%% file: chapters/05_handover/handoversim2real/tables/table_sim2sim_singlecol.tex
\begin{table}[t!]

 \centering
 \small

 \resizebox{1.0\columnwidth}{!}{%
 \begin{tabular}{l|l|cccc}
  \hline
  \multicolumn{6}{c}{Sim-to-Sim} \\
  \hline
  & & \multirow{2}{*}{success (\%)} &  \multicolumn{3}{c}{failure (\%)} \\
  & & & contact & drop & timeout  \\
  \hline
  \parbox[t]{2mm}{\multirow{4}{*}{\rotatebox[origin=c]{90}{Sequential}}}

  & GA-DDPG \cite{wang2021goal} &  19.44         & ~~\textbf{4.86} & 47.22          & 28.47      \\
  & \gc GA-DDPG \cite{wang2021goal} finetuned  & \gc 11.81      & \gc ~~6.25 &  \gc 68.75          &  \gc 13.19      \\
  
  &  Ours & 44.21  & ~~9.49 & 40.51 & ~~5.79   \\ 
  
  & \gc Ours w/o grasp pred. & \gc \textbf{54.40} & \gc ~~7.87 & \gc \textbf{33.34} & \gc ~~\textbf{4.40}  \\

  \hline

  \parbox[t]{2mm}{\multirow{4}{*}{\rotatebox[origin=c]{90}{Simult.}}}
  
  & GA-DDPG \cite{wang2021goal} & 11.11           & 15.97          & 48.61          & 24.31       \\
  & \gc GA-DDPG \cite{wang2021goal} finetuned  & \gc 16.67   & \gc ~~9.72 &  \gc 63.89          & \gc ~~9.72     \\
 
  & Ours    & 39.58 & ~~\textbf{9.03} & 43.75  & ~~7.64    \\
  & \gc Ours w/o grasp pred. & \gc \textbf{47.92} & \gc 10.65 & \gc \textbf{35.88} & \gc ~~\textbf{5.56}   \\ 
  
\bottomrule
  
 \end{tabular}
 }

 \caption{\textbf{Sim-to-Sim Experiment}. We evaluate sim-to-sim transfer of the learning-based method to Isaac Gym \cite{makoviychuk2021isaac}, Our method shows better transfer capabilities than GA-DDPG \cite{wang2021goal}. }

 \label{tab:h2r:sim2sim_singlecol}
\end{table}

%% file: chapters/05_handover/handoversim2real/figures/pilot_overview_l.tex
\begin{figure*}[t!]
 \centering
  \begin{minipage}{0.495\linewidth}
  \centering
  011\_banana
  \\~\\ \vspace{-3mm}
  \begin{minipage}{0.320\linewidth}
   \centering
   \includegraphics[width=\linewidth,trim={270 264 420 170},clip]{\dir/handoversim2real/figures/supp_pilot_compressed/1_01_1_IMG_9578.JPEG}
   \\ \vspace{-1mm}
   pose 1
  \end{minipage}~
  \begin{minipage}{0.320\linewidth}
   \centering
   \includegraphics[width=\linewidth,trim={270 264 420 170},clip]{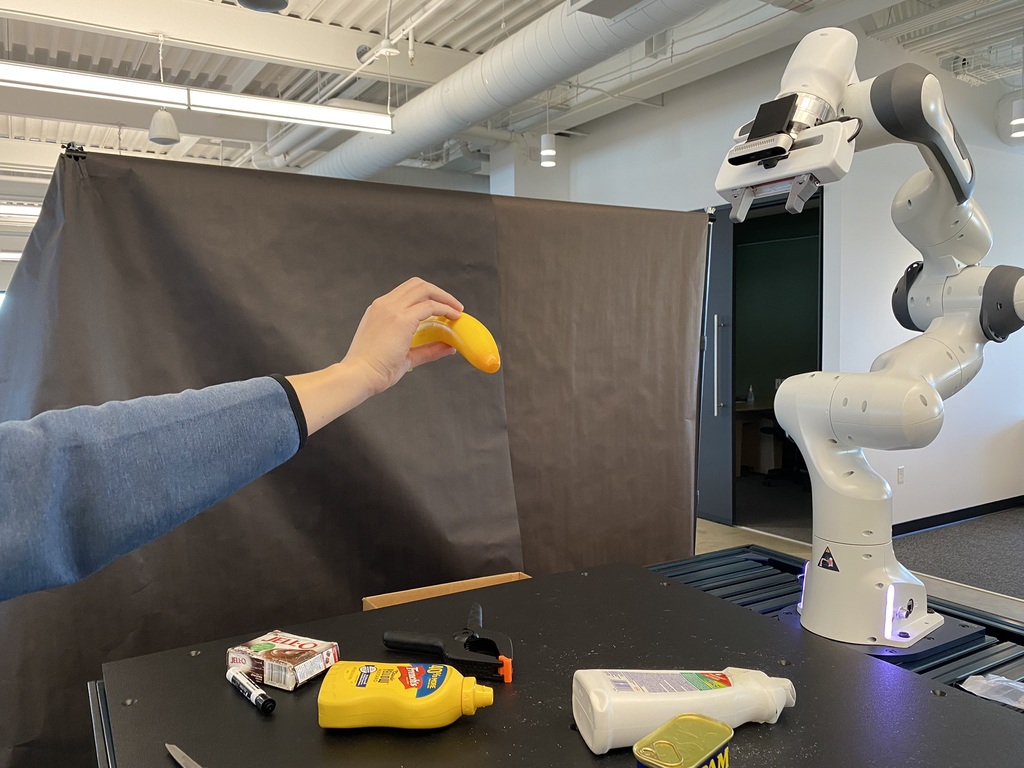}
   \\ \vspace{-1mm}
   pose 2
  \end{minipage}~
  \begin{minipage}{0.320\linewidth}
   \centering
   \includegraphics[width=\linewidth,trim={250 244 440 190},clip]{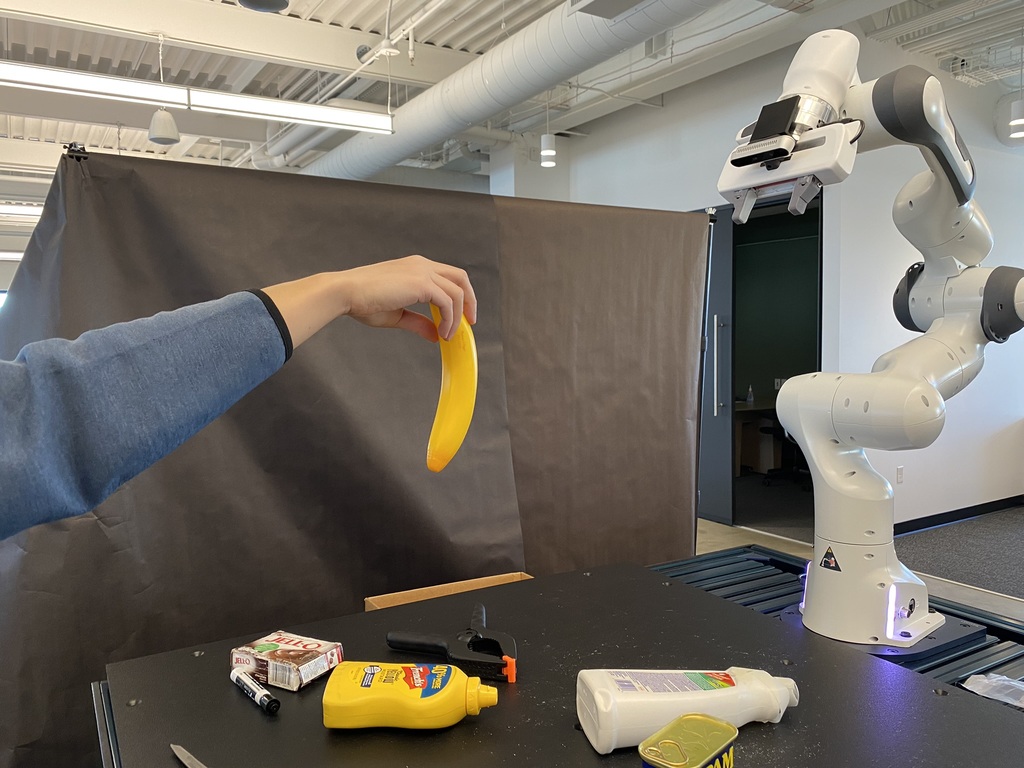}
   \\ \vspace{-1mm}
   pose 3
  \end{minipage}
  \\ \vspace{2mm}
 \end{minipage}~
 \begin{minipage}{0.495\linewidth}
  \centering
  037\_scissors
  \\~\\ \vspace{-3mm}
  \begin{minipage}{0.320\linewidth}
   \centering
   \includegraphics[width=\linewidth,trim={250 204 440 230},clip]{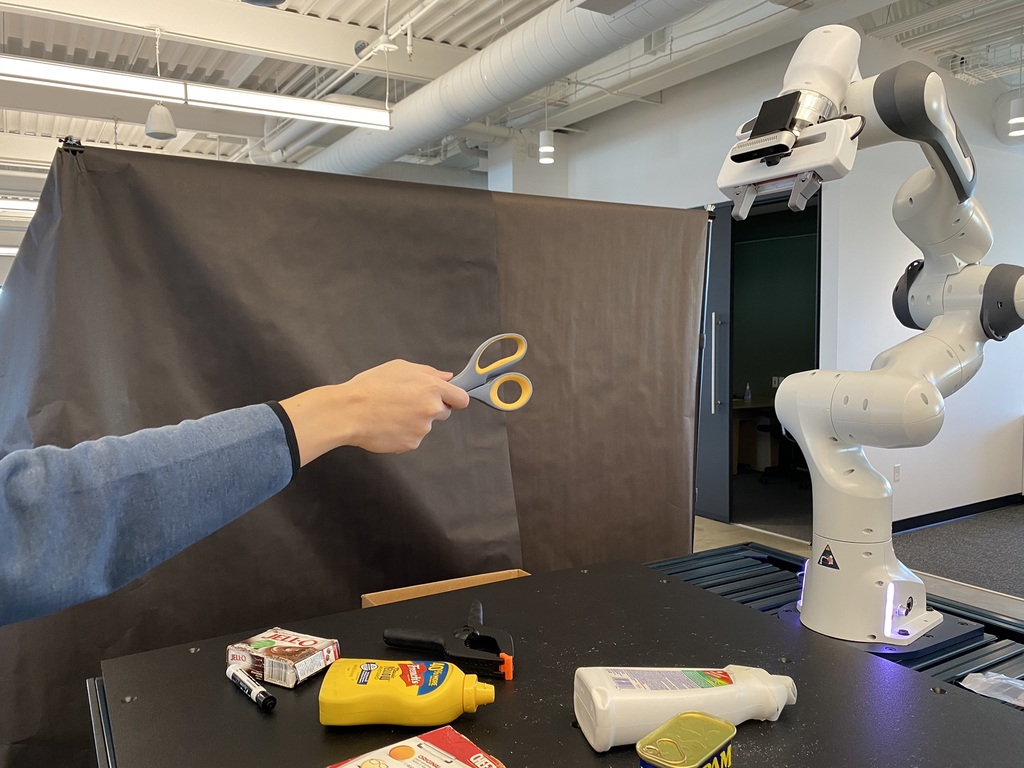}
   \\ \vspace{-1mm}
   pose 1
  \end{minipage}~
  \begin{minipage}{0.320\linewidth}
   \centering
   \includegraphics[width=\linewidth,trim={250 244 440 190},clip]{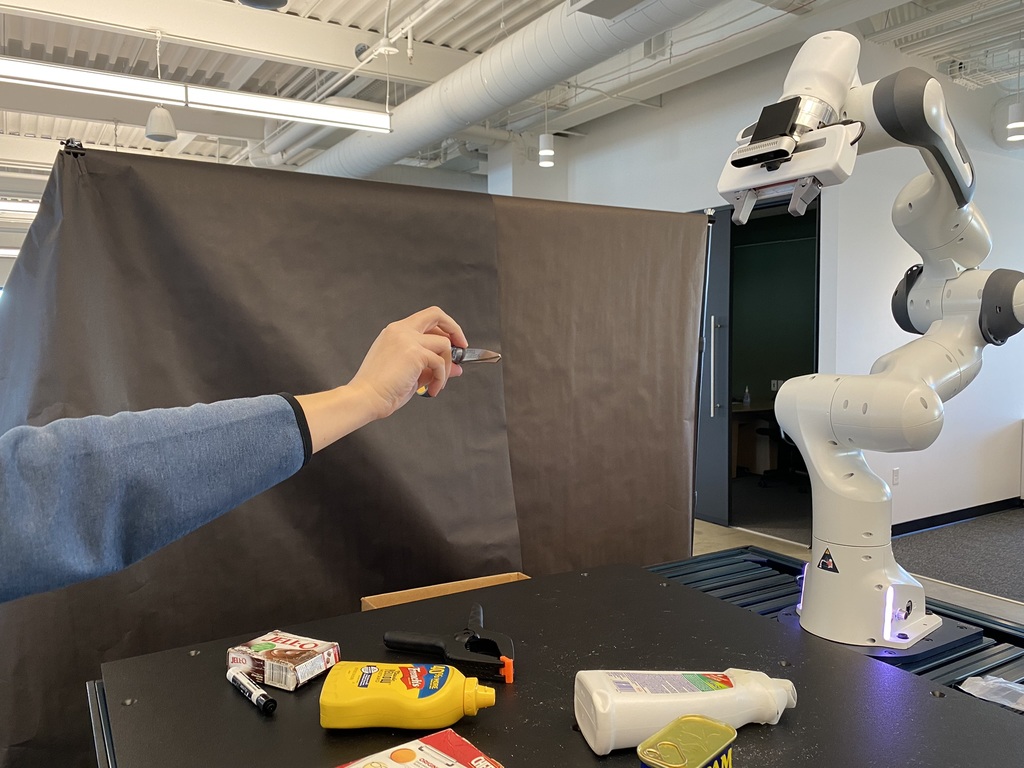}
   \\ \vspace{-1mm}
   pose 2
  \end{minipage}~
  \begin{minipage}{0.320\linewidth}
   \centering
   \includegraphics[width=\linewidth,trim={250 264 440 170},clip]{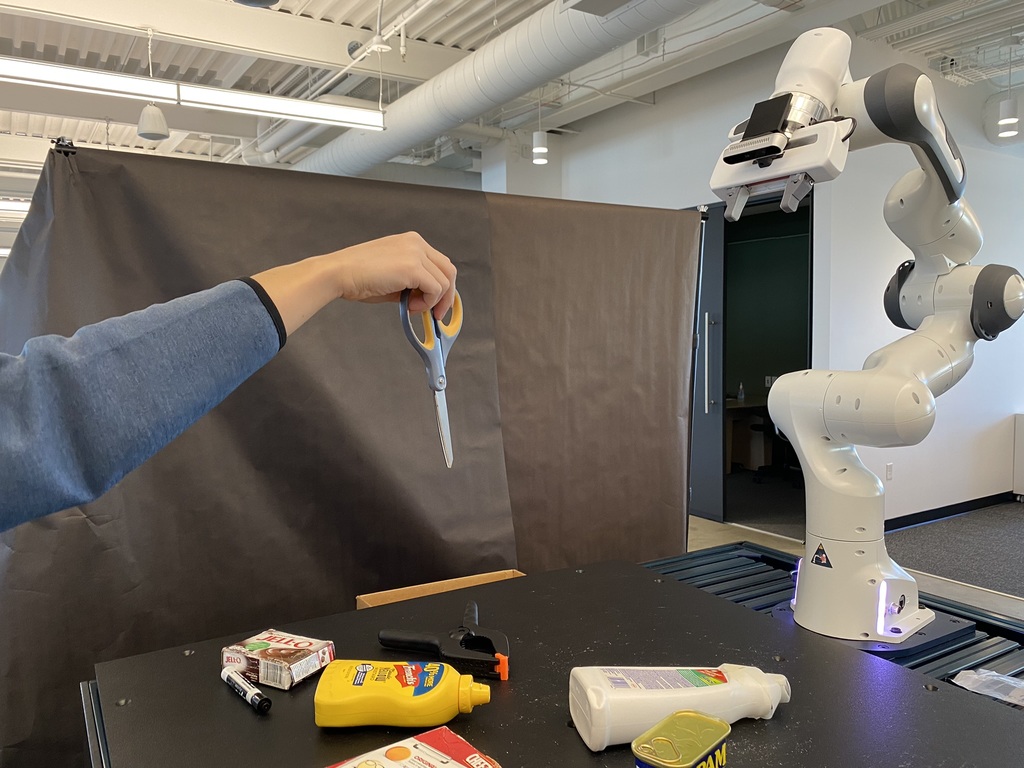}
   \\ \vspace{-1mm}
   pose 3
  \end{minipage}
  \\ \vspace{2mm}
 \end{minipage}
 \\ \vspace{3mm}
 \begin{minipage}{0.495\linewidth}
  \centering
  006\_mustard\_bottle
  \\~\\ \vspace{-3mm}
  \begin{minipage}{0.320\linewidth}
   \centering
   \includegraphics[width=\linewidth,trim={270 204 420 230},clip]{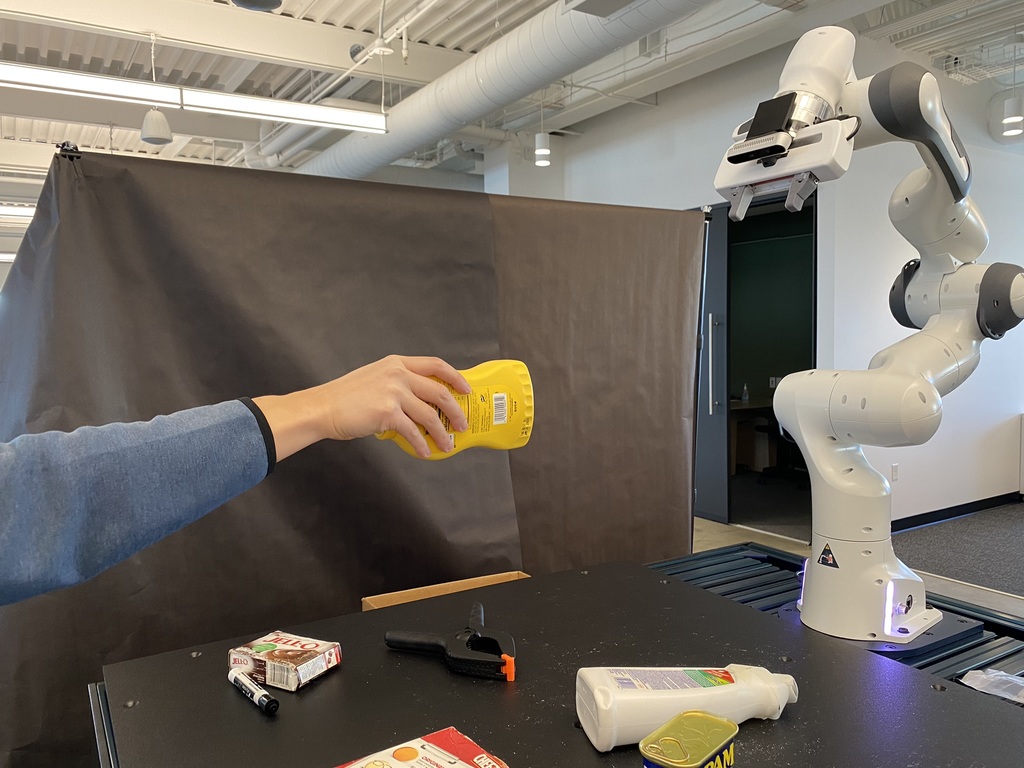}
   \\ \vspace{-1mm}
   pose 1
  \end{minipage}~
  \begin{minipage}{0.320\linewidth}
   \centering
   \includegraphics[width=\linewidth,trim={270 284 420 150},clip]{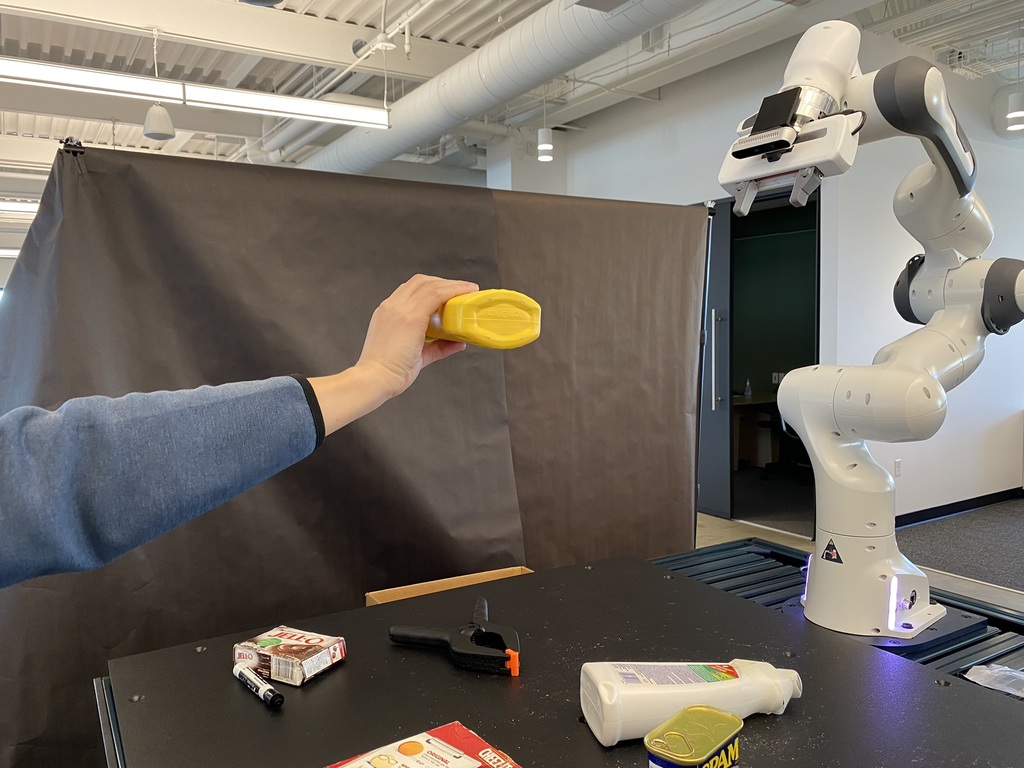}
   \\ \vspace{-1mm}
   pose 2
  \end{minipage}~
  \begin{minipage}{0.320\linewidth}
   \centering
   \includegraphics[width=\linewidth,trim={270 264 420 170},clip]{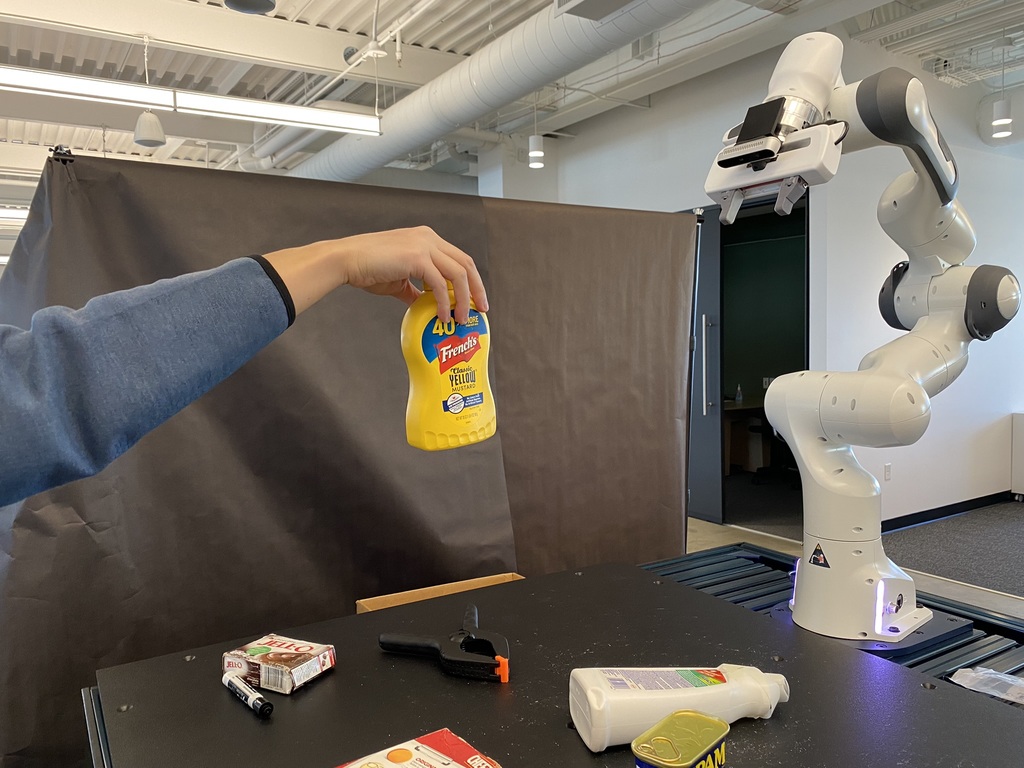}
   \\ \vspace{-1mm}
   pose 3
  \end{minipage}
  \\ \vspace{2mm}
 \end{minipage}~
 \begin{minipage}{0.495\linewidth}
  \centering
  024\_bowl
  \\~\\ \vspace{-3mm}
  \begin{minipage}{0.320\linewidth}
   \centering
   \includegraphics[width=\linewidth,trim={290 164 400 270},clip]{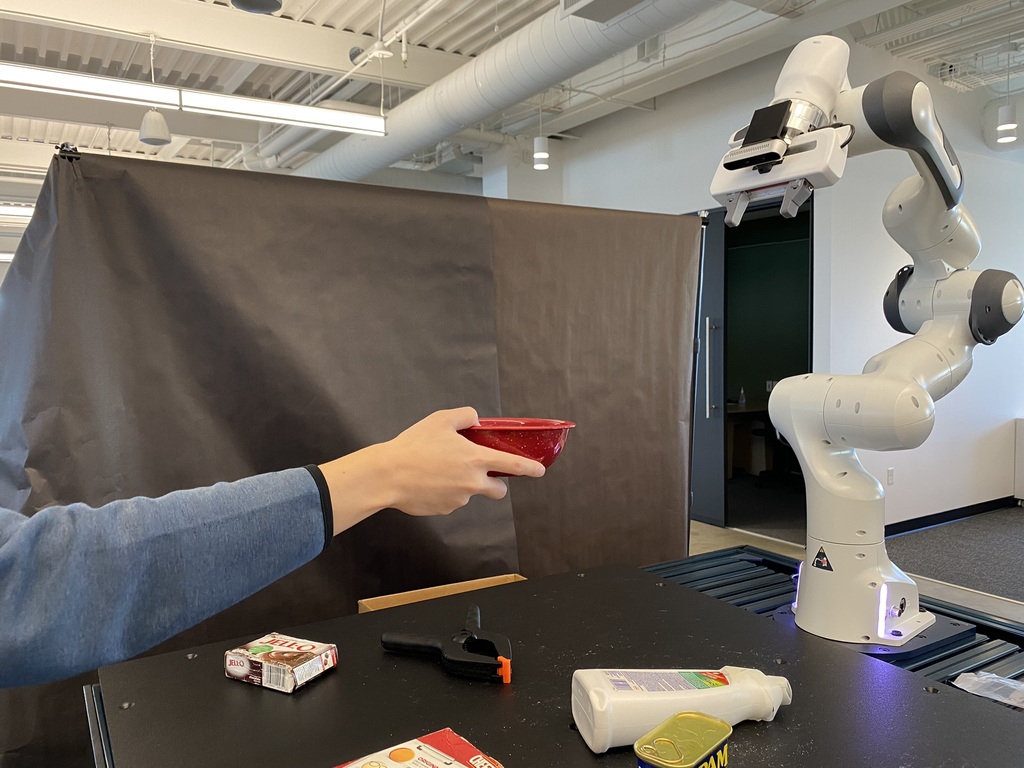}
   \\ \vspace{-1mm}
   pose 1
  \end{minipage}~
  \begin{minipage}{0.320\linewidth}
   \centering
   \includegraphics[width=\linewidth,trim={270 304 420 130},clip]{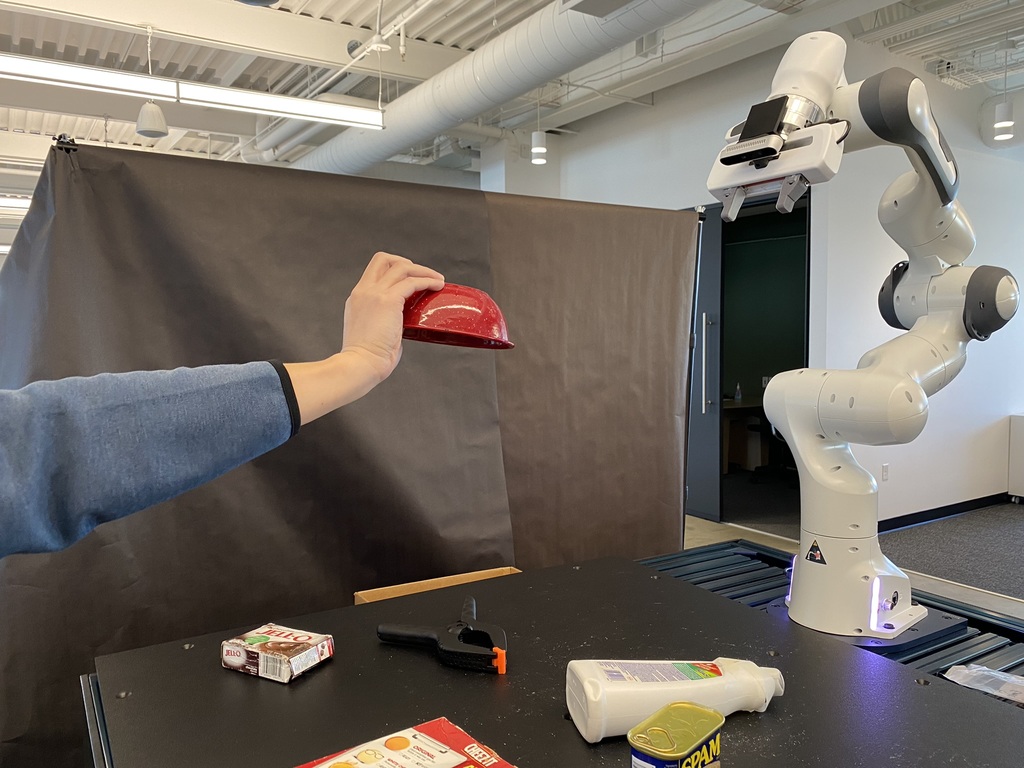}
   \\ \vspace{-1mm}
   pose 2
  \end{minipage}~
  \begin{minipage}{0.320\linewidth}
   \centering
   \includegraphics[width=\linewidth,trim={270 304 420 130},clip]{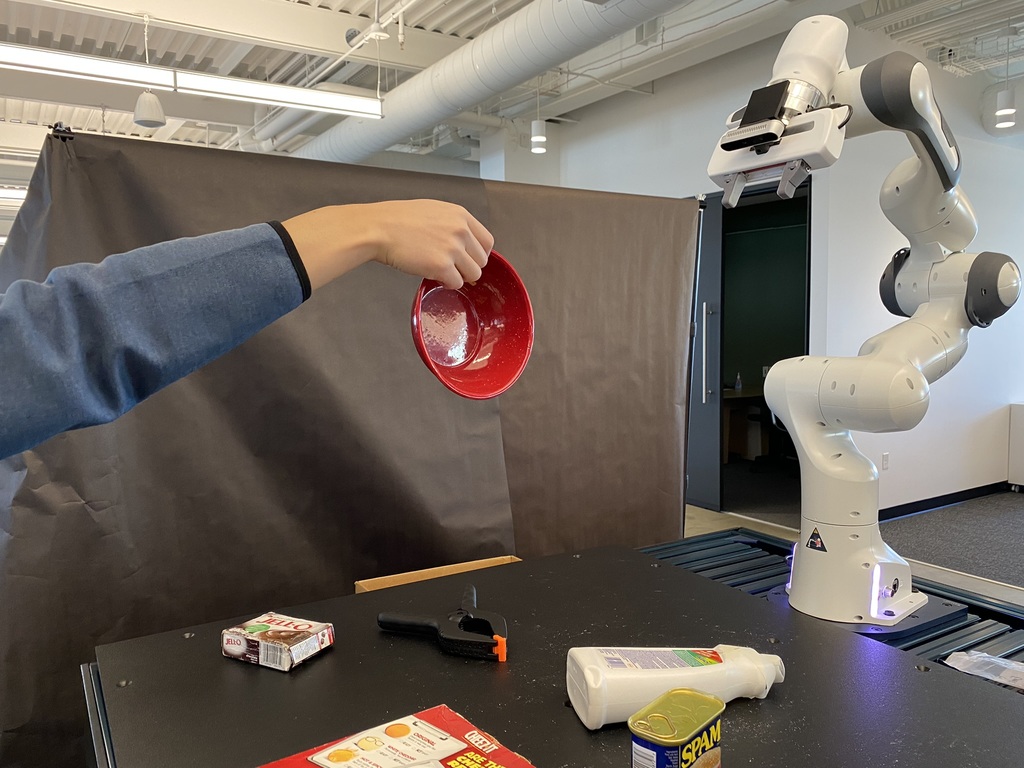}
   \\ \vspace{-1mm}
   pose 3
  \end{minipage}
  \\ \vspace{2mm}
 \end{minipage}
 \\ \vspace{3mm}
 \begin{minipage}{0.495\linewidth}
  \centering
  040\_large\_marker
  \\~\\ \vspace{-3mm}
  \begin{minipage}{0.320\linewidth}
   \centering
   \includegraphics[width=\linewidth,trim={270 224 420 210},clip]{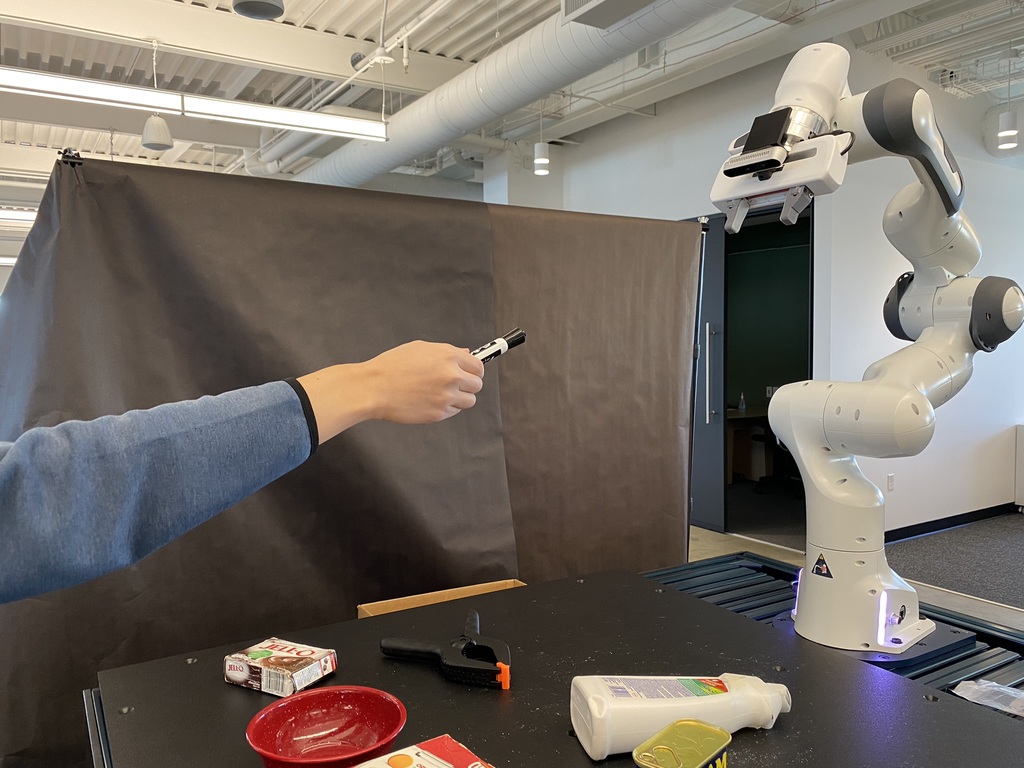}
   \\ \vspace{-1mm}
   pose 1
  \end{minipage}~
  \begin{minipage}{0.320\linewidth}
   \centering
   \includegraphics[width=\linewidth,trim={270 284 420 150},clip]{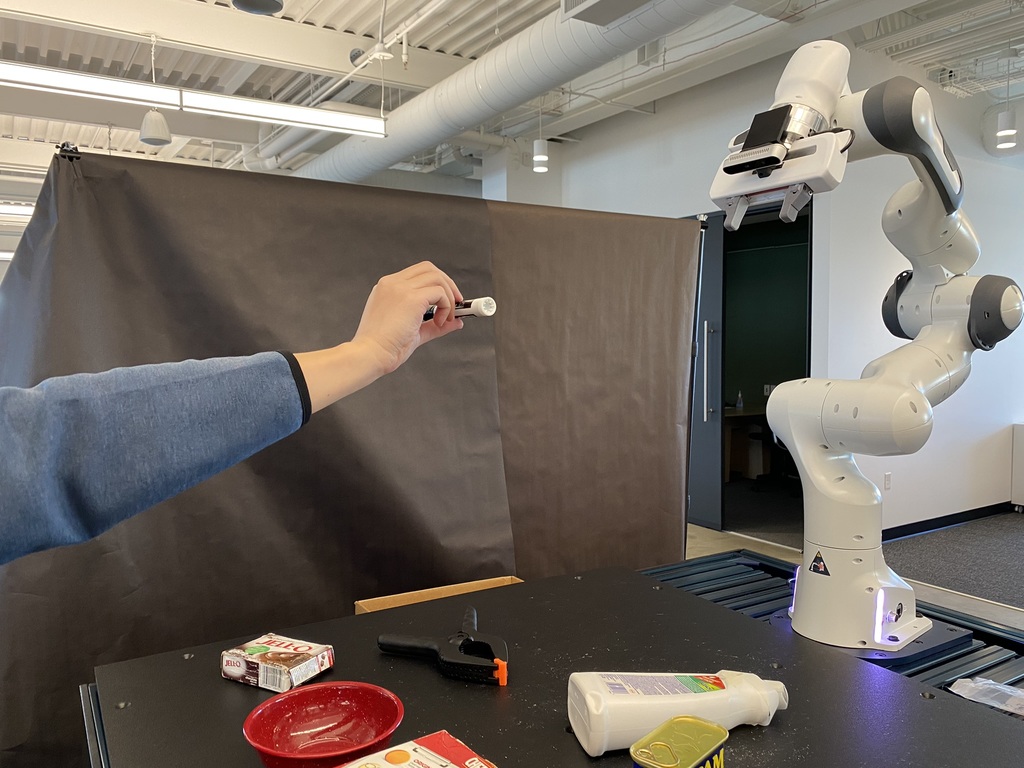}
   \\ \vspace{-1mm}
   pose 2
  \end{minipage}~
  \begin{minipage}{0.320\linewidth}
   \centering
   \includegraphics[width=\linewidth,trim={270 324 420 110},clip]{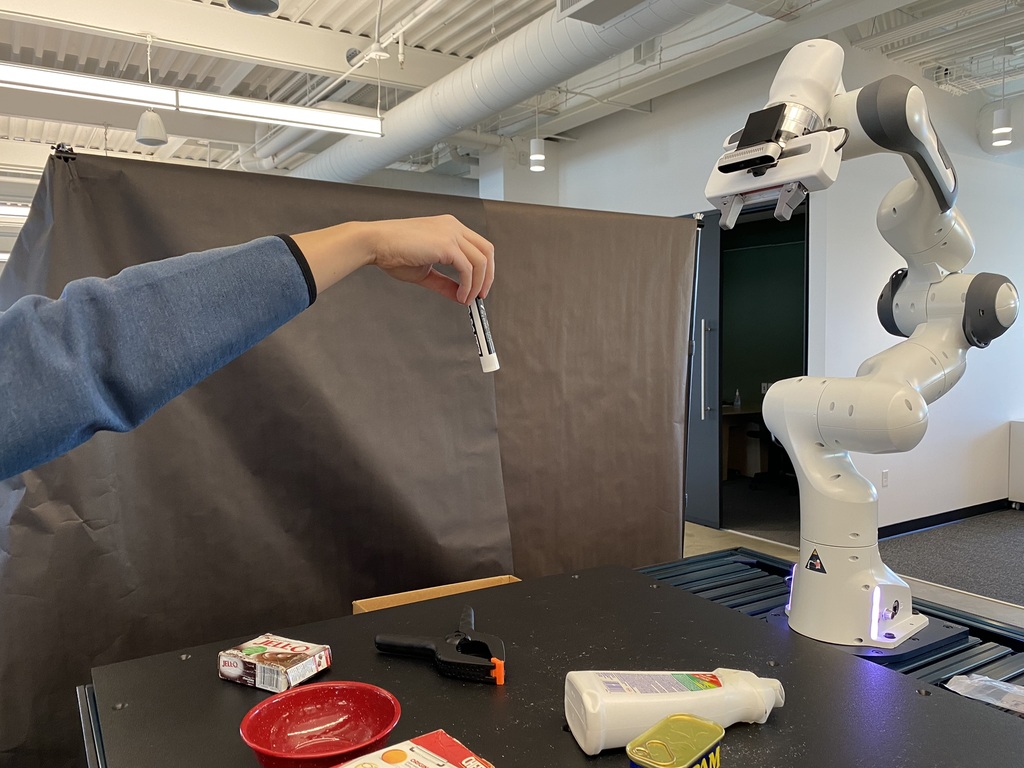}
   \\ \vspace{-1mm}
   pose 3
  \end{minipage}
  \\ \vspace{2mm}
 \end{minipage}~
 \begin{minipage}{0.495\linewidth}
  \centering
  003\_cracker\_box
  \\~\\ \vspace{-3mm}
  \begin{minipage}{0.320\linewidth}
   \centering
   \includegraphics[width=\linewidth,trim={270 264 420 170},clip]{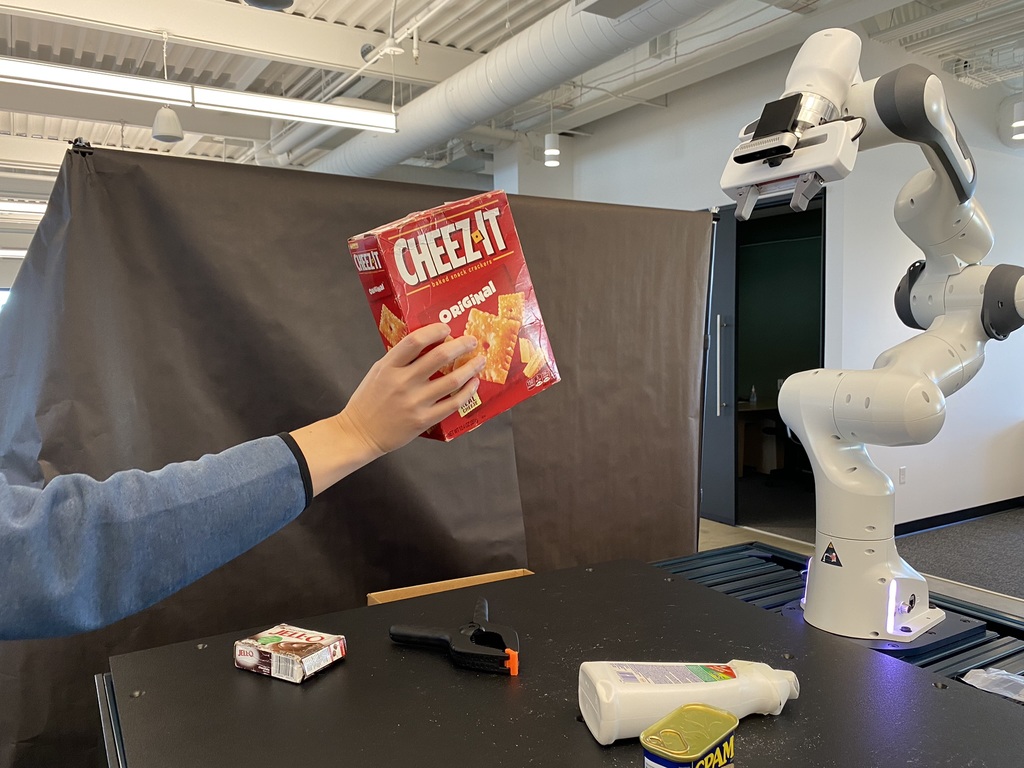}
   \\ \vspace{-1mm}
   pose 1
  \end{minipage}~
  \begin{minipage}{0.320\linewidth}
   \centering
   \includegraphics[width=\linewidth,trim={330 264 360 170},clip]{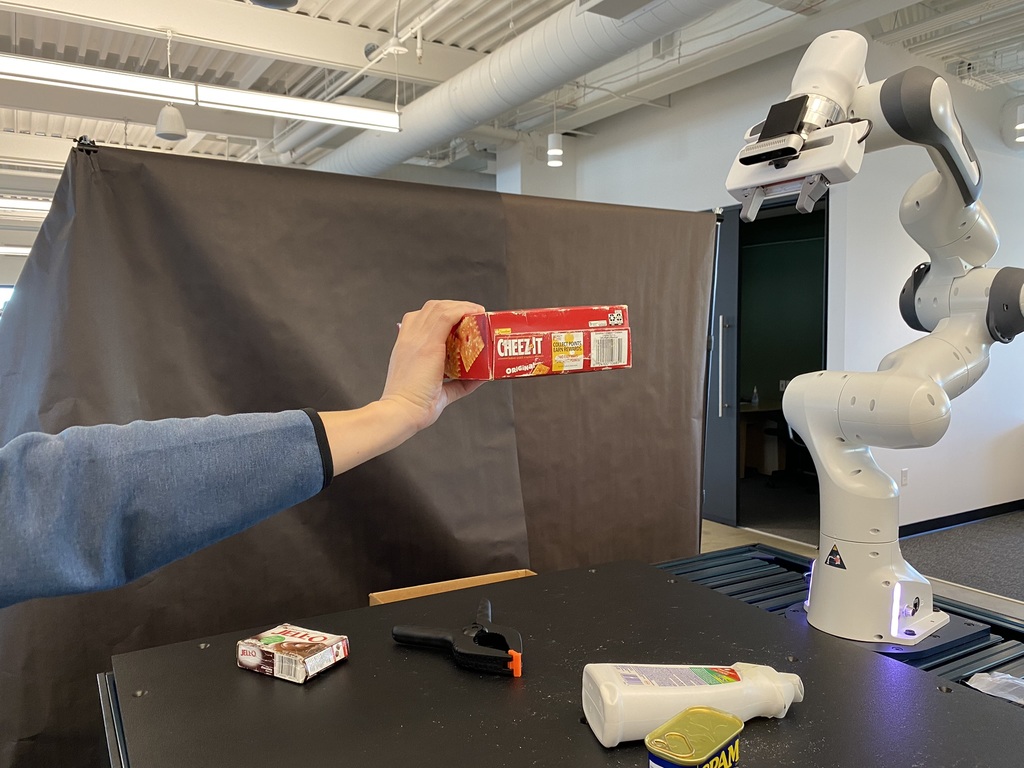}
   \\ \vspace{-1mm}
   pose 2
  \end{minipage}~
  \begin{minipage}{0.320\linewidth}
   \centering
   \includegraphics[width=\linewidth,trim={270 224 420 210},clip]{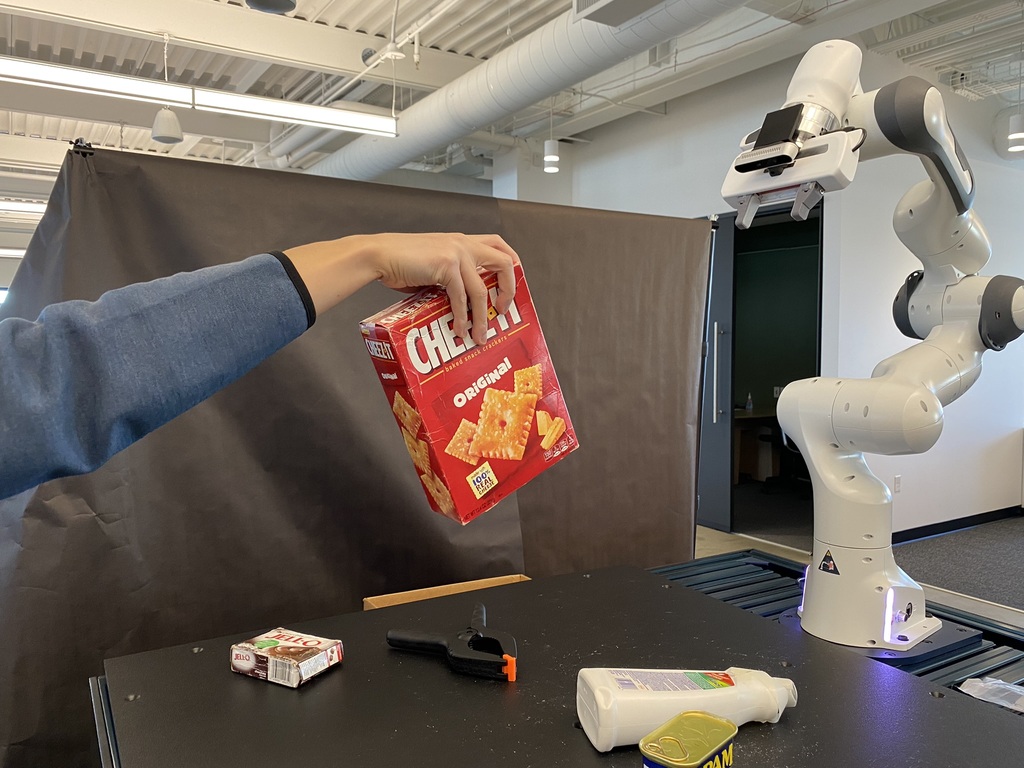}
   \\ \vspace{-1mm}
   pose 3
  \end{minipage}
  \\ \vspace{2mm}
 \end{minipage}
  \\ \vspace{2mm}
 \begin{minipage}{0.495\linewidth}
  \centering
  052\_extra\_large\_clamp
  \\~\\ \vspace{-3mm}
  \begin{minipage}{0.320\linewidth}
   \centering
   \includegraphics[width=\linewidth,trim={310 204 380 230},clip]{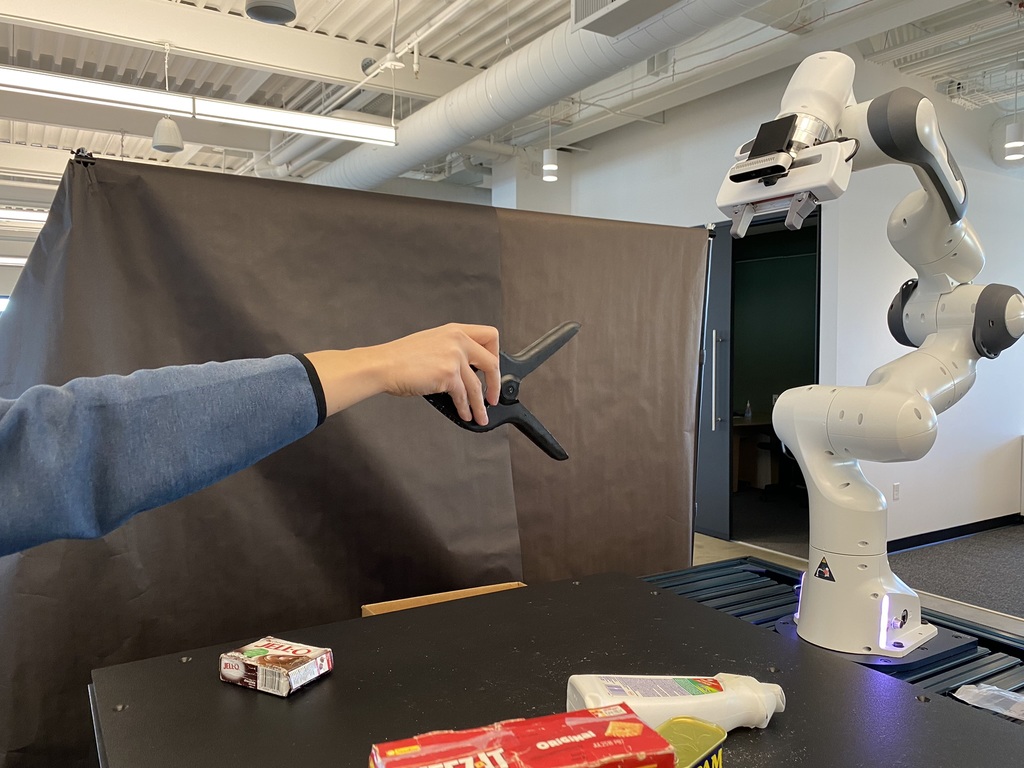}
   \\ \vspace{-1mm}
   pose 1
  \end{minipage}~
  \begin{minipage}{0.320\linewidth}
   \centering
   \includegraphics[width=\linewidth,trim={310 224 380 210},clip]{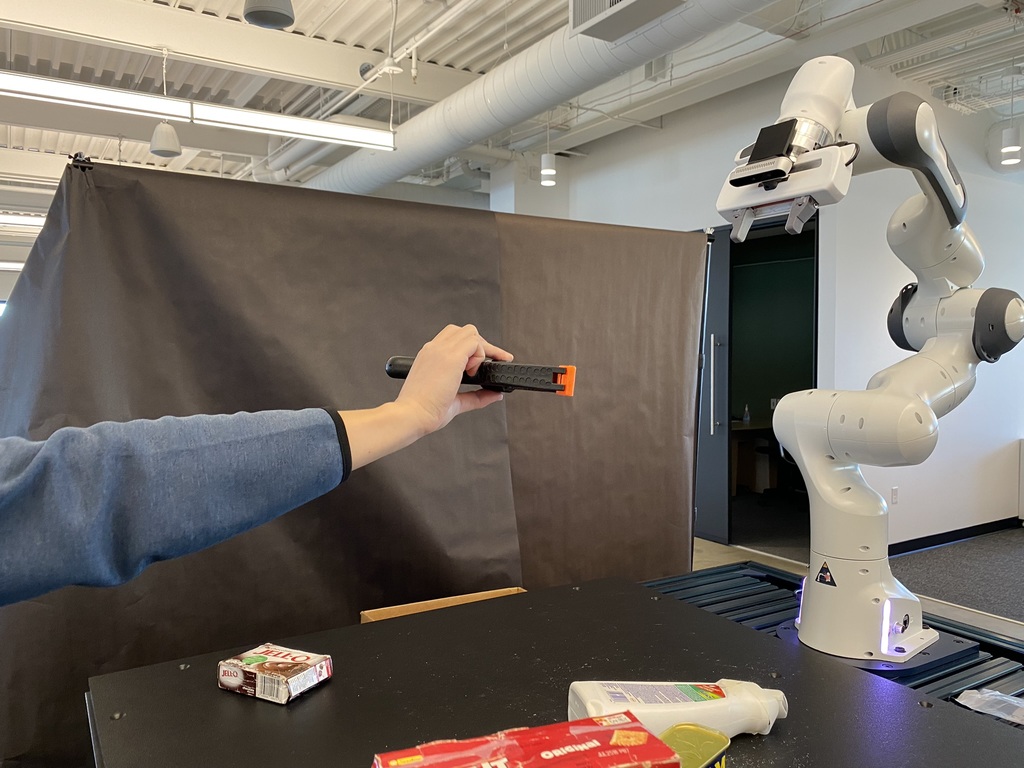}
   \\ \vspace{-1mm}
   pose 2
  \end{minipage}~
  \begin{minipage}{0.320\linewidth}
   \centering
   \includegraphics[width=\linewidth,trim={290 264 400 170},clip]{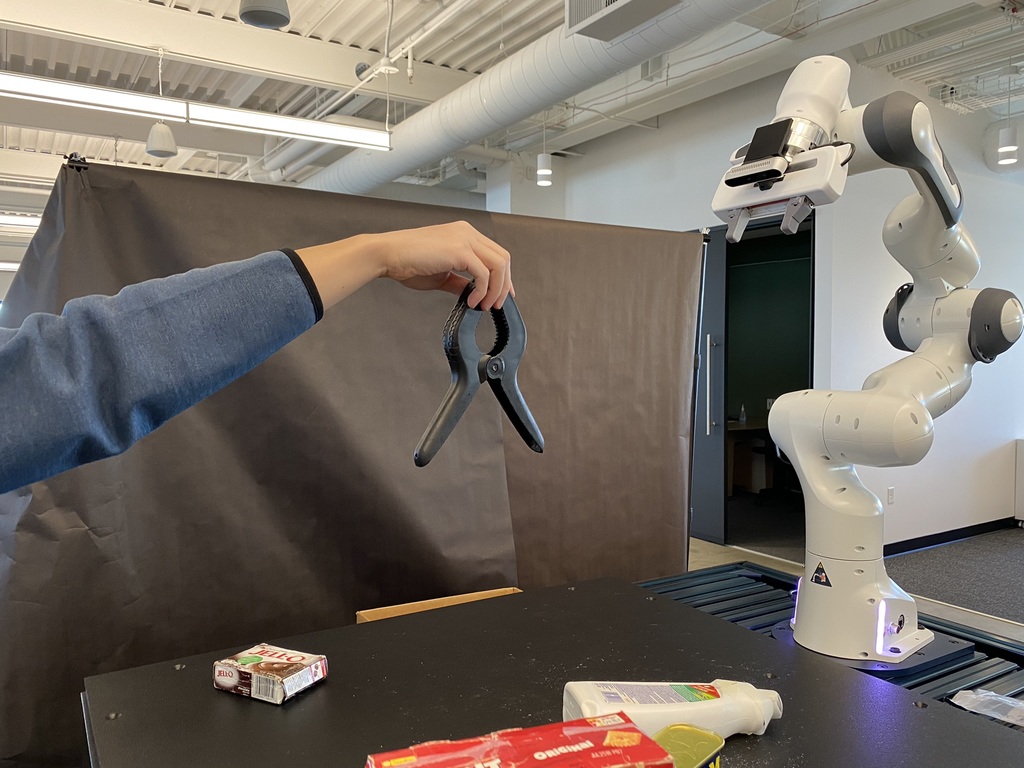}
   \\ \vspace{-1mm}
   pose 3
  \end{minipage}
  \\ \vspace{2mm}
 \end{minipage}~
 \begin{minipage}{0.495\linewidth}
  \centering
  008\_pudding\_box
  \\~\\ \vspace{-3mm}
  \begin{minipage}{0.320\linewidth}
   \centering
   \includegraphics[width=\linewidth,trim={270 174 420 260},clip]{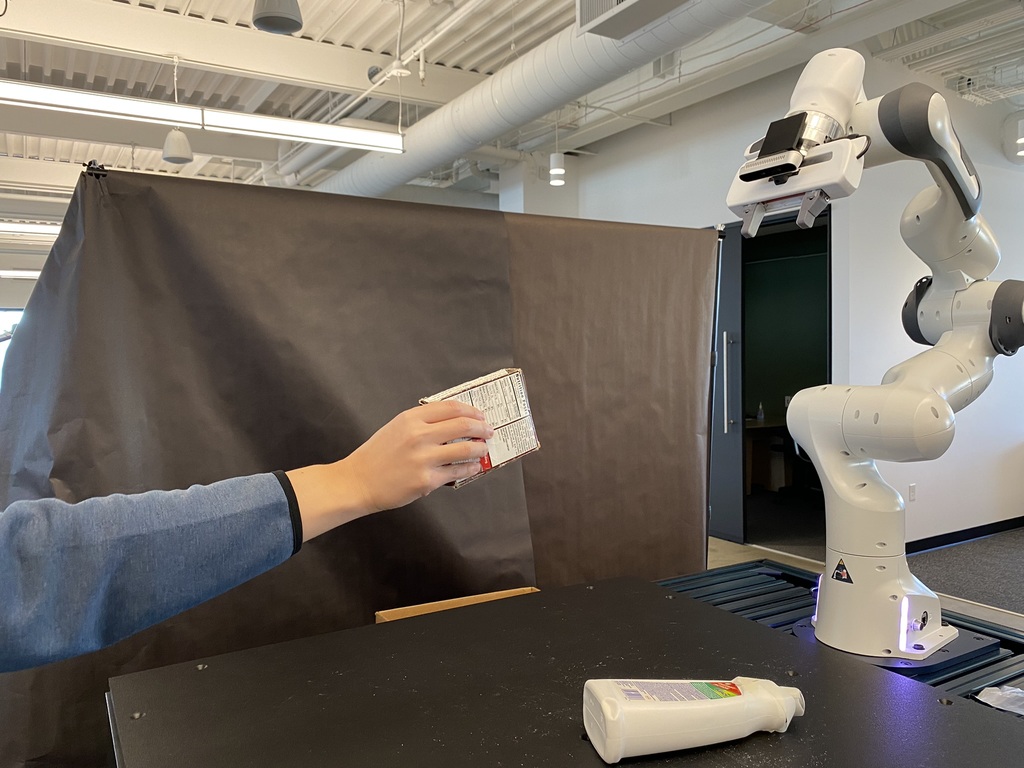}
   \\ \vspace{-1mm}
   pose 1
  \end{minipage}~
  \begin{minipage}{0.320\linewidth}
   \centering
   \includegraphics[width=\linewidth,trim={270 224 420 210},clip]{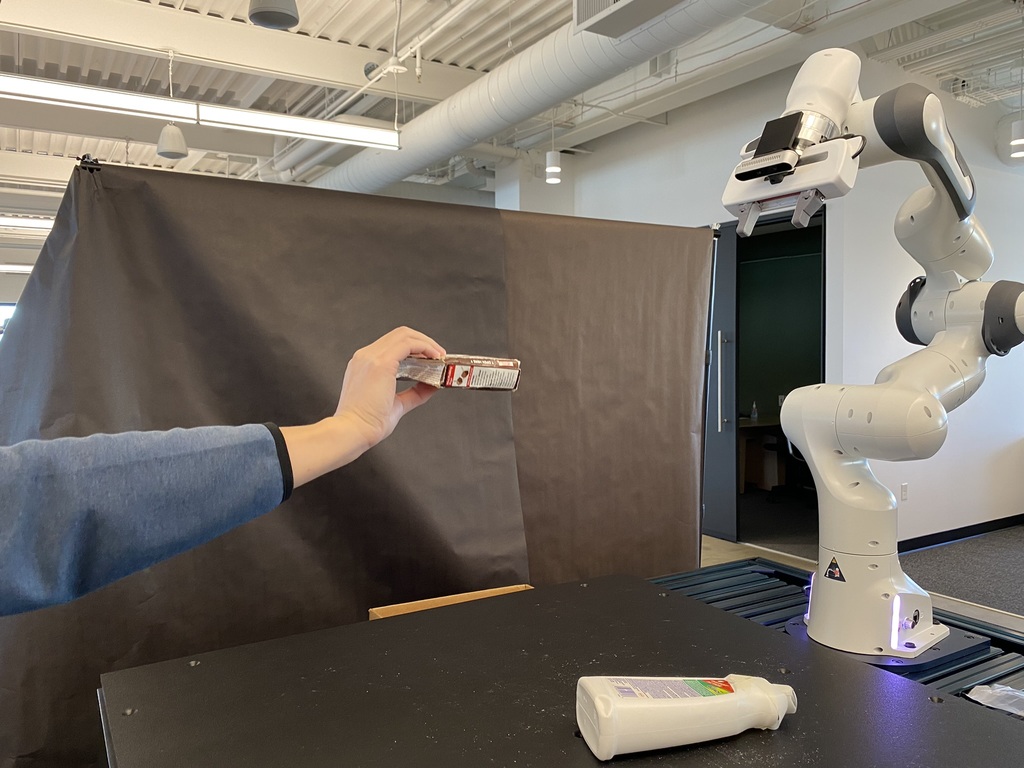}
   \\ \vspace{-1mm}
   pose 2
  \end{minipage}~
  \begin{minipage}{0.320\linewidth}
   \centering
   \includegraphics[width=\linewidth,trim={270 264 420 170},clip]{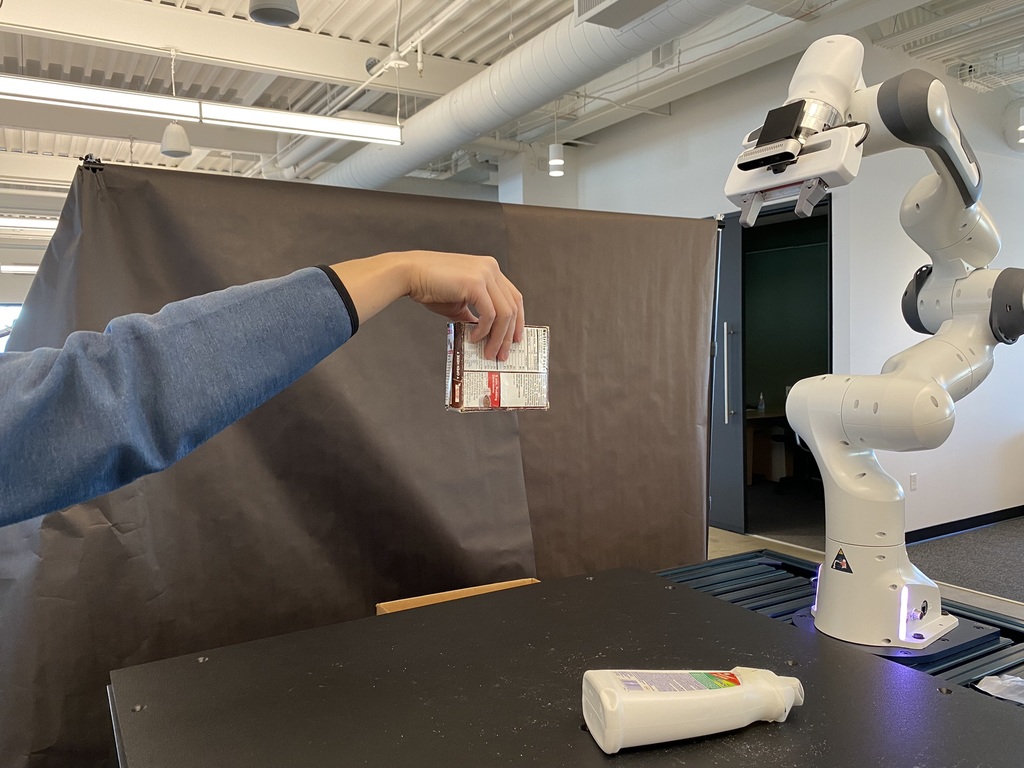}
   \\ \vspace{-1mm}
   pose 3
  \end{minipage}
  \\ \vspace{2mm}
 \end{minipage}
 \begin{minipage}{0.495\linewidth}
  \centering
  010\_potted\_meat\_can
  \\~\\ \vspace{-3mm}
  \begin{minipage}{0.320\linewidth}
   \centering
   \includegraphics[width=\linewidth,trim={280 254 430 200},clip]{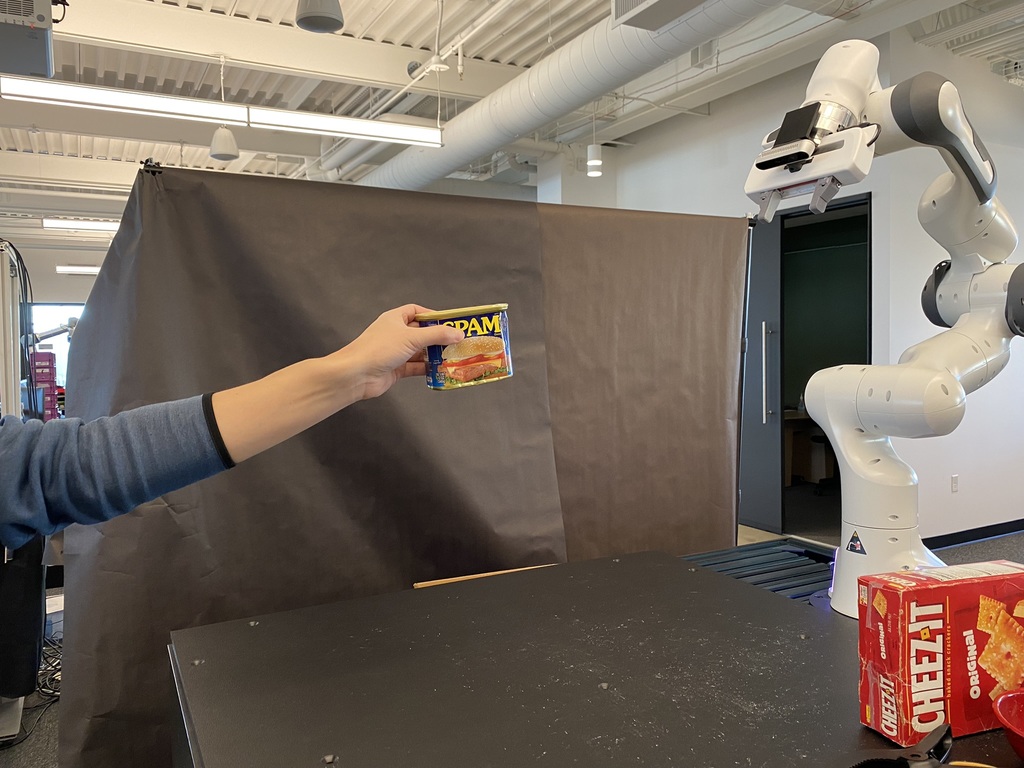}
   \\ \vspace{-1mm}
   pose 1
  \end{minipage}~
  \begin{minipage}{0.320\linewidth}
   \centering
   \includegraphics[width=\linewidth,trim={300 274 410 180},clip]{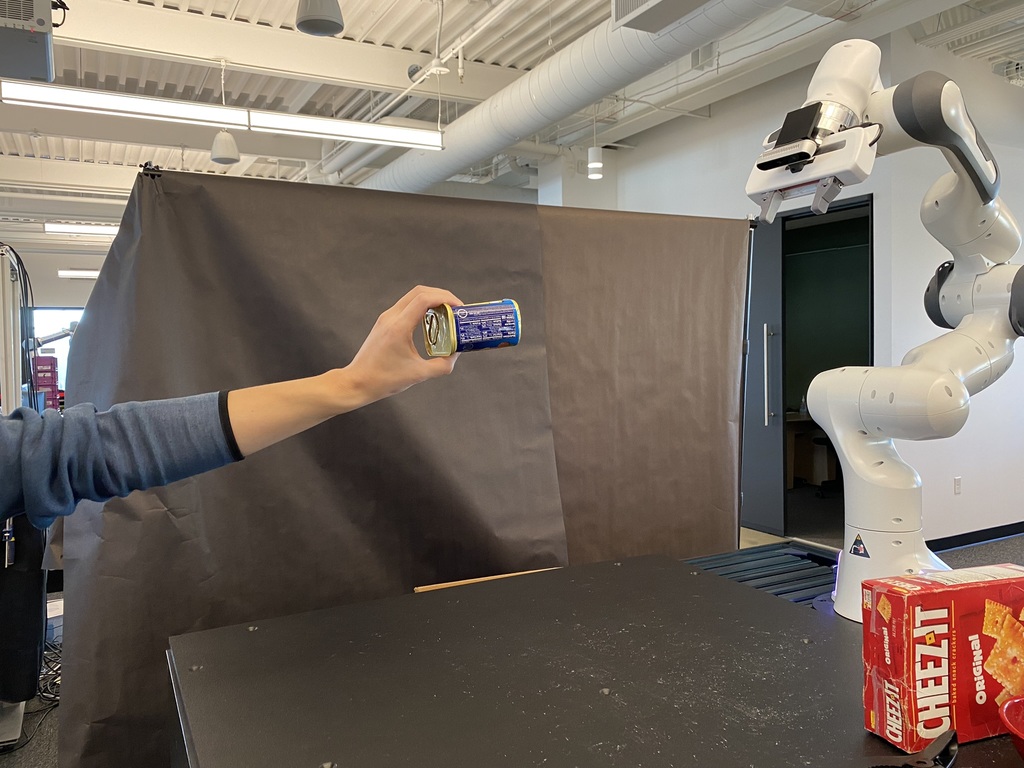}
   \\ \vspace{-1mm}
   pose 2
  \end{minipage}~
  \begin{minipage}{0.320\linewidth}
   \centering
   \includegraphics[width=\linewidth,trim={320 274 390 180},clip]{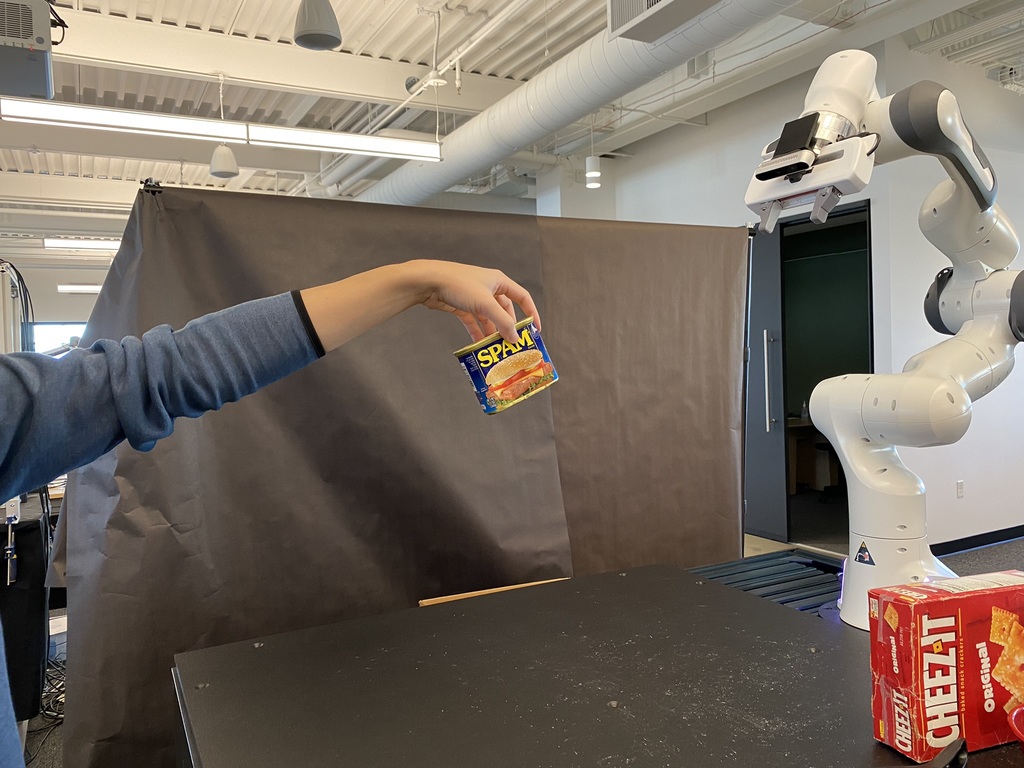}
   \\ \vspace{-1mm}
   pose 3
  \end{minipage}
  \\ \vspace{2mm}
 \end{minipage}~
 \begin{minipage}{0.495\linewidth}
  \centering
  021\_bleach\_cleanser
  \\~\\ \vspace{-3mm}
  \begin{minipage}{0.320\linewidth}
   \centering
   \includegraphics[width=\linewidth,trim={300 254 410 200},clip]{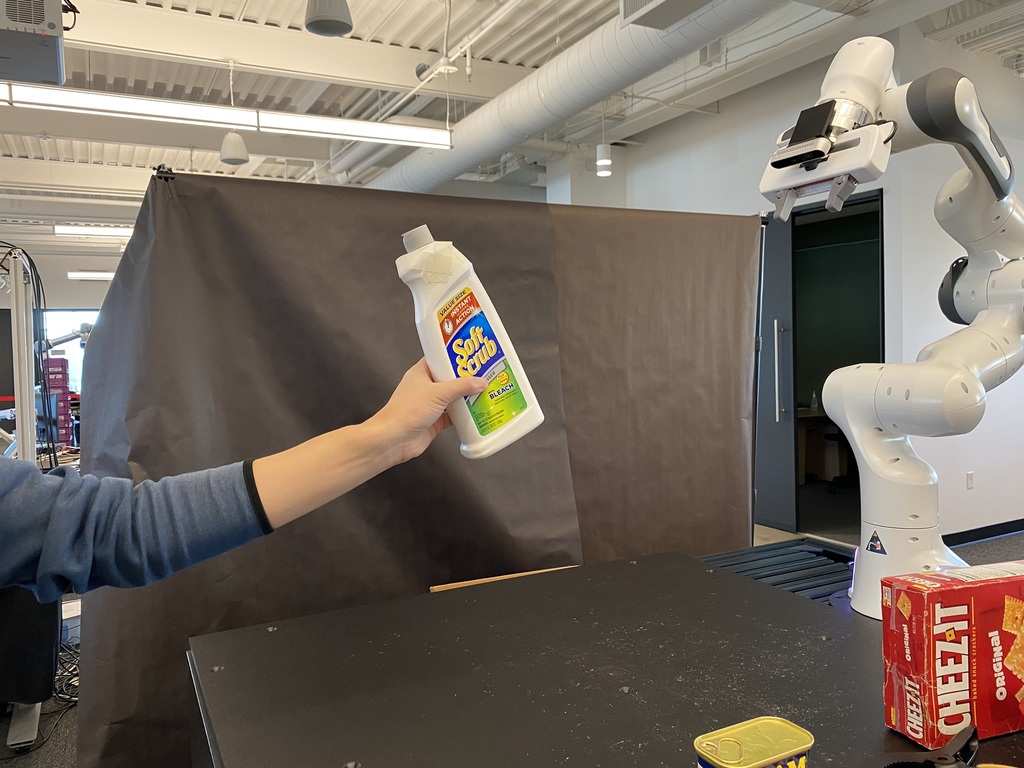}
   \\ \vspace{-1mm}
   pose 1
  \end{minipage}~
  \begin{minipage}{0.320\linewidth}
   \centering
   \includegraphics[width=\linewidth,trim={320 294 390 160},clip]{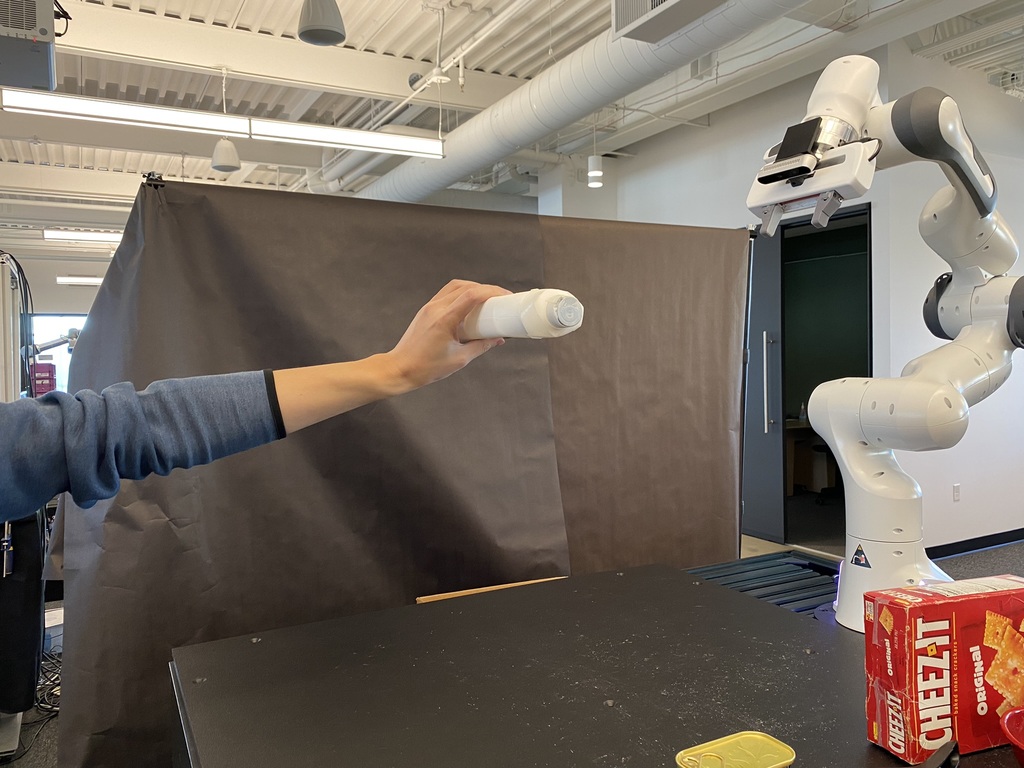}
   \\ \vspace{-1mm}
   pose 2
  \end{minipage}~
  \begin{minipage}{0.320\linewidth}
   \centering
   \includegraphics[width=\linewidth,trim={320 274 390 180},clip]{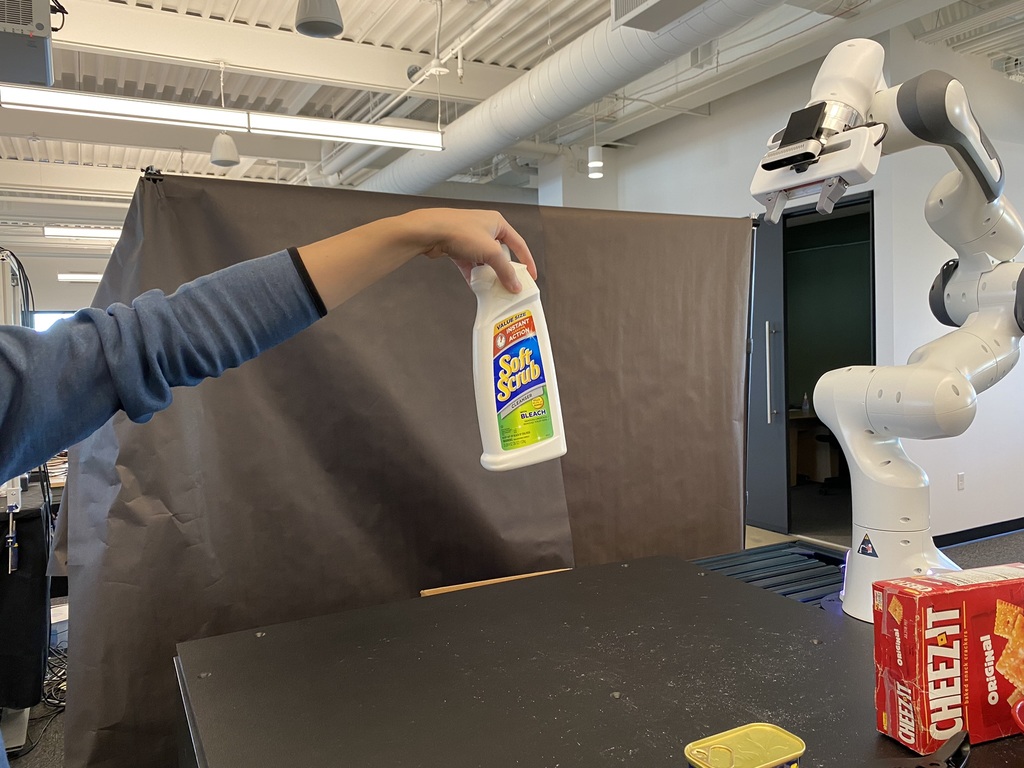}
   \\ \vspace{-1mm}
   pose 3
  \end{minipage}
  \\ \vspace{2mm}
 \end{minipage}
 \caption{\textbf{Controlled handover poses.} The instructed handover poses for 5 out of the 10 selected objects in the pilot study. We pre-select 3 handover poses per object.}
 \vspace{-2mm}
 \label{fig:h2r:pilot_left}
\end{figure*}

%% file: chapters/05_handover/handoversim2real/tables/table_pilot.tex
\begin{table}[]
 \centering
 \footnotesize
 \setlength{\tabcolsep}{2pt}
  \resizebox{0.9\columnwidth}{!}{%
 \begin{tabular}{l|C{1.27cm}C{1.27cm}|C{1.27cm}C{1.27cm}}
   \hline
                           & \multicolumn{2}{c}{Subject 1}                  & \multicolumn{2}{|c}{Subject 2}                 \\
  \cline{2-5}
                           & GA-DDPG                & Ours & GA-DDPG                & Ours \\
                         
  \hline
  011\_banana              & ~~3 / ~~6              & \textbf{~~6 / ~~6}    & \textbf{~~6 / ~~6}     & ~~5 / ~~6             \\
  \rowcolor{Gray}
  037\_scissors            & ~~2 / ~~6              & \textbf{~~5 / ~~6}    & ~~3 / ~~6              & \textbf{~~5 / ~~6}    \\
  006\_mustard\_bottle     & ~~1 / ~~6              & \textbf{~~3 / ~~6}    & ~~2 / ~~6              & \textbf{~~4 / ~~6}    \\
  \rowcolor{Gray}
  024\_bowl                & ~~3 / ~~6              & \textbf{~~4 / ~~6}    & \textbf{~~3 / ~~6}     & \textbf{~~3 / ~~6}    \\
  040\_large\_marker       & ~~0 / ~~6              & \textbf{~~4 / ~~6}    & ~~4 / ~~6              & \textbf{~~5 / ~~6}    \\
  \rowcolor{Gray}
  003\_cracker\_box        & \textbf{~~3 / ~~6}     & ~~2 / ~~6             & ~~0 / ~~6              & \textbf{~~2 / ~~6}    \\
  052\_extra\_large\_clamp & ~~1 / ~~6              & \textbf{~~4 / ~~6}    & \textbf{~~5 / ~~6}     & \textbf{~~5 / ~~6}    \\
  \rowcolor{Gray}
  008\_pudding\_box        & ~~3 / ~~6              & \textbf{~~6 / ~~6}    & \textbf{~~4 / ~~6}     & \textbf{~~4 / ~~6}    \\
  010\_potted\_meat\_can   & \textbf{~~2 / ~~6}     & \textbf{~~2 / ~~6}    & ~~3 / ~~6              & \textbf{~~4 / ~~6}    \\
  \rowcolor{Gray}
  021\_bleach\_cleanser    & ~~3 / ~~6              & \textbf{~~5 / ~~6}    & ~~3 / ~~6              & \textbf{~~4 / ~~6}    \\
  \hline
  \textbf{total}                      & 21 / 60                & \textbf{41 / 60}      & 33 / 60                & \textbf{41 / 60}   \\
    \hline
 \end{tabular}
 }

 \caption{\textbf{Sim-to-Real Experiment}. Success rates of the pilot study. Our method outperforms GA-DDPG~\cite{wang2021goal} for both subjects.}
 \label{tab:h2r:pilot_study}
\end{table}

%% file: chapters/05_handover/handoversim2real/figures/fig_user.tex
\begin{figure*}[t!]
 \centering
 \includegraphics[width=0.191\linewidth]{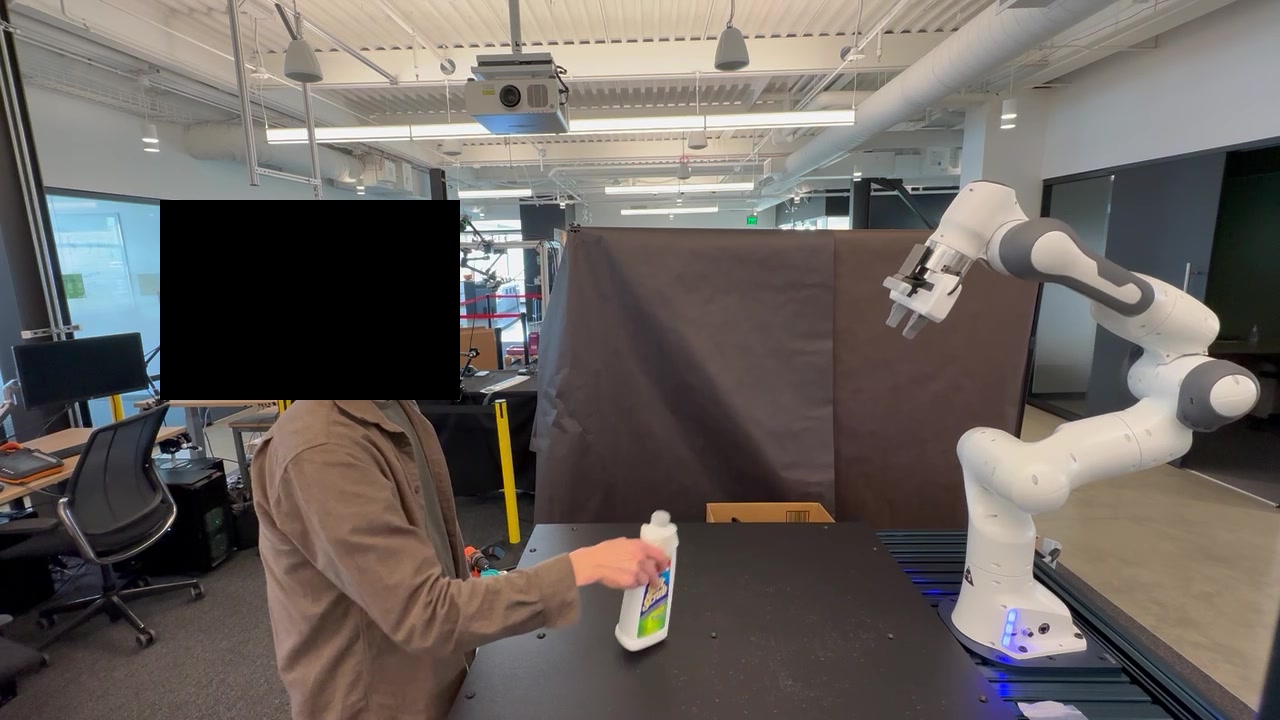}~
 \includegraphics[width=0.191\linewidth]{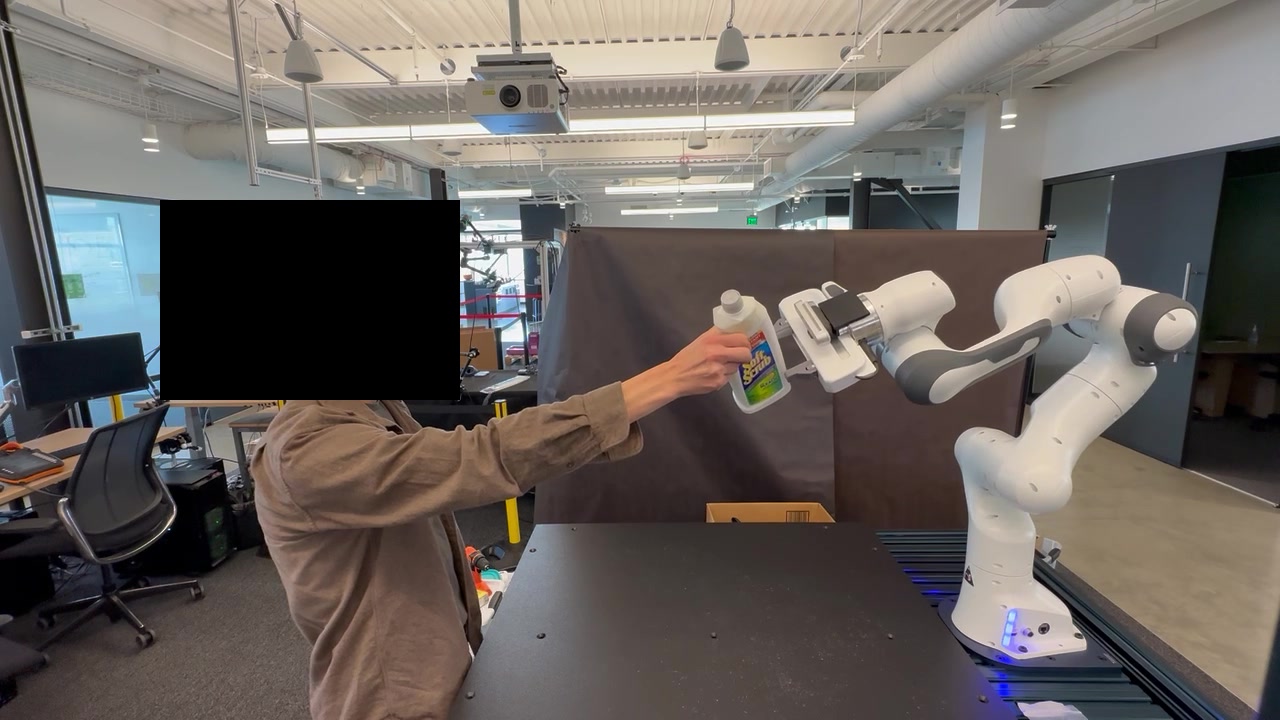}~
 \includegraphics[width=0.191\linewidth]{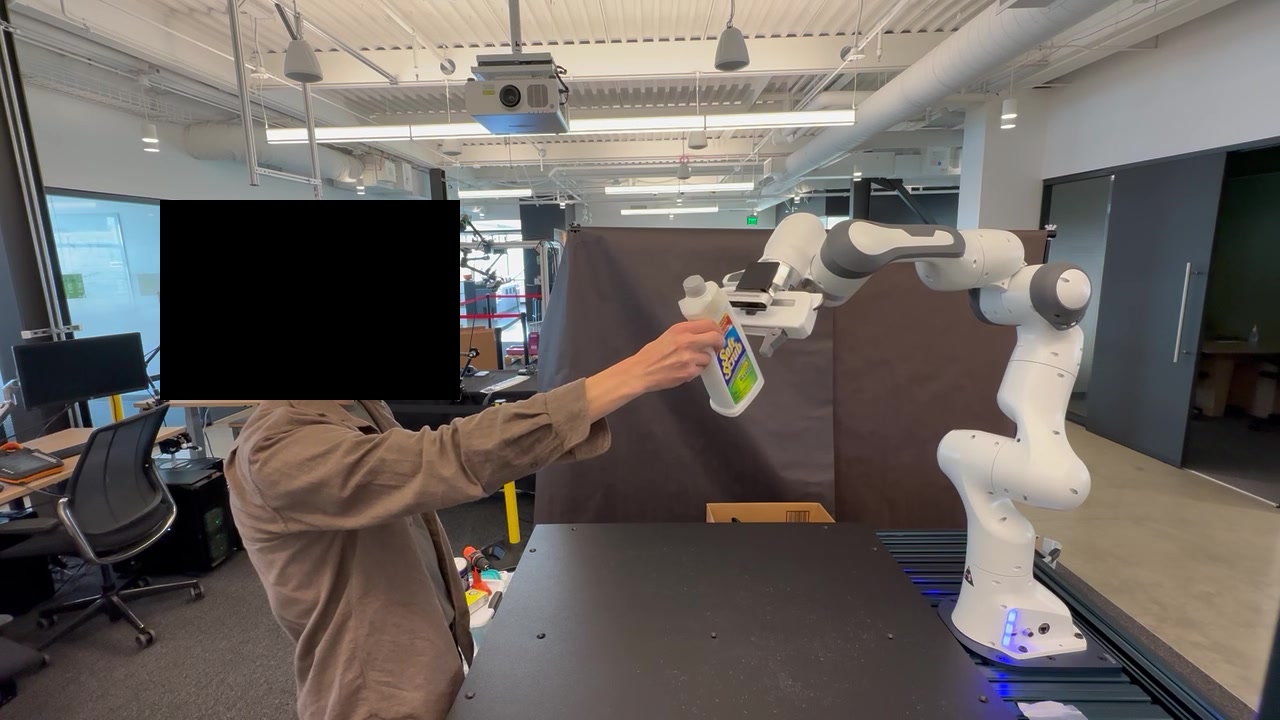}~
 \includegraphics[width=0.191\linewidth]{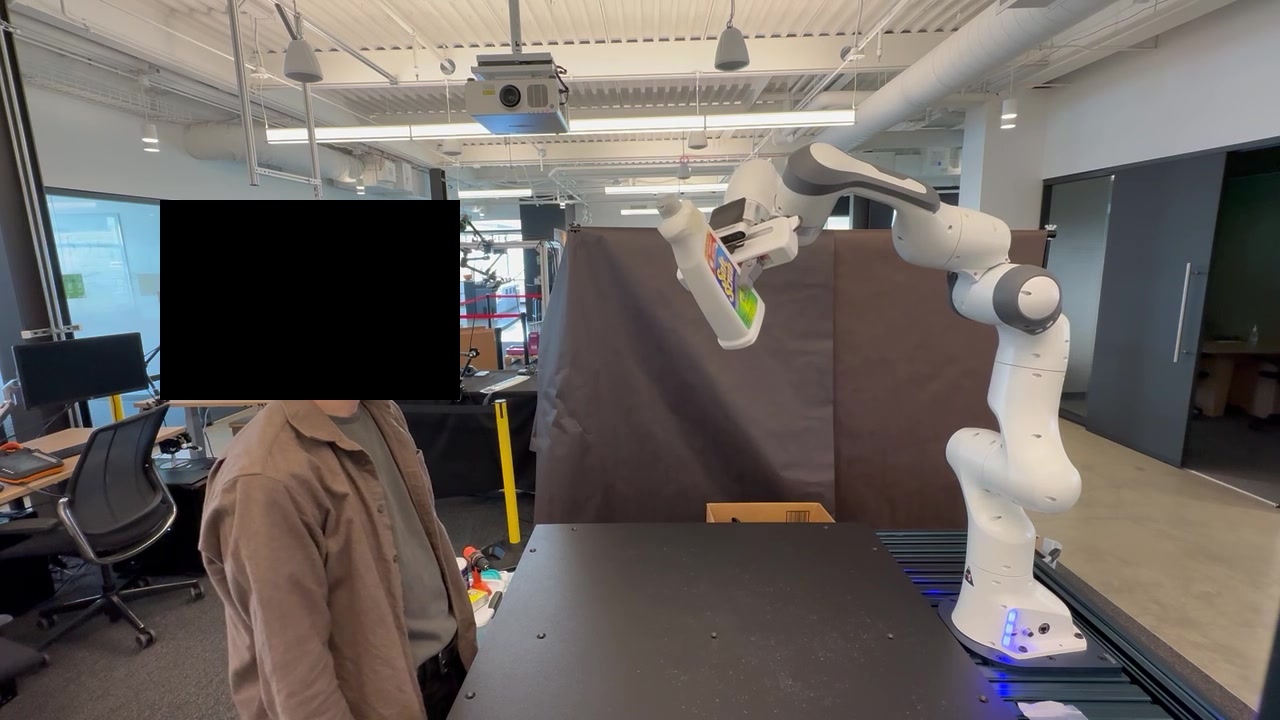}~
 \includegraphics[width=0.191\linewidth]{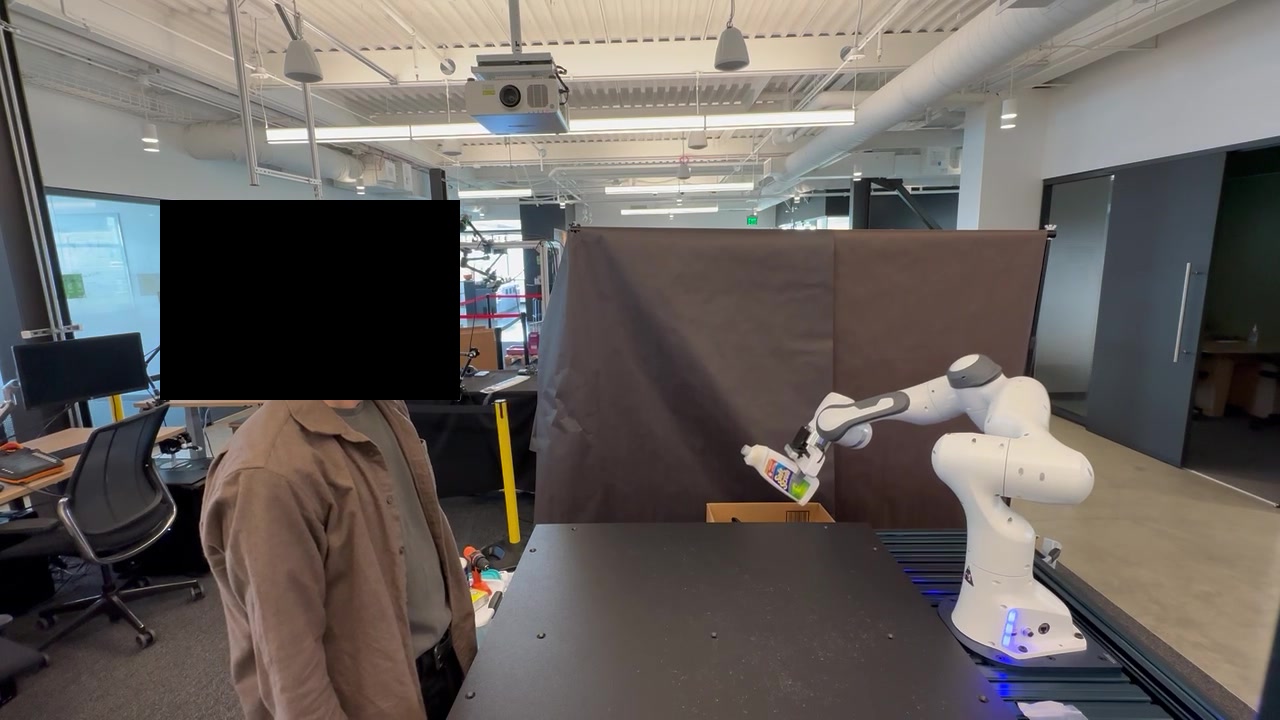}
 \\ \vspace{1mm}
 \includegraphics[width=0.191\linewidth]{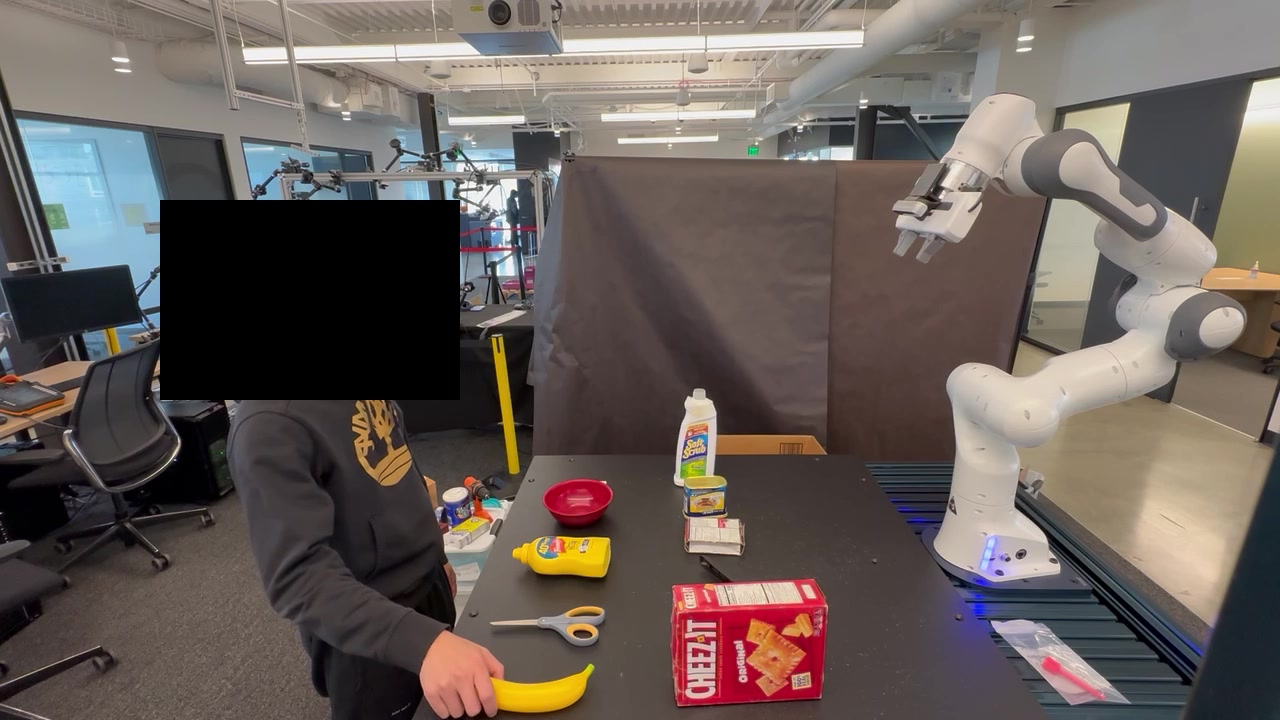}~
 \includegraphics[width=0.191\linewidth]{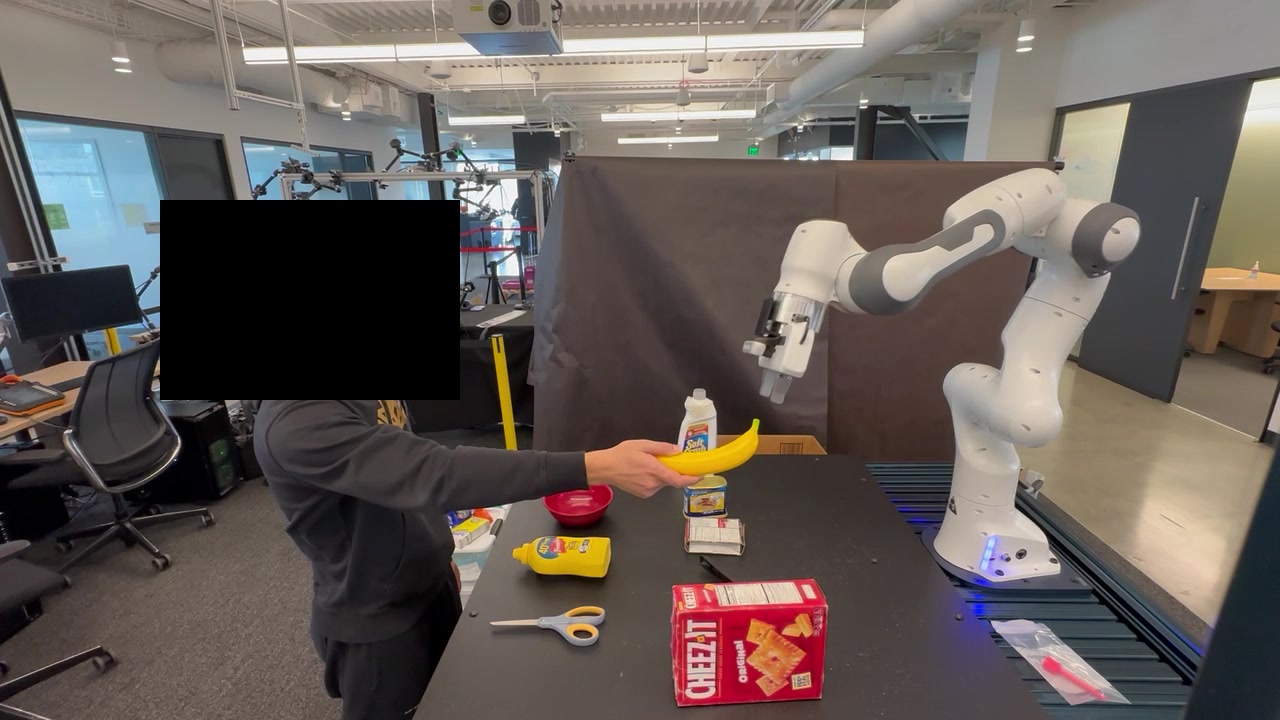}~
 \includegraphics[width=0.191\linewidth]{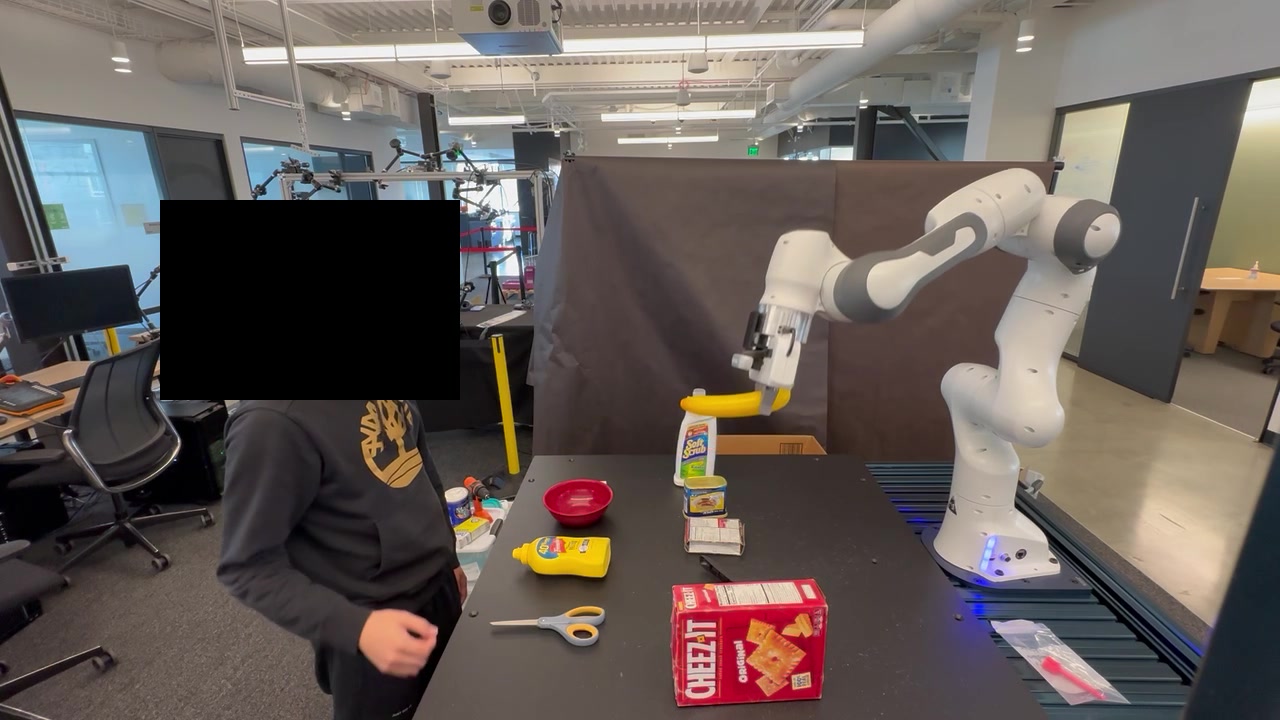}~
 \includegraphics[width=0.191\linewidth]{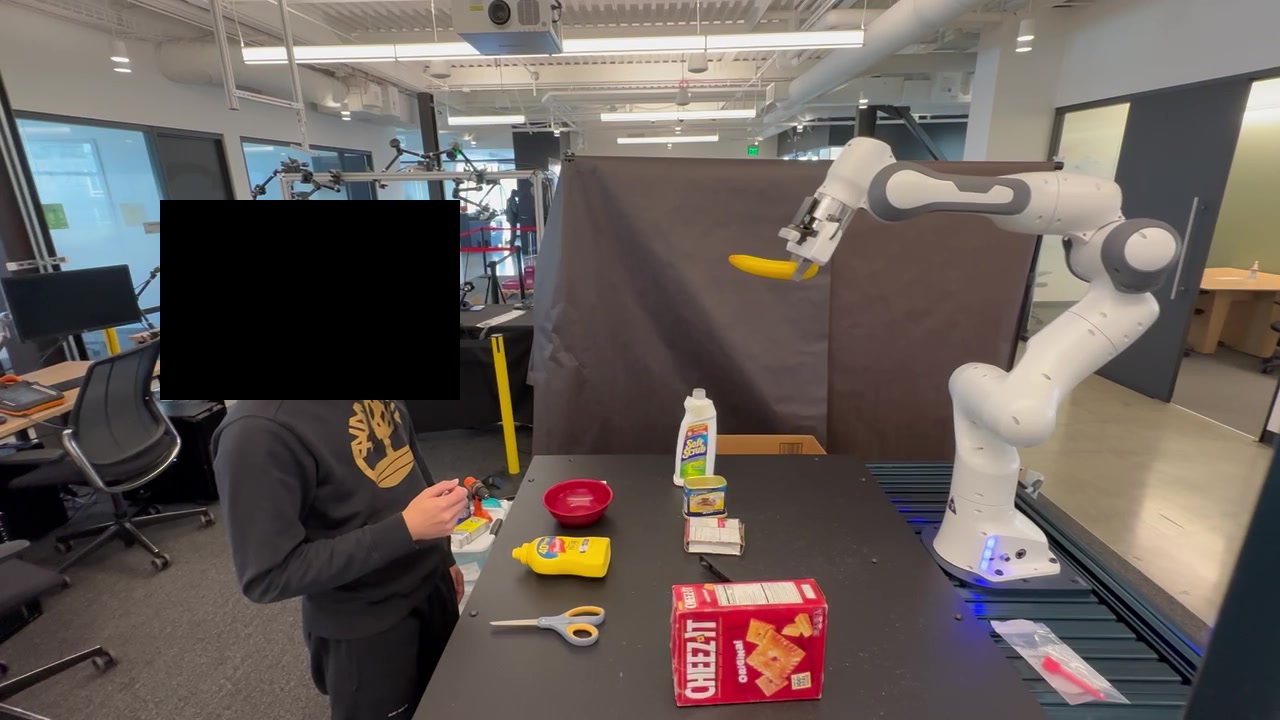}~
 \includegraphics[width=0.191\linewidth]{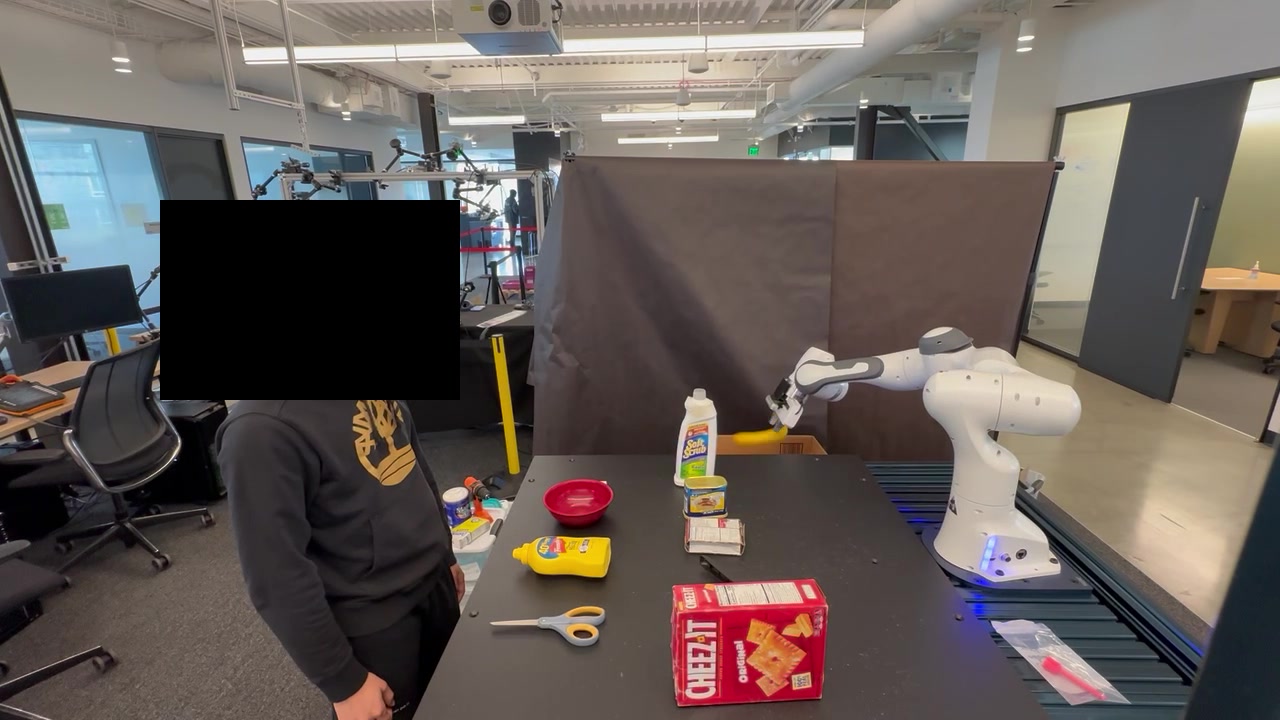}
 \\ \vspace{1mm}
 \includegraphics[width=0.191\linewidth]{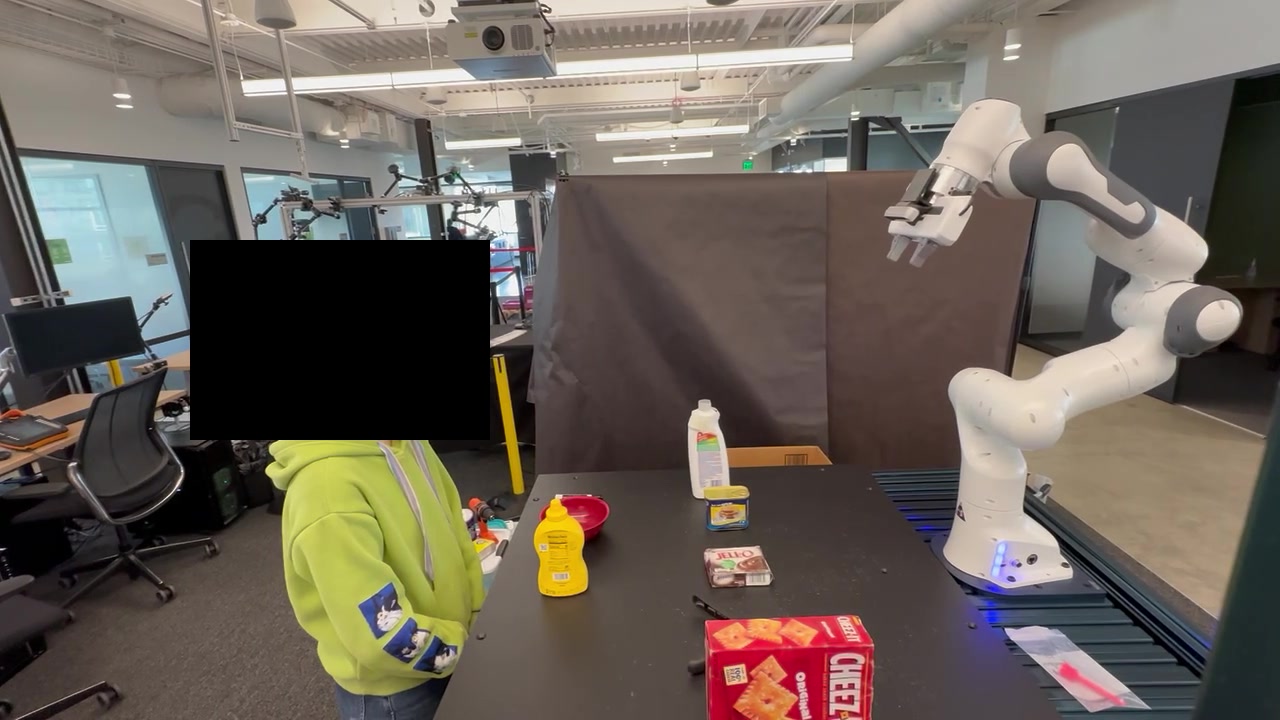}~
 \includegraphics[width=0.191\linewidth]{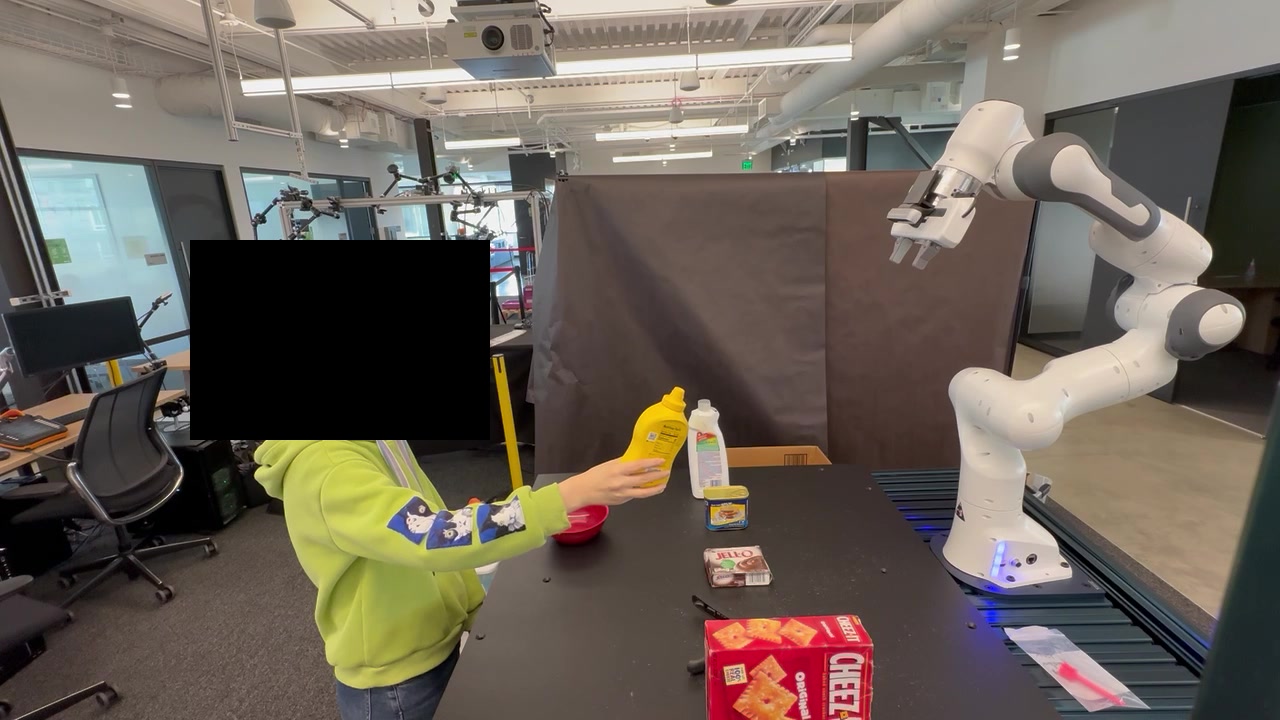}~
 \includegraphics[width=0.191\linewidth]{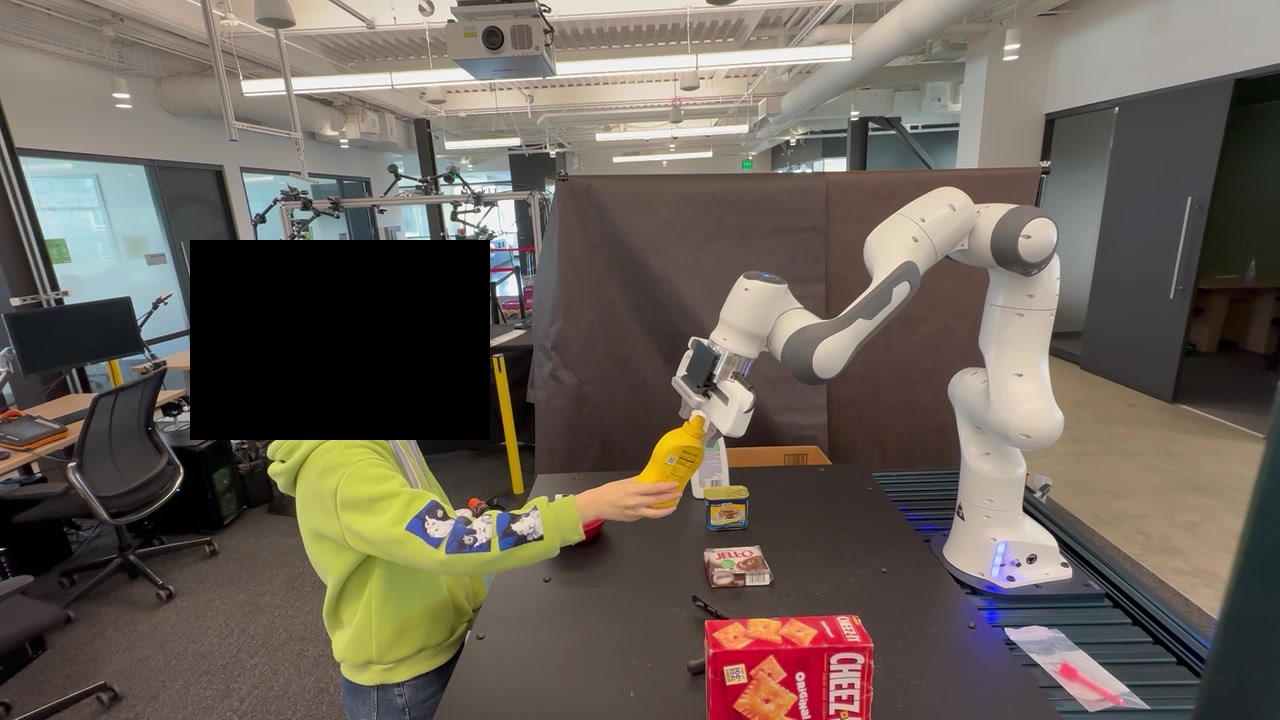}~
 \includegraphics[width=0.191\linewidth]{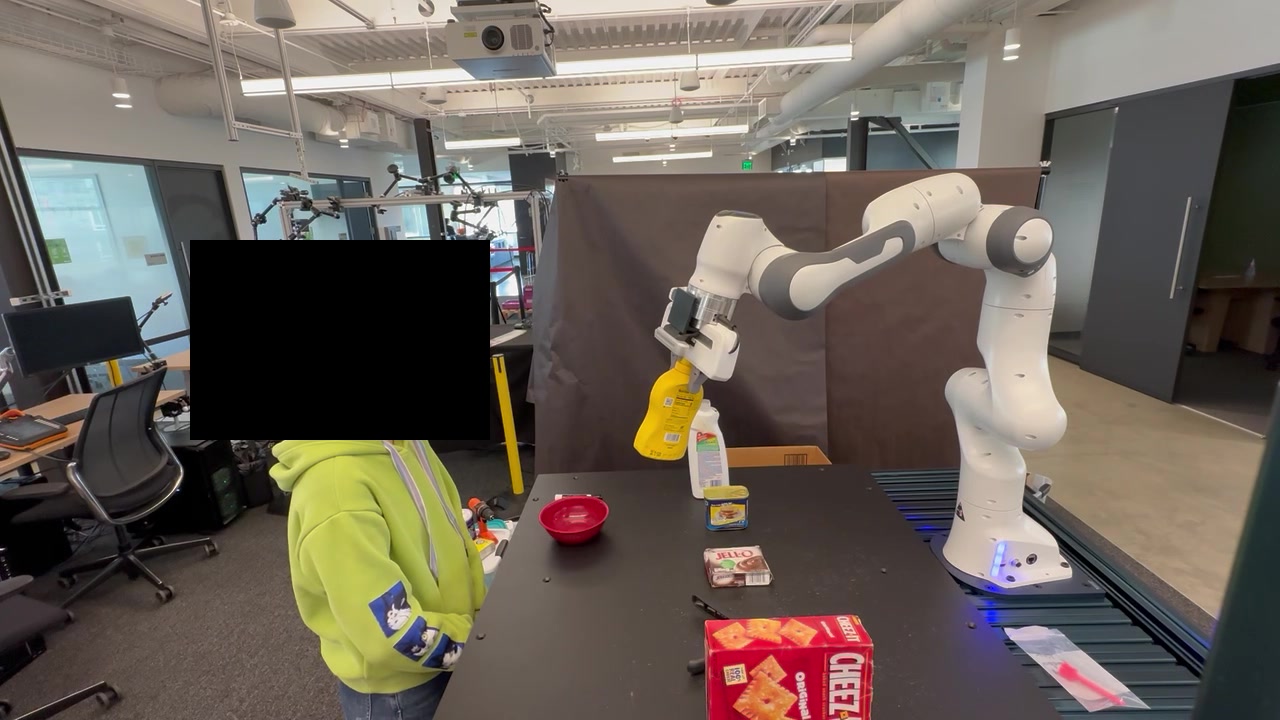}~
 \includegraphics[width=0.191\linewidth]{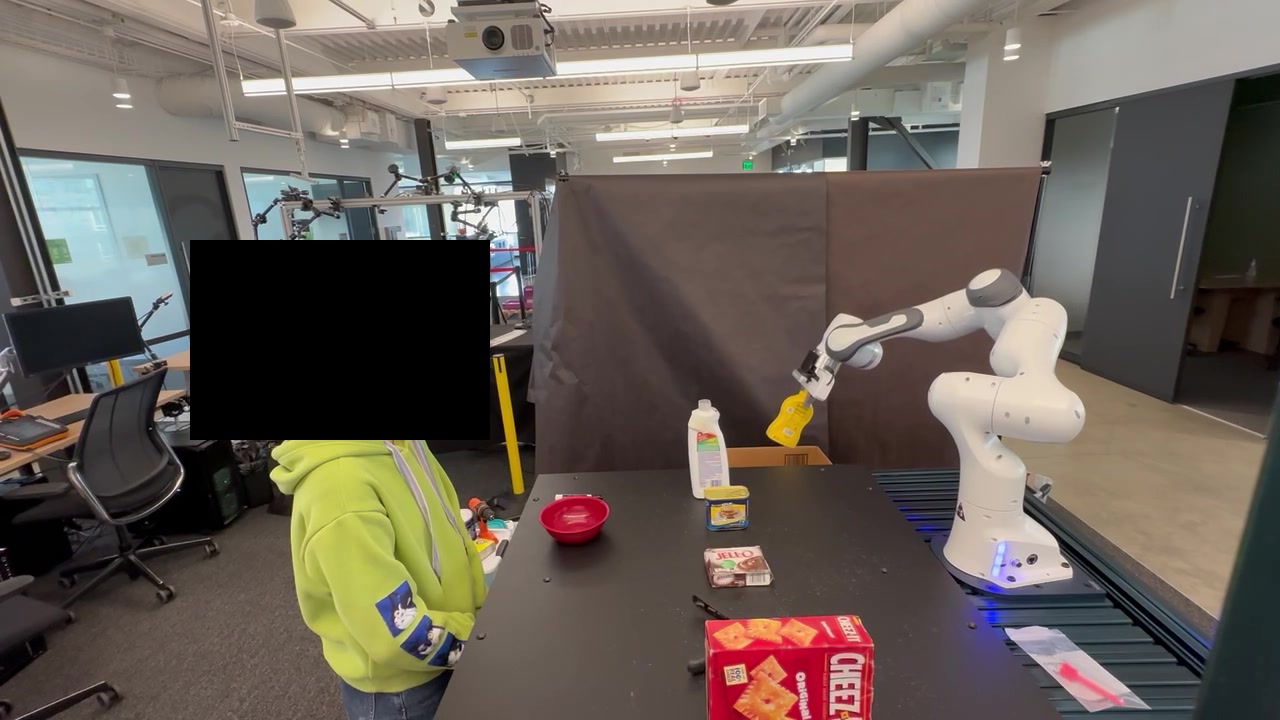}
 \\ \vspace{1mm}
 \includegraphics[width=0.191\linewidth]{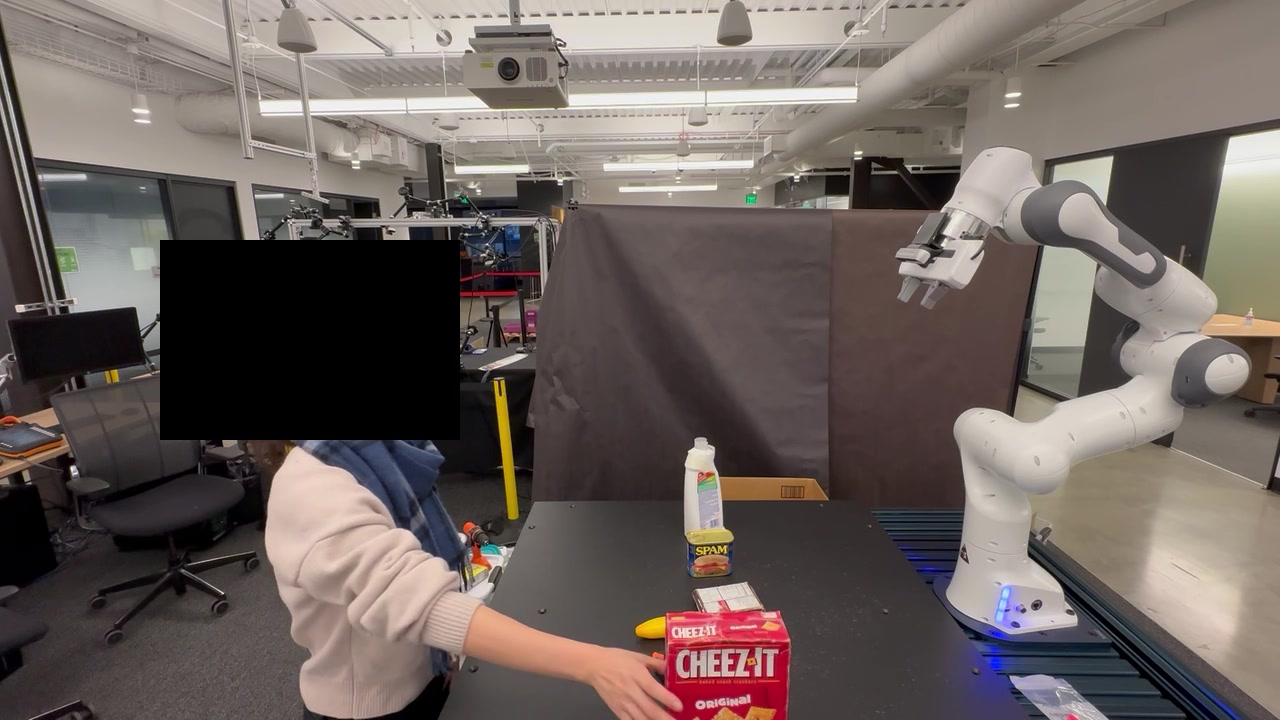}~
 \includegraphics[width=0.191\linewidth]{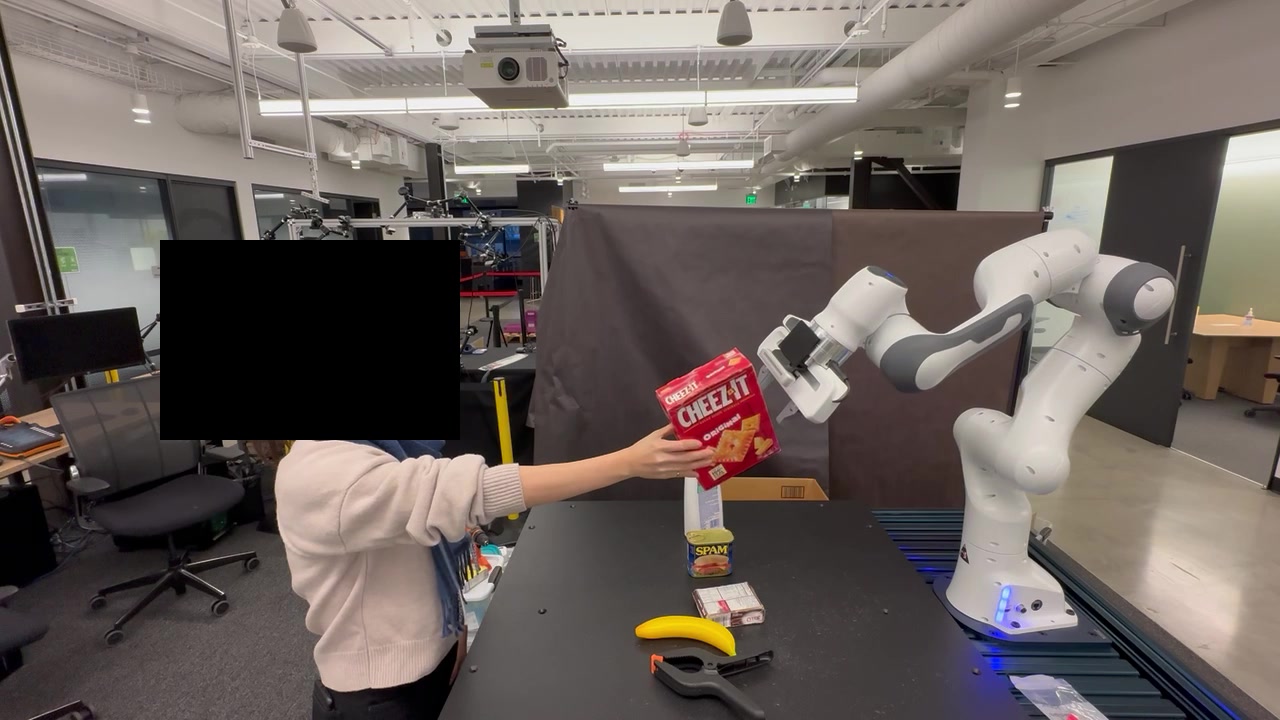}~
 \includegraphics[width=0.191\linewidth]{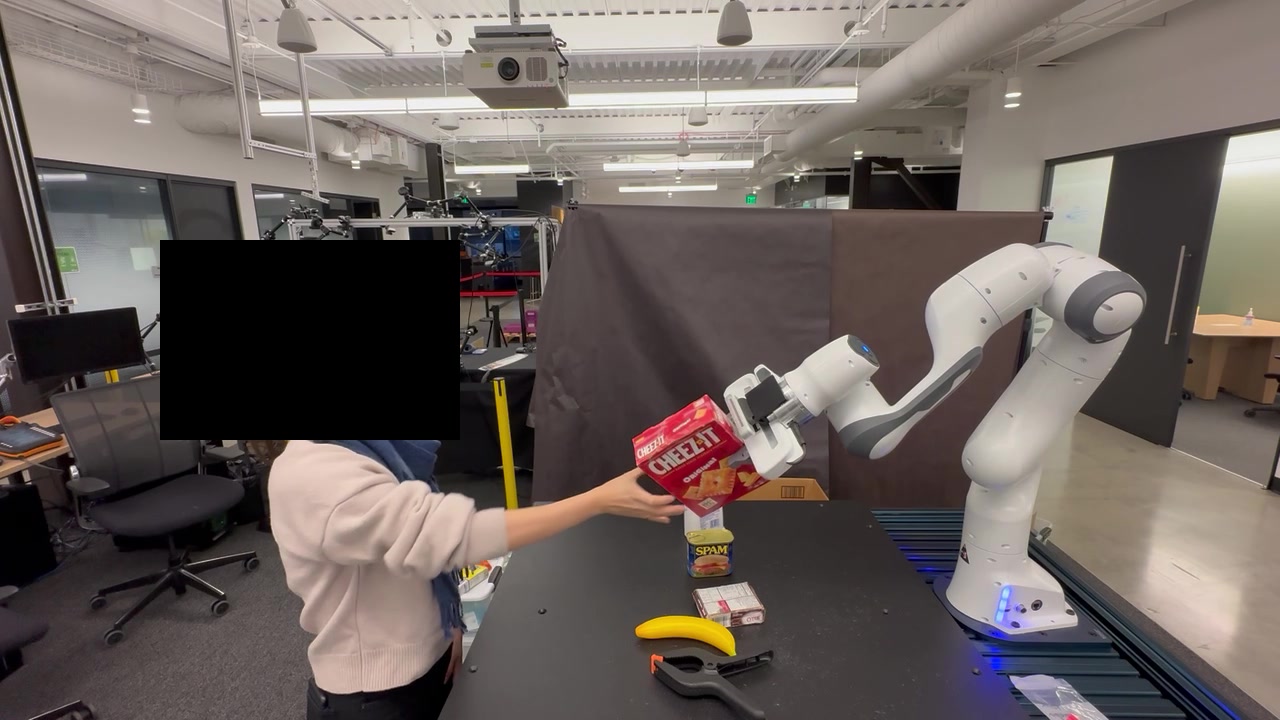}~
 \includegraphics[width=0.191\linewidth]{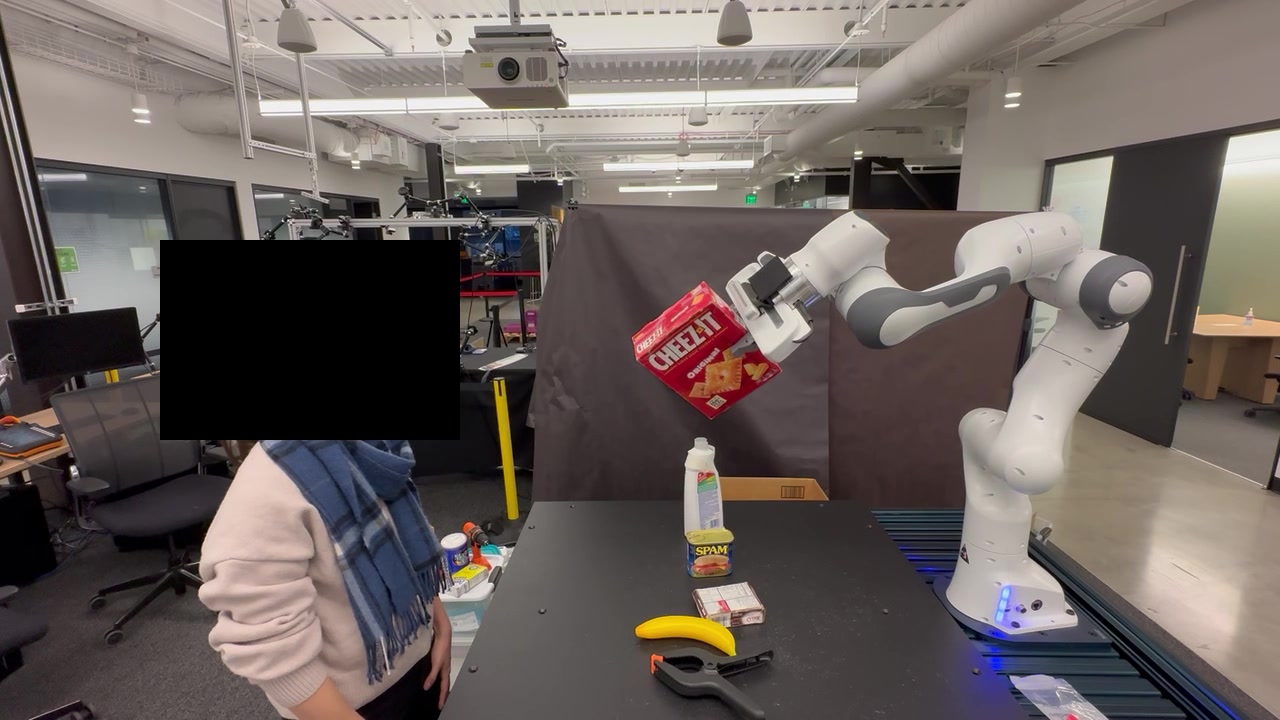}~
 \includegraphics[width=0.191\linewidth]{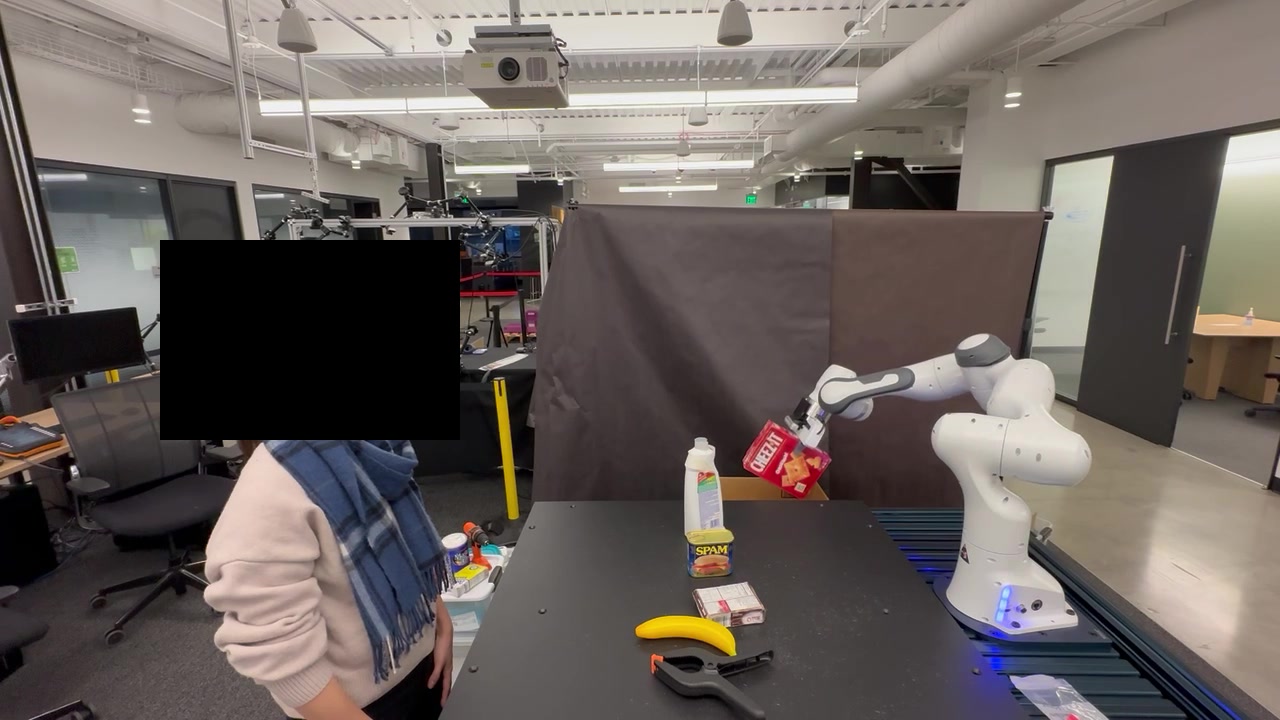}
 \\ \vspace{1mm}
 \includegraphics[width=0.191\linewidth]{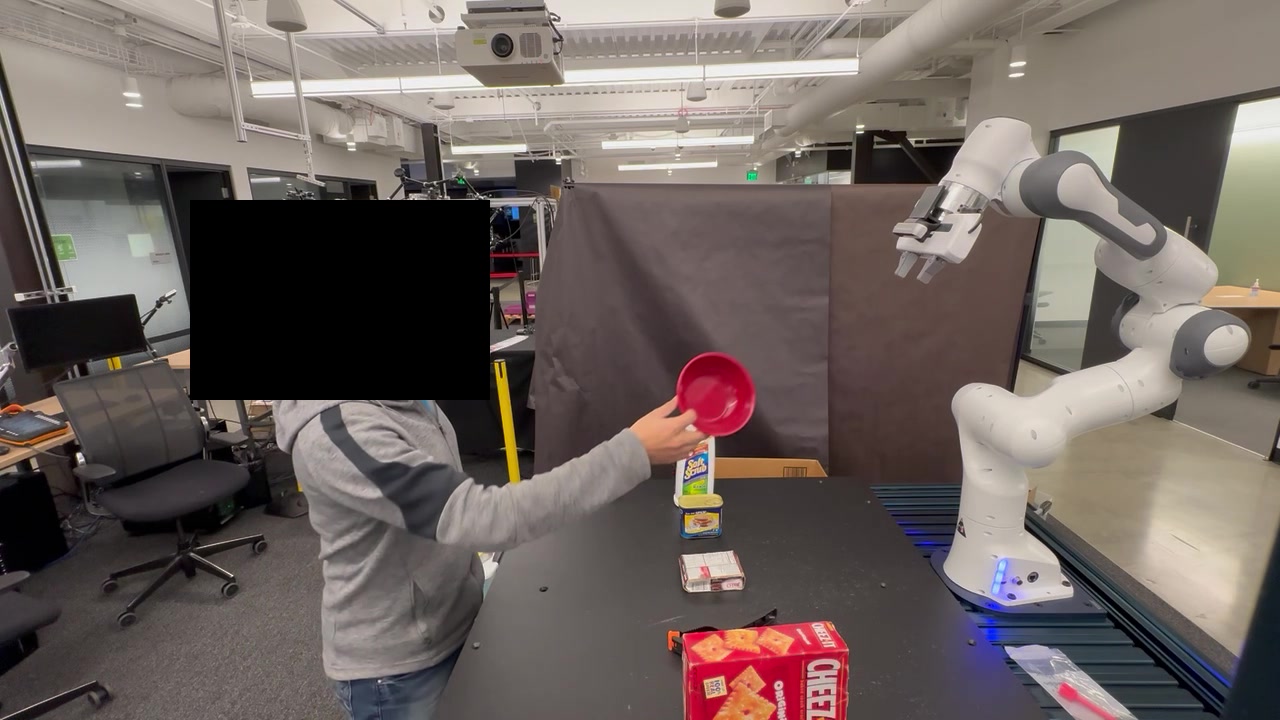}~
 \includegraphics[width=0.191\linewidth]{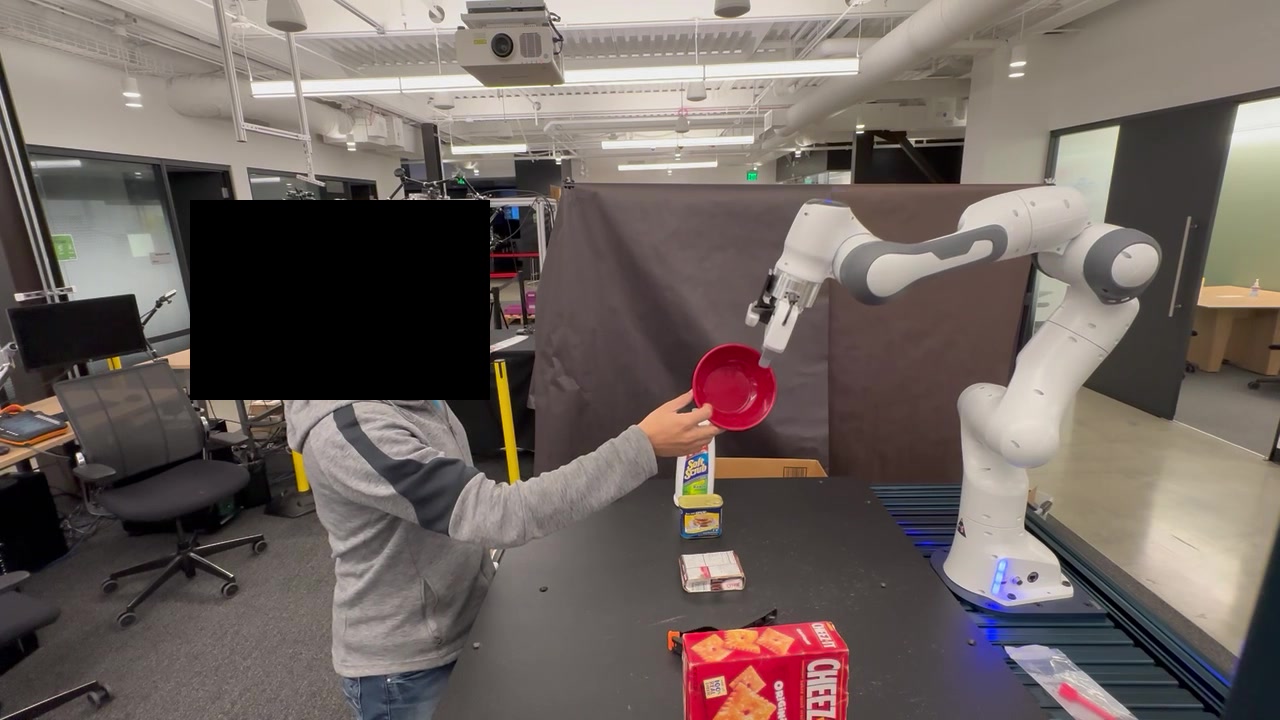}~
 \includegraphics[width=0.191\linewidth]{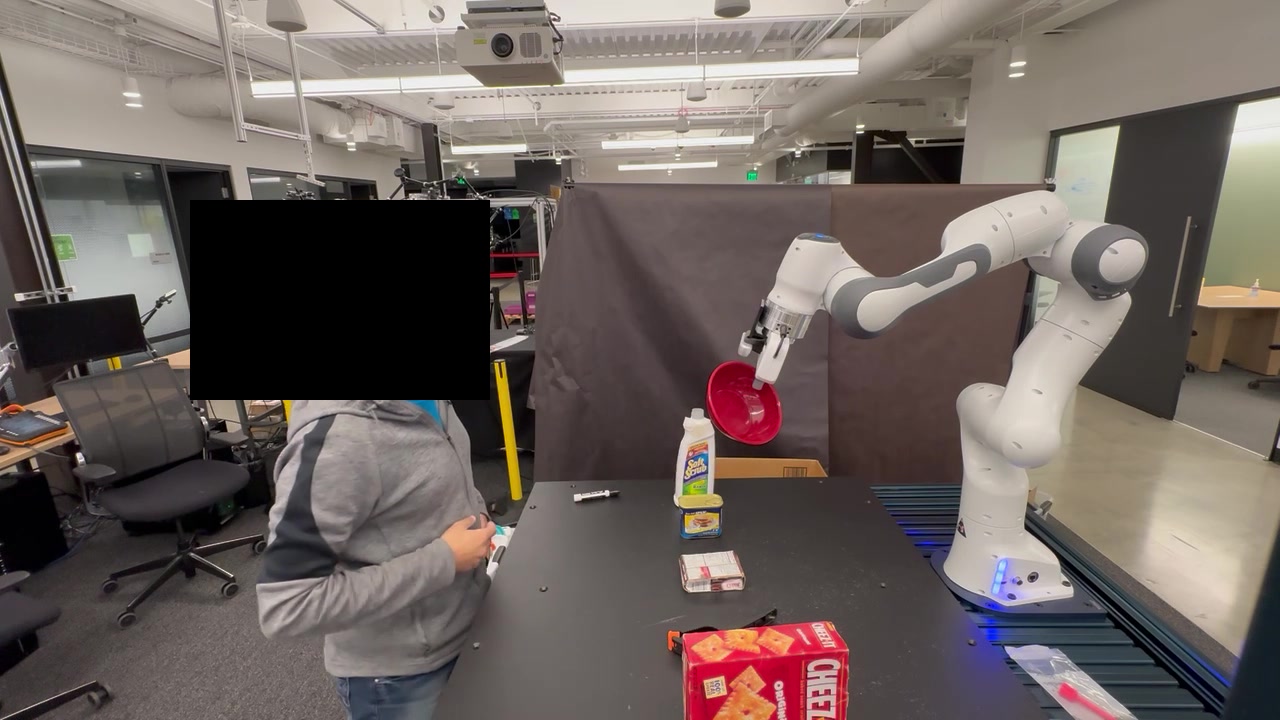}~
 \includegraphics[width=0.191\linewidth]{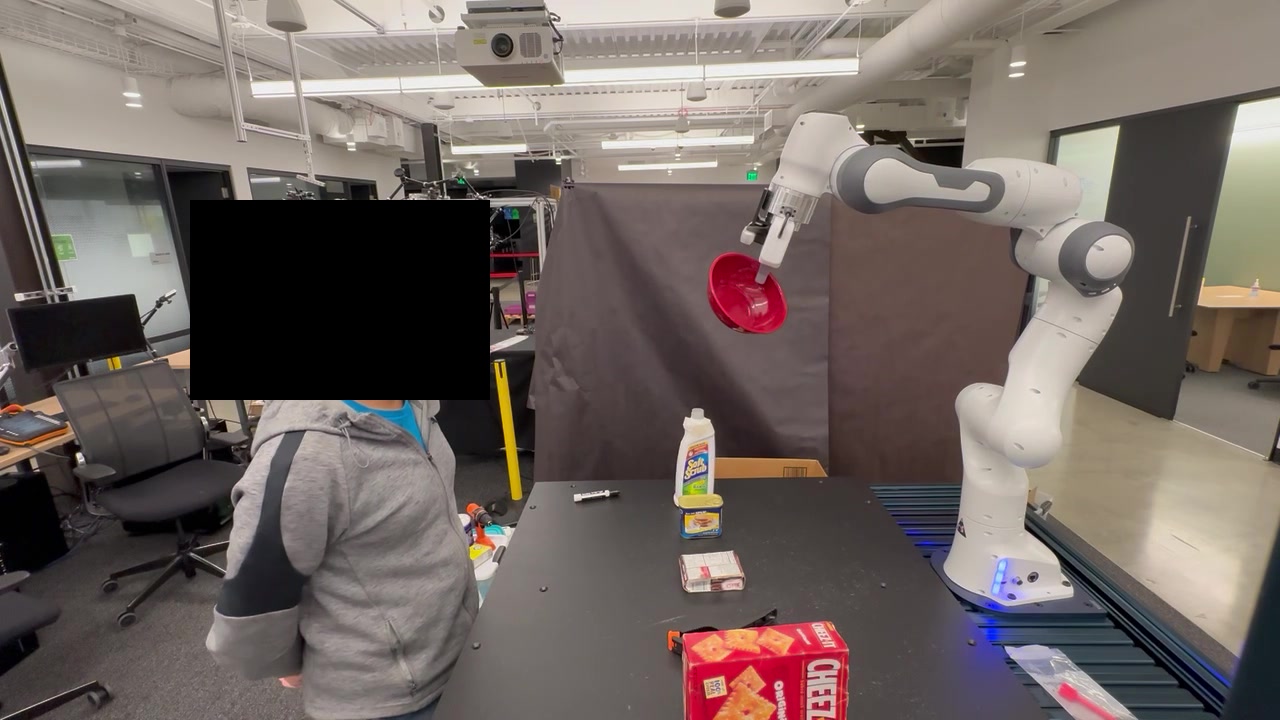}~
 \includegraphics[width=0.191\linewidth]{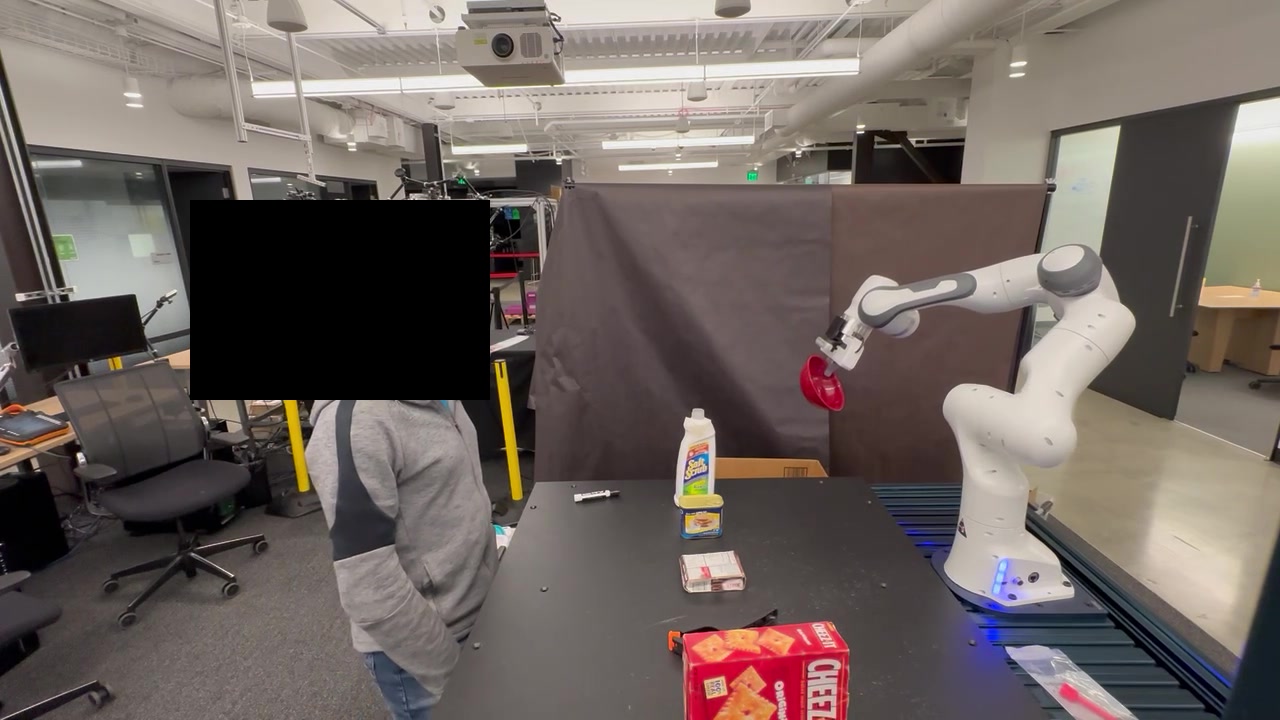}
 \\ \vspace{1mm}
 \includegraphics[width=0.191\linewidth]{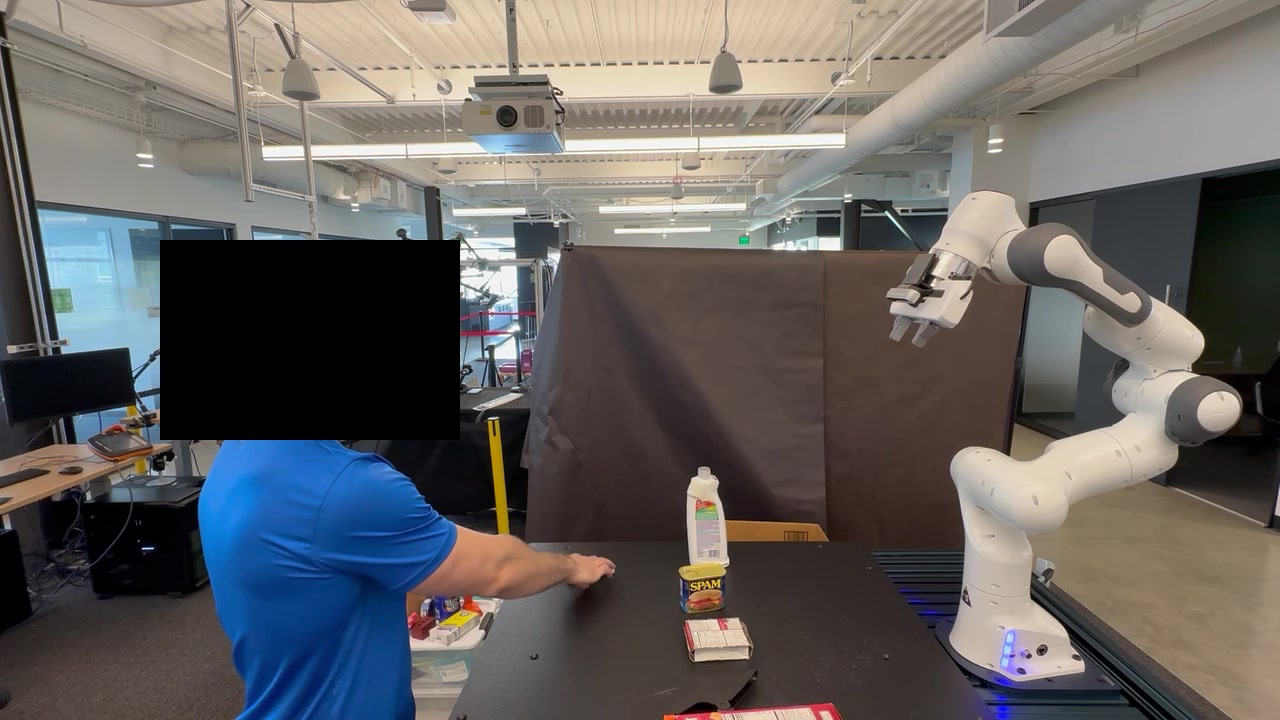}~
 \includegraphics[width=0.191\linewidth]{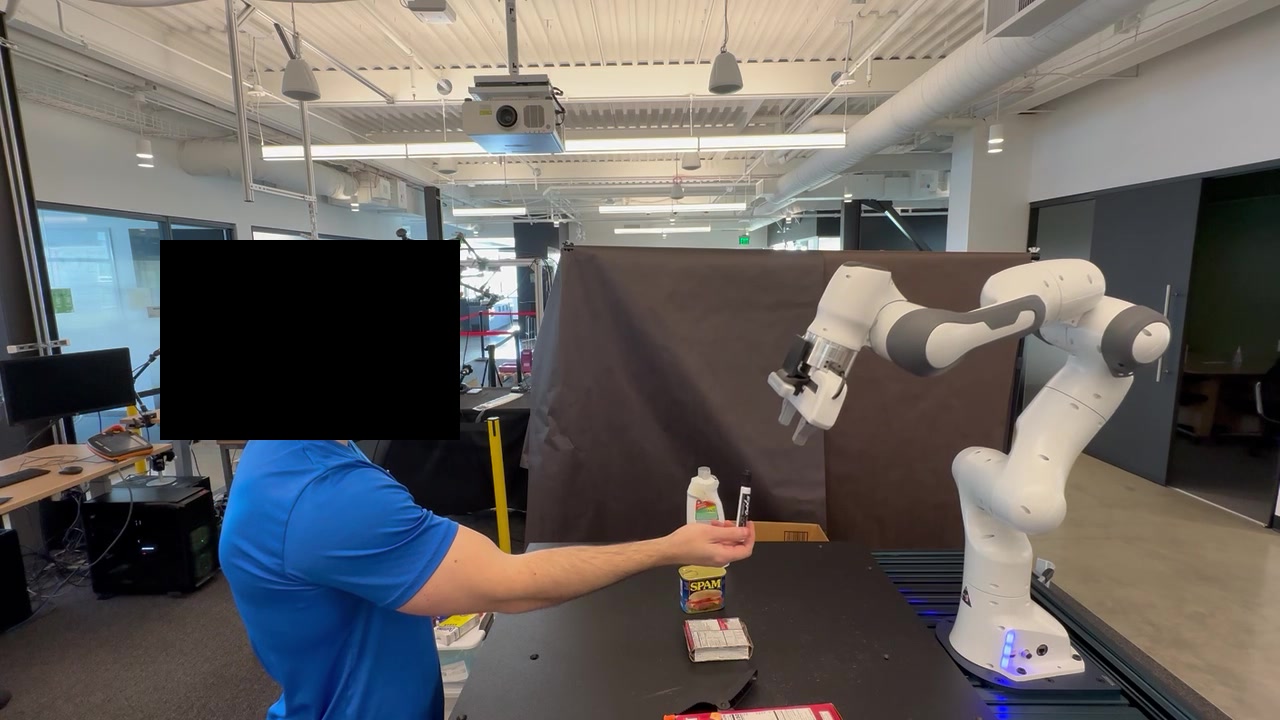}~
 \includegraphics[width=0.191\linewidth]{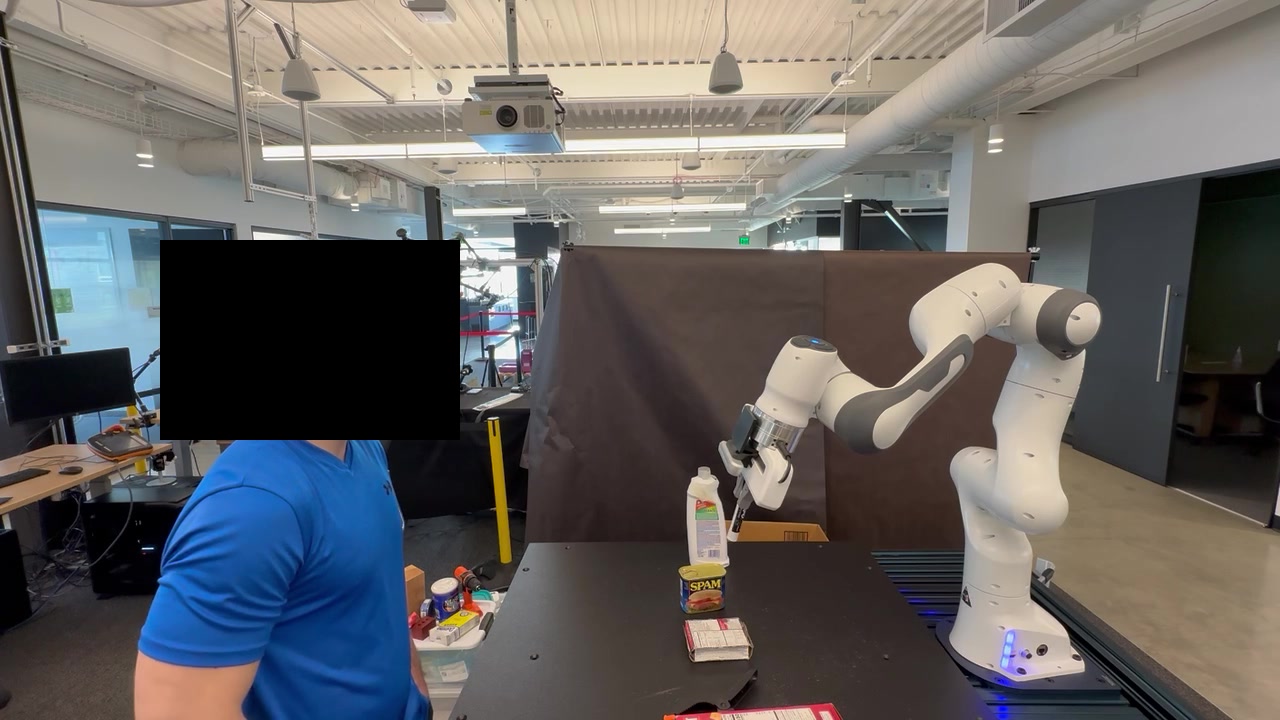}~
 \includegraphics[width=0.191\linewidth]{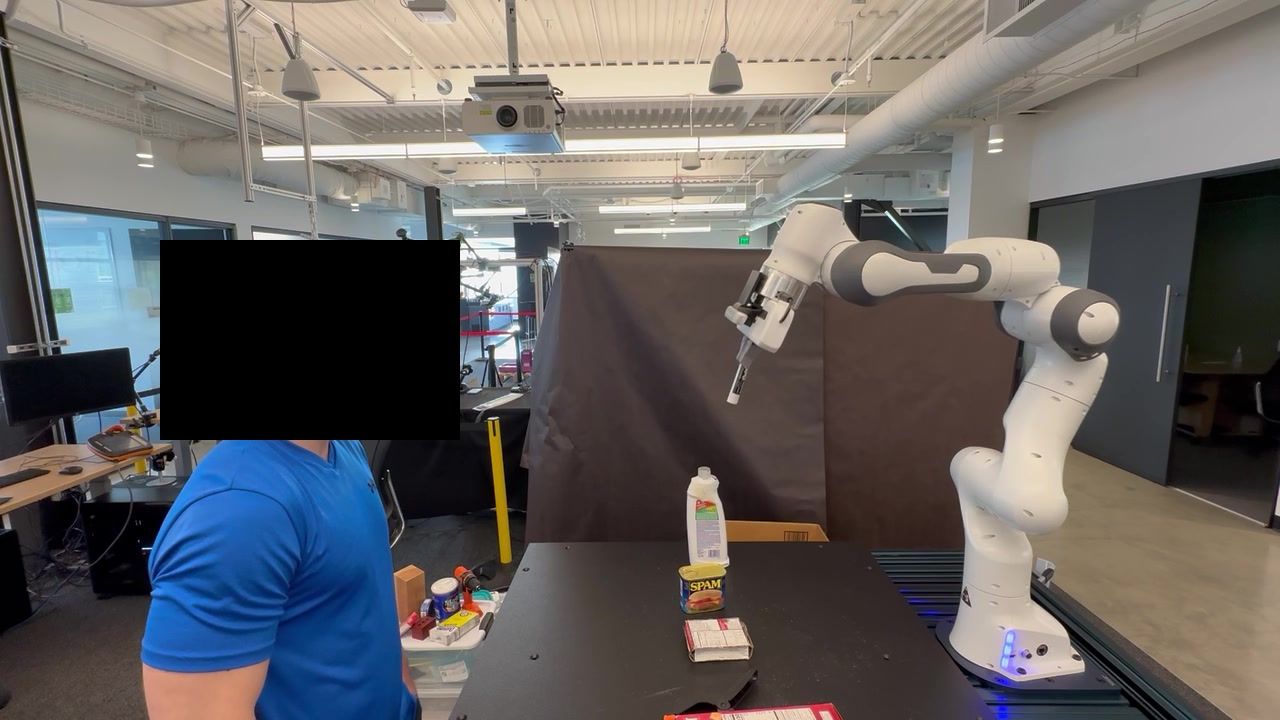}~
 \includegraphics[width=0.191\linewidth]{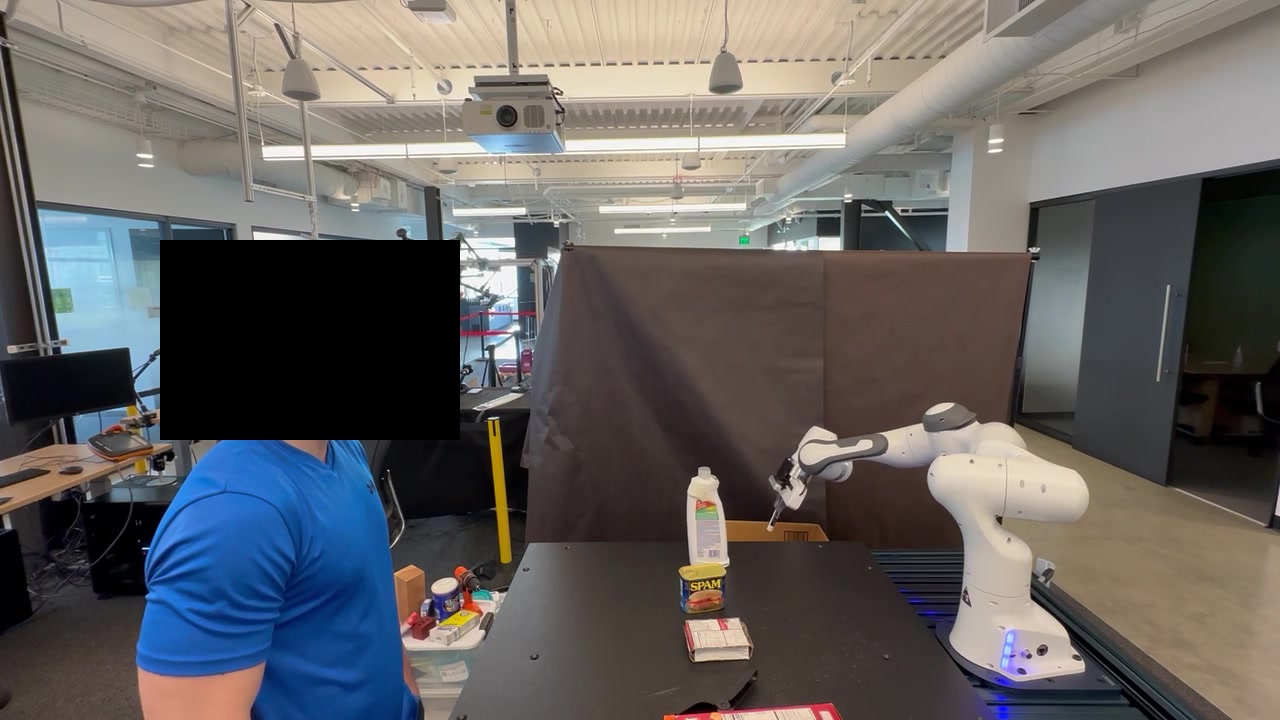}
 \caption{\small We conduct a user evaluation with 6 users by allowing the users to perform handovers freely. The images depict sequences (from left to right) of different users handing over a variety of objects to the robot.}

 \label{fig:h2r:user}
\end{figure*}

%% file: chapters/05_handover/handoversim2real/figures/fig_failures.tex
\begin{figure*}[h]
 \centering
 \includegraphics[width=0.20\linewidth]{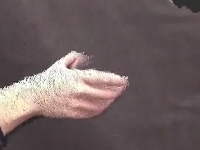}~
 \includegraphics[width=0.20\linewidth]{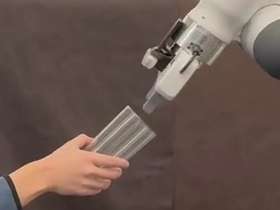}
 \includegraphics[width=0.20\linewidth]{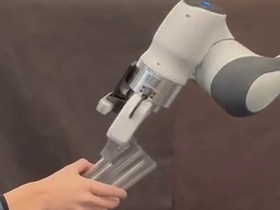}
 \includegraphics[width=0.20\linewidth]{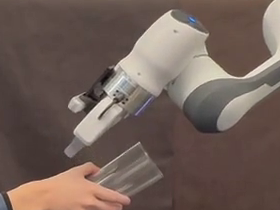}
 \\ \vspace{1mm}
 \includegraphics[width=0.20\linewidth]{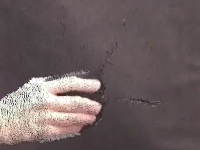}~
 \includegraphics[width=0.20\linewidth]{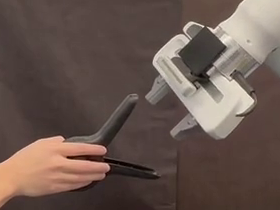}
 \includegraphics[width=0.20\linewidth]{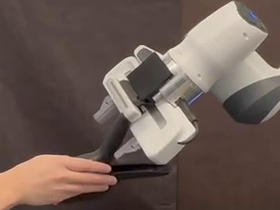}
 \includegraphics[width=0.20\linewidth]{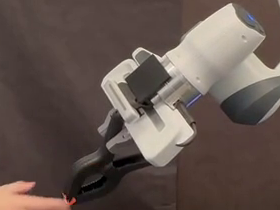}
 \\
\vspace{-1mm}
 \caption{\textbf{Real World Failures}. \textnormal{Left:} Missing/sparse point cloud of transparent/dark objects in real world perception. \textnormal{Right:} Handover policy behavior.}

 \label{fig:h2r:real_failure}
\end{figure*}

%% file: chapters/05_handover/handoversim2real/sections/06_conclusion.tex
\section{Conclusion}
In this chapter, we have presented a learning-based framework for human-to-robot handovers from vision input with a simulated human-in-the-loop. To establish a human counterpart for training the robot, we leverage a recent simulation environment that includes human handover motions from a capture system \cite{chao2022handoversim}. To train the robot, we have introduced a two-stage teacher-student training procedure. In a first stage, the robot acts in a simplified environment, where the human comes to a standstill before the robot starts moving. We have used a combination of behavior cloning and reinforcement learning to train the robot policy in this stage. The expert demonstrations can be obtained online from optimal motion and grasp planning \cite{wang2020manipulation}. In the second stage, the human and robot move at the same time. We cannot rely on expert demonstrations in this stage, hence we have fine-tuned our policies from the first stage with reinforcement learning and regularization techniques. In our experiments we have shown that our method outperforms baselines by a significant margin on the HandoverSim benchmark \cite{chao2022handoversim}. Furthermore, we have demonstrated that our approach is more robust when transferring to a different physics simulator and a real robotic system. While this presents the first step towards a learning-based handover framework, the available human handover motion data is limited both in the amount of objects and diversity of motions. In the next chapter, we will demonstrate how we can learn a more general policy by leveraging synthetic generation of handover motions.

%% file: chapters/05_handover/synh2r/synh2r.tex
\chapter{SynH2R: Synthesizing Hand-Object Motions for Learning Human-to-Robot Handovers}
\chaptermark{SynH2R: Synthesizing Hand-Object Motions for Handovers}
\label{ch:handovers:synh2r}

\contribution{
In the previous chapter, we created a novel framework to train human-to-robot handovers from point clouds. The robot is trained jointly with simulated human handover motions that were obtained from a captured dataset and modeled in simulation. However, the amount of available human motions and objects in existing datasets is limited. One option to add more diversity to the data is to collect more data, which is expensive and does not scale well. To address this, we propose to use dynamic grasp synthesis proposed in our first two chapters to generate data at large scale. We specifically focus on adding diversity in number of objects. However, to achieve human-like handover motions, we need to improve our grasp synthesis method such that it generalizes to unseen objects and can control the direction of approaching the robot. To achieve this, we propose a scalable grasp reference generation method that generalizes well to unseen objects. These references are passed to an improved version of our D-Grasp framework. We generate a synthetic train and test set with more than 1,000 objects each. The train set is integrated into the handover simulation framework which we use to train robotic handover policies. In our simulation experiments on the HandoverSim benchmark, we demonstrate that our method performs on par with the baseline that uses high-quality motion capture data for training. On the synthetic test set, our method significantly outperforms the baseline. A study on the real system shows that users cannot distinguish between the system trained purely on synthetic data and the system trained with captured human motions.
}

\input{\dir/synh2r/sections/01_introduction.tex}
\input{\dir/synh2r/sections/04_method.tex}
\input{\dir/synh2r/sections/05_experiments.tex}
\input{\dir/synh2r/sections/06_conclusion.tex}

%% file: chapters/06_conclusion/conclusion.tex
\def\dir{chapters/06_conclusion}

\input{\dir/content/summary.tex}
\input{\dir/content/future_work.tex}

%% file: chapters/06_conclusion/content/summary.tex
\chapter{Conclusion}
\label{ch:summary}

In this dissertation, we explored the synthesis of hand-object interactions in 4D (3D space + 1D time), which have the potential to facilitate user interactions with intelligent systems across various domains, including AR/VR, pose estimation, and human-robot interaction. In contrast to previous research, which has primarily focused on the generation of static hand-object interactions, we took a first step towards synthesizing entire sequences of hand-object interactions. We approached this challenge by introducing the novel problem setting of dynamic hand-object interaction synthesis and proposing a framework to solve it. We then investigated the specific application of our generated hand-object motions in the context of human-to-robot handovers. We will now provide an overall conclusion of this dissertation.

\section*{Synthesis of Hand-Object Interaction}
We presented two novel tasks for the synthesis of 4D hand-object interactions; dynamic grasp synthesis and bi-manual manipulation of articulated objects. These tasks introduced several challenges, which we described in Section \ref{sec:intro:hoi_challenges}. In particular, the generated sequences of hand and object poses need to be human-like, temporally coherent, and physical plausible (\textbf{C1.1}). Additionally, the model should be robust and capable of generalizing across different objects (\textbf{C1.2}). Finally, our framework should handle complex manipulation scenarios such as bi-manual manipulation and articulated objects (\textbf{C1.3}). To address these challenges, we formulated three hypotheses, which we explored in our work. First, we suggested that reinforcement learning within a physical simulation can learn a policy that generates temporally coherent and physically plausible hand-object interactions (\textbf{H1.1}). We further proposed incorporating static hand-object grasp references into the model’s state space and reward function to facilitate generalization to diverse objects and achieve human-like grasping behavior (\textbf{H1.2}). We explored these two hypotheses in D-Grasp (Chapter \ref{ch:hoi_generation:dgrasp}). Finally, we hypothesized that bi-manual manipulation and articulation of objects can be achieved by adapting our model to incorporate articulation information in both the state space and reward function, supported by a training curriculum and a learning-based control strategy for the global hand poses (\textbf{H1.3}). We investigated this in ArtiGrasp (Chapter \ref{ch:hoi_generation:artigrasp}). We will briefly discuss the technical contributions and experimental results of these works in more detail.

\subsection*{1. \dgraspTitle}
In D-Grasp (Chapter \ref{ch:hoi_generation:dgrasp}), we fist introduced the novel problem setting of synthesizing hand-object interactions in 4D. To address this task, we employed reinforcement learning within a physical simulation. As additional input to the state of the hand and object, the model receives a static hand-object grasp reference. Our key insight was the introduction of a two-stage model: the first stage establishes a stable grasp, and the second stage models the motion of the object to a target pose. To learn a grasping policy, our state space and reward function are designed around goal features that indicate the distance between the current hand pose and the target grasp reference. Simultaneously, we incentivize different objectives for grasp stability within the reward function. Our model therefore learns to balance imitating the grasp reference with achieving stable grasping. We ablated our framework by training a one-stage model with reinforcement learning and demonstrated that the grasp references, along with the goal features, are critical components in achieving stable and human-like grasping. Therefore, we can accept hypothesis \textbf{H1.2}. Our qualitative evaluations showed that our outputs appear natural and physically plausible, validating hypothesis \textbf{H1.1}. 

\subsection*{2. \artigraspTitle}
In ArtiGrasp (Chapter \ref{ch:hoi_generation:artigrasp}), we extended our framework to the bi-manual manipulation of articulated objects. Building on the solution in D-Grasp, we introduced several modifications to our model. To make the policy aware of the articulation of the object, we rewarded reaching target articulation angles and included articulation information in the observation space. To handle the complexity of two-handed coordination and fine-grained grasping for articulation, our curriculum first trains in a simpler setting where objects are stationary and the hands operate independently, then continues in an environment where both hands interact jointly with a non-stationary object. We demonstrated that ArtiGrasp can generate motions of bi-manual grasping and articulation, such as relocating a laptop with both hands and then opening it. In our comparison to baselines, we found that D-Grasp struggles with learning object articulation. Additionally, we showed that we can compose longer sequences of multi-object manipulation. Our ablation studies justified the contributions proposed in our hypothesis. These experiments therefore demonstrate that hypothesis \textbf{H1.3} holds, i.e., that we can adapt our framework to achieve bi-manual manipulation of articulated objects. 

\subsection*{Implications}
Based on simple task goals, such as a 6D target object pose or an articulation angle, our framework enables the generation of physically plausible object interactions in 4D. Although our model requires a grasp reference frame to achieve human-like grasping, we demonstrated that these references can be obtained either from static grasp synthesis or image-based pose estimation. Our works on such synthesis opens up several potential use-cases in downstream tasks. For instance, we could augment hand-object pose estimation models by combining our generated motions with photorealistic rendering techniques. In interactive AR/VR scenarios, our method could animate more realistic grasping behavior, which currently often relies on heuristics to attach objects to virtual hands. In human-robot interaction, the ability to simulate natural human motion behavior has the potential to enable robot training with humans in the loop. We explored this further in the context of human-to-robot handovers. 

\section*{Human-to-Robot Handovers}
Having established a framework for generating 4D hand-object interactions, we explored its applicability to the downstream task of human-to-robot handovers. In this task, a robot must perceive a human giver in order to safely grasp and retrieve the object. Previous solutions to this task require stationary humans and handcrafted modules. Progress in developing reactive systems has been hindered by the difficulty of accurately simulating humans and a learning framework that can be trained with humans in the loop. To address this, we outlined several challenges related to reactive human-to-robot handovers in Section \ref{sec:intro:handovers}. The first challenge is building a learning framework for human-to-robot handovers in simulation that can be transferred to a real robotic platform (\textbf{C2.1}). The second challenge is to train a model that generalizes well to dynamic human motions and object geometry (\textbf{C2.2}). To address this, we hypothesized that embedding human motions into the physical simulation and explicitly considering them in the model design and training procedure would enable the learning of a policy that effectively transfers to real systems with real human interactions (\textbf{H2.1}). To address the second challenge, we suggested that hand-object interaction synthesis can be used to generate more diverse human handover motions, thereby increasing the generalizability of handover policies (\textbf{H2.2}). We explored hypothesis \textbf{H2.1} in Chapter \ref{ch:handovers:handoversim2real} and hypothesis \textbf{H2.2} in Chapter \ref{ch:handovers:synh2r}. 

\subsection*{1. \handoverTitle}
In Chapter \ref{ch:handovers:handoversim2real}, we developed a learning environment that supports training with a simulated human in the loop. The simulation of human motion was achieved by replaying captured human handover motions in the physical simulation. To train a robot policy for grasping the object from the humans, we proposed a two-stage teacher student training framework. In the first stage, the robot moves only after the human has reached a handover pose. In this setting, we leveraged grasp and motion planning to provide expert demonstrations and used an imitation learning approach to distill the expert data into a student policy. In the second stage, the robot and human move simultaneously, thereby rendering grasp and motion planning unsuitable. We therefore fine-tuned our policy with reinforcement learning. By using point cloud representation, we were able to mitigate the domain gap between simulation and real. We performed experiments in simulation and on the real system. We found that compared to both learning-based and non-learning based solutions, our framework can achieve significantly better performance. In our qualitative user study, participants predominantly preferred our system over the closest baseline. These experiments empirically validate our hypothesis that simulation with a human in the loop enables the training of effective human-aware grasping policies (\textbf{H2.1}).

\subsection*{2. \synhrTitle}
In SynH2R (Chapter \ref{ch:handovers:synh2r}), we addressed the challenge of generalizing to a wide variety of object geometries. The limited amount of human handover motions captured in the real world hinders the training of the robot with more objects. To overcome this, we leveraged our work on hand-object motion synthesis to generate human handover motions. To this end, we had to extend our D-Grasp framework to scale to a large number of unseen objects and to control the direction of approach, as humans typically hand over objects facing the robot. Therefore, we proposed a grasp generator that synthesizes grasp references on unseen objects, which can be conditioned on a target direction of approach. These references were then used as input to our D-Grasp pipeline, which generates dynamic grasping motions. This enabled us to generate a large train and test set of synthetic human handover motions, with a scale 100 times larger in terms of the number of objects compared to the captured dataset. We used this human motion data for training the framework introduced in the previous chapter. Our experiments revealed that our system trained purely with synthetic data performs on-par with the system trained on real human motions on the HandoverSim benchmark (20 objects) \cite{chao2022handoversim}. On our newly introduced synthetic test set (1k+ objects), the method trained on synthetic data achieves significantly better performance than the system trained on real human motion data, indicating that synthetically generated handover motions increase the generalizability of handover policies (\textbf{H2.2}).

\subsection*{Implications}
Our works on human-to-robot handovers establishes a foundation and novel approach for training human-aware robotic policies. We demonstrated that realistically embedding human motions into physical simulation creates a suitable learning environment for robots. As our experimental validations showed, the system can be transferred to a real robotic platform with real human users. This highlights the potential of our approach and paves the way for more complex human-robot interaction tasks. A key insight from SynH2R is that users were unable to distinguish between a system trained with real captured human motions and one trained with synthetically generated human motions. This finding has significant implications: if we can realistically simulate human-like behavior, we can eliminate the need for expensive motion capture to train human-robot interaction algorithms.
We will further discuss the outlook and future work in the next chapter.

\section*{Thesis Conclusion}
In this thesis we outlined several challenges involved with generating hand-object interactions in 4D. To address these challenges, we introduced two novel tasks, dynamic grasp synthesis and bi-manual articulation of articulated objects. To solve these tasks, we proposed a framework based on reinforcement learning within physical simulation. Our experiments demonstrated that our generated outputs are natural and human-like. To explore an application of our models, we investigated human-to-robot handovers, an important task in human-robot interaction. We designed a framework that allows training human-aware robotic policies for such handovers. Our key insight was to embed human motions into physical simulation to train the robot with a human in the loop. Starting from captured data of human handover motions, we increased the diversity in handover motions by leveraging our model for hand-object interaction synthesis. Our experiments demonstrated that robotic policies can be trained solely on synthetic human handover motions and successfully be transferred to the real system with real human users.

%% file: chapters/06_conclusion/content/future_work.tex
\chapter{Outlook and Future Work}
\label{ch:outlook}

This dissertation introduced frameworks to generate human-object interactions and demonstrated the applicability of simulated humans to human-robot interaction. While we have taken a first step in these directions, more exploration is required. In this chapter, we outline and discuss two different directions we find promising; what we believe the next steps in human-robot interaction are and the potential of our methods to advance dexterous robotic grasping.

\section{The Future of Human-Robot Interaction}
Despite the progress in human-robot interaction, many challenges and problems that make it deployable into the real world remain. We discuss the future of human-robot interaction from several perspectives; looking at the tasks that we deem important for the community to solve, the improvements that need to be made with the current frameworks for both the robot and human to adhere to these tasks. These discussions are complimentary to the technical improvements suggested for hand-object interaction synthesis and human-to-robot handovers in Chapter \ref{ch:hoi:conclusion} and Chapter \ref{ch:handover:conclusion}, respectively. 

\subsection{Future Task Settings}
\label{sec:future_tasks}
We discuss three potential directions of novel task settings. First, we describe direct extensions of our current human-to-robot handover setting. Next, we discuss longer horizon tasks. Finally, we depict tasks where the robot and human must cooperate to solve a task.

\paragraph{Making Human-to-Robot Handovers more Realistic}
Human-to-robot handovers offer a controlled testing environment that integrates several important aspects of robotics, including control, sensing, and human-robot interaction. However, the scope of current handover tasks and settings could be broadened. One limitation in our task setup is the assumption of brief interactions where the object is pre-determined and the human immediately initiates the handover. In real-world scenarios, humans often manipulate objects before deciding to hand them over to another person or decide to abort the handover. Extending the problem setting to include prediction of human intent and decision-making could provide more realistic interaction scenarios. Moreover, our user study revealed that humans assist the robot if they anticipate failures, indicating that the human is actively monitoring the robot's actions. In human-to-human handovers, however, minimal attention is required, as both parties rely on learned behaviors. Humans can even perform successful handovers in the absence of vision-based cues by relying on other sensor capabilities such as touch. Therefore, enhancing robot models (see \secref{sec:future_robot_models}) to achieve similar reliability is a critical goal for future work.

\paragraph{Long Horizon Tasks}
If we imagine robots integrated into our daily lives, they could assist humans in a wide range of tasks that extend beyond handovers. These robots would need to be flexible, versatile, and capable of adapting to human needs. To achieve this, exploring longer, more interactive scenarios is essential. For example, a robot might serve as a cooking assistant, helping a person prepare a meal, or assist a child in building a Lego castle. Such complex tasks demand adaptations across multiple aspects of existing frameworks, especially the integration of task planning and diverse low-level interactions, with handovers being just one of them. These interactions could include both robot-to-human and human-to-robot handovers, as well as subtasks like handling specific manipulation steps that are delegated to the robot. We discuss potential approaches to handling such tasks from both the perspective of the human and the robot model in \secref{sec:future_human_models} and \secref{sec:future_robot_models}, respectively. 

\paragraph{Shared Control in Collaborative Tasks}
The aforementioned tasks involve both robots and humans interacting in a shared environment, where either party can handle certain subtasks, like stirring a pot or chopping vegetables. Other tasks demand direct collaboration with shared control over an object, such as carrying a heavy table or moving a cabinet across a room. Understanding how such collaborative interactions can be achieved between robots and humans is a promising area of research. There have been recent datasets \cite{zhang2024core4d} capturing such tasks between humans, which could be leveraged for learning human-robot interactions. As with the other mentioned scenarios, the human behavior model needs refinement to better simulate longer and more realistic interactions, which we discuss in more detail in \secref{sec:future_human_models}. On the other hand, the robot model should anticipate human behavior and adjust accordingly, which we discuss in \secref{sec:future_robot_models}.

\subsection{Human Models}
\label{sec:future_human_models}

\paragraph{Going beyond Disembodied Hands}
While focusing solely on the human hand has been sufficient for tasks like human-robot handovers, more complex or longer-horizon tasks (see \secref{sec:future_tasks}) may require modeling the entire human body. Our exploration in whole-body human grasping, as demonstrated in our recent works  \cite{braun:2023:physically, luo:2024:grasping}, represents a step in this direction. A key aspect of these models is that they enable conditioning on specific trajectories, such as those of the pelvis or objects. While these works primarily focused on grasping and subsequent object motion, future research can build on this foundation by developing more versatile models capable of performing multiple tasks, including locomotion, grasping, and broader environment interactions. 

\paragraph{Improved Frameworks} To move beyond disembodied hand models, we must refine model architectures and training procedures, as demonstrated in our work on whole-body hand-object synthesis \cite{braun:2023:physically, luo:2024:grasping}. Further improvements are necessary to align these models with intuitive human inputs, such as natural language commands, enabling a human to instruct a robot on the next task to execute. Insights from recent studies on human motion synthesis based on language descriptions, particularly through diffusion models \cite{tevet2023human, karunratanakul2023guided}, could be adapted to enhance human-object interaction scenarios \cite{christen:2024:diffh2o}.  

Our current framework primarily synthesizes short-term object interactions with the exception of a proof-of-concept longer sequence in ArtiGrasp (Chapter \ref{ch:hoi_generation:artigrasp}). As outlined in Section \ref{sec:future_tasks}, exploring long horizon and collaborative tasks presents an exciting challenge. Addressing this will require enhancements to human models, potentially through a hierarchical model structure. In such a structure, tasks are decomposed into varying levels of abstraction. For instance, in a collaborative cooking task, the overall goal is to prepare a meal. This goal entails several subtasks, such as preparing vegetables or boiling water, which can be further broken down into smaller subtasks like grasping a knife, cutting vegetables, and putting them into the pan. A more general framework for long-term interactions could integrate a planning layer, that understands the task structure and determines the sequence of subtasks  (e.g., cutting vegetables, boiling water, etc.), taking into consideration the robot assistant as well. The selected subtask could then invoke pre-existing low-level modules responsible for specific interactions (e.g., grasping the knife or cutting the vegetables). In this model architecture, our frameworks D-Grasp and ArtiGrasp could be utilized to generate the respective low-level interactions.

The long-horizon motion generation showcased in ArtiGrasp demonstrates the potential of this hierarchical approach, however, the individual subtasks are manually defined and the transitions between objects rely on heuristics. Future research should aim to develop high-level models capable of autonomously prompting the correct subtasks based on task progress and overall goals.
A key research question will be determining the ideal level of abstraction and the number of hierarchical levels, a topic extensively studied in hierarchical reinforcement learning \cite{sutton1999tempabstr, dayan2000feudalrl, sasha2017fun, schmidhuber1991learningTG}. This presents several technical challenges, such as how to optimize for the number of abstraction levels and subtasks, as well as how to incorporate prior task knowledge and the robot assistant into the model architecture. 

\paragraph{Reactive Human Models}
Up until now, we have assumed that humans act passively in their environment, such as merely moving an object to a handover pose. Similarly, by focusing only on short-term interactions, we have neglected the role of humans in reacting to the robot's actions at a task-planning level. However, this assumption is unrealistic in real-world scenarios. As demonstrated in our real-system experiments (Chapter \ref{ch:handovers:handoversim2real}), humans tend to assist robots, especially when they anticipate a potential failure in the robot's grasp during a handover. This observation highlights the need to model humans as active, reactive agents rather than passive participants.

As discussed in the previous section, long-term tasks often require collaboration between humans and robots. In hierarchical frameworks, higher levels of abstraction determine which subtasks to execute next. The human may adapt their plan or choose a subtask based on the robot's behavior, leading to dynamic interaction patterns. Therefore, an important step in developing new frameworks is to model the human in a reactive, rather than passive, manner. This is crucial for both low-level interactions, like human-to-robot handovers, and high-level interactions, such as selecting the optimal next subtask to execute, especially in the context of extended tasks. 

However, developing reactive human models presents several challenges. It is difficult to collect human-robot interaction data that accurately reflects how humans respond to robots, particularly for low-level interactions. Additionally, humans may adapt their behavior over time, requiring human behavior models that include such continual learning \cite{hadsell2020embracing}. For low-level tasks, such as human-to-robot handovers, one approach could be to use multi-agent reinforcement learning, allowing a human policy and robot optimize their behavior jointly \cite{langerak2022marlui}. Ideally, such an approach could lead to emerging human behavior that closely resembles real human responses.

To gather data for high-level task planning in human-robot collaboration, we could use pre-learned robot behaviors to fulfill subtasks and then allow a human to interact with the robot to solve a high-level task together. For example, a human could give the robot instructions on which subtask to address next, providing valuable data on collaborative decision-making processes.

\subsection{Robot Models}
\label{sec:future_robot_models}

\paragraph{Going beyond Two-Fingered Grippers}
In our work, as well as in much of the existing research on human-robot interaction, a basic two-fingered gripper is commonly used as the robot's end-effector. While this design is adequate for simple handover tasks, more advanced and interactive scenarios will require robots with dexterous manipulation capabilities. The development of anthropomorphic hands therefore represents an important step forward, enabling robots to perform fine-grained, human-like motor skills and manipulate a large variety of objects. 

\paragraph{Improved Frameworks}
As discussed in Section \ref{sec:future_human_models}, just as human models need to become more general to meet the demands of long-term interaction and collaboration, so too must robot models. Our current framework for handovers models the transfer of object between human and robot. While this is a fundamental interaction that needs to work robustly before tackling more complex tasks, handovers should ultimately be viewed as just one of many low-level skills within a holistic task context. For example, in the collaborative cooking scenario, the robot might need to pass a vegetable to the human or vice-versa—an action that, while crucial, is only one part of the larger task of meal preparation. To perform effectively in such scenarios, robots must be versatile and capable of executing a wide range of interactions. 

An interesting direction for future work in robot models is to explore hierarchical models, as proposed for human models in \secref{sec:future_human_models}.  These models could integrate task and motion planning at higher levels of abstraction, determining which subtasks to perform next and which sub-skills are needed for each subtask. For instance, a subtask might involve grasping an object from a human, while the overarching goal could be something more complex, like preparing a meal.

Another important aspect to improve in robot models is their ability to adapt to unforeseen human behavior. Humans may act in ways the robot has not previously encountered, hence the model must be robust and flexible in responding to such behavior. One approach to addressing this challenge is to enhance the robot's ability to perceive and anticipate human behavior or intent, such as predicting human movements and integrating priors about human behavior and task knowledge. For example, the task prior might include all the steps necessary to prepare a dish, whereas the human prior could include the tasks a person prefers to handle independently versus those they would like the robot's assistance with. This capability would enable the model to predict the optimal robot action to support the human effectively. 

Importantly, optimal robot behavior may vary depending on the user's preferences, making it essential for robots to adapt their behavior accordingly \cite{hwang2024promptable}. For instance, one person might prioritize the robot's efficiency in cutting vegetables, while another person might value minimizing the mess made in the kitchen. For humans to communicate their preferences directly to the robot or allowing the robot to query the human when uncertain about how to assist, integrating language into the robot's model could lead to better support. Additionally, incorporating continual learning techniques \cite{hadsell2020embracing} would enable robots to learn and adapt to best support humans over the long term.

\section{Dexterous Robotic Grasping}
Very related to our work on generating human grasping behavior is dexterous robotic grasping. Achieving robust dexterous grasping is an open area of research, and both hardware and algorithms need to improve towards this goal. In this section, we outline the steps we believe are required to achieve this on the algorithm side.

\subsection{Modeling Robotic Systems and Hand Morphologies}
The frameworks we introduced focused on a disembodied hand, however, on real robotic systems, hands are typically integrated with a robotic arm. Future work could extend our approach by incorporating a robotic arm, which introduces workspace constraints due to the added degrees of freedom. A straightforward approach is to directly apply our method with these additional constraints, where the global hand control is achieved through the control of the robotic arm. For instance, the global hand pose could still be predicted, followed by inverse kinematics to compute the arm's joint configuration. If this approach does not succeed, novel strategies on the control or learning side of the framework are interesting directions to explore.

Furthermore, our current frameworks model the human hand, whereas robotic hands typically exhibit distinct morphologies and degrees of freedom. For example, replicating the versatility of the human thumb remains challenging in robotic hand designs \cite{shadowrobotshadow}. Given these differences, further exploration is needed to adapt our methodology for robotic hands. One approach is to focus on retargeting from the human hand to a robotic hand, either by adjusting the grasp references our method relies on or by leveraging the trained policy for the human hand. In the first scenario, the grasp references would be retargeted to a robotic hand, followed by training our method with the robotic hand. In the second scenario, the trained policy for the human hand could serve as a teacher model and be distilled into a policy for the robotic hand. Another approach is to eliminate the dependency on human grasp references altogether and develop a general framework to directly train robust grasping with a robotic hand. We explored this strategy in our GraspXL work \cite{zhang:2024:graspxl}, where we demonstrate successfully grasping of 500k+ objects across various hand morphologies. However, without human grasp references, we fall short of achieving functional grasping which we presented in D-Grasp or ArtiGrasp. Consequently, an interesting direction for future research lies in integrating human-like grasp priors with general grasping techniques.

\subsection{Transfer to Real-World Sensing and Actuation Capabilities} In our works, D-Grasp and ArtiGrasp, the policies rely on perfect observations of object pose, velocity, and force sensor data. However, obtaining such accurate observations in the real world is challenging. For instance, tracking object pose and velocity requires techniques like multi-view pose estimation \cite{labbe2020cosypose} or monocular video methods such as BundleSDF \cite{wen2023bundlesdf}. Despite these techniques, some noise in the tracked position is inevitable, which can lead to grasping failures if the object pose is slightly off. To mitigate this, visual feedback could be used to verify whether an object has been grasped correctly and whether the hand pose needs adjustment. Additionally, tactile sensing plays a crucial role in dexterous grasping, yet remains challenging in robotic manipulation both from hardware and algorithmic perspectives \cite{yousef2011tactile}. In our frameworks, we utilized perfect contact and force information and passed it to the policy. Our frameworks assume perfect contact and force information, but in reality, a domain gap exists that must be accounted for. This gap also extends to actuators, where the torques applied in the real world may differ from those in simulation, leading to discrepancies in outcomes.

Addressing the sim-to-real gap has been approached in two predominant ways \cite{hwangbo2019learning}. The first approach is to increase simulation fidelity, making the sensors and actuators in simulation more closely match their real-world counterparts. The second approach is to improve the system's robustness robust to imperfect simulation. This can be achieved through domain randomization, adding noise or perturbations to sensor data in simulation, or using a stochastic policy. Given the recent success of diffusion models in denoising \cite{ho2020denoising, sohl2015deep}, this technique could also be explored for cleaning noisy sensor observations. Another promising method involves learning a mapping between simulated and real-world data. For example, a mapping can be learned in a self-supervised manner, as demonstrated in \cite{hwangbo2019learning}, where policy actions are mapped to torques via a neural network using real-world data. This can then be used during training in simulation to produce realistic torques that allow a policy to generate actions suitable for direct input to the real world robot controller. A similar approach could be applied to sensors by learning a mapping between the simulated and real world force sensing. 

Another interesting approach worth exploring is the elimination of privileged information. To tackle this, a teacher-student framework could be employed, by training a teacher policy with access to privileged information and distilling it into a student policy that only has access to realistic information \cite{chen2021simple, wan2023unidexgrasp++}. Alternatively, all privileged information could be encoded into a latent vector, which is then predicted in a second stage based on the history of states and actions. A policy that uses a predicted latent as input could be fine-tuned before deployment to the real system. While this has been validated for legged robots \cite{kumar2021rma}, applying a similar principle is an interesting direction for future work.

%% file: frontbackmatter/bibliography.tex
\manualmark%
\markboth{\spacedlowsmallcaps{\bibname}}{\spacedlowsmallcaps{\bibname}} 
\refstepcounter{dummy}
\addtocontents{toc}{\protect\vspace{\beforebibskip}} 
\addcontentsline{toc}{chapter}{\tocEntry{\bibname}}
\label{app:bibliography}
{%
  \emergencystretch=1em%
  \printbibliography%
}

\newcommand{\bibstyleheader}[1]{%
  \section*{\normalsize #1}  
  \markright{#1} 
  }

\bibstyleheader{Generative AI Used}
The following tool has been used during the writing of this dissertation. Prompts were, or in similar spirit to, "Correct the grammar and syntax [paragraph]", or "Give feedback on [paragraph]." Furthermore, they have been used to support the translatation of the English abstract into German. 
\begin{enumerate}
    \item \fullcite{openai2023chatgpt}
\end{enumerate}